%% file: main.tex
\RequirePackage{silence}
\documentclass[%
	paper=A4,					
	twoside=true,				
	openright,					
	parskip=full,				
	chapterprefix=true,			
	11pt,						
	headings=normal,			
	bibliography=totoc,			
	listof=totoc,				
	titlepage=on,				
	captions=tableabove,		
	draft=false,				
]{scrreprt}%

\newcommand{\thesisTitle}{On the Stability of Neural Networks in Deep Learning
}
\newcommand{\thesisName}{Blaise Delattre}
\newcommand{\thesisSubject}{Machine Learning}

\newcommand{\thesisUniversity}{\protect{Universit\'e Paris-Dauphine-PSL}}

\usepackage[utf8]{inputenc}		
\usepackage[english]{babel} 

\usepackage[
    figuresep=colon,
    sansserif=false,
    hangfigurecaption=false,
    hangsection=true,
    hangsubsection=true,
    colorize=full,
    colortheme=bluemagenta,
    bibsys=bibtex,             
	bibfile=references,%
	bibstyle=authoryear, 
]{cleanthesis}

\bibliography{bibliography/springer,bibliography/neurips,bibliography/linear_algebra,bibliography/aaai,bibliography/icml,bibliography/iclr,bibliography/iccv,bibliography/eccv,bibliography/cvpr,bibliography/arxiv,bibliography/other}

\hypersetup{					
	pdftitle={\thesisTitle},	
	pdfsubject={\thesisSubject},
	pdfauthor={\thesisName},	
	plainpages=false,			
	colorlinks=false,			
	pdfborder={0 0 0},			
	breaklinks=true,			
	bookmarksnumbered=true,		%
	bookmarksopen=true			%
}

\usepackage{notation}
\usepackage{amsthm}
\usepackage{amsmath}
\usepackage{mathtools}
\usepackage{float}
\usepackage{tabularx}
\usepackage{minitoc}
\usepackage{algorithm}
\usepackage[noend]{algpseudocode}

\usepackage{stmaryrd}
\usepackage[space]{grffile}
\usepackage{booktabs}
\usepackage{bibentry}
\usepackage{scrhack}
\usepackage{algpseudocode}

\usepackage{graphicx}
\usepackage{caption}
\usepackage{pifont}
\usepackage{subcaption}

\usepackage{listings}

\usepackage{venndiagram}

\usepackage{tikz}
\usetikzlibrary{matrix,calc}

\usepackage{xspace}
\usepackage{xcolor}
\usepackage{pgf}
\definecolor{myorange}{RGB}{247,214,157}
\definecolor{burstred}{RGB}{235,50,35}
\definecolor{burstgreen}{RGB}{140,206,89}
\definecolor{burstblue}{RGB}{75,174,234}
\usepackage{xspace}
\usepackage{enumitem}
\usepackage{varwidth}
\usepackage{multirow}
\usepackage{wrapfig}
\usetikzlibrary{fit}
\usetikzlibrary{calc}
\usetikzlibrary{shapes}
\usetikzlibrary{arrows.meta}
\usetikzlibrary{decorations.text}
\usetikzlibrary{decorations.pathreplacing}
\usetikzlibrary{decorations,calligraphy}

\usepackage{lmodern} 

\usepackage{bbm}
\usepackage{multirow}

\usepackage{psl-cover}
\pslassetspath{figures}

\newcolumntype{Y}{>{\centering\arraybackslash}X}

\title{\thesisTitle}

\author{\thesisName}

\institute{\thesisUniversity}
\doctoralschool{\'Ecole Doctorale SDOSE}{543}
\specialty{Informatique}
\date{20 Juin 2025}

\jurymember{1}{Francis BACH}{Directeur de recherche CRNS, ENS-PSL}{Président}
\jurymember{2}{Rémi GRIBONVAL}{Directeur de recherche, INRIA Lyon}{Rapporteur}
\jurymember{3}{Audrey REPETTI}{Associate Professor, Maxwell Institute for Mathematical Sciences Edinburgh}{Rapporteure}
\jurymember{4}{Claire BOYER}{Professeure des universités
Institut de Mathématiques d'Orsay, Université Paris-Saclay}{Examinatrice}
\jurymember{5}{Franck MAMALET}{Senior Expert in Artificial Intelligence, IRT St-Exupery}{Examniateur}
\jurymember{6}{Quentin BARTHELEMY}{Chercheur, Foxstream Lyon}{Encadrant de thèse}
\jurymember{7}{Alexandre ALLAUZEN}{Professeur, Paris-Dauphine-PSL}{Directeur de thèse}

\frabstract{
L’apprentissage profond a connu des succès remarquables dans de nombreux domaines, mais ses modèles restent fragiles et instables : de petites perturbations des données d’entrée peuvent entraîner de fortes variations des prédictions, et l’optimisation est souvent affectée par des paysages de perte trop abrupts. Cette thèse aborde ces problèmes à travers le prisme unificateur de l’analyse de sensibilité, qui étudie la réponse des réseaux de neurones aux perturbations des entrées et des paramètres.

Nous explorons les réseaux Lipschitziens comme moyen de contrôler la sensibilité aux perturbations des entrées, améliorant ainsi la généralisation, la robustesse face aux attaques adversariales et la stabilité de l’entraînement. En complément, nous introduisons des techniques de régularisation basées sur la courbure de la fonction de perte, afin de favoriser des paysages plus lisses et de réduire la sensibilité aux paramètres. Le lissage aléatoire est également étudié comme méthode probabiliste pour renforcer la robustesse au voisinage des frontières de décision.

En combinant ces approches, nous proposons un cadre unifié reliant la continuité Lipschitzienne, le lissage aléatoire et la régularisation par courbure pour traiter les défis fondamentaux de la stabilité. La thèse apporte à la fois des analyses théoriques et des méthodes pratiques, incluant le calcul efficace de normes spectrales, de nouvelles couches contraignant la constante de Lipschitz, et des procédures de certification améliorées.
}

\enabstract{
Deep learning has achieved remarkable success across a wide range of tasks, but its models often suffer from instability and vulnerability: small changes to the input may drastically affect predictions, while optimization can be hindered by sharp loss landscapes. This thesis addresses these issues through the unifying perspective of sensitivity analysis, which examines how neural networks respond to perturbations at both the input and parameter levels.

We study Lipschitz networks as a principled way to constrain sensitivity to input perturbations, thereby improving generalization, adversarial robustness, and training stability. To complement this architectural approach, we introduce regularization techniques based on the curvature of the loss function, promoting smoother optimization landscapes and reducing sensitivity to parameter variations. Randomized smoothing is also explored as a probabilistic method for enhancing robustness at decision boundaries.

By combining these perspectives, we develop a unified framework where Lipschitz continuity, randomized smoothing, and curvature regularization interact to address fundamental challenges in stability. The thesis contributes both theoretical analysis and practical methodologies, including efficient spectral norm computation, novel Lipschitz-constrained layers, and improved certification procedures.
}

\frkeywords{Stabilité, Réseaux Lipschitz, Robustesse, Norme spectrale, Lissage aléatoire, Régularisation de Hessienne, Apprentissage profond}
\enkeywords{Stability, Lipschitz networks, Robustness, Spectral norm, Randomized smoothing, Hessian regularization, Deep learning}

\usepackage{setspace}
\begin{document}

\renewcaptionname{english}{\figurename}{Fig.}
\renewcaptionname{english}{\tablename}{Tab.}

\pslcover{}

\pagenumbering{roman}			
\pagestyle{empty}				
\cleardoublepage%
\pagestyle{plain}				
\cleardoublepage%
\input{content/acknowledgement} 
\cleardoublepage%
\cleardoublepage%

\dominitoc%

\setcounter{tocdepth}{2}		
\tableofcontents				
\cleardoublepage%

\cleardoublepage%

\pagenumbering{arabic}			
\setcounter{page}{1}			
\pagestyle{maincontentstyle} 	

\input{content/chapter-introduction}
\input{content/chapter-background}
\input{content/chapter-spectral_norm_computation}
\input{content/chapter-lipschitz_computation}
\input{content/chapter-robustness}
\input{content/chapter-risk_management.tex}
\input{content/chapter-regularization}

\input{content/chapter-conclusion}

\cleardoublepage%
\appendix
\counterwithin{figure}{section}
\counterwithin{table}{section}
\renewcommand{\thetable}{\thesection.\arabic{table}}
\renewcommand{\thefigure}{\thesection.\arabic{figure}}
\cleardoublepage%

\input{content/appendix/appendix-spectral_norm.tex}
\input{content/appendix/appendix-lipschitz.tex}
\input{content/appendix/appendix-robustness.tex}
\input{content/appendix/appendix-risk_management.tex}
\input{content/appendix/appendix-regularization.tex}

%

{%
	\setstretch{1.1}
	\renewcommand{\bibfont}{\normalfont\small}
	\setlength{\biblabelsep}{0pt}
	\setlength{\bibitemsep}{0.5\baselineskip plus 0.5\baselineskip}
	\begin{refcontext}[sorting=none]
		\printbibliography[nottype=online]
		\printbibliography[heading=subbibliography,title={Webseiten},type=online,prefixnumbers={@}]
	\end{refcontext}
}

\cleardoublepage%

\listoffigures
\cleardoublepage%

\listoftables
\cleardoublepage%



\end{document}

%% file: content/acknowledgement.tex
%
\pdfbookmark[0]{Remerciements}{Remerciements}
\chapter*{Funding and Grants}\label{sec:acknowledgement}
\vspace*{-10mm}

This doctoral research was conducted within the Machine Intelligence and LEarning Systems (MILES) team of the LAMSADE research lab at Université Paris-Dauphine / Université PSL, between April 2022 and March 2025, as part of a CIFRE collaboration with the company Foxstream.
It was supported by the French National Research Agency (ANR) through the SPEED project (ANR-20-CE23-0025) and by the French Government via the France 2030 program (ANR-23-PEIA-0008, SHARP).
Part of this work was performed using high-performance computing resources provided by GENCI-IDRIS (Grant 2023-AD011014214R1).

\chapter*{Remerciements}\label{sec:thankss}
\vspace*{-10mm}

Je souhaite exprimer ma profonde reconnaissance à l’ensemble des membres du jury pour l’honneur qu’ils m’ont fait en acceptant d’évaluer ce travail.
Je remercie Francis Bach, Président du jury, ainsi que Claire Boyer et Franck Mamalet, examinateurs, pour l’attention portée à cette soutenance et pour la pertinence de leurs remarques et questions.
J’adresse une gratitude toute spéciale à Rémi Gribonval et à Audrey Repetti, rapporteurs de cette thèse, pour la rigueur et la précision de leur lecture, ainsi que pour la richesse et la profondeur de leurs commentaires. Leurs observations détaillées ont grandement contribué à améliorer la qualité de ce manuscrit.

J’exprime aussi mes remerciements les plus sincères à mes encadrants.
Je remercie tout particulièrement Quentin Barthélemy pour son encadrement soutenu et bienveillant, pour sa disponibilité, ainsi que pour m’avoir fait découvrir le monde de la recherche.
Je suis également profondément reconnaissant à Alexandre Allauzen pour son regard avisé, mais aussi pour sa bienveillance et sa bonne humeur, qui ont contribué à rendre ces années de thèse particulièrement stimulantes et enrichissantes.
Je tiens enfin à remercier Alexandre Araujo pour son accompagnement précieux au début de ma thèse et pour l’aide qu’il m’a apportée.

Je suis reconnaissant à l’ensemble des membres de l’entreprise Foxstream pour leur accueil chaleureux et leur sympathie tout au long de cette collaboration.
Je souhaite aussi remercier l’équipe MILES, que j’ai eu beaucoup de plaisir à rencontrer et à côtoyer au quotidien. Discussions, collaborations et moments de vie partagés ont largement contribué à rendre ces années riches et agréables.
Je pense en particulier à Yann, Benjamin et Jamal, mais aussi aux doctorants et post-doctorants Alexandre, Paul, Florian, Erwan et Lucas, avec qui j’ai eu la chance de collaborer et de partager de nombreux moments de camaraderie.

Enfin, je souhaite remercier mes amis, ma famille et ma compagne pour tout ce qu’ils m’apportent jour après jour et plus particulièrement durant ces années de thèse.
Je pense en particulier à mes parents, qui m’ont toujours encouragé dans ce projet depuis le début, et à mon père, qui m’a transmis le goût de la recherche et avec qui j’ai aujourd’hui le plaisir de le partager.

%% file: content/chapter-introduction.tex
\chapter{Introduction}\label{chapter:intro}
\minitoc%
\section{Success of deep learning}
Deep learning has driven groundbreaking progress across various domains, fundamentally transforming modern technology. In computer vision, its impact has been profound, beginning with early applications in digit recognition and image classification~\citep{lecun1998gradient}. The field reached a milestone with the success of AlexNet, which achieved a dramatic breakthrough on ImageNet by significantly outperforming traditional methods using deep convolutional neural networks (CNNs)~\citep{krizhevsky2012imagenet}. This success has inspired advancements in diverse fields, from healthcare and biology to natural language processing and robotics, showcasing deep learning’s transformative potential.

In the medical domain, deep learning has revolutionized imaging, with U-Net~\citep{ronneberger2015unet} enabling precise segmentation of biomedical structures, advancing cancer detection, and automating diagnostic tasks~\citep{falk2019unet}. Similarly, AlphaGo’s victory against world champions in the board game Go demonstrated deep learning’s capacity for strategic decision-making through reinforcement learning~\citep{silver2016alphago}.
Biology has also witnessed transformative advancements, most notably in protein structure prediction. AlphaFold~\citep{jumper2021alphafold} resolved a decades-old challenge, achieving near-experimental accuracy in predicting protein structures and paving the way for advances in drug discovery and molecular biology.
Generative models have expanded the boundaries of creativity and utility across numerous domains. GANs~\citep{goodfellow2014generative} have set the standard for high-quality image synthesis, enabling applications such as photorealistic image generation, artistic creation, and deepfake technology. Variational Autoencoders (VAEs)~\citep{kingma2014auto} have advanced probabilistic modeling and latent space representation, while diffusion models~\citep{ho2020denoising} have emerged as state-of-the-art for generating realistic and diverse outputs in fields like molecular design and scientific simulations.
CNN-based architectures are fundamental to computer vision. Models like YOLO~\citep{redmon2016you, yolov5} are essential for real-time object detection, combining accuracy with low latency for tasks such as autonomous driving. Modern CNNs such as ConvNeXt~\citep{liu2022convnet} continue to deliver state-of-the-art performance with computational efficiency. Vision Transformers (ViTs)~\citep{dosovitskiy2020image} have further extended deep learning’s capabilities by leveraging self-attention mechanisms~\citep{vaswani2017attention} to excel in large-scale image classification tasks.
%
%
In natural language processing, attention has enabled significant advancements, replacing recurrent neural networks~\citep{hochreiter1997long, cho2014learning} as the dominant architecture for sequence modeling, with early attention mechanisms first introduced in neural machine translation~\citep{bahdanau2015neural} and later generalized to self-attention~\citep{vaswani2017attention}.
This shift has led to the development of large language models (LLMs) that have driven remarkable breakthroughs. BERT~\citep{devlin2019bert} introduced bidirectional pre-training, enabling context-aware embeddings that revolutionized tasks such as question answering and sentiment analysis. Generative models like GPT~\citep{brown2020language} have scaled autoregressive training to generate coherent and contextually relevant text, transforming applications such as conversational AI, machine translation, and content creation. Notably, LLMs exhibit remarkable emergence phenomena~\citep{wei2022emergent}, including zero-shot and in-context learning capabilities~\citep{brown2020language}, allowing them to perform tasks without explicit task-specific fine-tuning.

The success of deep learning across vision and language tasks has been largely driven by scaling—deeper networks, larger datasets, and increased computational power—facilitated by open-source frameworks like PyTorch~\citep{paszke2019pytorch} and TensorFlow~\citep{abadi2016tensorflow},
which provide efficient tools for automatic differentiation to streamline gradient computation and accelerate model development~\citep{kaplan2020scaling, brown2020language}. A key enabler of this scaling is the parallelization of computations on GPUs, which dramatically speeds up matrix operations and enables the efficient training of large models on massive datasets. Modern GPUs, along with distributed training frameworks such as TensorFlow's and PyTorch’s distributed module, have made it possible to train models with billions of parameters by distributing computations across multiple devices, reducing training times from weeks to hours.
This ability to efficiently scale model depth and size enables neural networks to learn intricate hierarchical representations, capturing patterns that shallower architectures cannot achieve~\citep{he2016deep, mallat2016understanding, dosovitskiy2020image}.
The increasing depth and parameter scale of modern neural networks has driven remarkable successes in deep learning. However, this depth comes at a cost, introducing significant challenges related to the stability of these models.
Key issues include adversarial robustness, training stability, and generalization.
%
%
\section{Seeking stability}
%
Stability is a foundational concept across multiple scientific disciplines, reflecting the ability of a system to maintain consistent behavior under perturbations.
The formal study of stability originated in the field of dynamical systems, with Lyapunov’s seminal work~\citep{lyapunov1892stability}, which introduced methods to assess whether systems return to equilibrium after disturbances. These concepts, initially developed for physical systems, remain highly relevant to modern machine learning, where neural networks can be viewed as discrete dynamical systems during both training and inference.

The work of~\citet{tikhonov1943regularization}, who studied the continuity of solutions to ill-posed problems, laying the groundwork for stability analysis in inverse problems. Tikhonov regularization was introduced to stabilize solutions to such problems, ensuring robustness to small perturbations in data. This interplay between stability and robustness has since influenced modern learning theory, where regularization techniques are used to control sensitivity to noise.
%

Numerical stability was also a concern in computational systems, as highlighted by~\citet{vonneumann1947numerical}, who introduced the concept of condition numbers to quantify the sensitivity of solutions to input perturbations.
%
Early foundations of statistical stability can be traced back to~\citet{cramer1946mathematical}, who formalized how perturbations in data affect estimators. His work introduced key principles such as consistency, ensuring estimators converge to the true parameter as the sample size grows, and asymptotic efficiency, which guarantees minimal variance in large-sample regimes. These properties help mitigate sensitivity to small fluctuations in data, forming a theoretical basis for stable statistical methods.
More recent advancements in robust statistics have further refined these stability concepts. In particular, the median-of-means (MoM) estimator~\citep{jerrum1986median, nemirovsky1983problem, lugosi2019mean} has become a key tool for robust mean estimation under heavy-tailed distributions. Unlike the empirical mean, which can be severely impacted by outliers, MoM partitions data into subsets, computes local means, and aggregates them using the median. This approach ensures strong concentration properties even in adversarial settings, making it a fundamental technique for stability in modern statistical learning.

Generalization stability refers to a model's ability to remain unaffected by small changes in the training data, such as variations due to sampling, ensuring that the learned function and its performance on unseen data are consistent.
%
\citet{vapnik1971uniform} introduced the concept of the Vapnik-Chervonenkis (VC) dimension, which measures the complexity of a hypothesis class by quantifying the largest number of points it can shatter—that is, classify in all possible ways. Their work provided fundamental theoretical insights into generalization by deriving bounds on the expected generalization error based on the VC dimension. These results established a direct connection between model complexity and generalization, demonstrating that overly complex models tend to overfit while simpler models generalize better.
%
In the late 1970s, \citet{devroye1979distribution_free} and \citet{devroye1979performance_bounds} explored the connection between generalization error and stability, introducing inequalities for leave-one-out and holdout error estimates. Their work implicitly addressed what \citet{kearns1999stability} later termed hypothesis stability,
formalizing how the variance of leave-one-out error could be bounded using properties of the learning algorithm. They also demonstrated a connection between finite VC dimension and stability, showing that stable algorithms are inherently generalizable.
%
The PAC framework~\citep{valiant1984pac} provided probabilistic guarantees for generalization, while the PAC-Bayes framework~\citep{mcallester1999pac} extended this approach by leveraging prior distributions to derive tighter bounds on the performance of learning algorithms.

Sensitivity analysis~\citep{saltelli2000sensitivity} focuses on quantifying the extent to which variations in inputs can influence a system's outputs. It has been widely applied across various disciplines, including statistics, control theory, and mathematical programming where it is referred as perturbation analysis~\citep{bonnans1996perturbation}.
The primary motivation for such analyses is to design robust systems that remain resilient to noise or disturbances affecting the inputs.
\citet{bousquet2002stability} proposed a sensitivity-based framework to analyze learning algorithms, focusing on how changes in the training set influence the function produced by the algorithm.
Unlike classical sensitivity analysis, their framework emphasizes the randomness introduced by sampling and its impact on generalization, bridging stability with probabilistic guarantees.
%
Differential privacy~\citep{dwork2006calibrating} represents a specialized application of sensitivity analysis, where the focus is on bounding the influence of any data point on the output of an algorithm to ensure privacy. By introducing the concept of global sensitivity, differential privacy quantifies the maximum change in the output due to changes in a data point and calibrates noise to this sensitivity. While its primary goal is privacy protection, the mechanisms of differential privacy inherently limit the dependence of the output on individual inputs, aligning it closely with robustness and stability.

Lipschitz continuity and smoothness emerged as key mathematical tools for analyzing function stability and optimization landscapes. While Lipschitz continuity ensures bounded rates of change in function values, smoothness (characterized by bounded Hessian norm) extends this to gradients~\citep{nesterov2003introductory}. These properties are fundamental to modern optimization theory, providing convergence guarantees for gradient-based methods and a framework for analyzing learning algorithm stability by ensuring that small input perturbations lead to controlled changes in both function values and their gradients.

\section{Challenges in deep learning}
\subsection{Stability to sampling noise in data : Generalization}
The Lipschitz constant of a neural network plays a crucial role in determining its generalization bounds~\citep{bartlett2017spectrally, sokolic2017robust, neyshabur2018pacbayesian}. This constant is often bounded by quantities such as the product of the spectral norms of the network's layers, which provides a measure of the network's sensitivity to input perturbations.
\begin{figure}[h]
    \centering
    \begin{tikzpicture}[scale=1.0]
        \draw[->] (-5,-0.5) -- (-5,4) node[anchor=east] {loss}; 
        \draw[->] (-5,-0.5) -- (5,-0.5) node[anchor=north] {parameters}; 

        \draw[thick] plot [smooth] coordinates {(-4,3) (-3,1) (-2,0) (-1,1) (0,3)}; 

        \draw[red, dashed] plot [smooth] coordinates {(-3.5,3) (-2.5,1.2) (-1.5,0) (-0.5,1.2) (0.5,3)}; 

        \filldraw[black] (-2,-0.7) node[anchor=north] {convex flat minimum};

        \draw[thick] (2,0) parabola (2.5,3); 
        \draw[thick] (2,0) parabola (1.5,3); 

        \draw[red, dashed] (2.5,0) parabola (3,3); 
        \draw[red, dashed] (2.5,0) parabola (2,3); 

        \filldraw[black] (2,-0.7) node[anchor=north] {sharp minimum};

        \filldraw[black] (-4,3) circle(0pt) node[anchor=south west] {train};
        \filldraw[red] (0.5,3) circle(0pt) node[anchor=south] {test};
    \end{tikzpicture}
    \caption{Depiction of convex flat and sharp minima in one dimension. Both types of minima have the same relative shift between train and test losses.
    }
    \label{fig:sharp_flat_minima}
\end{figure}
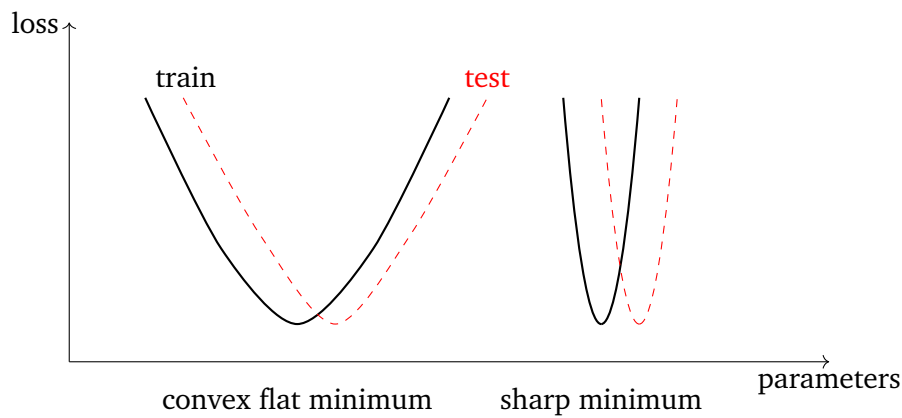
Beyond Lipschitz continuity, sharpness-based metrics offer additional insights into generalization. Sharpness has long been regarded as a strong indicator of generalization, initially defined as the size of the connected region around the minimum where the training loss remains low~\citep{hochreiter1997flat}. Sharpness, captured by the curvature of the loss landscape's Hessian with respect to parameters, reflects the steepness of the loss landscape. Smaller Hessian values have been empirically associated with improved robustness and generalization~\citep{hochreiter1997flat, keskar2017large-batch}.
Flatter minima, characterized by broader and shallower regions in the loss landscape, confer robustness to parameter perturbations and are less sensitive to data distribution shifts. Regularization techniques like Sharpness-Aware Minimization~\citep{foret2021sharpnessaware} explicitly reduce sharpness by minimizing the maximum loss within a local neighborhood of the model’s parameters. Other standard methods, such as Weight Decay (WD)~\citep{krogh1992simple} have also been shown to encourage flatter minima.
Despite these empirical successes, sharpness and flatness metrics are not universally tied to generalization. These measures are sensitive to parameterization and re-scaling effects~\citep{zhang2017rethinking}. Furthermore, their relevance heavily depends on both the data and the model~\citep{andriushchenko2022understanding}. This underscores the need for nuanced approaches when interpreting sharpness-based heuristics.
Nevertheless, techniques that reduce sharpness, including SAM and Hessian-based regularization, remain effective practical tools for achieving generalization. Figure~\ref{fig:sharp_flat_minima} illustrates the differences between sharp and flat minima, highlighting that flat minima are less sensitive to shifts in the data distribution.

We aim to explore a foundational challenge in neural network design: What principles and regularization techniques can guide the design of neural networks that generalize well to unseen data? While existing methods such as Lipschitz-based regularization and sharpness-aware minimization have shown promise, understanding their interplay and limitations remains an active area of research.
%
\subsection{Stability to input: Adversarial robustness}
\begin{figure}[t]
    \centering
    \includegraphics[width=1.0\textwidth]{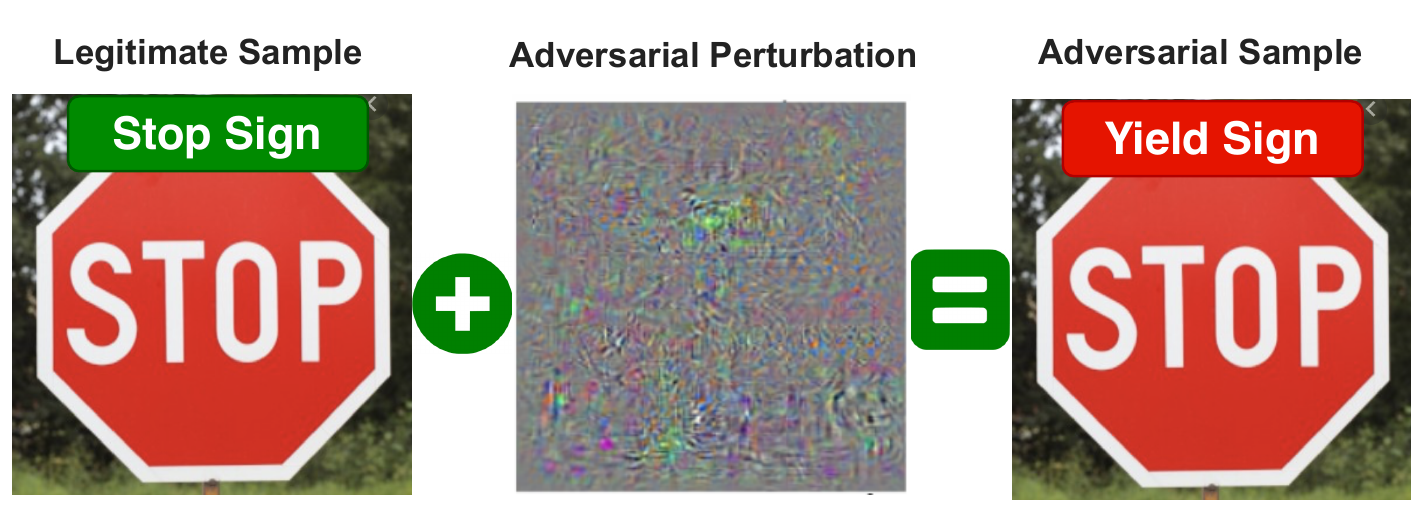}
    \caption{Example of an adversarial attack on a stop sign. The perturbation is imperceptible to the human eye but causes a misclassification by the network. In this case, the stop sign is misclassified as a yield sign. Figure taken from~\citet{ahmad2022developing}.}
    \label{fig:stop_sign_adv}
\end{figure}

The high capacity and complexity of deep networks make them particularly vulnerable to adversarial examples which are small, carefully crafted perturbations in the input that lead to significant deviations in output~\citep{szegedy2013intriguing, goodfellow2014explaining}. These perturbations highlight the sensitivity of models to input changes, which becomes increasingly problematic as networks grow larger and more complex.
For images, an adversarial attack takes the form of a perturbation imperceptible to the human sight added to the input image.
We formulate adversarial robustness as the ability of a model to maintain consistent predictions at inference under input perturbations, ensuring that small changes in the input do not lead to significant changes in the output.
Addressing adversarial robustness is critical, particularly in applications where input perturbations can lead to catastrophic failures, see Figure~\ref{fig:stop_sign_adv} for example, where the adversarial perturbation causes a stop sign to be misclassified as a yield sign.
The most popular defense to adversarial attack is adversarial training~\citep{madry2018towards} where the network is trained on adversarial examples. This method has been shown to improve robustness to adversarial attacks, but it is computationally expensive and does not provide theoretical guarantees against all types of attacks.
Over the past decade, research in adversarial robustness has often resembled a cat-and-mouse game, with increasingly powerful attacks being developed to exploit model vulnerabilities. Notable works in this domain include those by~\citet{goodfellow2014explaining, kurakin2016adversarial, carlini2017towards, croce2020reliable}.

In response to this game, achieving certified adversarial robustness has been a significant focus, with notable contributions such as~\citet{raghunathan2018certified, wong2018provable}.
To achieve this, researchers have primarily focused on two categories of methods.
The first approach, Randomized Smoothing (RS), comes from differential privacy and is based on randomization~\citep{lecuyer2018certified, li2018secondorder, cohen2019certified, pinot2019theoretical, salman2019provably}, where noise is injected to the input.
This method is the only certified defense on the ImageNet dataset offering non-trivial guarantees, but it suffers from large computational overheads that limit its scalability.
%
The second approach involves controlling the relationship between the Lipschitz constant of the network and the prediction margins~\citep{tsuzuku2018lipschitz}, following this work involves constructing 1-Lipschitz layers using specific linear transformations~\citep{cisse2017parseval, li2019preventing, anil2019sorting, li2020implicit, trockman2021orthogonalizing, singla2021improved, singla2021skew} relying on that the Lipschitz constant of a function quantifies the maximum sensitivity of the function's output to changes in its input. By bounding the Lipschitz constant of neural networks, one can compute a certification radius around any given point by establishing maximum sensitivity around the input for a given perturbation's budget. While these methods offer a more direct and theoretically grounded approach, they are hindered by limited performance and scalability.

Given these limitations, a critical question arises: How can we improve existing methods for certified adversarial robustness to address their computational and scalability challenges while maintaining strong guarantees?

\subsection{Training stability}
Stability during training can be defined as low sensitivity to perturbations in the input, parameters, or gradients noise, ensuring that small changes do not lead to drastic variations in the training's output. Training stability is crucial for achieving consistent performance across different runs and datasets, as well as for enabling efficient optimization and convergence to good solutions.

The increasing scale of deep networks, in terms of depth and parameter count, often exacerbates training instability. Training deep networks is inherently challenging due to issues such as exploding and vanishing gradients~\citep{pascanu2013difficulty}, as well as sensitivity to initialization~\citep{glorot2010understanding}. These challenges become especially pronounced in large-scale architectures, where instability can hinder convergence and lead to suboptimal solutions.
One of the earliest breakthroughs in addressing these issues was the development of improved activation functions. Early architectures relied on sigmoid and tanh activations, which often suffered from saturation issues, resulting in vanishing gradients~\citep{hochreiter1997long}. The introduction of ReLU (Rectified Linear Unit)~\citep{nair2010rectified} marked a turning point by allowing gradients to flow more effectively. Subsequent variants, such as Leaky ReLU~\citep{maas2013rectifier}, Parametric ReLU~\citep{he2015delving}, and GELU (Gaussian Error Linear Unit)~\citep{hendrycks2016gelu}, further improved gradient flow and facilitated the training of deeper networks.
Building on these advances, residual connections, introduced with ResNet~\citep{he2016deep}, offered another transformative solution to training instability. While normalization layers such as batch normalization~\citep{ioffe2015batch}, layer normalization~\citep{ba2016layer}, and group normalization~\citep{wu2018group} help stabilize activations and address the vanishing gradient problem, residual connections tackle a related but distinct challenge. By enabling gradients and signals to bypass one or more layers, residual connections mitigate the degradation of the noise-to-signal ratio as information propagates through deep networks~\citep{balduzzi2017shattered}. This mechanism preserves critical information during both forward and backward passes, helping stabilize training in very deep architectures and enabling faster convergence with higher performance.

In addition to architectural innovations, advancements in optimization techniques have been pivotal in improving training stability. Optimizers such as Adam~\citep{kingma2014adam} and its variants (e.g., AdamW~\citep{loshchilov2019decoupled} and LAMB~\citep{you2020large}) have become the standard for training deep models, providing faster convergence and better handling of large parameter spaces. Techniques like learning rate schedules, momentum~\citep{polyak1964some}, and adaptive gradient methods (e.g., Adagrad~\citep{duchi2011adaptive} and RMSProp~\citep{tieleman2012lecture}) further enhance stability and efficiency. Additionally, normalization layers like batch normalization~\citep{ioffe2015batch} not only stabilize activations but also enable the use of larger learning rates, further boosting training dynamics in deep networks.

While these techniques improve training stability, they often rely on complex heuristics and tuning. This underscores the need for architectures that are inherently stable by design, reducing reliance on external tricks.
Lipschitz regularization or normalization techniques, such as spectral normalization~\citep{miyato2018spectral}, have been widely used to stabilize GAN training by constraining the gradient norms, preventing exploding gradients, and ensuring smoother updates for the generator and discriminator.

Lipschitz networks provide robustness to input perturbations and ensure stable gradient flow that does not explode, making them strong candidates for improving stability by design. This naturally raises the question: How can we build efficient Lipschitz neural networks?

\section{Contributions and outline}
\label{sec:contributions_and_outline}

This thesis adopts sensitivity analysis as both a guiding framework and a practical tool to address stability challenges and answer the questions raised in the previous section. By leveraging derivative-based methods, such as Lipschitz constant estimation and Hessian analysis, alongside variance-based techniques like noise injection, we quantify and characterize sensitivity at various levels of deep models, proposing actionable solutions through architecture design that inherently minimizes sensitivity.

Specifically, this work develops Lipschitz networks to control output sensitivity to input perturbations, offering solutions to challenges in generalization, adversarial robustness, and training stability. Furthermore, we introduce Hessian-bounded loss regularization to smooth the loss landscape, reducing parameter sensitivity and improving optimization dynamics. Randomized smoothing is employed to enhance robustness at decision boundaries. Finally, we explore the theoretical and practical connections between these approaches and the Lipschitz continuity of neural networks, unifying them under a common framework.

We aim to address those challenges in the following chapters:
\begin{itemize}
    \item \textbf{Chapter~\ref{chapter:background}} This chapter provides an overview of Lipschitz networks, certified robustness in the context of Lipschitz networks, randomized smoothing, and regularization techniques aimed at improving generalization.

    \item \textbf{Chapter~\ref{chapter:spectral_norm_estimation} }
          This chapter focuses on spectral norm computation, a key step in constructing Lipschitz networks. We introduce \textit{Gram iteration}, an efficient algorithm for estimating the spectral norm of dense and convolutional layers under different padding schemes. Additionally, we establish theoretical connections between the spectral norms of circular and zero-padding convolutions. These results lay the foundation for Lipschitz-constrained architectures.

    \item \textbf{Chapter~\ref{chap:lipschitz_computation} }
          In this chapter, we explore novel approaches for computing and regularizing the Lipschitz constant of convolutional layers, including the development of novel by design Lipschitz layers, spectrally rescaled, based on \textit{Gram iteration}.
          In the same line of work of~\cite{salman2019provably}, we view the randomized smoothing procedure as a way to obtain Lipschitz networks through Weierstrass transform and explore its connections with Lipschitz continuity.

    \item \textbf{Chapter~\ref{chapter:robustness} }
          In this chapter, we propose methods to certify robustness by designing Lipschitz networks using spectrally rescaled layers. Furthermore, we demonstrate the effective integration of randomized smoothing with Lipschitz networks, leveraging their inherent stability to provide better probabilistic guarantees.
          Additionally, we extend the simplex mapping in randomized smoothing to a scaled simplex mapping, enabling improved margin-to-Lipschitz ratio and robustness certification for a broader range of perturbations.

    \item \textbf{Chapter~\ref{chap:risk_management} }
          Randomized smoothing introduces additional stochasticity and variance, necessitating careful management to ensure robust performance. This chapter focuses on risk management in randomized smoothing, with an emphasis on statistical certification procedures and computational efficiency.
          We tackle the Lipschitz-variance-margin tradeoff for Randomized Smoothing (LVM-RS), demonstrating how a balanced approach can achieve both low variance and high margins for robust models. To this end, we propose new techniques such as LVM-RS and class partitioning methods (CPM) for confidence intervals, which enhance robustness while significantly reducing computational overhead.

    \item \textbf{Chapter~\ref{chap:regularization} }
          In this chapter, Lipschitz regularization is used to enhance generalization. We then introduce Hessian curvature regularization as a means to promote flat minima to improve generalization, leveraging the Weierstrass transform applied to the gradient of the loss with respect to parameters.
          Furthermore, we establish a connection between Lipschitz networks and flat minima in their loss landscape, showing how these architectures naturally promote broader, flatter solutions. Finally, we propose the Activation Decay method as a deterministic approach to loss smoothing, complementing Lipschitz regularization.
\end{itemize}

The work presented in this thesis is based on results published in the following peer-reviewed conferences, as well as preprints under review:
\begin{contribframe*}[Contributions included in the manuscript]
    {\quad}
    \begin{my_list_item}
        \item In Chapter~\ref{chapter:spectral_norm_estimation} and Chapter~\ref{chap:regularization}:\\
        B.Delattre, Q.Barthélemy, A.Araujo, and A.Allauzen,
        "Efficient Bound of Lipschitz Constant for Convolutional Layers by Gram Iteration,"
        Proceedings of the 40th International Conference on Machine Learning (ICML), 2023.

        \item In Chapter~\ref{chapter:spectral_norm_estimation} and Chapter~\ref{chap:lipschitz_computation}:\\
        B.Delattre, Q. Barthélemy, and A.Allauzen,
        "Spectral Norm of Convolutional Layers with Circular and Zero Paddings,"
        (under review) arXiv preprint, 2024.

        \item In Chapter~\ref{chap:lipschitz_computation}, Chapter~\ref{chap:risk_management}, and Chapter~\ref{chapter:robustness}:\\
        B.Delattre, A.Araujo, Q.Barthélemy, and A.Allauzen,
        "The Lipschitz-Variance-Margin Tradeoff for Enhanced Randomized Smoothing,"
        International Conference on Learning Representations (ICLR), 2024.\\
        %
        B.Delattre, P.Caillon, Q.Barthélemy, E.Fagnou, and A.Allauzen,
        "Bridging the Theoretical Gap in Randomized Smoothing,"
        Proceedings of the 28th International Conference on Artificial Intelligence and Statistics (AISTATS), 2025.

        \item In Chapter~\ref{chap:regularization}:\\
        B.Delattre, Q.Barthélemy, P.Caillon, E.Fagnou, and A.Allauzen,
        "Regularization by Activation Decay to Enhance Generalization,"
        (under review), 2025.
    \end{my_list_item}
\end{contribframe*}

\section{Additional contributions}
Building on Lyapunov's stability theory, \citet{haber2018stable} interpreted deep learning as a parameter estimation problem in nonlinear dynamical systems. They proposed architectures inspired by ordinary differential equations (ODEs) to address numerical instabilities, such as exploding and vanishing gradients and designed well-posed architectures for arbitrarily deep networks.

Inspired by this perspective, we developed a unified framework for constructing 1-Lipschitz Neural Networks by interpreting Residual Networks as continuous-time dynamical systems. This framework reveals that many existing methods are specific instances of this approach. Moreover, we showed that ResNet flows derived from convex potentials naturally define 1-Lipschitz transformations, leading to the introduction of the \textit{Convex Potential Layer (CPL)}.
This method enables $\ell_2$-provable defenses against adversarial examples and allows for the stable training of extremely deep ResNets, with over 1000 layers, without relying on techniques such as batch normalization or gradient clipping.

We further expanded this work by unifying recent advancements in 1-Lipschitz neural networks, including methods based on convex potentials and other concurrent approaches, under a novel algebraic framework. By leveraging a common semidefinite programming (SDP) condition, we demonstrated that existing techniques can be derived as special cases and introduced new parameterizations for Lipschitz layers using the Gershgorin circle theorem. This approach, termed \textit{SDP-based Lipschitz Layers (SLL)}, generalizes convex potential layers.

These additional works have been disseminated in peer-reviewed venues:
\begin{contribframe*}[Contributions not included in the manuscript]
    {\quad}
    \begin{my_list_item}
        \item L.Meunier$^*$, B.Delattre$^*$, A.Araujo$^*$, and A.Allauzen,\\
        \textit{``A Dynamical System Perspective for Lipschitz Neural Networks,''}\\
        Proceedings of the 39th International Conference on Machine Learning (ICML), 2022.

        \item A.Araujo$^*$, A.J.Havens$^*$, B.Delattre, A.Allauzen, and B.Hu,\\
        \textit{``A Unified Algebraic Perspective on Lipschitz Neural Networks,''}\\
        International Conference on Learning Representations (ICLR), 2023.
    \end{my_list_item}
\end{contribframe*}



%% file: content/chapter-background.tex
%
\chapter{Background}
\label{chapter:background}
\minitoc%

%

We define a feed-forward neural network \(f\) with \(L\) layers by
\begin{equation}
  \label{eq:neural_network_composition}
  f \;=\; f^{(L)} \,\circ\, f^{(L-1)} \,\circ \cdots \circ\, f^{(1)},
\end{equation}
where each layer consists of a weight matrix \( \mathbf{W}^{(l)} \) followed by a nonlinearity \( \nonlin^{(l)} \), typically ReLU or GELU,
\[
  f^{(l)}(\mathbf{z}) \;=\;
  \begin{cases}
    \nonlin^{(l)}\!\bigl(\mathbf{W}^{(l)}\,\mathbf{z}\bigr),
     & l=1,\dots,L-1, \\
    \mathbf{W}^{(L)}\,\mathbf{z},
     & l=L.
  \end{cases}
\]
We denote the spectral norm of \( \mathbf{W}^{(l)} \) by \( \|\mathbf{W}^{(l)}\|_2 \).
For simplicity, we omit biases, which are implicitly accounted for.
We note $\theta$ as the set of all parameters in the network, $\theta = \{ \mathbf{W}^{(l)} \}_{l=1}^L$.
For an input \(\mathbf{x}\), define intermediate outputs inductively:
\[
  h^{(0)}(\mathbf{x}) \;=\; \mathbf{x},
  \quad
  h^{(l)}(\mathbf{x}) \;=\; f^{(l)}\bigl(h^{(l-1)}(\mathbf{x})\bigr),
  \quad
  l=1,\dots, L.
\]
Then \(f(\mathbf{x}) = h^{(L)}(\mathbf{x})\).

A key property of neural networks is their Lipschitz constant, which quantifies how much the output of the network can change with respect to small changes in input. Formally, the Lipschitz constant of \( f \) is defined as:

\[
  \Lip(f) = \sup_{\vx \neq \vx^\prime} \frac{\|f(\vx) - f(\vx^\prime)\|}{\|\vx - \vx^\prime\|} \ ,
\]
and provides an upper bound on how much the output can change relative to variations in input. A high Lipschitz constant implies large output variations under small input changes, making networks more susceptible to adversarial attacks~\citep{szegedy2013intriguing, goodfellow2014explaining} and training instabilities.
This property is directly linked to the maximum gradient norm of the network, as for differentiable functions:
\(
\Lip(f) \geq \sup_{\vx} \|\nabla f(\vx)\| \ .
\)
Excessive gradient magnitudes w.r.t parameters $\theta$
can lead to unstable optimization, known as the exploding gradient problem~\citep{pascanu2013difficulty}. Regularizing the Lipschitz constant mitigates such issues by constraining gradient growth, promoting smoother and more stable training.
Beyond stability, Lipschitz continuity plays a crucial role in generalization by limiting function complexity and reducing overfitting~\citep{bartlett2017spectrally}. It also provides theoretical guarantees on model behavior under input perturbations~\citep{scaman2018lipschitz}, with applications in adversarial robustness~\citep{tsuzuku2018lipschitz}, generalization bounds~\citep{bartlett2017spectrally}, and optimization~\citep{fazlyab2019efficient}.

In this chapter, we begin by presenting techniques for estimating the Lipschitz constant of neural networks, analyzing their strengths and limitations. This is followed by methods for enforcing Lipschitz continuity through regularization during training, indirectly constraining model behavior. Additionally, we highlight architectural strategies that ensure Lipschitz continuity by design, removing the need for post-training adjustments.
The convolution with Gaussian kernel also known as Weierstrass transform is also explored as a smoothing mechanism to control the Lipschitz constant of the network. \\
Next, we examine certified robustness, which guarantees model stability under adversarial perturbations. We review how Lipschitz networks achieve deterministic certification by leveraging their bounded Lipschitz constant. In parallel, we discuss randomized smoothing, a probabilistic method for robustness certification, showing how smoothing techniques complement Lipschitz constraints to improve resilience. \\
The final section of this chapter explores regularization techniques for generalization which aim to control model complexity.
Lipschitz regularization constrains sensitivity to input perturbations through spectral norm penalties, projection methods, and architectural constraints.
Flat minima are also closely linked to better generalization, with techniques like sharpness-aware minimization promoting smoother loss landscapes by reducing sharpness.
Finally, loss smoothing and noise-based regularization introduce stochasticity, leveraging techniques such as dropout and perturbed gradient descent.

\section{Lipschitz constraint}

Lipschitz constraints can be expressed under various norms, each inducing different geometric and robustness properties. In this work, we adopt the conventional but ultimately arbitrary choice of the $\ell_2$-norm. This selection is primarily motivated by its correspondence with the spectral norm for linear transformations, which provides a tractable and principled proxy for controlling layerwise sensitivity. Moreover, Gaussian noise—ubiquitous in both training procedures and certified defenses—spreads isotropically, naturally aligning with the $\ell_2$ metric. These factors make the $\ell_2$-norm a practical and theoretically grounded foundation for studying Lipschitz regularization.
However, this choice is not universal. In adversarial robustness, the $\ell_\infty$-norm captures worst-case, high-frequency perturbations~\citep{wong2018scaling}, and several works have developed Lipschitz architectures explicitly tailored to this norm to obtain stronger certified guarantees~\citep{zhang2022rethinking, zhang2022boosting, zhang2021towards}. More broadly, \citet{gouk2021regularisation} show that alternative norms such as $\ell_1$ and $\ell_\infty$ can be effective for promoting robustness and stability, depending on the inductive bias or application domain. The choice of norm should thus reflect the structure of the task and the form of perturbations one aims to guard against.

Formally, the Lipschitz constant of a function \( f \) with respect to the \( \ell_2 \)-norm is denoted as
\[
  \Lip(f) = \sup_{\vx \neq \vy} \frac{\|f(\vx) - f(\vy)\|_2}{\|\vx - \vy\|_2} \ .
\]
For a linear transformation with weight matrix \( \mW \), this constant coincides with the \emph{spectral norm} of \( \mW \), defined as
\[
  \|\mW\|_2 = \max_{\|\vx\|_2 = 1} \|\mW \vx\|_2 \ ,
\]
which is equal to the largest singular value of \( \mW \).

The spectral norm is closely related to the \emph{spectral radius} of a matrix, defined as
\[
  \rho(\mW) = \max \{ |\lambda| : \lambda \ \text{is an eigenvalue of } \mW \} \ .
\]
In general, we have \( \rho(\mW) \le \|\mW\|_2 \), with equality when \( \mW \) is normal (i.e., \( \mW^\top \mW = \mW \mW^\top \)). While the spectral norm governs the worst-case amplification of vector norms, the spectral radius measures the asymptotic growth rate of powers of \( \mW \), a property relevant for stability analysis in iterative processes.

In Chapter~\ref{chapter:spectral_norm_estimation}, we present existing methods and introduce \emph{Gram iteration}, a novel method for estimating spectral norms that improves upon existing techniques. Most non-linear activation functions, such as \(\mathrm{ReLU}\), sigmoid, or \(\mathrm{GELU}\), are 1-Lipschitz~\citep{CombettesPesquet2020_VI, CombettesPesquet2020_LipCert}.

\paragraph{Lipschitz networks require increased capacity:}

Scaling the number of parameters is essential for Lipschitz networks to balance smoothness constraints with expressivity. By increasing model width or depth, one can compensate for the representational limitations introduced by enforcing strict Lipschitz bounds, thus preserving the network's ability to fit complex functions while maintaining robustness and stability guarantees.
While Lipschitz continuity provides strong theoretical guarantees for robustness and generalization, it limits the function space, requiring significantly larger models for competitive performance.
\citet{bubeck2021law, bubeck2021universal} showed that smooth decision boundaries under Lipschitz constraints require \(d\)-times more parameters than non-smooth ones, where \(d\) is the data dimension. Thus, achieving robustness and certification necessitates scaling model capacity to maintain expressivity.
Current Lipschitz-based certification methods often fail to certify points that are in fact robust—i.e., inputs for which the model’s prediction remains constant under all admissible perturbations within a norm ball~\citep{yang2020closer}.
Addressing these challenges requires scaling Lipschitz networks to fully leverage their theoretical advantages.
\citet{bethune2024lipschitz} also explore the expressivity of Lipschitz networks, focusing on orthogonal architectures. They highlight that these architectures require more parameters than standard ones to achieve equivalent expressivity.

\paragraph{Lipschitz beyond adversarial robustness and generalization}
While its primary utility lies in generalization and adversarial robustness, robustness to perturbations is also of great use in several applications such as reinforcement learning~\citep{brunke2022safe}, control algorithms~\citep{shi2018neural}, or other applications that leverage Lipschitz continuity, such as one-class classification with signed distance functions~\citep{bethune2023robust}. For instance, Wasserstein Generative Adversarial Networks (WGANs)~\citep{arjovsky2017wasserstein} leverage the Wasserstein-1 distance to improve GAN training stability, which requires the discriminator to be 1-Lipschitz.
Enforcing this constraint, critical for valid distance computation, is typically achieved through techniques such as weight clipping~\citep{arjovsky2017wasserstein} or gradient penalty~\citep{gulrajani2017improved}.
Beyond GANs, Lipschitz constraints have been used in tasks requiring differential privacy.
For example, \citep{bethune2024dpsgd} propose Lipschitz-SGD to enforce differential privacy by bounding gradient sensitivity, while \citet{ghazanfari2024lipsim} introduces a provably Lipschitz continuous image similarity measure to enhance the robustness of DreamSim~\citep{fu2023dreamsim} under adversarial attacks.
Also, Lipschitz networks are used as a backbone architecture for normalizing flow models~\citep{verine2023expressivity}, where the Lipschitz constraint ensures the invertibility and stability of the flow.

\subsection{Estimating the Lipschitz constant}
Computing the exact Lipschitz constant of a neural network is NP-hard~\citep{scaman2018lipschitz} (even for a 2-layer multi-layer perceptron), primarily due to the exponential proliferation of piecewise linear regions created by activations like ReLU. Each ReLU introduces new subregions in the input space, and finding the global maximum over all such regions quickly becomes intractable as networks grow in size.
While some methods exist to estimate or to bound it~\citep{fazlyab2019efficient,latorre2020lipschitz}, their computational cost makes them impractical for deep architectures. The complexity of modern networks, with their exponentially growing number of parameters and layers, further exacerbates this challenge.
Approximation techniques, such as bound propagation~\citep{zhang2019recurjac,jordan2020exactly,shi2022efficiently}, have been proposed to address scalability. However, these methods often provide loose or overly conservative bounds, limiting their utility during training. Thus, efficiently estimating and controlling Lipschitz constants in deep learning remains an open and active area of research.

\paragraph{Measuring Local Lipschitzness}
\label{sec:local_lipschitzness}
Local Lipschitzness quantifies how much a function's output can change within a small neighborhood of the input. In contrast to the global Lipschitz constant—which provides a uniform upper bound over the entire domain and can be overly conservative—local Lipschitz constants offer a more refined measure of sensitivity in restricted regions.

\begin{definition}[Local Lipschitz constant]
  Let \( f : \R^d \to \R^m \) be a function and let \( \| \cdot \| \) be a norm on \( \R^d \). Given an open set \( \mathcal{B} \subset \R^d \), the local Lipschitz constant of \( f \) over \( \mathcal{B} \) is defined as
  \begin{equation}\label{eq:def_lip_local}
    \Lip(f, \mathcal{B}) = \sup_{\substack{\vx, \vx' \in \mathcal{B} \\ \vx \neq \vx'}} \frac{\lVert f(\vx) - f(\vx') \rVert}{\lVert \vx - \vx^\prime \rVert} \enspace .
  \end{equation}
  We say that \( f \) is locally Lipschitz over \( \mathcal{B} \) if \( \Lip(f, \mathcal{B}) < \infty \). When \( \mathcal{B} = \R^d \), we write \( \Lip(f) \coloneqq \Lip(f, \R^d) \) and refer to it as the global Lipschitz constant.
\end{definition}
\textbf{Remark}
The restriction \( \vx \neq \vx' \) ensures that the denominator is non-zero. In the limit \( \vx' \to \vx \), the Lipschitz ratio approaches the norm of the Jacobian when \( f \) is differentiable. In particular, one has
\[
  \Lip(f, \mathcal{B}) \geq \sup_{\vx \in \mathcal{B}} \|J_f(\vx)\| \, ,
\]
where \( \|J_f(\vx)\| \) denotes the operator norm of the Jacobian at \( \vx \).

To evaluate the local sensitivity of a neural network \( f \), one can estimate its local Lipschitz constant over a dataset. Let \( \mathcal{D} \) be the data distribution, and let \( \mathcal{S} = \{ \vx_i \}_{i=1}^n \subset \mathbb{R}^d \) denote an empirical dataset drawn i.i.d. from \( \mathcal{D} \).
Following~\citet{yang2020closer}, the local Lipschitz constant at a point \( \vx \sim \mathcal{D} \), relative to a perturbation set \( \mathcal{B}(\vx) \subset \mathbb{R}^d \), is defined in expectation as:
\[
  \Lip(f, \mathcal{B}) = \mathbb{E}_{\vx \sim \mathcal{D}} \left[ \sup_{\vx' \in \mathcal{B}(\vx)} \frac{\|f(\vx) - f(\vx')\|}{\|\vx - \vx'\|} \right],
\]
where \( \mathcal{B}(\vx) \) typically denotes a norm-bounded neighborhood around \( \vx \), such as \( \mathcal{B}(\vx) = \{ \vx + \perturb \in \mathbb{R}^d ~|~ \|\perturb\|_2 \leq \epsilon \} \).
In practice, this expectation is approximated by an empirical mean over the dataset \( \mathcal{S} \), yielding the empirical Lipschitz estimate:
\begin{equation}
  \label{eq:empirical_lipschitz}
  \Lip_{\text{emp}}(f, \mathcal{B}) = \frac{1}{n} \sum_{i=1}^{n} \max_{\vx'_i \in \mathcal{B}(\vx_i)} \frac{\|f(\vx_i) - f(\vx'_i)\|}{\|\vx_i - \vx'_i\|}.
\end{equation}
This quantity serves as a tractable surrogate for measuring the local smoothness of \( f \) around points in the dataset.
To approximate the inner maximization in~\eqref{eq:empirical_lipschitz}, \citet{yang2020closer} employ a projected gradient ascent procedure similar to PGD attacks (see Algorithm~\ref{algo:pgd_l2}). While effective in practice, this approach does not yield a guaranteed upper bound on the local Lipschitz constant, but only an empirical lower bound based on approximate maximizers.



\paragraph{A product upper bound for Lipschitz constant}
The Lipschitz constant of a neural network arises from the composition of its layers (Equation~\ref{eq:neural_network_composition}), the composition property of Lipschitz functions offers a starting point: the product of individual layer Lipschitz constants provides an upper bound for the global Lipschitz constant.
\begin{definition}[Product Upper Bound ($\PUB$) for Lipschitz constant]
  A simple upper bound for the Lipschitz constant of a neural network is given by the product of individual layer Lipschitz constants:
  \begin{align}
    \label{eq:prod_upper_bound}
    \Lip(f) & \leq \PUB(f) = \prod_{l=1}^\nblayers \Lip(f^{(l)}) \ .
  \end{align}
  Here, $\Lip(f)$ is the Lipschitz constant of the entire network, and $\Lip(f^{(l)})$ is the Lipschitz constant of the $l$-th layer.
\end{definition}
For ReLU networks, the product of layer-wise Lipschitz constants generally overestimates the true Lipschitz constant. While linear layers can in principle align to amplify the same direction, ReLU layers often zero out parts of the input, disrupting this alignment and preventing each layer’s maximal scaling factor from fully accumulating.
Even for linear networks, the product bound holds strictly only under the perfect alignment of transformations—for instance, in homothetic (scaling) layers or orthogonal transformations which is rarely the case with standard networks. \\
The product upper bound was first used in~\citep{szegedy2013intriguing} and has since been applied in generalization bounds~\citep{bartlett2017spectrally} and the design of Lipschitz networks~\citep{tsuzuku2018lipschitz, trockman2021orthogonalizing, meunier2022dynamical, araujo2023a}.
While this bound is often loose, it serves as the foundation for designing scalable techniques to estimate and regularize Lipschitz constants, with applications in stability, robustness, and generalization. Using this bound, one can leverage the spectral norm of each layer to control the network's Lipschitz constant or design Lipschitz layers directly.
The $\PUB$ serves as a starting point for the design and regularization of Lipschitz layers.

\paragraph{Lipschitz constant of attention}
The $\PUB$ can be used when the Lipschitz constant of each layer is tractable to compute, such as for linear layers or layers with known Lipschitz constants.
However, for self-attention blocks, which form the core of transformer architectures~\citep{vaswani2017attention}, deriving Lipschitz properties is less straightforward.
The self-attention score is given by
\[
  \mX \mapsto = \mathrm{softmax} \left( \frac{\mW_K \mX (\mW_Q \mX)^\top}{\sqrt{d}} \right) \ ,
\]
where the  input vector $\vx$ is reshaped into a matrix $\mX$, and $\mW_K, \mW_Q$ are respectively the key and query weight matrices.
While the $\mathrm{softmax}$ function itself is 1-Lipschitz~\citep{gao2018propertiessoftmaxfunctionapplication}, the quadratic form $\mW_K \mX (\mW_Q \mX)^\top$ introduces a bilinear dependency on \(\mX\), leading to potential unbounded growth in the operator norm.
This makes it difficult to control the Lipschitz constant of the self-attention layer, as the spectral norm of \(\mX \mapsto \mW_K \mX (\mW_Q \mX)^\top\) can scale with \(\|\mX\|^2\).
Unlike linear layers, where the Lipschitz bound is determined by the spectral norm of a single weight matrix, the interaction between \(\mW_K, \mW_Q\) and the input \(\mX\) complicates the analysis.
The work of~\citet{kim2021the} explores alternative Lipschitz-constrained formulations of the attention block, while several studies investigate the Lipschitz properties of self-attention for bounded input domains~\citep{havens2024fine} or with respect to Wasserstein distance~\citep{vuckovic2021regularityattention, castin2024how}.
However, deriving Lipschitz bounds for attention mechanisms remains an open challenge, with implications for the robustness and stability of transformer models.

\paragraph{Lipschitz Constant  of Normalization Layers.}
Normalization layers such as BatchNorm~\citep{ioffe2015batch}, LayerNorm~\citep{ba2016layer}, and GroupNorm~\citep{wu2018group} are widely used in deep learning to stabilize training and improve convergence. For BatchNorm, each activation is normalized as
\[
  \operatorname{BN}(\vx) = \gamma \frac{\vx - \mu_j}{\sigma_j} + \beta,
\]
where \(\mu_j\) and \(\sigma_j\) are the empirical mean and standard deviation across the batch for feature dimension \(j\), and \(\gamma, \beta\) are learnable affine parameters.
\citet{santurkar2018does} showed that BatchNorm alters the geometry of the loss by modifying gradient norms. For a loss \(\mathcal{L}\) and pre-activation \(\vy_i\) of the \(i\)-th input in a mini-batch of size \(m\), the gradient norm of the transformed loss \(\hat{\mathcal{L}}\) satisfies
\[
  \left\| \nabla_{\vy_i} \hat{\mathcal{L}} \right\|^2 \le \frac{\gamma^2}{\sigma_j^2} \left( \left\| \nabla_{\vy_i} \mathcal{L} \right\|^2 - \frac{1}{m} \langle \mathbf{1}, \nabla_{\vy_i} \mathcal{L} \rangle^2 - \frac{1}{m} \langle \nabla_{\vy_i} \mathcal{L}, \hat{\vy}_j \rangle^2 \right),
\]
where \(\hat{\vy}_j\) denotes the normalized activation direction. This reveals that BatchNorm rescales gradients by \(\gamma / \sigma_j\) while suppressing components aligned with the constant vector and the activation axis, effectively reducing the local Lipschitz constant and stabilizing training. Second-order analysis confirms that the Hessian norm is similarly attenuated, bounded by \(\gamma^2 / \sigma_j^2\).

However, despite these benefits, BatchNorm does not ensure Lipschitz continuity during training. Because \(\mu_j\) and \(\sigma_j\) are computed from the current mini-batch, the transformation becomes a stochastic function that depends on the entire batch. If the variance \(\sigma_j^2\) is small, the scaling factor \(\gamma / \sigma_j\) may become arbitrarily large, leading to unbounded local Lipschitz constants. Moreover, the interaction between examples introduces coupling effects that complicate theoretical analysis. The discrepancy between training (with batch statistics) and inference (with frozen statistics) further breaks Lipschitz consistency across phases. Similar caveats apply to other normalization schemes like LayerNorm or GroupNorm, although they normalize across different axes. In summary, while normalization layers improve gradient flow and training stability, they complicate explicit Lipschitz control and may be incompatible with Lipschitz-regularized objectives.

In the following sections, we present regularization methods for enforcing Lipschitz constraints on neural networks.

\subsection{Lipschitz through regularization}

\paragraph{Gradient Norm Penalty}
Gradient norm penalties~\citep{drucker1992double, ross2018improving, gulrajani2017improved} offer an indirect way to control the Lipschitz constant by regularizing the norm of the input gradients during training. This approach has been successfully applied in adversarial robustness and Wasserstein GANs. A typical regularization term, introduced by~\citet{gulrajani2017improved}, penalizes deviations of the gradient norm from unity:
\begin{equation*}
  \label{eq:gradient_penalty}
  \mathcal{L}_{\text{GP}} = \frac{1}{m} \sum_{i=1}^m \left( \left\| \nabla_{\vx} f(\vx_i) \right\|_2 - 1 \right)^2 \,,
\end{equation*}
where the expectation is approximated by the empirical average over a batch of \(m\) input points. This regularizer encourages the function to be approximately 1-Lipschitz in regions traversed during training, without explicitly constraining the spectral norm of the weights.
However it is computationally expensive because it requires a double backpropagation which involves a large computational graph. \\
To circumvent this cost regularization on individual layers has been proposed by~\cite{yoshida2017spectral} by proposing spectral normalization which directly constrains the Lipschitz constant of the layer by bounding the spectral norm of the weight matrix.
$$ \mathcal{L_{\text{SN}}} = \sum_{i=1}^\nblayers \sigma_1(\mW^{(l)}) $$
Other techniques, such as soft regularization  promote orthogonality~\citep{wang2020orthogonal, huang2020controllable}, with regularization loss like:
\begin{equation*}
  \mathcal{L}_{\text{ortho}} = \sum_{l=1}^\nblayers \norm{\mW^{(l)} \mW^{(l)\top} - \mI}_F^2 \ .
\end{equation*}

The work of \citet{arjovsky2017wasserstein} uses weight clipping to ensure the Lipschitz constraint of the discriminator in Wasserstein GANs. This method is simple and effective, but it can lead to optimization issues and model instability.
Other approaches rely on projection $\proj$ after weights updates but do not alter gradient computation, for instance,
\citet{salimans2016weight} used weight normalization to constrain the Lipschitz constant of the network,
$$
  \proj(\mW^{(l)}) = \frac{\mW^{(l)}}{\norm{\mW^{(l)}}_\frob} \ .
$$
But Frobenius norm is a loose bound on the spectral norm which is the Lipschitz constant of the layer.
\citet{miyato2018spectral} refines this method by using spectral normalization to enforce Lipschitz constraints on the discriminator of GANs,
$$
  \proj(\mW^{(l)}) = \frac{\mW^{(l)}}{\sigma_1(\mW^{(l)})} \ .
$$
In the same flavor \citet{gouk2021regularisation} proposed a projection-based method to enforce Lipschitz constraints on individual layers.
enforce the Lipschitz constraint on a layer, \citet{gouk2021regularisation} applies a projection $\proj$ onto the set of matrices with a Lipschitz constant less than or equal to \(\lambda\).
This projection is applied to the weight matrices \(\mW^{(l)}\) after each training update:
\[
  \proj(\mW^{(l)}) = \frac{\mW^{(l)}}
  {\max\left(1, \frac{\|\mW^{(l)}\|_p}{\lambda} \right)}
\]
With $\lambda$ a hyperparameter controlling the layer Lipschitz constant, and for $p=1, 2, \infty$.

Lipschitz constant approximation methods~\citep{fazlyab2019efficient} provide tighter bounds by incorporating upper bounds of the constants into the loss function using semidefinite programming, though their high computational cost makes them less practical for large-scale models.
Those normalization methods are efficient from a computational standpoint, but they can be too restrictive, and lead to vanishing gradients, the maximum direction is controlled but the average direction can collapse to zero.

Margin-based regularization~\citep{tsuzuku2018lipschitz} has been explored to improve robustness by maximizing the margin between decision boundaries. The work of \citet{leino21gloro, leino2021relaxing} extends this paradigm by introducing Lipschitz bound ($\PUB$) regularization directly integrated into the loss function.
Those approaches conjugate regularization training with by design Lipschitz layers to ensure robustness and training stability.

\subsection{Lipschitz by design}
\label{section:related_works:by_design_lipschitz_layers}
By design Lipschitz layers allow for certified robustness and stability without the need for post-hoc adjustments or regularization during training. These layers are designed to be 1-Lipschitz, ensuring that the network as a whole is also 1-Lipschitz leveraging the $\PUB$:
$$
  \Lip(f) \leq \prod_{i=1}^\nblayers \Lip(f^{(l)})  =1 \ .
$$
The main difficulty with Lipschitz layers lies in deriving such layers for complex architectures like convolutional layers, residual layers, and attention mechanisms.
Dense layers are more straightforward to derive and analyze in terms of their Lipschitz constants as we can study directly the matrix representing the transformation.
\begin{figure}[h]
  \centering
  \includegraphics[width=0.7\textwidth]{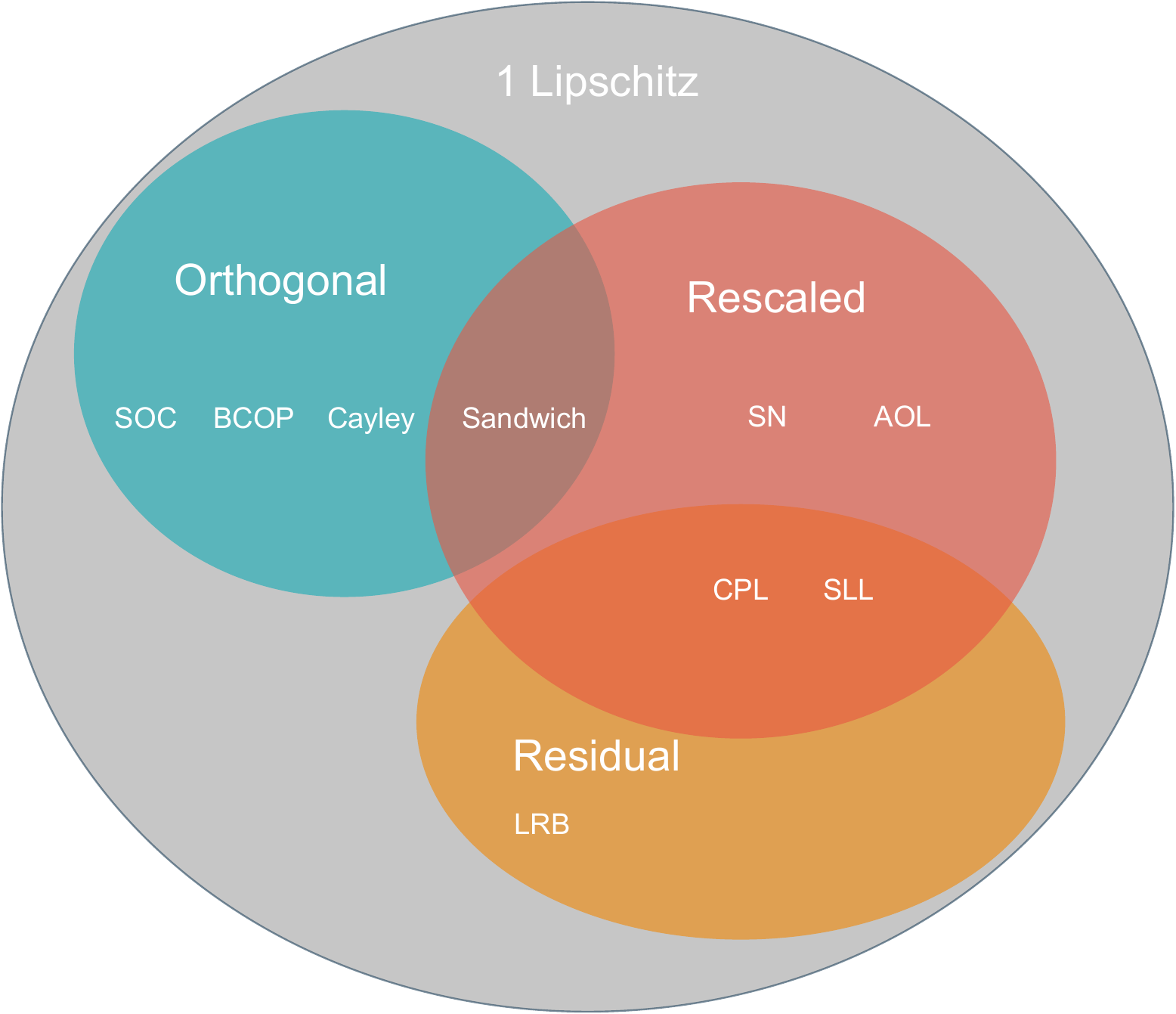}
  \caption{Venn diagram illustrating the categorization of Lipschitz layers, including scaled layers, orthogonal layers, and residual layers.}
  \label{fig:lipschitz_layers_venn}
\end{figure}
First, we consider a linear layer,
$$\vx\mapsto \mW \vx$$
where $\mW$ is the weight matrix and input $\vx \in \R^d$.
Here $\mW$ can be a dense matrix or a sparse matrix such as for the convolutional layer.
Here the bias is implicitly included.

\paragraph{Orthogonal layers}
One convenient way to ensure the Lipschitz constant is to enforce orthogonality on the weight matrix, it is a good property for training properties as the layer is norm preserving.
This can be done by optimizing the weight matrix onto the orthogonal manifold~\citep{anil2019sorting, ablin2022fast}.
Another method is to rely on orthogonal layers by design, these methods aim to project weight matrices onto an orthogonal space.
A key challenge arises with convolutional layers, where the operation cannot be directly expressed as matrix multiplication, because the matrix representing the convolutional operation is usually too large to be stored in memory: the size of the matrix grows with $O(n^4)$ where $n$ is the input dimension ($\vx \in \mathbb{R}$).
For images from CIFAR-10 or ImageNet $n=3\times 32 \times 32$ and respectively $3\times 224 \times 224$, which is infeasible to store in memory.

To address this, Reshaped Kernel Orthogonalization (RKO) methods~\citep{li2019preventing} enforce orthogonality on the reshaped convolutional filter \(\tK\). However, making \(\tK\) orthogonal does not guarantee that the convolutional layer itself is orthogonal, as the convolution operation introduces additional dependencies.
Instead, the Lipschitz constant of the convolutional layer is upper bounded by the Lipschitz constant of the reshaped kernel, up to a scaling factor.
However, it empirically gives singular values closer to 1 than spectral normalization.

Orthogonalization techniques specifically designed for convolutional layers with circular padding have been formally derived. The most notable methods for enforcing orthogonality in circular convolutional layers include:
\begin{itemize}
  \item Block Convolution Orthogonal Parameterization (BCOP)~\citep{li2019preventing} utilizes the
        iterative algorithm of \citet{bjorck1971iterative} to orthogonalize the linear transformation in convolutional layers.
        This transformation is applied directly to the convolutional filter \(\tK\). The base implementation of BCOP
        is computationally expensive due to the use of block convolution.
  \item
        Skew Orthogonal Convolutions (SOC)~\citep{singla2021skew} leverage the key property that the
        matrix exponential of a skew-symmetric matrix is always orthogonal, i.e.,
        \(
        \exp(\mA) = \mW, \quad \text{where} \quad \mA = -\mA^\top.
        \)
        To enforce this structure in convolutional layers, SOC constructs the skew-symmetric matrix as
        \(
        \mA = \mQ^\top - \mQ,
        \)
        where \(\mQ\) is a learnable parameter. The orthogonal transformation \(\mW\) is then obtained via a
        finite Taylor series approximation of the matrix exponential : it requires multiple
        propagations through the base convolutional layer making it computationally expensive.
  \item Cayley Transform~\citep{trockman2021orthogonalizing}: Orthogonalizes weight matrices via \( \mW  = (\mI - \mA)^{-1} (\mI + \mA) \), where \(\mA\) is skew-symmetric. For convolutional layers, this transform is performed in the Fourier domain, but it requires matrix inversion, which scales with the input size.
\end{itemize}

A review of orthogonal convolutional layers is presented in~\citep{boissin2025adaptive}, highlighting both their strengths and limitations. While these approaches have demonstrated strong performance on datasets such as CIFAR-10, their scalability remains restricted to such datasets.
Also, the work by~\citet{boissin2025adaptive} has significantly improved the efficiency of BCOP and SOC methods.
However, enforcing equal singular values can be overly restrictive in certain cases, often requiring significantly more parameters to maintain the same level of expressivity~\citep{bethune2024lipschitz}.
Several works based on rescaling layers have been proposed in parallel and mitigate those issues.

\paragraph{Rescaled layers}
Spectral normalization (SN) ~\citep{yoshida2017spectral, miyato2018spectral, farnia2019generalizable} is a widely used technique to enforce control over the Lipschitz constant of linear layers. Unlike earlier methods, it directly leverages the Lipschitz property, offering deterministic guarantees. Specifically, weight matrices can be normalized by their largest singular values to ensure \(1\)-Lipschitz layers~\citep{miyato2018spectral, farnia2018generalizable, anil2019sorting}.
Formally, a spectral-normalized layer is defined as:
\[
  \vx \mapsto \mW \mR \vx,
\]
where \(\mR = \frac{1}{\norm{\mW}_2}\) acts as a rescaling factor based on the spectral norm \(\norm{\mW}_2\). The spectral norm is typically computed using power iteration. Throughout this work, we assume that \(\norm{\mW}_2 \neq 0\) to ensure numerical stability. \\
During training, one iteration of Power Iteration (PI) is performed per step, with the estimated singular vector being stored in a buffer for each weight matrix. This allows for an efficient spectral norm estimation, as the weight matrices change gradually during training, making a single iteration sufficient to track the dominant singular value. This approach, introduced in \cite{miyato2018spectral}, keeps the computational cost low while maintaining stability. \\
However $\PI$ can be slow to converge to the spectral norm, is non-deterministic, and does not provide a strict upper bound on the spectral norm, which can be critical in certified robustness application or normalizing flow where inversion requires a precise upper bound on the spectral norm of layers.
In Chapter~\ref{chapter:spectral_norm_estimation} we present a novel method for estimating spectral norms that improves upon existing techniques: faster convergence and upper bound on the spectral norm. \\
Another limitation of SN is stacking multiple rescaled layers with spectral normalization can lead to vanishing gradients, as all singular values except the dominant one may collapse to zero, effectively reducing the rank of the transformation and impairing gradient flow.

To address those issues, the Almost-Orthogonal Layer (AOL)~\citep{prach2022almost} provides a balanced approach by normalizing layers to approximate 1-Lipschitz behavior while promoting soft orthogonality.
This design helps maintain stable gradient flow during training, mitigating the vanishing gradient problem observed in stacked spectral-normalized layers.
The fully connected AOL layer is defined using a diagonal rescaling matrix \(\mR\), given by \footnote{This assumes that each column of \(\mW\) has at least one non-zero entry, ensuring that \eqref{eq:DAOL} is well-defined.}
\begin{equation}\label{eq:DAOL}
  \mR = \textstyle \diag \left( \sum_j | \mW^\top \mW |_{ij} \right)^{-\frac{1}{2}}.
\end{equation}
This formulation guarantees a strictly 1-Lipschitz layer, unlike spectral normalization, which only provides an approximation of the Lipschitz constant.
The rescaling matrix can also be computed efficiently for convolutional layers.
Moreover, empirical results indicate that after training, the network's Jacobian (with respect to \(\vx\)) remains nearly orthogonal,  deeper architectures without excessive gradient attenuation—at least to a greater extent than spectral normalization.

Several works~\citep{anil2019sorting, singla2022improved, huang2021local} have focused on constraining Lipschitz constants through activation functions, particularly in the setting of linear orthogonal layers. These approaches often enhance performance by maintaining a balance between orthogonality and non-linear activation behavior.
More generally, designing Lipschitz layers beyond linear layers is a challenging task, as the Lipschitz constant is not directly tied to the spectral norm of the weight matrix. \citet{wang2023direct} propose layer combinations that are inherently Lipschitz, drawing inspiration from the SDP condition of~\citet{fazlyab2019efficient} and related works~\citep{meunier2022dynamical, araujo2023a}.
They also rely on the work of~\citet{trockman2021orthogonalizing} using Cayley transform to design convolutional layers.
They introduce the Sandwich Layer, the transformation applied within the layer is given by
\[
  \vx \mapsto \sqrt{2} \mA^\top \mR^{-1} \rho \left(\sqrt{2} \mR \mB \vx\right),
\]
where the matrices \(\mA\), \(\mR\), and \(\mB\) are constructed as
\[
  \mR = \mathrm{diag}(\exp(\vq))^{-1}, \quad
  \begin{bmatrix} \mA^\top \\ \mB^\top \end{bmatrix}
  = \mathrm{Cayley} \left( \begin{bmatrix} \mW \\ \mV \end{bmatrix} \right),
\]
with
\[
  \mathrm{Cayley}
  \left(
  \begin{bmatrix} \mW \\ \mV \end{bmatrix}
  \right)
  =
  \begin{bmatrix} (I + \mZ)^{-1}(I - \mZ) \\ -2\mV (I + \mZ)^{-1} \end{bmatrix},
\]
where \(
\mZ = \mW - \mW^\top + \mV^\top \mV.
\) Here, \(\mW\) and \(\mV\) represent the weight matrices of the layer, while \(\vq\) is a vector that parameterizes the transformation.

However, in practice, scaling very deep Lipschitz architectures remains challenging, even when stacking layers: performance is not always improved by adding more layers and the training issue can be exacerbated when the network is too deep.
To mitigate this issue, residual connections can be incorporated.

\paragraph{Residual layers}
Residual connections are fundamental to modern neural network architectures, as highlighted in the seminal work of \citet{he2016deep}. From a training dynamics perspective, residual networks are easier to optimize than plain networks, as learning residual mappings with an identity initialization is more efficient than directly learning the identity function. These connections enhance the signal-to-noise ratio and simplify the training of very deep networks by allowing information to flow directly across layers~\citep{balduzzi2017shattered}.

The work of \citet{meunier2022dynamical} interprets residual blocks through the lens of dynamical systems, framing residual networks as discretized continuous flows. They show that parameterizations based on the Cayley transform~\citep{trockman2021orthogonalizing} and SOC~\citep{singla2021skew} correspond to two specific discretization schemes for the skew-symmetric component of these flows. Additionally, they introduce a novel discretization approach for the symmetric component, derived from a convex potential, ensuring a 1-Lipschitz property. By discretizing this flow, they construct a new layer, the \textit{Convex Potential Layer} (\(\CPL\)), which is residual, nonlinear, and inherently 1-Lipschitz:
\begin{equation}
  \label{eq:cpl_layer}
  \vx\mapsto \mathbf{x} - 2  \mathbf{W} \mR^2 \nonlin(\mathbf{W}^\top\mathbf{x} + \mathbf{b}),
\end{equation}
where $\mR = \frac{1}{\|\mathbf{W}\|_2}$,  \(\|\mathbf{W}\|_2\) is the spectral norm of \(\mathbf{W}\), and \(\nonlin\) is the \(\relu\) activation function.
This layer can also be interpreted through the lens of Maximal Monotone Operator (MMO) theory~\citep{bauschke2017convex}, where the update corresponds to a discretized reflected resolvent step of a monotone operator.

The \(\CPL\) layer is 1-Lipschitz and can be stacked to form deep networks, providing a scalable and robust alternative to standard architectures.
Its bias towards the identity makes it easy to train without any additional regularization(batch norm, gradient clipping). They demonstrate successful training with up to $1000$ layers on CIFAR-10 and CIFAR-100 outperforming standard architectures in terms of certified robustness. The recent survey of~\citet{prach2024lipschitz} places as a good tradeoff the \(\CPL\) layer in the broader context of Lipschitz networks, highlighting its potential for scalable and robust training.

\citet{araujo2023a} build upon the work of \citet{meunier2022dynamical}, their work extends this approach by developing a general framework that unifies and generalizes multiple existing Lipschitz layers, including SN, $\CPL$, $\AOL$, and Sandwich Layers.
Specifically, they propose the SDP Lipschitz Layer (\(\SLL\)), leveraging semidefinite programming (SDP) theory to ensure the 1-Lipschitz constraint while maintaining expressivity. This layer provides a structured way to derive and modify Lipschitz layers, including spectral normalization layers, $\AOL$ layers, and $\CPL$ layers.
A key contribution of \citet{araujo2023a} is the introduction of a new residual layer, which retains the $\CPL$ structure but incorporates $\AOL$-based rescaling for improved stability and flexibility. The \(\SLL\) block serves as a generalized version of both $\CPL$ and $\AOL$, where the transformation matrix \(\mR\) is adapted from $\AOL$’s construction to allow broader parameterization:
\begin{equation}\label{equation:lipschitz_layer}
  \mR = \diag\left(\sum_{j=1}^n | \mW^\top \mW|_{ij} \frac{\vq_j}{\vq_i}  \right)
  ^{-\frac{1}{2}},
\end{equation}
where  $\vq$ represents a vector of positive elements.

The work of~\citet{hu2023scaling} introduced the simpler but effective \emph{Residual Linear Block} (RLB), where the weight matrix is biased toward the identity \(\mW = \mI + \mQ\). They compute the spectral norm of $\mW$ with power iteration, however, this approach does not account for nonlinearities, which are essential for capturing complex patterns in data.

All those residual architectures have converged towards a convolutional stack where the dimension of the input is preserved, and after that, a linear MLP classifier is applied. This architecture is 1-Lipschitz and has been shown to be competitive with standard architectures on various datasets~\citep{meunier2022dynamical, araujo2023a, hu2023scaling, hu2024recipe}.

\begin{figure}[t]
  \centering
  \includegraphics[width=0.5\textwidth]{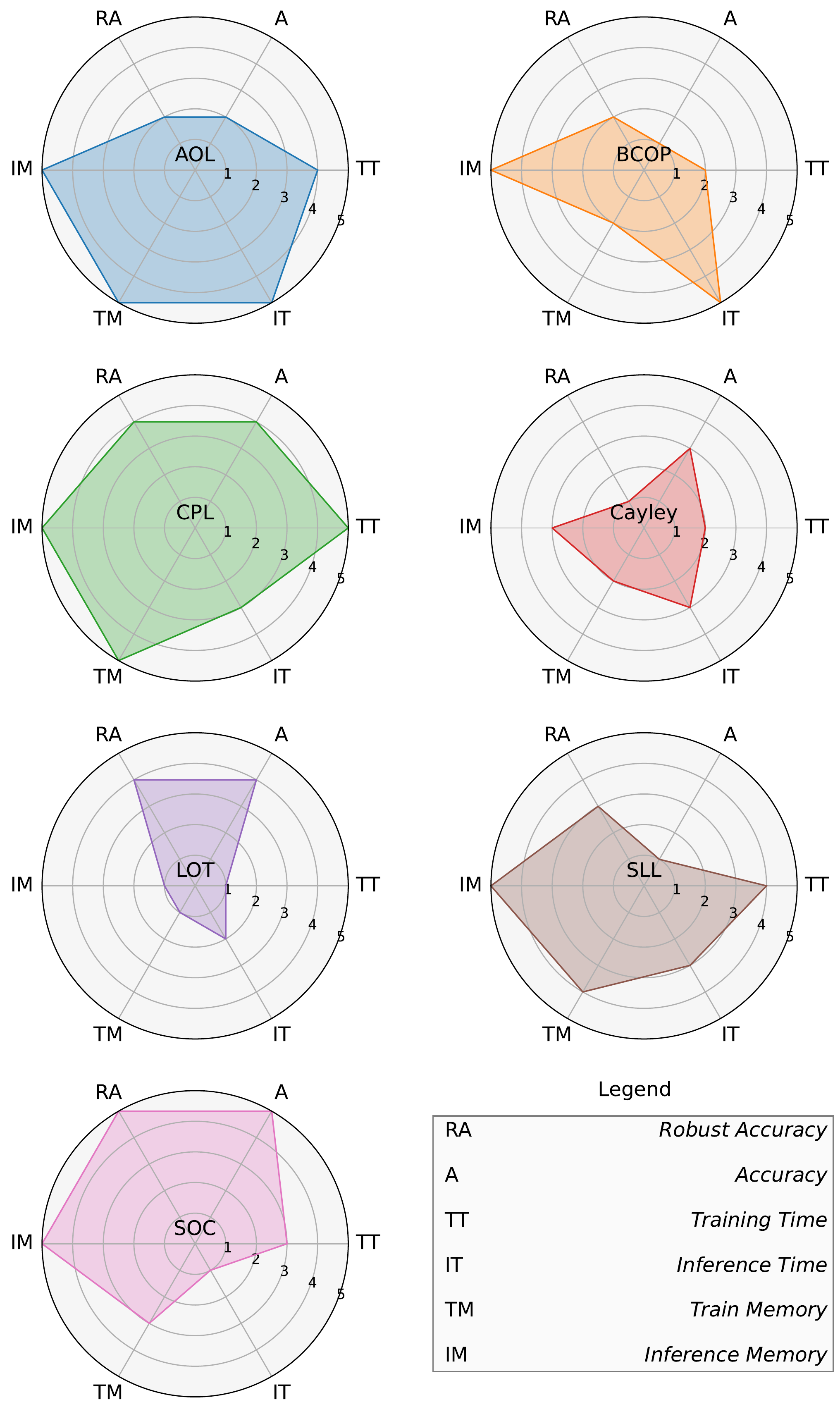}
  \caption{Figure taken from~\citet{prach2024lipschitz} comparing different Lipschitz layers w.r.t to different criteria.
    Scores ranged from $1$ (worst) to 5 (best) for every layers.}
  \label{fig:radar_plot}
\end{figure}
Training orthogonal layers is typically computationally intensive. The Cayley method requires matrix inversion involving the image input size, which is costly, and SOC involves either singular value decomposition (\(\SVD\)) or an iterative Taylor expansion. The review from \citet{prach2024lipschitz} compares the performance of different Lipschitz layers, including BCOP, SOC, \(\CPL\), and AOL, across various criteria, such as robustness, scalability, and computational efficiency.
Considering all evaluated metrics (summarized in Figure~\ref{fig:radar_plot}), \(\CPL\) emerges as the most favorable option due to its superior performance and lower computational cost. When ample computational resources are available and strict timing constraints are not a concern during training and inference, the SOC layer may be a viable choice, given its slightly improved performance. On the other hand, applications where inference time is critical may benefit from \(\AOL\) or BCOP, as they introduce no additional runtime overhead compared to standard convolution. Additionally, for higher-resolution images, \(\CPL\) appears to be the most promising approach.

\paragraph{Limitation of Lipschitz layers}
Independently of the choice of layer design, the \(\PUB\) is used to bound the overall Lipschitz constant of such \(1\)-Lipschitz networks. However, the \(\PUB\) is often loose because it fails to consider the interactions between layers. For deep networks, this leads to overly conservative estimates, as the product of individual Lipschitz constants inflates the bound and does not reflect the true sensitivity of the network. Additionally, decomposing neural networks into individual layers overlooks the potential smoothing effects across layers, where dependencies may naturally reduce sensitivity.
Moreover, the Lipschitz property of certain non-linear layers, such as attention mechanisms, can be challenging to determine~\cite{kim2021the}.
To address those limitations we are going to present the Weierstrass transform in the next section, which is a smoothing technique that can be used to regularize neural networks and improve their Lipschitz properties.

\subsection{Weierstrass transform}
To address limitations of $\PUB$, others techniques exist to improve stability and robustness.
One such method is the \emph{Weierstrass transform}~\citep{zayed1996handbook}, a classical smoothing technique from functional analysis. Also known as Gaussian convolution, it consists in convolving a function with a Gaussian kernel, the Weierstrass transform enhances regularity, making the function continuous and differentiable. This smoothing effect not only mitigates sensitivity to perturbations but also induces global regularity by smoothing decision boundaries. As stated in \citet{salman2019provably}, \emph{randomized smoothing} (RS)~\citep{lecuyer2018certified, cohen2019certified} is an application of the Weierstrass transform in the context of neural networks leveraging its probabilistic interpretation.
Originating from differential privacy, RS derives its certified robustness guarantees from a probabilistic perspective, as introduced by \citet{lecuyer2018certified}. It applies the principle of convolution with a predefined probability distribution to the input, commonly relying on Gaussian perturbations, enabling scalability to large datasets like ImageNet. In practice, it relies commonly on Gaussian perturbations on the input rather than transformations of the entire network.

\begin{definition}[Weierstrass transform~\citep{zayed1996handbook}]
  Let \( f: \mathbb{R}^d \to \mathbb{R} \) be a measurable function integrable with respect to the Gaussian kernel. The Weierstrass transform of \( f \) is defined as:
  \[
    \tilde{f}(\vx) = (f * \phi_\sigma)(\vx) = \int_{\mathbb{R}^d} f(\vx - \delta) \, \phi_\sigma(\delta) \, d\delta,
  \]
  where \( \phi_\sigma(\delta) = \frac{1}{(2\pi \sigma^2)^{d/2}} \exp\left( -\frac{\|\delta\|^2}{2\sigma^2} \right) \) is the isotropic Gaussian kernel. Since \( \phi_\sigma \) is symmetric, this expression is equivalent to
  \[
    \tilde{f}(\vx) = \int_{\mathbb{R}^d} f(\vx + \delta) \, \phi_\sigma(\delta) \, d\delta,
  \]
  which corresponds to the probabilistic interpretation used in randomized smoothing.
\end{definition}
In the following, we always assume that $f$ is measurable and that the integral defining $\tilde{f}$ is well-defined.

Alternatively, the Weierstrass transform can be interpreted probabilistically.
Let \(\mathbb{\delta} \sim \mathcal{N}(\vzero, \sigma^2 \mI)\) be a random vector following a multivariate Gaussian distribution with mean zero and covariance matrix \(\sigma^2 \mI\).
Then, the Weierstrass transform is the expectation of \( f \) over this Gaussian distribution:

\[
  \tilde{f}(\vx) = \mathbb{E}_{\mathbb{\delta} \sim \mathcal{N}(\vzero, \sigma^2 \mI)} \left[ f(\vx + \mathbb{\delta}) \right].
\]

This probabilistic interpretation highlights that the Weierstrass transform smooths \( f \) by averaging its values over Gaussian perturbations of the input \(\vx\). When the integral is not explicitly computable, this allows its estimation through Monte Carlo integration by sampling from the Gaussian distribution as done in RS.
This transform smooths
\( f \) by averaging its values over neighborhoods determined by the parameter \( \sigma \), enhancing the regularity of \( f \) and making \( \tilde{f} \) infinitely differentiable (\( C^{\infty} \)) because the Gaussian kernel allows differentiation under the integral sign~\citep{stein1970singular}.

\begin{lemma}[Stein's Lemma \citep{stein1970singular}]
  \label{lemma:stein}
  Let \( \sigma > 0 \), and let \( f : \mathbb{R}^d \to \mathbb{R} \) be a  function.
  Then \( \tilde{f} \) is differentiable, and its gradient is given by:
  \[
    \nabla \tilde{f}(\vx) = \frac{1}{\sigma^2} \, \E_{\delta \sim \mathcal{N}(\vzero, \sigma^2 \mI)} \left[ \delta  f(\vx + \delta) \right].
  \]
\end{lemma}
Using Stein's Lemma, we can further characterize the regularity properties of \( \tilde{f} \) under an additional boundedness assumption.
\begin{lemma}[Lipschitz continuity of the Weierstrass transform for bounded functions~\citep{salman2019provably}]\label{lemma:lip_weierstrass_transform}
  Let \( \sigma > 0 \), and let \( f : \mathbb{R}^d \to \mathbb{R} \) be a  function satisfying \( |f(\vx)| \leq 1 \) for all \( \vx \in \mathbb{R}^d \).
  Then, \( \tilde{f} \) is Lipschitz continuous with respect to the \( \ell_2 \)-norm, and :
  \[
    \Lip\left(\tilde{f}\right) \leq \sqrt{\frac{2}{\pi \sigma^2}}.
  \]
\end{lemma}
Note that when the function \( f \) is Lipschitz continuous, the Weierstrass transform \( \tilde{f} \) inherits this property, as shown in the following lemma:
\begin{lemma}[Lipschitz continuity of the Weierstrass transform for Lipschitz functions~\citep{nesterov2017random}]
  \label{lemma:lip_weierstrass_transform_lipschitz}
  Let \( f : \mathbb{R}^d \to \mathbb{R} \) be a Lipschitz continuous function with Lipschitz constant \( \Lip(f) \). Then \( \widetilde{f} \) is Lipschitz continuous with respect to the \( \ell_2 \)-norm and its Lipschitz constant is given by:
  \(
  \Lip\left(\widetilde{f}\right) \leq \Lip(f).
  \)
\end{lemma}

Coming from randomized smoothing~\citet{salman2019provably}, a stronger Lipschitz bound can be derived by considering the composition of quantile function with a smoothed classifier. This involves considering the Lipschitz constant of the following function:
\[
  (\quant \circ \tilde{f})(\vx) = \quant \left( \E_{\delta \sim \mathcal{N}(\vzero, \sigma^2 \mI)} \left[ f(\vx + \delta) \right] \right),
\]
where \( \quant \) is the Gaussian quantile, defined as the inverse of the Gaussian cumulative distribution function. The function \( \quant \circ \tilde{f} \) is Lipschitz continuous with respect to the \( \ell_2 \)-norm, with its Lipschitz constant given by:
\begin{lemma}[Lipschitz bound for quantile-composed smoothed classifier~\citep{salman2019provably,cohen2019certified}]
  \label{lemma:lip_quantile_smoothed_classifier}
  \( \quant \circ \tilde{f} \) is Lipschitz continuous with respect to the \( \ell_2 \)-norm, and its Lipschitz constant is given by:
  \[
    \Lip\left(\quant \circ \tilde{f}\right) = \frac{1}{\sigma}.
  \]
\end{lemma}
If we suppose that the function \( f \) is Lipschitz continuous it is not straightforward to derive a bound on the Lipschitz constant of the quantile-composed smoothed classifier which intervenes $\Lip(f)$.
Indeed, the quantile function is not Lipschitz continuous, and the composition of a Lipschitz function with a non-Lipschitz function does not guarantee a Lipschitz function.

In the next section, we are going to see how Lipschitz networks can be used to provide certification in the context of adversarial robustness.

\section{Certified robustness}

Consider a classification problem where the input space is \( \mathcal{X} \subset \mathbb{R}^d \), and the label space is \( \mathcal{Y} = \{1, \dots, c\} \), representing \( c \) distinct classes.
Let \( \mathcal{D} = \{(\mathbf{x}_i, y_i)\}_{i=1}^{n_{\mathrm{train}}} \) denote a dataset of \( n_{\mathrm{train}} \) labeled examples, where each \( \mathbf{x}_i \in \mathcal{X} \) is an input, and \( y_i \in \mathcal{Y} \) is its corresponding label. A function \( f: \mathcal{X} \to \mathbb{R}^c \) maps each input \( \mathbf{x} \in \mathcal{X} \) to a vector of scores \( f(\vx) = [f_1(\mathbf{x}), \dots, f_c(\mathbf{x})]^\top \), where \( f_k(\mathbf{x}) \) represents the predicted confidence for class \( k \). The final classifier is defined as:
\[
  \hat{y} = \argmax_k f_k(\mathbf{x}),
\]
where \( \hat{y} \) is the predicted label.
$f_y(\vx)$ is the score for the true label \( y \) of the input \( \vx \).

\begin{figure}[t]
  \centering
  \includegraphics[width=1.0\textwidth]{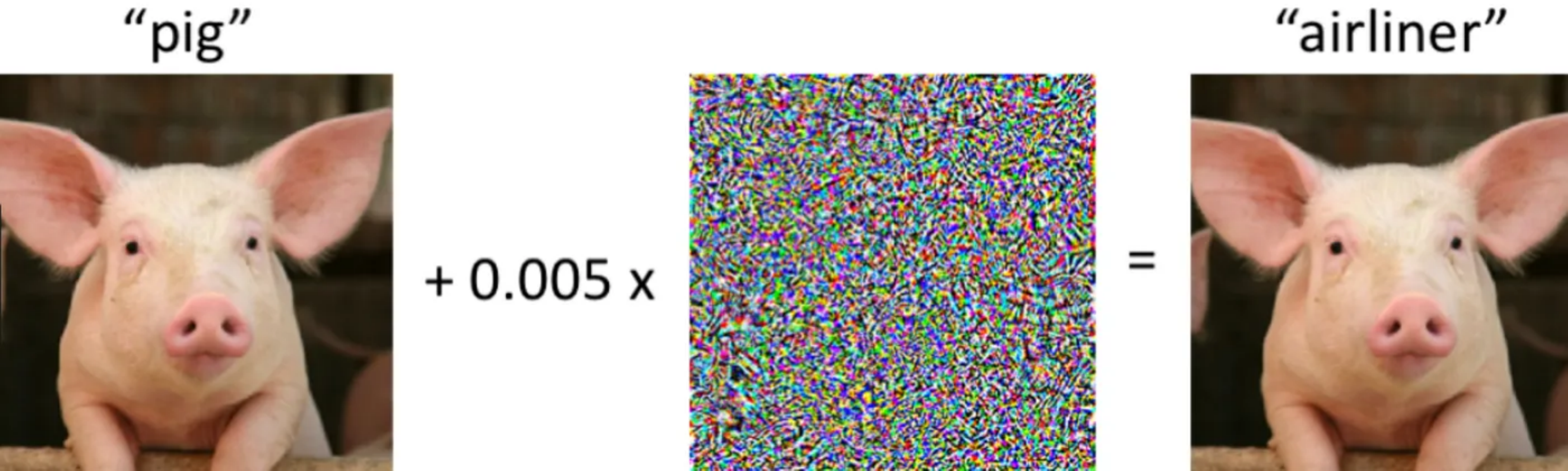}
  \caption{Example of an adversarial attack on an image classifier. The original image (left) is correctly classified as a pig, but the perturbed image (right) is misclassified as an airliner. The perturbation is imperceptible to the human eye but causes the classifier to make an incorrect prediction. The added noise is scaled by a factor of $0.005$ for visualization purposes.
    Example was taken from the blogpost of~\citet{haldar2020adversarial}.
  }  \label{fig:adversarial_pig}
\end{figure}

In a deep learning context, \( f \) is parameterized as a neural network, trained on a labeled dataset to minimize a classification loss. When trained effectively, \( f \) achieves high accuracy on test data, successfully predicting the correct label \( y \) for most inputs \( \mathbf{x} \)~\citep{krizhevsky2012imagenet}.
Despite this success, deep neural networks are highly vulnerable to adversarial attacks~\citep{szegedy2013intriguing, goodfellow2014explaining}, where small, carefully designed perturbations to the input can cause incorrect predictions. This instability raises concerns in critical applications such as autonomous driving and healthcare. In Figure~\ref{fig:adversarial_pig}, we illustrate an adversarial attack on an image classifier, where an imperceptible perturbation causes the classifier to misclassify an image of a pig as an airliner.
%
\begin{definition}[Adversarial attacks~\citep{szegedy2013intriguing}]
  Let \(\vx \in \mathcal{X}\) be an input, \(y \in \mathcal{Y}\) its true label, and \(f\) a classifier. An adversarial perturbation at level \(\varepsilon\) is a vector \(\perturb\) such that \(\lVert \perturb \rVert \leq \varepsilon\) and:
  \begin{equation*}
    \argmax_k f_k(\vx + \perturb) \neq y.
  \end{equation*}
\end{definition}
The level $\epsilon$ is also called adversarial budget, and it quantifies the maximum perturbation allowed to the input.
In this work and commonly in the literature
we evaluate adversarial~\citep{carlini2017towards,madry2018towards} robustness under \(\ell_2\)-bounded perturbations \(\perturb\) such that
\(
\|\perturb\|_2 \leq \varepsilon,
\)
where \(\varepsilon\) is expressed in pixel scale units, i.e., relative to the input domain \([0, 255]\) (image range pixel intensity). In normalized input space \([0, 1]\), this corresponds to
\(
\varepsilon_\text{normalized} = \frac{\varepsilon_\text{raw}}{255}.
\)

To construct and simple and effective adversarial attack one can use the Projected Gradient Descent (PGD) attack~\citep{madry2018towards}.
The PGD attack iteratively updates an adversarial perturbation \(\perturb\) to maximize the classifier's loss while ensuring the perturbation remains within the allowed \( \ell_2 \)-norm constraint, see Algorithm~\ref{algo:pgd_l2}.
\begin{algorithm}[h]
  \caption{PGD Attack under \(\ell_2\)-norm Constraint}
  \label{algo:pgd_l2}
  \begin{algorithmic}[1]
    \State \textbf{Input}: Classifier \( f \), input \( \vx \), true label \( y \), step size \( \alpha \), perturbation budget \( \budget \), number of iterations \( T \).
    \State \textbf{Initialize}: \(\perturb_0 \gets \text{random unit vector} \cdot \budget\)
    \For{\(t = 0\) to \(T-1\)}
    \State Compute gradient: \( g \gets \nabla_{\vx} \mathcal{L}(f(\vx + \perturb_t), y) \)
    \State Normalize gradient: \( g \gets g / \|g\|_2 \)
    \State Update perturbation: \( \perturb_{t+1} \gets \perturb_t + \alpha g \)
    \State Project onto \(\ell_2\)-ball: \( \perturb_{t+1} \gets \budget \cdot \frac{\perturb_{t+1}}{\max(\budget, \|\perturb_{t+1}\|_2)} \)
    \EndFor
    \State \textbf{Return} \( \vx_{\text{adv}} = \vx + \perturb_T \)
  \end{algorithmic}
\end{algorithm}
Several methods have been proposed to enhance the robustness of classifiers. Adversarial training~\citep{madry2018towards} strengthens models by incorporating adversarial examples into the training process. However, the ongoing interplay between attacks and defenses has been likened to a \emph{cat-and-mouse game}, where new attacks continually challenge existing defenses~\citep{athalye2018obfuscated}. This dynamic nature highlights the need for defenses with formal, certified guarantees of robustness.
\begin{figure}[t]
  \centering
  \includegraphics[width=1.0\textwidth]{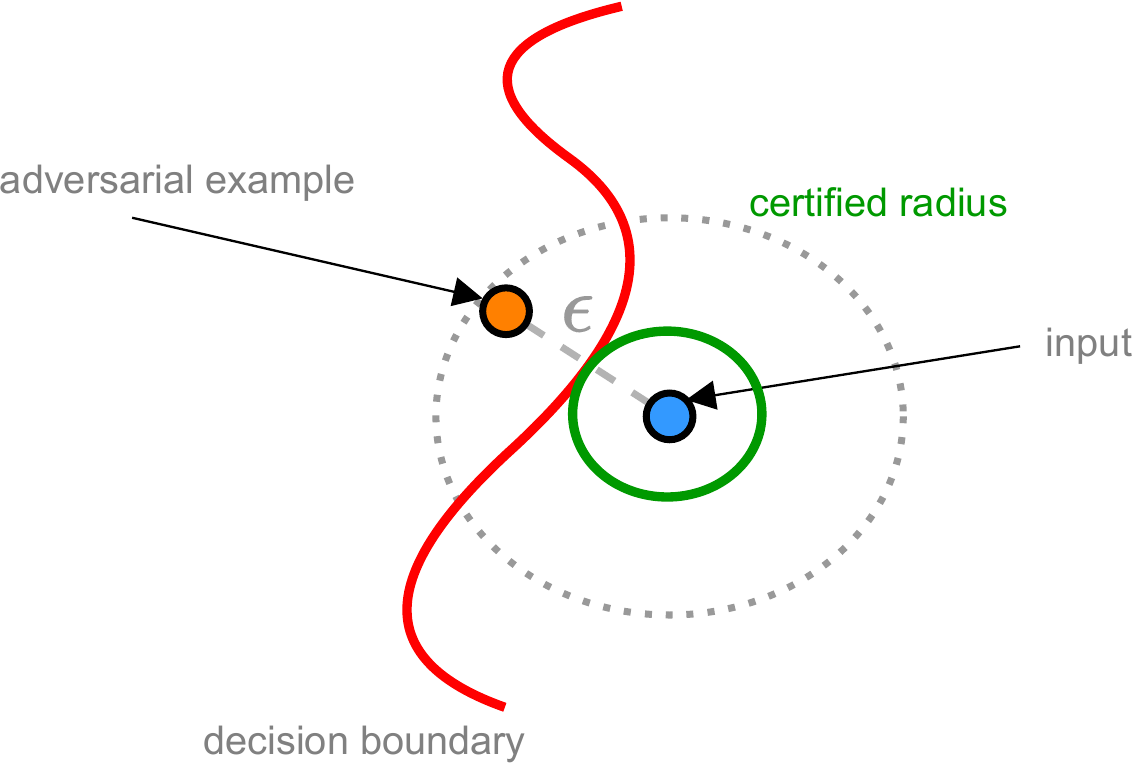}
  \caption{Illustration of an adversarial attack $\vx_{\text{adv}}$ of budget \( \budget \) on an input \( \vx \) with classifier boundaries and the certified radius in green. No adversarial perturbation within the certified radius can cause the classifier to change its decision.}
  \label{fig:certified_robustness}
\end{figure}
Certified defenses, such as Lipschitz continuity~\citep{cisse2017parseval, tsuzuku2018lipschitz} and \emph{randomized smoothing} (RS)~\citep{lecuyer2018certified, li2018second, cohen2019certified}, have emerged to address the challenges posed by adversarial attacks. A key metric for evaluating these defenses is the \emph{certified robust radius}, which quantifies the maximum allowable perturbation to an input \( \vx \) that ensures the classifier’s decision remains stable. Larger certified radii indicate greater robustness.
\begin{definition}[Certified radius~\citep{tsuzuku2018lipschitz}]
  \label{def:certified_radius}
  For a classifier \( f: \mathcal{X} \to \mathbb{R}^c \), an input \( \vx \in \mathcal{X} \), and its label \( y \), the certified radius \( R(f, \vx, y) \) is defined as:
  \begin{align*}
    \mathrm{R}(f, \vx, y) := \inf \left\{ \budget \mid \budget > 0, \ \exists \perturb \in B_2(\vzero, \budget), \ \argmax_{k} f_{k}(\vx + \perturb) \neq \argmax_{k} f_{k}(\vx) \right\}.
  \end{align*}
  where \( B_2(\vzero, \budget) = \{ \perturb \in \mathbb{R}^d ~|~ \norm{\perturb}_2 \leq \budget \} \).
\end{definition}
Figure~\ref{fig:certified_robustness} illustrates the concept of certified robustness, where the certified radius quantifies the maximum allowable perturbation \( \budget \) to an input \( \vx \) that ensures the classifier's decision remains stable.
To measure the robustness of a classifier, we define the \emph{certified accuracy} as the proportion of inputs for which the classifier is guaranteed to be robust against any perturbation within a given radius \( \budget \). The certified accuracy provides a quantitative measure of the classifier's robustness, reflecting the proportion of inputs for which the classifier's decision remains stable under perturbations.

\begin{definition}[Certified accuracy for $\ell_2$-norm perturbations~\citep{tsuzuku2018lipschitz}]
  \label{def:certified_accuracy}
  The \emph{certified accuracy} of a classifier at a perturbation radius \( \budget \) is the proportion of inputs for which the classifier is guaranteed to be robust against any perturbation within the \(\ell_2\)-norm ball of radius \( \budget \).
  The certified accuracy at radius \( \budget \) is defined as:
  \[
    \frac{1}{|\mathcal{D}|} \sum_{(\vx, y) \in \mathcal{D}} \mathds{1}_{\left( \argmax_{k} f_{k}(\vx + \perturb) = y, \ \forall \ \|\perturb\|_2 \leq \budget \right)} \ .
  \]
\end{definition}

\subsection{Lipschitz bounded networks}
\label{sec:lipschitz_networks_certified_robustness}

The concept of Lipschitz continuity complements the certified robust radius by introducing the Lipschitz constant, which quantifies the sensitivity of the network \( f \) to input perturbations \( \perturb \). Specifically, a function \( f \) is Lipschitz continuous if there exists a constant \( \Lip(f) > 0 \) such that:
\[
  \norm{f(\vx + \perturb) - f(\vx)} \leq \Lip(f) \norm{\perturb}.
\]
A smaller Lipschitz constant implies that the network \( f \) exhibits slower variations in its output with respect to changes in its input, thus reducing the risk of instability under perturbations. In parallel, the prediction margin,
\[
  \margin(f(\vx), y) = \max(0, f_y(\vx) - \max_{k \neq y} f_k(\vx)),
\]
quantifies the confidence of the classifier \( f \) in assigning the label \( y \) to the input \( \vx \). A larger margin indicates greater confidence and resilience to perturbations.
Together, the Lipschitz constant and the margin are interconnected quantities that directly influence the certified robust radius \( \mathrm{R}(f, \vx, y) \).

A simple lower bound on the certified radius can be derived from the Lipschitz constant of the network. This bound is expressed as:
\begin{equation}
  \label{eq:first_radius}
  \mathrm{R}(f, \vx, y)  \geq \min_{y^\prime \neq y} \frac{\max (f_y(\vx) - f_{y^\prime}(\vx), 0)}{\Lip(f_{y} - f_{y^\prime})}.
\end{equation}

The Lipschitz constant \( \Lip(f_{y} - f_{y^\prime}) \) can be bounded by \( \Lip(f_y) + \Lip(f_{y^\prime}) \), giving the expression:
\[
  \min_{y^\prime \neq y} \frac{\max (f_y(\vx) - f_{y^\prime}(\vx), 0)}{\Lip(f_{y}) + \Lip(f_{y^\prime})}.
\]

This bound reflects the sensitivity of individual scores to input perturbations. When all logits share the same Lipschitz constant, robustness depends on the logit margin scaled by this sensitivity:
\begin{align}
  \label{eq:rcoord_bound}
  \Rcoord(f, \vx, y) = \frac{\margin(f(\vx), y)}{2 \Lip(f_{y})}.
\end{align}
Since robustness is determined by the smoothness of each logit rather than the entire classifier, this justifies the term coordinate-wise bound and derived radius $\Rcoord$.

\begin{figure}
  \includegraphics[width=1.0\textwidth]{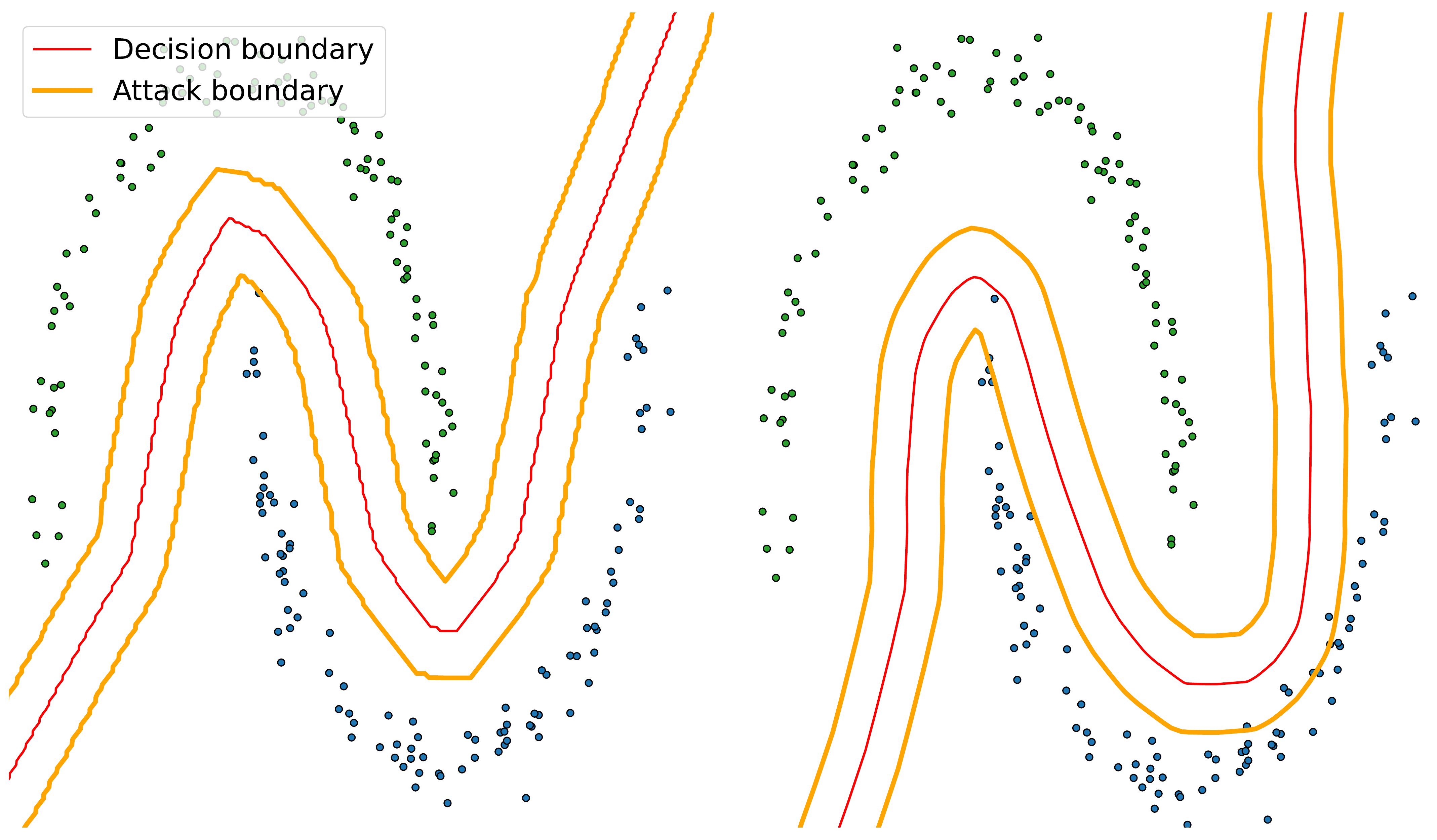}
  \caption{Illustration of the attack perimeter (orange) for a given attack budget for a MLP classifier with two classes (green and blue dots). The margin is the distance between the decision boundary (red) and the closest point to the decision boundary.
    The left figure shows the decision boundary with larger margins than the right figure ensuring robustness to the attack perimeter in orange.}
  \label{fig:boundary_margin}
\end{figure}

This expression shows that the certified radius is directly proportional to the margin and inversely proportional to the Lipschitz constant. Larger margins and smaller Lipschitz constants lead to greater robustness. In Figure~\ref{fig:boundary_margin}, we illustrate the attack perimeter for a given budget and the same Lipschitz constant for both classifiers, illustrating that larger margins increase robustness.
Note that both classifiers perform with perfect accuracy on the test set.

For the \(\ell_2\)-norm, \cite{tsuzuku2018lipschitz} derived a tighter bound for \( \Lip(f_{y} - f_{y^\prime}) \).
\begin{proposition}[Lipschitz-margin lower bound on the certified radius \citep{tsuzuku2018lipschitz}]
  Given a Lipschitz continuous network \( f \) under the \(\ell_2\)-norm, and given a perturbation level \( \varepsilon > 0 \), \( \vx \in \mathcal{X} \), and \( y \in \mathcal{Y} \) as the label of \( \vx \), if the margin satisfies:
  \[
    \margin(f(\vx), y) > \sqrt{2} \Lip(f) \varepsilon,
  \]
  then for every \( \perturb \) such that \( \lVert \perturb \rVert_2 \leq \varepsilon \), we have \( \argmax_{k} f_k(\vx + \perturb) = y \).
\end{proposition}
Reworking this proposition, a lower bound for the certified radius is given by
\begin{align}
  \label{eq:radius_tsuzuku}
  \Rglobal(f, \vx, y) = \frac{\margin(f(\vx), y)}{\sqrt{2} \Lip(f)}.
\end{align}
Since it relies on the global Lipschitz property, it is referred to as the $\Rglobal$.
Indeed, for Lipschitz-by-design networks, \( \Lip(f) \) can be bounded explicitly, making it straightforward to estimate the certified radius by evaluating the margin at the input \( \vx \).
These results highlight the critical role of Lipschitz continuity in achieving certified robustness, demonstrating that controlling \( \Lip(f) \) and maximizing the margin \( \margin(f(\vx), y) \) are key strategies for designing robust models.

\subsection{Randomized smoothing}
\label{sec:randomized_smoothing}

Consider the set \( \Delta^{c-1} = \left\{ \vp \in \mathbb{R}_+^c \mid \mathbf{1}^\top \vp = 1 \right\} \) defines the \((c-1)\)-dimensional probability simplex.
Let \( \tau: \mathbb{R}^c \mapsto \Delta^{c-1} \) represent a mapping onto this simplex, typically corresponding to functions like \( \mathrm{softmax} \) or \( \mathrm{hardmax} \).
For a logit vector \( \vz \in \mathbb{R}^c \), the mapping onto \( \Delta^{c-1} \) is denoted by \( \tau(\vz) \).
A specific case of this mapping is the \( \mathrm{hardmax} \), where for each component \( k \), we have \( \tau_k(\vz) = \mathds{1}_{\arg\max_{i} \vz_i = k} \).

We note  \( f: \mathcal{X} \mapsto \mathbb{R}^c \) as the network, which produces the logits before applying \( \tau \).
$\mathrm{hardmax}$ is a common choice for \( \tau \) in the context of randomized smoothing, as it assigns all the probability mass to the highest logit, softmax is also used in the work of~\citet{levine2019certifiably}.
The soft classifier $F : \mathbb{R}^d \to \Delta^{c-1}$ is defined as:
\begin{equation}
  \label{eq:soft_classifier}
  F(\vx) = \tau(f(\vx)) = \left[ F_1(\vx), \ldots, F_c(\vx) \right]^\top,
\end{equation}
which outputs a probability distribution over the \( c \) classes. The final decision is given by the
hard classifier $\Fhard : \R^d \to \mathcal{Y}$, defined as:
$$\Fhard(\vx) = \arg\max_{k \in \mathcal{Y}} F_k(\vx),$$
which returns the predicted label \( \hat{y} = \Fhard(\vx) \).

Randomized smoothing (RS) was initially proposed by~\cite{li2018second-order, lecuyer2019certified} and further developed in key works~\citep{cohen2019certified, salman2019provably}. The core idea behind RS is to improve the robustness of a classifier by averaging its predictions over Gaussian perturbations of the input. This gives a hard smoothed classifier $\tilde{F}^H$, which is more resistant to adversarial attacks than the classifier $F^H$.
The soft smoothed classifier $\tilde{F}$ is defined as the Weierstrass transform of $F$:
\[
  \tilde{F}(\vx) = \mathbb{E}_{\mathbf{\delta} \sim \mathcal{N}(0, \sigma^2 \mI)} \left[ F(\vx + \mathbf{\delta}) \right] \ .
\]
This means that $\Ftildehard = \argmax_k \tilde{F}_k(\vx)$   returns the class $k$ that has the highest probability when Gaussian noise $\mathbf{\delta}$ is added to the input $\vx$.

\begin{figure}[t]
  \centering
  \includegraphics[width=1.0\textwidth]{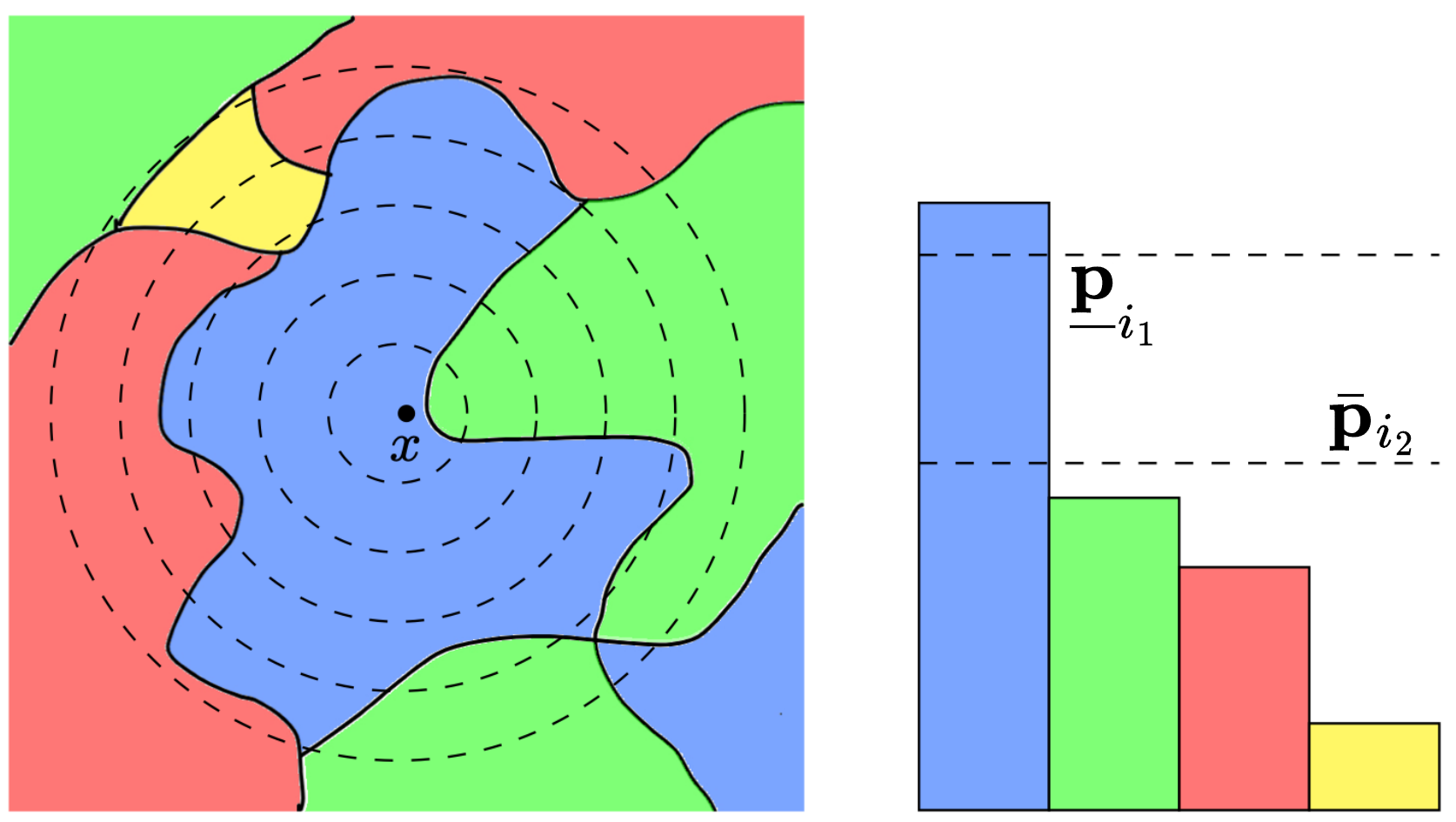}
  \caption{Evaluating the smoothed classifier at a given input \( \vx \). Left: the decision regions of the base classifier \( F \) are shown in different colors. The dotted lines represent the level sets of the Gaussian distribution \( \mathcal{N}(\vx, \sigma^2 \mathbf{I}) \). Right: the probability distribution of \( F(\mathcal{N}(\vx, \sigma^2 \mathbf{I})) \). Here, \( \underline{\vp}_{i_1} \) is a lower bound on the probability of the predicted (top) class, and \( \overline{\vp}_{i_2} \) is an upper bound on the probability of each other class. In this example, the smoothed classifier assigns the input to the "blue" region.
    Figure taken from~\cite{cohen2019certified}}.
  \label{fig:randomized_smoothing}
\end{figure}

Note that if \(\tau = \mathrm{hardmax}\), then for all \(k \in \mathcal{Y}\), we have:
\[
  \tilde{F}_k(\mathbf{x})
  \;=\;
  \mathbb{P}_{\mathbf{\delta}\sim \mathcal{N}(0,\sigma^2 I)}
  \Bigl[\arg\max_{i}\, f_i(\mathbf{x}+\mathbf{\delta}) \;=\; k\Bigr],
\]
which is the usual formulation in the literature~\citep{cohen2019certified} based on
the probabilistic interpretation of the smoothed classifier. This quantity corresponds to the probability that class \( k \) wins the majority vote under Gaussian perturbations.
However, we retain
the more general notation to accommodate other simplex mappings~\(\tau\) besides
\(\mathrm{hardmax}\) in the remainder of this thesis.
For ease of notation, let's denote $\vp = \Ftilde (\vx)$, the vector of output probabilities given by the smoothed classifier, with $\vp_k$ its $k-$th component.
A key contribution of RS is that it allows us to compute a lower bound on the certified radius.
Its estimation is based on the top two class probabilities, denoted as $\vp_{i_1}$ and  $\vp_{i_2}$ where $ i_1 = \argmax_k \vp_k(\vx)$ and $i_2 = \argmax_{k \neq i_1} \vp_k(x)$. In practice the lower bound on the radius is obtained using upper bound $\overline{\vp}_{i_2}$ and respectively lower bound ($\underline{\vp}_{i_1}$) on the probabilities  $\vp_{i_2}$ and respectively $\vp_{i_1}$.
The Figure~\ref{fig:randomized_smoothing} illustrates the decision regions of the base classifier \( F \) and the probability distribution of \( F(\mathcal{N}(\vx, \sigma^2 \mathbf{I})) \) for a given input \( \vx \).
Two main approaches have been proposed to derive the certified radius. The first relies on a probabilistic proof using the Neyman-Pearson lemma~\citep{cohen2019certified}. The second employs a deterministic proof based on the Weierstrass transform regularization to enforce Lipschitz continuity, which is then used with the coordinate-wise certified radius \(\Rcoord\)~\citep{salman2019provably}.
The bound on certified radius is given in the following Theorem~\ref{thm:certified_radius_rs_cohen}:
\begin{theorem}[Certified radius for randomized smoothing~\citep{cohen2019certified}]
  \label{thm:certified_radius_rs_cohen}
  Let \( f : \mathbb{R}^d \to \R^c \) be any deterministic or random function, its associated soft classifier $F = \mathrm{hardmax} \circ f$ , and let \( \delta \sim \mathcal{N}(0, \sigma^2 \mathbf{I}) \). Define the smoothed classifier \( \tilde{F} \) as:
  \[
    \tilde{F}(\vx) = \E(F(\vx + \delta)).
  \]
  We note $\vp = \tilde{F}(\vx)$, suppose \( i_1 \in \mathcal{Y} \) and
  \( \underline{\vp}_{i_1}, \overline{\vp}_{i_2} \in [0, 1] \) satisfy:
  \[
    \vp_{i_1} \geq \underline{\vp}_{i_1} \geq \overline{\vp}_{i_2} \geq \vp_{i_2}.
  \]
  Then, \( \tilde{F}^\mathrm{H}(\vx + \delta) = i_1 \) for all \( \|\perturb\|_2 < \mathrm{R} \), where the certified radius \( \mathrm{R} \) is given by:
  \begin{align}
    \label{eq:certified_radius_rs}
    \mathrm{R} = \frac{\sigma}{2} \left( \Phi^{-1}( \underline{\vp}_{i_1}) - \Phi^{-1}(\overline{\vp}_{i_2}) \right),
  \end{align}
  and \( \Phi^{-1} \) denotes the inverse cumulative distribution function of the standard Gaussian distribution.
\end{theorem}
In practice most works~\citep{cohen2019certified,salman2019provably, yang2020randomized}
used $\underline{\vp}_{i_1} = \vp_{i_1}$ and $\overline{\vp}_{i_2} = 1 - \vp_{i_1}$ to simplify the computation of the certified radius.
Resulting in the following mono-class certified radius, smaller than Equation~\ref{eq:certified_radius_rs}:
\begin{align}
  \label{eq:certified_radius_rs_mono}
  \Rmono(\vp) = \sigma \Phi^{-1}(\vp_{i_1}) \leq \mathrm{R}.
\end{align}
This mono-class certified radius \(\Rmono\) is non trivial only if the winning class probability \(\vp_{i_1}\) is greater than \(\frac{1}{2}\), otherwise it is equal to zero.

This radius provides a quantifiable measure of the classifier’s robustness.
The probabilities $\vp$ are estimated using a Monte Carlo (MC) approach. Given $n$ Gaussian samples $\mathbf{\delta}_i \sim \mathcal{N}(0, \sigma^2 \mI)$, the empirical estimate of $\vp$ is:
\[
  \hat{\vp} = \frac{1}{n} \sum_{i=1}^n F(\vx + \delta_i) \ .
\]


Since the certified radius \(\mathrm{R}\) is derived from the probabilities \(\vp_{i_1}\) (for the winning class) and \(\vp_{i_2}\) (for any other class), we estimate these values using Monte Carlo sampling. This estimation introduces uncertainty due to the limited number of samples.
To control this uncertainty and to compute \(\mathrm{R}\) at an exact risk level \(\alpha\), confidence interval methods are employed. In particular, lower and upper bounds, \(\underline{\vp}_{i_1}\) and \(\overline{\vp}_{i_2}\), are obtained via concentration inequalities (see, e.g., \cite{lecuyer2019certified, levine2019certifiably}) or the Clopper-Pearson method (as in \cite{cohen2019certified}). These bounds provide a conservative margin that is then converted into the certified radius \(\mathrm{R}\).
The overall randomized smoothing process is summarized in Figure~\ref{figure:mind_map_cohen}.
\input{figures/mind_map_cohen.tex}

Randomized smoothing relies on injecting noise at inference time to improve the robustness of deep neural networks.
Notably, randomized smoothing can be interpreted as a form of regularization, as it  aims to control the model's sensitivity to input perturbations. In the next section, we explore regularization techniques that explicitly enforce smoothness for improved generalization.

\section{Regularization for generalization}
\label{sec:regularization_techniques}

Regularization techniques are essential for enhancing the generalization capability of neural networks by controlling model complexity, reducing overfitting, and stabilizing training.
Generalization is the ability of a model to perform well on unseen data.
It is often associated with the Lipschitz constant of the network~\citep{bartlett2017spectrally} and the flatness of the minima found during training~\citep{hochreiter1997flat} as highlighted by empirical studies~\citep{keskar2017large-batch,chaudhari2019entropy}, where sharp minima are linked to poorer generalization.

For a given label $ \vy $, the loss is denoted by $\mathcal{L}(\vy, \vtheta)$ in short, where
$
  \vtheta = \mathrm{vec}\left( \{ \mathbf{W}^{(l)} \}_{l = 1, \dots, L} \right) \ .
$
For brevity, we write the loss function as \( \mathcal{L}(\tens{\theta}) \), and assuming it is differentiable, we denote respectively
\( \nabla_{\tens{\theta}} \mathcal{L} \) and \( \nabla^2_{\tens{\theta}} \mathcal{L} \) as its gradient and Hessian with respect to the parameters \( \tens{\theta} \).

\subsection{Lipschitz constant}
\label{sec:lipschitz_network}
Lipschitz constant regularization has emerged as a significant technique for improving the generalization capabilities of deep neural networks. By constraining the Lipschitz constant, these methods aim to control the sensitivity of the model to input perturbations, leading to smoother decision boundaries and enhanced robustness.~\citet{finlay2018lipschitz} demonstrated that input gradient regularization, which implicitly enforces a Lipschitz constraint, results in improved generalization and adversarial robustness.
The work of~\citet{yoshida2017spectral} introduces a regularization scheme that penalizes the sum of the spectral norms of weight matrices by adding a corresponding term to the loss function.
$$
  \min_\vtheta \mathcal{L}(\vtheta) + \lambda \sum_{l=1}^{\nblayers} \|\mW_l\|_2,
$$
where \( \lambda \) is a hyperparameter controlling the strength of the regularization term.
In the same flavor \citet{gouk2021regularisation} introduced a projection method  to enforce Lipschitz continuity in neural networks for different norms. Instead of relying on regularization, they enforce a hard constraint.
Furthermore,~\citet{scaman2018lipschitz} provided an analysis of Lipschitz continuity in deep neural networks and proposed an efficient estimation technique, contributing to the understanding of how Lipschitz constraints can be applied to enhance generalization.

\subsection{Flat minima}
\label{section:background_and_related_workt:subsection:flat_minima}

The relationship between loss landscapes, generalization, and stochastic gradient descent (SGD) has been a central topic in machine learning research for years~\citep{hochreiter1997flat}. For instance, it has been shown that in overparameterized models, local minima of the loss function are often close to the global minima~\citep{choromanska2015loss}. Additionally,~\citet{xing2018walk} demonstrates that SGD exhibits an implicit bias, favoring regions of the loss landscape resembling a valley.
The sharpness is captured by the magnitude  of the singular values of the Hessian of the loss w.r.t parameters,
$\|\nabla^2 \mathcal{L}(\vtheta) \|_2 $ (worst sharpness) or $\mathrm{Tr}(\nabla^2 \mathcal{L}(\vtheta))$ (average sharpness)
and reflects the curvature of the loss landscape.
A pivotal study by~\citet{keskar2017large-batch} shows that large-batch training tends to converge to sharp minima, which correlates with worse generalization compared to the flatter minima achieved with small-batch training, as Hessian measure are hard to compute in practice, other metrics have been proposed to assess sharpness, such as
\begin{equation}
  \label{eq:sharpness}
  \frac{\max_{\mathbf{\Delta} \in \mathcal{B}}\loss(\vtheta + \mathbf{\Delta)} - \loss(\vtheta)}{1 + \loss(\vtheta)} \ ,
\end{equation}
where \( \mathcal{B} \) is a ball around the current parameters \( \vtheta \).
However,~\citet{dinh2017sharp} point out that common sharpness metrics, such as the spectral norm of the Hessian of \( \mathcal{L}(\vtheta) \) and proxies, are sensitive to re-scaling.
It is important to note that flatter minima do not always guarantee better generalization.
Counterexamples exist where sharp minima generalize well and flat minima perform poorly~\citep{andriushchenko2022understanding, zhang2017rethinking}.
Nevertheless,~\citet{jiang2020fantastic} show that sharpness-based metrics often outperform other complexity metrics for evaluating generalization, such as the metric from Equation~\ref{eq:sharpness}.
Sharpness-Aware Minimization (SAM), introduced by \citet{foret2021sharpnessaware}, is a widely studied loss smoothing technique that explicitly promotes flat minima by penalizing sharp regions of the loss landscape.
SAM minimizes a robust smoothed objective:
\begin{equation}
  \label{eq:sam}
  \min_{\vtheta} \max_{\mathbf{\Delta} \in \mathcal{B}(0, \rho)} \mathcal{L}(\vtheta + \mathbf{\Delta}) \ ,
\end{equation}
where \( \mathcal{B}(0, \rho) \) represents a ball of radius \( \rho \) around zero, and \( \mathbf{\Delta} \) is the perturbation applied to the parameters. By introducing an adversarial component into the optimization process, SAM effectively reduces sharpness in the loss landscape, which improves generalization across many computer vision tasks.
Flat minima, known to enhance generalization, can be achieved through various regularization strategies, including noise injection.

\subsection{Loss smoothing and noise injection}
Noise injection methods add stochasticity during training, either to inputs, activations, or weights. It improves robustness and prevents overfitting.
The most widely known is probably the dropout~\citep{srivastava2014dropout}, which randomly drops units during training, reducing co-adaptation of neurons and improving generalization, it is applied through noise injection on activations.
However, its stochastic nature introduces additional variance in performance, requiring careful tuning to prevent underfitting, particularly when applied to deep models~\citep{liu2023dropout}. The increased training variability may also result in unstable training dynamics under certain conditions.
Perturbed Gradient Descent (PGD)~\citep{jin2017escape} introduces noise into the weight updates to help models escape saddle points and better explore the optimization landscape.
Building on PGD, ~\citep{orvieto2022anticorrelated} modifies the noise structure and proves it promotes convergence to flatter minima by controlling the curvature of the loss landscape.
The resulting optimization program can be summarized with the smoothed loss:
$$\min_{\vtheta}
  \mathbb{E}_{\mathbf{\Delta} \sim \mathcal{N}(0, \sigma^2 \mI)}\left[ \mathcal{L}(\vtheta + \mathbf{\Delta}) \right].
$$
The resulting optimization landscape becomes smoother, reducing the impact of sharp regions. However, tuning the noise parameter \( \sigma \) is critical, as improper choices can lead to instability with exploding variance or insufficient regularization.

While noise-based regularization techniques have been widely adopted, they often introduce undesirable variance into the training process, which can hinder performance.
The work of~\citet{bishop1995training} interprets Gaussian noise on inputs as Tikhonov regularization on parameters also known as weight decay, establishing a connection between noise injection and deterministic loss smoothing. This encourages smaller weights and promotes simpler solutions that generalize better~\citep{krogh1992simple}. Such regularization flattens the loss landscape, leading to convergence toward wider, flatter minima that correlate with improved generalization~\citep{hochreiter1997flat}. The regularized loss function can be expressed as:
$$\min_{\mathbf{\theta}} \mathcal{L}(\vtheta) + \frac{\sigma^2}{2} \|\vtheta\|_2^2.$$
Notable other works proposed layer-wise regularization such as margin constraint~\citep{elsayed2018large} or activation regularization~\citep{yashwanth2024minimizing}.

\section{Conclusion}
\label{sec:backgrond_conclusion}
In this chapter, we provided an overview of key concepts in neural network robustness and generalization, focusing on Lipschitz continuity, certified robustness, and regularization techniques. First, we introduced Lipschitz continuity as a fundamental property of neural networks, impacting stability, adversarial robustness, and generalization. We examined methods for estimating the Lipschitz constant, including product upper bounds and local Lipschitz estimation techniques. Additionally, we discussed architectural constraints, such as orthogonal, rescaled, and residual layers, designed to enforce Lipschitz constraints explicitly.
We then explored certified robustness, highlighting how Lipschitz-bounded networks provide deterministic certification guarantees, while randomized smoothing offers a probabilistic alternative. We established the connection between the Weierstrass transform and randomized smoothing, showing how both techniques introduce controlled smoothing to certify robustness.
Finally, we examined regularization techniques for improving generalization, including Lipschitz-based constraints, flat minima optimization, and noise-based methods such as dropout and loss smoothing. These approaches aim to control model complexity, reduce overfitting, and improve stability during training.

Overall, this chapter established the theoretical foundations necessary for understanding stability in deep learning, with a focus on Lipschitz continuity. In the following chapters, we delve deeper into practical implementations of Lipschitz networks, loss smoothing techniques, and their applications to certified robustness and generalization.
A key challenge in Lipschitz constant estimation, Lipschitz regularization, and Lipschitz layer design is the computation of the spectral norm for linear layers, which is crucial for the implementation of the methods presented in the subsequent chapters. To address this, we introduce in the next chapter Gram iteration, a novel approach for spectral norm estimation that is faster, deterministic, and provides strict upper bound guarantees for both convolutional and dense layers.

%% file: figures/mind_map_cohen.tex
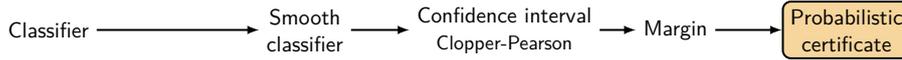
\begin{figure*}[t]
  \centering
  \scalebox{0.75}{%
    \input{figures/figurev3_cohen}
  }%
  \caption{
    \cite{cohen2019certified}~smooths soft classifier $F$ to create the soft smoothed classifier $\tilde{F}$. The risk factor $\alpha$ is then estimated using the Clopper-Pearson interval to provide a probabilistic certificate using the Neyman-Pearson lemma.
  }
  \label{figure:mind_map_cohen}
\end{figure*}

%% file: figures/figurev3_cohen.tex
\tikzset{%
>={Latex[width=0mm,length=0mm]},
base/.style = {fill=white, font=\sffamily},
inputs0/.style={
    rounded corners,
    draw=black,
    line width=1pt,
    fill=burstblue!50,
  },
inputs/.style={
    rounded corners,
    fill=none,
    draw=none,
    line width=1pt,
    minimum width=2cm,
    minimum height=1cm,
  },
crops/.style={
    draw=black,
    rounded corners=0,
    line width=1pt,
    minimum width=0.75cm,
    minimum height=0.56cm
  },
/.style={
  },
rectangleinput/.style={
    rounded corners,
    fill=myorange,
    draw=black,
    line width=1pt,
    minimum width=3.5cm,
    minimum height=1cm
  },
rectangle1/.style={
    rounded corners,
    fill=myorange,
    draw=black,
    line width=1pt,
    minimum width=2cm,
    minimum height=1cm
  },
rectangle2/.style={
    rounded corners,
    fill=white,
    draw=black,
    dashed,
    line width=1pt,
    minimum width=2cm,
    minimum height=1cm
  },
descr/.style={
    fill=white,
    inner sep=2.5pt
  },
connector/.style={
    -latex,
    color=black,
    line width=1pt,
  },
rectangle connector/.style={
    connector,
    to path={(\tikztostart) |- ++(3.1,0.25) \tikztonodes -- (\tikztotarget)},
    pos=0.5
  },
}
\def\shifta{+0.25}
\def\shiftb{-0.50}
\def\shiftc{-1.30}
\begin{tikzpicture}[every node/.style=base, align=center, scale=5]
  \node[base] (classifier)  at (+0.8, +0.35+\shiftb) {Classifier};
  \node[base] (smooth1)     at (+1.7, +0.35+\shiftb) {Smooth \\ classifier};
  \node[base] (risk_shift1) at (+2.4, +0.35+\shiftb) {Confidence interval \\ {\small Clopper-Pearson}};
  \node[base] (margin2)     at (+3.0, +0.35+\shiftb) {Margin};
  \node[rectangle1] (certificate2) at (+3.6, +0.35+\shiftb) {Probabilistic \\ certificate};

  \draw[connector] (classifier) -- (smooth1);
  \draw[connector] (smooth1) -- (risk_shift1);
  \draw[connector] (risk_shift1) -- (margin2);
  \draw[connector] (margin2) -- (certificate2.west);

\end{tikzpicture}%

%% file: content/chapter-spectral_norm_computation.tex
%
\chapter{Spectral norm computation}\label{chapter:spectral_norm_estimation}


\minitoc%

The spectral norm of a matrix, defined as its largest singular value, is a fundamental quantity in numerical linear algebra, optimization, and control theory~\citep{golub2012matrix}.
Accurate and efficient computation of the spectral norm is essential for tasks ranging from stability analysis and system design to matrix approximations and data regularization~\citep{lanczos1950iteration, arnoldi1951principle, kublanovskaya1962algorithms, lehoucq1996deflation}.
In deep learning, it plays a key role in understanding model stability, robustness, and generalization. Constraining the spectral norm of neural network layers has been shown to improve training stability~\citep{miyato2018spectral}, enhance generalization~\citep{bartlett2017spectrally}, and provide certified robustness~\citep{scaman2018lipschitz, tsuzuku2018lipschitz}.

Despite its importance, efficiently computing the spectral norm remains a challenge, particularly for deep neural networks where layers may be high-dimensional or structured, such as convolutional layers. Existing approaches, such as power iteration (\(\PI\)), provide estimates but lack strict guarantees on approximation quality, slow in convergence (linear),
and can be sensitive to initialization.$\SVD$ methods offer exact computation but are computationally expensive on GPUs.
This limits their effectiveness in applications requiring certified bounds, such as adversarial robustness and constrained optimization.

In this chapter, we introduce a novel method for spectral norm estimation that provides a strict upper bound at every iteration, addressing the limitations of prior techniques while achieving super geometric convergence, contrary to PI it is not a matrix-free method.
Our approach extends beyond fully connected layers to convolutional layers, including both circular and zero-padding cases.
While the spectral norm is traditionally defined via the explicit weight matrix, we also provide a matrix-free formulation that operates directly on the convolutional filter, even when the corresponding matrix is not explicitly constructed.

Unlike power iteration, our method is deterministic, ensuring stability and reproducibility, and integrates seamlessly into neural network training pipelines for spectral regularization.
Indeed, we provide an explicit gradient formulation for backpropagation, enabling spectral regularization during training.
Additionally, we leverage GPU acceleration to maintain computational efficiency, making our approach scalable to large-scale models.
Code is available at this link \url{https://github.com/blaisedelattre/lip4conv}.

\section{Spectral norm estimation for matrices}
The Frobenius norm of \( \mW \) is represented as \( \|\mW\|_\frob \). A matrix norm \( \|\cdot\| \) is said to be \emph{consistent} if it satisfies the submultiplicative property: for any matrices \( \mA \) and \( \mB \),
\[
  \|\mA \mB\| \leq \|\mA\| \|\mB\|.
\]
The Kronecker product of two matrices \( \mA \) and \( \mB \) is denoted as \( \mA \otimes \mB \).
The conjugate transpose of a matrix \( \mW \) is denoted by \( \mW^* \).

The spectral norm of a matrix \( \mW \in \C^{p \times q} \), which represents a linear operator, is defined as its largest singular value. Formally, the spectral norm \( \|\mW\|_2 \) is given by:
\[
  \|\mW\|_2 = \sup_{\|\vx\|_2 = 1} \|\mW \vx\|_2,
\]
where \( \vx \) is any vector in the input space. This norm quantifies the maximum amplification of a unit-norm vector under the transformation induced by \( \mW \). Consequently, computing the spectral norm requires evaluating the singular values of the matrix.
\begin{definition}[Singular value decomposition and singular values]
  For a matrix \( \mW \in \mathbb{C}^{p \times q} \), the Singular Value Decomposition ($\SVD$) is a factorization of the form:
  \[
    \mW = \mU \mSigma \mV^\ast,
  \]
  where \( \mU \in \mathbb{C}^{p \times p} \) and \( \mV \in \mathbb{C}^{q \times q} \) are unitary matrices, and
  \( \mSigma \in \mathbb{R}^{p \times q} \) is a diagonal matrix with non-negative entries \( \sigma_1 \geq \sigma_2 \geq \cdots \geq \sigma_{\min(p, q)} \geq 0 \), called the singular values of \( \mW \).
\end{definition}
The spectral norm $\norm{\mW}_2$ equals the largest singular value $\sigma_1(\mW)$.
Among the most widely used methods for computing the spectral norm are the Singular Value Decomposition  ($\SVD$) via $\QR$ method and Power Iteration ($\PI$).
\paragraph{Singular value decomposition} provides an exact computation of the spectral norm by directly extracting the largest singular value~\citep{golub1996matrix}. However, its computational complexity is \( O(pq^2) \) for a matrix of size $p \times q$,
making it expensive for large matrices. Iterative methods, such as the Lanczos~\citep{lanczos1950iteration} or Arnoldi~\citep{arnoldi1951principle} methods, offer more efficient alternatives but remain computationally intensive.
The QR method with shift is another iterative approach used to compute the $\SVD$~\citep{golub1996matrix}. This method transforms the matrix into a simpler form (e.g., upper triangular) and uses shifts to accelerate convergence. While QR methods are numerically stable and converge faster with shifts, they are not implemented on GPUs in frameworks like PyTorch, which instead rely on the divide-and-conquer algorithm for $\SVD$ computation. This divide-and-conquer approach, although accurate, can be slower for large matrices.
Moreover, $\QR$ methods are known to be ill-conditioned for differentiation, making them unsuitable for applications requiring backpropagation through $\SVD$~\citep{wang2022robust}. Ill-conditioning arises from the sensitivity of singular values and singular vectors to small perturbations in the input matrix, causing instability during gradient computation.

That is why we focus on iterative methods, which are more suitable for backpropagation and can be implemented on GPUs for efficient parallel computation.

\paragraph{Power iteration} ($\PI$) is a classical solution that serves as the basis for many others, and it is described in Algorithm~\ref{algo:power_iteration}.
The method iteratively applies the matrix $\mW$ to a random vector $\vu$ and normalizes the result to obtain the largest singular value $\sigma_1(\mW)$. The process is repeated for a fixed number of iterations $N_\text{iter}$ until convergence. The convergence rate of the $\PI$ is linear, with a rate of $\sigma_1 / \sigma_2$, where $\sigma_1$ and $\sigma_2$ are the largest and second-largest singular values of $\mW$, respectively. The method is not fully deterministic, as it starts from a randomly generated vector, and intermediate iterations are not guaranteed to be a strict upper bound on the spectral norm.
PI has been used to compute the dominant principal component efficiently for high-dimensional data~\citep{halko2011finding}, Google’s PageRank algorithm relies on power iteration to compute the leading eigenvector of the hyperlink matrix~\citep{page1999pagerank}, used to extract dominant spectral features of adjacency matrices in graph learning and spectral clustering~\citep{von2007tutorial}.
\begin{algorithm}[h]
  \caption{Power\_iteration$(\mW, N_\text{iter})$}\label{algo:power_iteration}
  \begin{algorithmic}[1]
    \State~\textbf{Inputs:} matrix $\mW$, number of iterations $N_\text{iter}$
    \State~\textbf{Initialization:} draw a random vector $\vu$
    \For{$i = 1$ to $N_\mathrm{iter}$}
    \State~$\vv \gets \mW \vu / \|\mW \vu\|_2$
    \State~$\vu \gets \mW^* \vv / \|\mW^* \vv\|_2$
    \EndFor~\State~$\sigma_1 \gets {(\mW \vu)}^* \vv$
    \State~\textbf{return} $\sigma_1$
  \end{algorithmic}
\end{algorithm}

To quantify the rate of convergence of the Power Iteration, we recall the notions of Q-linear and R-linear convergence.
\begin{definition}[Q-linear and R-linear convergence~\citep{nocedal2006numerical}]
  Suppose the sequence $(b_k)$ converges to a limit $L$. The sequence is said to converge \emph{Q-linearly} if, for some $\mu \in (0, 1)$:
  \begin{align*}
    \lim_{k \to \infty} \frac{|b_{k+1} - L|}{|b_k - L|} = \mu.
  \end{align*}
  Moreover, the sequence is said to converge \emph{R-linearly} if there exists another sequence $(\epsilon_k)$ such that:
  \begin{align*}
    |b_k - L| \leq \epsilon_k, \quad \forall k,
  \end{align*}
  and $(\epsilon_k)$ converges Q-linearly to $0$, i.e.,
  \begin{align*}
    \lim_{k \to \infty} \frac{\epsilon_{k+1}}{\epsilon_k} = \mu.
  \end{align*}
\end{definition}

For $\PI$ the convergence is Q-linear with a convergence rate of $\mu = \sigma_1 / \sigma_2$. This ratio can be close to $1$ and other variations of $\PI$ improve it~\citep{lanczos1950iteration,arnoldi1951principle, kublanovskaya1962algorithms} and more recently in~\citet{lehoucq1996deflation}.
However, these methods suffer from important drawbacks. First, many iterations are required to ensure  convergence, up to zero numerical precision, which can be computationally expensive.
Second, the method is not fully deterministic, as it starts from a randomly generated vector. Finally, intermediate iterations are not guaranteed to be a strict upper bound on the spectral norm.

\paragraph{Lower bound by Power Iteration}
Here we explain why $\PI$ provides a lower bound on the spectral norm for square matrices (without loss of generality).
Let $\mW \in \R^{n \times n}$ and $\|\mW\|_2 = \max_{\|u\|_2=1} \|\mW u\|_2$.
Power iteration generates
\[
  \vu_{k+1} = \frac{\mW \vu_k}{\|\mW \vu_k\|_2}, \qquad \|\vu_k\|_2 = 1.
\]
At each step,
\[
  \|\mW \vu_k\|_2 \;\leq\; \|\mW\|_2,
\]
so $\|\mW \vu_k\|_2$ is a lower bound on the spectral norm.
Moreover, since $\vu_k \to \vv_1$ (dominant eigenvector),
\[
  \lim_{k \to \infty} \|\mW \vu_k\|_2 \;=\; \|\mW\|_2.
\]

\section{Gram iteration}\label{sec:spectral_norm:gram_iteration}

To remedy those issues, we present \emph{Gram iteration} ($\GI$), a novel method for estimating an upper bound on the spectral norm of a matrix optimized for GPU acceleration.
It is based on repetitive Gram matrix computation.
\begin{definition}[Gram matrix]
  Given a matrix \(\mW \in \mathbb{C}^{p \times q}\), the Gram matrix associated with \(\mW\) is defined as
  \[
    \mW^\star \mW \in \mathbb{C}^{q \times q} \ ,
  \]
  where \(\mW^\star\) denotes the transpose (or Hermitian transpose in the complex case) of \(\mW\).
\end{definition}

Detailed proofs can be found in Appendix~\ref{app:sec:spectral_norm_computation}. We refer to the iterative method outlined in Theorem~\ref{thm:gram_iteration_main_result} as Gram iteration due to its structure: at each step, the matrix $\mW^{(t+1)}$ is the Gram matrix of $\mW^{(t)}$, and
$\sigma_1(\mW^{(t+1)}) = \sigma_1(\mW^{(t)})^2$
\begin{equation}
  \label{eq:gram_iterate}
  \begin{cases}
    \mW^{(1)} = \mW \ , \\
    \mW^{(t+1)} = {\mW^{(t)}}^\star {\mW^{(t)}} \ .
  \end{cases}
\end{equation}
$\mW^{(t)}$ is called the $t$-th iterate of the Gram iteration,
Gram iteration algorithm is presented in Algorithm~\ref{algo:gram_iteration_naive} for any matrix norm $\norm{.}$. Note that this algorithm does not include rescaling to avoid overflow, which is presented in Algorithm~\ref{algo:gram_iteration}.
\begin{algorithm}[h]
  \caption{Gram\_iteration\_naive$(\mW, N_\text{iter})$}
  \label{algo:gram_iteration_naive}
  \begin{algorithmic}[1]
    \State~\textbf{Inputs:} matrix $\mW$, number of iterations $N_\text{iter}$
    \For{$i = 1$ to $N_\mathrm{iter}$}
    \State~$\mW \gets \mW^* \mW$
    \EndFor
    \State~$\sigma_1 \gets \|\mW\|^{2^{-N_\text{iter}}}$
    \State~\textbf{return} $\sigma_1, \mW$
  \end{algorithmic}
\end{algorithm}
At the end of the Gram iteration, the dominant singular vectors of \( \mW \) can be estimated. The final matrix \( \mW \), obtained through repeated applications of \( \mW^* \mW \), acts as a projection onto the subspace of the largest singular value. As iterations increase, \( \mW \) converges to a rank-one projector aligned with the dominant singular vector.
Thus, any nonzero column \( \mW_{:, i} \) is proportional to the largest right-singular vector \( \vu \), which can be estimated as:
\begin{equation}
  \vu \leftarrow \frac{\mW_{:, i}}{\|\mW_{:, i}\|_2}.
\end{equation}
where \( i \) is any column index such that \( \mW_{:, i} \neq 0 \).

\paragraph{Largest eigenpair by squaring iteration}
Let \( \mW \) be a square complex matrix, \( \mW \in \mathbb{C}^{p \times p} \).
We aim to compute its largest eigenpair \( (\lambda_1, u_1) \), where \( \lambda_1 \) is the largest eigenvalue and \( u_1 \) is the corresponding eigenvector.
Algorithm~\ref{algo:gram_iteration_naive} can be modified to extract the dominant eigenpair by replacing Step 3 with squaring the matrix \( \mW \) at each iteration:
\begin{equation}
  \mW \gets \mW \mW
\end{equation}
Given the eigenvalue decomposition \( \mW = Q \Lambda Q^{-1} \), the \( t \)-th iterate satisfies
\begin{equation}
  \mW^{(t)} = \mW^{2^{t-1}} = Q \Lambda^{2^{t-1}} Q^{-1}
\end{equation}
As \( t \to \infty \), the largest eigenvalue \( \lambda_1 \) dominates (if $\lambda_1 > \lambda_2$), and \( \mW^{(t)} \) converges to a rank-one approximation aligned with the dominant eigenvector.
Since \( \mW^{(t)} \) approximates a rank-one projector, any nonzero column \( \mW_{:, i} \) is proportional to \( \vu_1 \), allowing its estimation as
\begin{equation}
  \vu_1 \gets \frac{\mW_{:, i}}{\|\mW_{:, i}\|_2}, \quad \text{where } i \text{ is any nonzero column index}
\end{equation}

\begin{definition}[Gelfand’s formula for spectral radius~\citep{horn2012matrix}]
  \label{def:gelfand_formula}
  Let \( \mW \) be a square matrix over \( \mathbb{C}^{p \times q} \) and \( \|\cdot\| \) any matrix norm. The spectral radius of \( \mW \), denoted \( \rho(\mW) \), is given by the limit
  \[
    \rho(\mW) = \lim_{t \to \infty} \|\mW^t\|^{1/t}.
  \]
  This identity, known as Gelfand’s formula, always holds in finite-dimensional spaces, regardless of the choice of norm. Moreover, if \( \|\cdot\| \) is sub-multiplicative, then the sequence \( \|\mW^t\|^{1/t} \) satisfies
  \[
    \rho(\mW) \leq \|\mW^t\|^{1/t}, \qquad \text{for all } t \in \mathbb{N},
  \]
  and thus converges to \( \rho(\mW) \) from above.
\end{definition}

Gram iteration uses Gelfand's formula~\ref{def:gelfand_formula} on Gram matrix, as $\norm{\mW}_2^2 = \rho(\mW^\star \mW)$,
and applies matrix squaring to generate a sequence of iterates that converge to the spectral norm.

\paragraph{Convergence analysis and upper bound guarantee}
To quantify the rate of convergence of the $\GI$, we recall the notions of Q-quadratic R-quadratic and super geometric convergence.
Linear convergence reduces the error proportionally, while quadratic convergence reduces the error proportionally to its square, making quadratic much faster as it approaches the limit. Super geometric convergence, on the other hand, implies that the error decays faster than any linear converging sequence, leading to even more rapid convergence but slower than quadratic.
%
\begin{definition}[Q-quadratic, R-quadratic, and supergeometric convergence]
  Suppose the sequence $(b_k)$ converges to a limit $L$.

  \begin{itemize}
    \item The sequence converges \emph{Q-quadratically} if there exists a constant $\mu>0$ such that
          \[
            \lim_{k\to\infty} \frac{|b_{k+1}-L|}{|b_k-L|^2} = \mu \, .
          \]

    \item The sequence converges \emph{R-quadratically} if there exists a nonnegative sequence
          $(\epsilon_k)$ with $|b_k-L|\le \epsilon_k$ for all $k$, and $(\epsilon_k)$ converges Q-quadratically to $0$.

    \item More generally, for $p>1$, we say $(b_k)$ converges \emph{R-superlinearly of order $p$} if there exists $K>0$ such that
          \[
            |b_{k+1}-L|\ \le\ K\,|b_k-L|^p \qquad \text{for all $k$ sufficiently large.}
          \]
          Quadratic convergence corresponds to $p=2$.

    \item Following Friedland~\citep{friedland1991revisiting}, the sequence is said to converge \emph{supergeometrically of order $p$ with rate $K\in(0,1)$} if
          for every $\varepsilon>0$ there exist constants $C>0$ and $N$ such that
          \[
            |b_k - L| \;\le\; C\,(K+\varepsilon)^{\,p^{\,k}}
            \qquad \forall k\ge N.
          \]
          This means the error decays faster than any geometric sequence $c^k$, and with a double-exponential speed in $p^k$.
  \end{itemize}
\end{definition}

%

%
These results are formalized in the following theorem.
\begin{theorem}[Gram iteration: convergence, upper bound, and supergeometric rate (Frobenius)]
  \label{thm:gram_iteration_main_result}
  Let $\|\!\cdot\!\|$ be a submultiplicative matrix norm and let $\mW\in\C^{p\times q}$.
  Define $\mW^{(1)}=\mW$ and $\mW^{(t+1)}={\mW^{(t)}}^{*}\mW^{(t)}$, and set
  \[
    s_t:=\|\mW^{(t)}\|^{\,2^{\,1-t}},\qquad t\ge 1.
  \]
  Then, for all $t\ge 2$:
  \begin{enumerate}
    \item (\emph{Convergence}) $s_t\to\sigma_1(\mW)$ as $t\to\infty$.
    \item (\emph{Upper bound}) For any matrix norm, $s_t\ge\sigma_1(\mW)$.
  \end{enumerate}
  Moreover, the following \emph{rate refinements} hold.
  \begin{enumerate}
    \item[(a)] (\emph{General norm}) In general, one has the R-linear bound
          \[
            0\le s_t-\sigma_1(\mW)\le C\,2^{-t}\quad\text{for all large }t,
          \]
          for some constant $C>0$ depending on the norm and the dimension.
    \item[(b)] (\emph{Frobenius readout: supergeometric and “order $\to 2$”})
          If the final readout is the Frobenius norm, then the error
          $e_t:=s_t-\sigma_1(\mW)$ is \emph{supergeometric} of order $2$ with rate $\mu=\sigma_2(\mW)/\sigma_1(\mW)\in(0,1)$:
          for every $\varepsilon>0$ there exist $C_\varepsilon>0$ and $T$ such that
          \[
            0\le e_t \le C_\varepsilon\,(\mu+\varepsilon)^{\,2^{t}}\qquad\forall t\ge T.
          \]
          In particular, the local (log–log) order
          \[
            p_t:=\frac{\log e_{t+1}}{\log e_t}
          \]
          satisfies $p_t=2-\Theta(t/2^{t})$; hence $p_t\uparrow 2$ and the
          convergence is \emph{superlinear} and \emph{arbitrarily close to quadratic}
          (as $t$ grows), but \emph{not} R-quadratic in general.
  \end{enumerate}
\end{theorem}

See proof in Appendix~\ref{section:appendix_gram_iteration_main_result}.
The first part of Theorem~\ref{thm:gram_iteration_main_result} was already noted
in~\citet{friedland1991revisiting}, but only for square matrices and without any
practical implementation or experimental validation (e.g.\ on GPUs).

It is also important to clarify the type of convergence obtained. In classical
numerical analysis, \emph{quadratic convergence} means that $e_{t+1}\le \mu e_t^2$
for some constant $\mu>0$, which is equivalent to the local order
$p_t=\log(e_{t+1})/\log(e_t)$ tending to $2$. In the Gram iteration with Frobenius
readout we established instead
\[
  e_t = \Theta\!\left(\frac{q^{2^t}}{2^t}\right), \qquad q=\tfrac{\sigma_2}{\sigma_1}<1,
\]
which is \emph{supergeometric} in the sense of Friedland (double–exponential decay).
The local order $p_t$ still tends to $2$, so the convergence is superlinear and
appears “almost quadratic,” but it is \emph{not} R–quadratic since
$e_{t+1}/e_t^2 \sim 2^t \to\infty$.

In practice, this provides a sharp contrast with the classical \emph{power iteration},
which converges only linearly with rate $\sigma_2/\sigma_1$. Gram iteration thus
achieves convergence that is faster than any linear rate and empirically nearly
indistinguishable from quadratic, but without the uniform guarantee of true quadratic convergence.


\paragraph{Algorithm and implementation}
We present in the following algorithm~\ref{algo:gram_iteration} a practical implementation of the Gram iteration method with Frobenius norm to have an upper bound on spectral norm at any iteration.
To avoid overflow, matrix $\mW$ is rescaled at each iteration, and scaling factors are accumulated in variable $r$ to unscale the result at the end of the method.
Unscaling is crucial as it is required to remain a strict upper bound on the spectral norm at each iteration of the method.
One can note that $\GI$ provides a proper norm at each iteration, which is a deterministic upper bound on the spectral norm, and which converges super geometrically. However, $\GI$ is not a matrix-free method, contrary to power iteration ($\PI$) and its variants.
\begin{algorithm}[h]
  \caption{Gram\_iteration$(\mW, N_\text{iter})$}\label{algo:gram_iteration_dense}
  \label{algo:gram_iteration}
  \begin{algorithmic}[1]
    \State~\textbf{Inputs}: matrix $\mW$, number of iterations $N_\text{iter}$
    \State~$r \gets 0$ \hfill \text{// Initialize rescaling}
    \For{$t = 1$ to $N_\text{iter}$}
    \State~$r \gets 2(r + \log \norm{\mW}_\frob)$ \hfill \text{// Accumulate rescaling}
    \State~$\mW \gets \mW/\norm{\mW}_\frob$ \hfill \text{// Rescale to avoid overflow}
    \State~$\mW \gets \mW^* \mW$ \hfill \text{// Gram iteration}
    \EndFor%
    \State~$\sigma_1 \gets \norm{\mW}^{2^{-N_\text{iter}}} \exp{(2^{-N_\text{iter}} r)}$
    \State~\textbf{return} $\sigma_1$
  \end{algorithmic}
\end{algorithm}

\paragraph{Explicit gradient computation}
In the context of differentiating through the spectral norm of a layer, ensuring the differentiability of the upper bound is necessary. For the Frobenius norm, an explicit gradient formulation can be derived. Interestingly, the bound sequence \( \big(\|\mW^{(t)}\|_\frob^{2^{1-t}}\big) \) aligns with the Schatten \( 2^{1-t} \)-norm of \( \mW \), providing a useful connection between norm approximations and their gradients.
\begin{proposition}[Gradient of the Gram iteration bound]
  \label{prop:gradient_gram_iteration}
  For the Frobenius norm, the bound sequence \( (\norm{\mW^{(t)}}_F^{2^{1 -t}}) \) is differentiable with respect to \( \mW \), and its gradient is given by:
  \begin{equation}
    \label{eq:gram_iteration_bound_gradient}
    \frac{\partial (\norm{\mW^{(t)}}_\frob^{2^{1 -t}})}{\partial \mW}
    =\frac{\mW {( \mW^* \mW)}^{2^{t-1} -1}}{{\norm{\mW^{(t)}}_\frob}^{2(1 - 2^{-t})}} \ .
  \end{equation}
\end{proposition}
This proposition is essential as it enables the direct computation of a gradient for optimization, allowing efficient backpropagation without the overhead of maintaining a large computational graph. However, computing this gradient involves matrix exponentiation, which can be numerically unstable. To mitigate this issue, a practical approach is to rescale by an estimate of the spectral norm, computed during the forward pass.

While our focus so far has been on linear layers, structured layers like convolutional layers are equally critical due to their widespread use in deep learning models such as convolution neural networks, particularly in computer vision~\citep{lecun1998gradient, krizhevsky2012imagenet, tan2021efficientnetv2}. Deriving spectral norm bounds for these structured layers is essential to extend the robustness and generalization guarantees of spectral regularization techniques to more complex architectures~\citep{tsuzuku2018lipschitz, sedghi2019singular}. In the following, we address this challenge and provide bounds adapted to convolutional layers.

\section{Related works on convolutional layers}\label{section:circulant_and_toeplitz_matrices}

Convolutional layers are the main building block of Convolutional Neural Networks (CNNs)~\citep{lecun1998gradient}, which have become pivotal in computer vision and achieve state-of-the-art performance~\citep{krizhevsky2012imagenet, tan2021efficientnetv2}.
The essence of CNNs lies in a cascade of convolutional layers, crucial for extracting hierarchical features.
Convolutional architecture gradually reduces the spatial dimensions of input while increasing the semantic dimension and maintaining the separability of classes~\citep{mallat2016understanding}.
CNNs are both sparse and structured: indeed the weight-sharing topology of such networks encodes inductive biases, like locality and translation-equivariance. From these biases arises a very powerful architecture for computer vision, in terms of speed, parameter efficiency, and performance.
Beyond visual tasks, convolutional layers contribute to applications like natural language and speech processing~\citep{conneau2017very, baevski20wav2vec, gu2022efficiently, li2022survey} (mainly with one-dimensional convolution), and to graph data~\citep{zhang2019graph} for applications in biology for instance.

Despite their long-standing use, CNNs are still the focus of lively research on many theoretical and empirical aspects.
For instance, the relationship between the dataset structure and CNN architectures is explored using spectral theory~\citep{pinson2023linear}.
Examining CNNs from a different perspective, some approaches investigate their properties through the Hessian rank of their matrix representation~\citep{singh2023hessian}.
Padding in convolutional layers is the subject of various studies, analyzing its impact on frequency~\citep{tang2023defects}, adversarial robustness~\citep{gavrikov2023interplay}, and encoded position information~\citep{islam2019position, kayhan2020translation}. The effects of stride on convolutional and pooling layers, particularly addressing aliasing issues, are considered and corrected within the framework of sampling theory~\citep{zhang2019making}. All these contributions showed that CNNs' layers and their design can still be improved, thanks to theoretical exploration.

\section{Related works on the spectral norm of a convolutional layer}
Let \( \mX \in \R^{\cin \times n \times n} \) be an input image, \( \vx \in \R^{\cin n^2} \) its vectorized form, and \( \tK \in \R^{\cout \times \cin \times k \times k} \) a convolution filter, and \( k \) the kernel size.
Throughout this section, unless stated otherwise, \( \tK \) is assumed to be padded to match the input size, i.e., \( \tK \in \R^{\cout \times \cin \times n \times n} \), using both zero padding and circular padding.
Applying maximum non-trivial zero padding extends the input \( \mX \in \R^{\cin \times n \times n} \) to \( \R^{\cin \times (n + k - 1) \times (n + k - 1)} \).

For \( 1 \leq j \leq \cout \) and \( 1 \leq u, v \leq n \), the convolution output \( \mY = \tK \star \mX \) is given by:
\begin{equation}
  \label{eq:convolution_product}
  \mY_{j,u,v} = \sum_{i=1}^{\cin} \sum_{k_1=1}^{n} \sum_{k_2=1}^{n} \mX_{i, u + k_1, v + k_2} \tK_{j, i, k_1, k_2}.
\end{equation}
In practice, the Im2Col~\citep{chellapilla2006high} representation is often used to express convolution as a matrix multiplication by patching the input \( \mX \) and vectorizing the filter \( \tK \), significantly improving computational efficiency and memory access patterns. However, this approach is not suited for spectral norm computation.
Instead we represent the convolution operator \( \star \) explicitly as a matrix-vector product, where \( \mW \in \R^{\cout n^2 \times \cin n^2} \) encodes the filter \( \tK \). This formulation explicitly illustrates the weight-sharing property of convolution~\citep{jain1989fundamentals}:
\begin{equation}
  \vy = \vector(\tK \star \mX) = \mW \vx,
\end{equation}
although it is highly inefficient for memory storage due to the large size of \( \mW \).




\begin{table*}[t]
  \caption{Qualitative summary of estimations of Lipschitz constant of convolutional layers. Precision and speed qualifications come from experimental results. "deter" abbreviates "deterministic". \citet{ryu2019plug} uses PI adapted to convolution to compute an estimate arbitrarily precise in a zero padding (a special case of constant padding) setting but suffers from slow convergence. \citet{araujo2021lipschitz} uses the Toeplitz matrix theory to devise a fast bound despite precision. Other approaches suppose circular padding for convolution which is a reasonable (comparable performance) and useful assumption to reduce computation complexity thanks to the Fourier transform. \citet{sedghi2019singular} provides exact computation of singular values, however, the method is very slow and does not scale. \citet{singla2021fantastic} proposes a faster using PI method to the detriment of precision. Methods using PI are not guaranteed to produce an upper bound on convolution spectral norm regarding constant or circular padding.
    Our method gathers the best of both worlds: being fast, precise, deterministic, well differentiable, and with a strict upper bound on the spectral norm.
  }
  \centering
  \vspace{0.1cm}
  \centering
  \resizebox{1.1\textwidth}{!}{ 
    \begin{tabular}{lcccccc}
      \toprule
      \textbf{Methods}                                           & \textbf{Precision} & \textbf{Speed} & \textbf{Padding} & \textbf{Upper Bound} & \textbf{Algorithm}       & \textbf{Convergence}             \\
      \midrule
      {\small \citeauthor{ryu2019plug,farnia2019generalizable}}  & ++                 & -              & zero             & \xmark               & PI: iterative, non-deter & linear                           \\
      {\small \citeauthor{sedghi2019singular,bibi2019deep}}      & ++                 & - -            & circular         & exact                & SVD: deter               & -                                \\
      {\small \citeauthor{singla2021fantastic,yi2020asymptotic}} & +                  & +              & circular         & \xmark               & PI: iterative, non-deter & linear                           \\
      {\small \citeauthor{araujo2021lipschitz}}                  & -                  & ++             & zero             & \cmark               & close-form, deter        & -                                \\
      \midrule
      \textbf{Ours}                                              & ++                 & ++             & circular         & \cmark               & GI: iterative, deter     & super geometric (near quadratic) \\
      \bottomrule
    \end{tabular}
  }
  \label{tab:review}
\end{table*}
Estimating the spectral norm of convolutional layers is challenging.
The matrix associated with a convolution has size $n^2\cin \times n^2\cout$
for an input $\vx \in \R^{\cin \times n \times n}$, making direct methods such as the $\SVD$ computationally infeasible.
The difficulty is compounded for zero-padded convolutions, where Fourier diagonalization no longer applies and no closed-form formula is available.
A first line of work develops \emph{matrix-free} estimators based on power iteration ($\PI$),
which only require applying the convolution as a matrix--vector product.
\citet{ryu2019plug} introduced a convolution-adapted $\PI$ whose runtime scales with the convolution itself
and can be efficiently accelerated on GPUs, but the linear convergence of $\PI$ limits accuracy.
\citet{singla2021fantastic} proposed a related method that computes an upper bound by taking the minimum spectral norm across four transformed matrices;
while efficient, this bound can be loose, especially for zero-padded convolutions~\citep{yi2020asymptotic}.
A second line of work exploits the structural properties of convolutions.
\citet{sedghi2019singular} represented convolutional layers as doubly-block circulant matrices,
which can be block-diagonalized with the Fourier transform;
applying $\SVD$ on the resulting blocks yields exact spectral norms under certain assumptions,
but the cost remains prohibitive for large filters.
\citet{bibi2019deep} similarly proposed exact but computationally expensive formulations.
More recent approaches aim at scalable \emph{upper bounds}.
\citet{araujo2021lipschitz} derived efficient bounds that are exact for certain filter shapes
(e.g.\ $\cout \times 1 \times k \times k$ or $1 \times \cin \times k \times k$) but do not generalize to all cases.
\citet{yi2020asymptotic} showed that such bounds remain valid even with zero padding.
However, methods relying on $\PI$~\citep{ryu2019plug, singla2021fantastic} do not guarantee upper bounds,
which restricts their applicability for robustness certification.

While recent methods have improved scalability, they often sacrifice precision or lack formal guarantees, see Table~\ref{tab:review} for a summary.
Estimating the spectral norm of convolutional layers efficiently remains an open challenge, balancing precision and computational cost. Our work addresses this gap by leveraging the \emph{Gram iteration} approach, which provides a scalable and certified method for bounding the spectral norm of convolutional layers.

\subsection{Zero-Padded convolutions}
In typical CNNs, convolutional layers are often applied with zero padding to ensure the output $\mY$ has the same spatial dimensions as the input, assuming a stride of 1. The shared weights of convolutional filters result in a Toeplitz structure~\citep{jain1989fundamentals}. This can be expressed as:
\begin{equation}
  \vector(\mY) = \mT \vx,
\end{equation}
where $\mT \in \mathbb{R}^{\cout n^2 \times \cin n^2}$ is a $\cout \times \cin$ block matrix. Each block $\mT_{j,i}$ is a $n^2 \times n^2$ banded doubly Toeplitz matrix formed from the kernel $\tK_{j,i}$. A detailed representation of the Toeplitz matrix is provided in Appendix~\ref{section:appendix_toeplitz_matrix}.
As noted in~\citep{yi2020asymptotic}, $\mT$ is a multi-level block Toeplitz matrix with $\cout \times \cin$ entries. The Lipschitz constant of the convolutional layer is determined by the maximum singular value of $\mT$, $\sigma_1(\mT)$. Direct computation of \(\sigma_1(\mT)\) is computationally expensive; hence, Fourier transform techniques are often used to simplify the analysis of Toeplitz matrices.

\subsection{Circular-Padded convolutions}
Let us consider first the case where \( \cin= \cout = 1 \) to simplify the analysis. When circular padding is applied, the convolution matrix no longer has a Toeplitz structure but instead becomes doubly-block circulant. Specifically, \( \mC \) is a doubly-block circulant matrix, where each row block is a circular shift of the previous one, and each block itself is circulant~\citep{jain1989fundamentals}.
Denoting \( \circulant(\cdot) \) as the operator that generates a circulant matrix from a vector, the structure of \( \mC \) is entirely determined by the convolutional kernel \( \tK \). A detailed representation can be found in Appendix~\ref{section:appendix_circulant_matrix}. This matrix can be expressed as:
\begin{equation}
  \mC = \begin{pmatrix}
    \circulant(\tK_1) & \circulant(\tK_2) & \cdots            & \circulant(\tK_n) \\
    \circulant(\tK_n) & \circulant(\tK_1) & \ddots            & \vdots            \\
    \vdots            & \ddots            & \circulant(\tK_1) & \circulant(\tK_2) \\
    \circulant(\tK_2) & \cdots            & \circulant(\tK_n) & \circulant(\tK_1)
  \end{pmatrix}.
\end{equation}

For simplicity, we denote $\mC = \bcirc(\tK_1, \dots, \tK_n)$. The choice of padding significantly impacts the convolution's properties. With zero padding, the matrix is doubly-block Toeplitz, and properties such as Fourier diagonalization do not apply. In contrast, the doubly-block circulant structure introduced by circular padding enables efficient computation using Fourier transforms.
$\mC$, it can be diagonalized using the discrete Fourier transform (DFT).

\begin{definition}[Discrete Fourier Transform Matrix]
  \label{def:dft_matrix}
  Let $\omega := e^{-2\pi \ci/n}$. The DFT matrix $\mU\in\C^{n\times n}$ is
  \[
    \mU_{u,v} \;=\; \omega^{(u-1)(v-1)}, \qquad 1\le u,v\le n,
  \]
  with inverse $\mU^{-1}=\tfrac{1}{n}\mU^*$.
\end{definition}

For a 2D input $\mX$, the 2D discrete Fourier transform can be expressed as:
\begin{equation}
  \vector(\mU \mX \mU^\top) = (\mU \otimes \mU) \vx,
\end{equation}
For brevity, we denote $\mF = \mU \otimes \mU$.
%
%
\begin{theorem}[Diagonalization of Circulant Matrices~\citep{jain1989fundamentals}]
  Let $\tK \in \mathbb{R}^{n \times n}$ be a convolutional kernel and $\mC \in \mathbb{R}^{n^2 \times n^2}$ be the doubly-block circulant matrix such that $\mC = \bcirc(\tK_1, \dots, \tK_n)$. Then, $\mC$ can be diagonalized as:
  \begin{equation}
    \mC = \mF \diag(\vlambda) \mF^{-1},
  \end{equation}
  where $\vlambda = \mF \vector(\tK)$ are the eigenvalues of $\mC$.
\end{theorem}
In the context of deep learning, convolutions are applied with multiple channels: $\cout$ convolutions are applied on input with $\cin$ channels.
Therefore, the matrix associated with the convolutional filter $\tK \in \R^{\cout \times \cin \times n \times n}$ becomes a block matrix, where each block is a doubly-block circulant matrix: $\mC = (\mC_{i,j})$, for $1 \leq i \leq \cout$ and $1 \leq j \leq \cin$, with $\mC_{i,j} = \bcirc(\tK_{i, j, 1}, \dots, \tK_{i, j, n})$.
As also studied by~\citet{sedghi2019singular} and~\citet{yi2020asymptotic},
this type of block doubly-block circulant matrix can be \emph{block} diagonalized as follows:

\begin{theorem}[Block diagonalization of multi-channel circular convolution~{\citep[Cor.~A.1.1]{trockman2021orthogonalizing}}]
  \label{theorem:block_diagonalization}
  Let $\tK \in \R^{\cout \times \cin \times n \times n}$ be a circular convolutional filter
  and let $\mC \in \R^{\cout n^2 \times \cin n^2}$ denote the corresponding convolution matrix.
  Let
  $\mP_\mathrm{out} \in \R^{\cout n^2 \times \cout n^2}$ and
  $\mP_\mathrm{in} \in \R^{\cin n^2 \times \cin n^2}$ be permutation matrices,
  and let $\mF = \mU \otimes \mU$ be the 2D DFT matrix.
  Then $\mC$ admits the block diagonalization
  \begin{equation}
    \mC \;=\; (\mI_{\cout} \otimes \mF)\;\mP_\mathrm{out}\;\mD\;\mP_\mathrm{in}^\top\;(\mI_{\cin} \otimes \mF^{-1}),
  \end{equation}
  where $\mI_{\cout}$ and $\mI_{\cin}$ are identity matrices of size $\cout$ and $\cin$, respectively,
  and $\mD$ is block diagonal with $n^2$ diagonal blocks of size $\cout \times \cin$.

  Each block of $\mD$ corresponds to one spatial frequency and collects the Fourier
  coefficients of the filters across channels. Explicitly,
  \begin{equation*}
    \mD \;=\; \mP_\mathrm{out}^\top
    \begin{pmatrix}
      \mF\,\vector(\tK_{1,1})     & \cdots & \mF\,\vector(\tK_{1,\cin})     \\
      \vdots                      &        & \vdots                         \\
      \mF\,\vector(\tK_{\cout,1}) & \cdots & \mF\,\vector(\tK_{\cout,\cin})
    \end{pmatrix}
    \mP_\mathrm{in}.
  \end{equation*}
\end{theorem}

\paragraph{Explicit permutations $\mP_\mathrm{out}$ and $\mP_\mathrm{in}$}
Let $\ve_k$ denote the $k$-th canonical basis vector.
We index outputs by $(i,f)$ with $1\le i\le\cout$ (channel) and $1\le f\le n^2$ (frequency),
so that the row index is $(i-1)n^2+f$.
Similarly, inputs are indexed by $(j,f)$ with $1\le j\le\cin$, giving column index $(j-1)n^2+f$.
This corresponds to the usual \emph{channel-major} ordering.
The permutation matrices $\mP_{\mathrm{out}}\in\R^{\cout n^2\times \cout n^2}$ and
$\mP_{\mathrm{in}}\in\R^{\cin n^2\times \cin n^2}$ reorder these indices as
\[
  (\mP_{\mathrm{out}})_{(f-1)\cout+i,\,(i-1)n^2+f}=1,\qquad
  (\mP_{\mathrm{in}})_{(f-1)\cin+j,\,(j-1)n^2+f}=1,
\]
with all other entries $0$. Equivalently,
\[
  \mP_{\mathrm{out}}\ve_{(i-1)n^2+f}=\ve_{(f-1)\cout+i},\qquad
  \mP_{\mathrm{in}}\ve_{(j-1)n^2+f}=\ve_{(f-1)\cin+j}.
\]
Thus $\mP_{\mathrm{out}}$ and $\mP_{\mathrm{in}}$ convert vectors from \emph{channel-major} to
\emph{frequency-major} order, with their transposes performing the inverse.

\begin{figure}[t]
  \centering
  \begin{subfigure}[b]{0.45\textwidth}
    \centering
    \includegraphics[width=\textwidth]{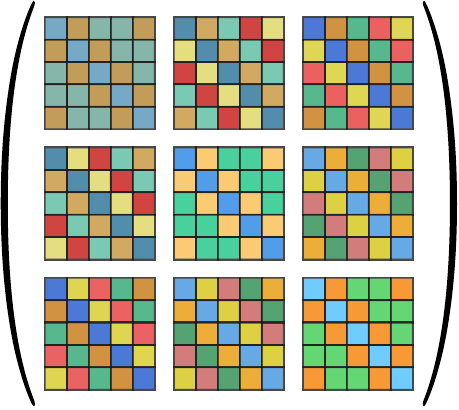}
    \caption{Matrix \( \mA \)}
    \label{fig:matrix_w}
  \end{subfigure}
  \hfill
  \begin{subfigure}[b]{0.45\textwidth}
    \centering
    \includegraphics[width=\textwidth]{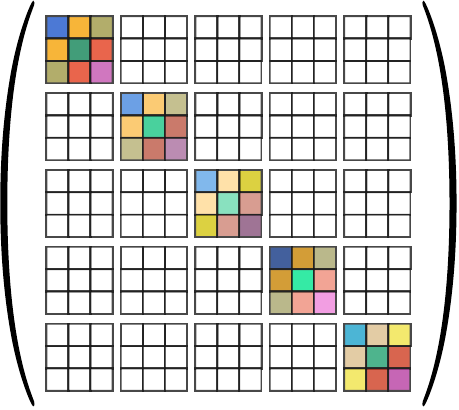}
    \caption{Matrix \( \mB \)}
    \label{fig:matrix_d}
  \end{subfigure}
  \caption{Illustration of permutation described in~\citep{henderson1981TheVM}. The matrix \(\mA\), with blocks where each block has constant diagonal and sub-diagonal values, can be reshaped into a block diagonal form \(\mB\) using permutation matrices $\mP_\mathrm{in}$ and $\mP_\mathrm{out}$, preserving singular values. The figure is inspired by~\citet{gnassounou2024multi}.}
  \label{fig:matrix_comparison}
\end{figure}

In the block-diagonalization of convolutional layers,
$\mP_\mathrm{out}$ and $\mP_\mathrm{in}$ simply reshuffle the matrix $\mD$
so that it becomes block diagonal. This corresponds to adopting an alternative
vectorization of the layer input~\citep{sedghi2019singular,yi2020asymptotic} while preserving
the singular values~\citep{henderson1981TheVM}; see Fig.~\ref{fig:matrix_comparison}.
As a result, the largest singular value $\sigma_1(\mC)$ can be obtained directly
from the block-diagonal matrix $\mD$.




%
Let ${(\mD_i)}_{i = 1, \dots, n^2}$ be the diagonal blocks of the matrix $\mD$, then:
\begin{equation}\label{equation:lip_conv_circu_expression}
  \sigma_1(\mC) = \sigma_1(\mD) = \max_i \sigma_1(\mD_i) \ .
\end{equation}
It requires calculating the spectral norm of each $\mD_i$, one could use $\SVD$ as in~\citet{sedghi2019singular} or $\PI$ as in~\citet{yi2020asymptotic}, however, these algorithms do not scale up nor are efficient. In the following section, we present a more precise, fast, and scalable approach based on previously introduced Gram iteration.

\section{Spectral norm of circular padding convolutional layers}
\label{ssec:spectral_norm_conv}
The spectral norm of circular convolutional layers can be estimated using $\GI$ with Equation~(\ref{equation:lip_conv_circu_expression}). The method applies Theorem~\ref{thm:gram_iteration_main_result} to sequences \((\mD_i^{(t)})\), derived via a 2D Fast Fourier Transform (FFT2) of the convolutional filter \(\tK\).
\begin{proposition}\label{prop:upper_bound_conv}
  The spectral norm of a circular convolutional layer is upper-bounded as:
  \begin{equation}\label{equation:bound_conv_gram}
    \sigma_1(\mC) \leq \max_{1 \leq i \leq n^2} \norm{\mD^{(t)}_{i}}_\frob^{2^{1-t}}  \underset{t \to +\infty}{\longrightarrow} \sigma_1(\mC) \ .
  \end{equation}
\end{proposition}
Here, \(\norm{\mD_i^{(t)}}_\frob^{2^{1-t}} \) is the $2^{t}$-Schatten norm, which is convex. The maximum over these norms defines a convex and subdifferentiable upper bound.
\begin{algorithm}[h]
  \caption{: norm2\_circ$(\tK, N_\text{iter})$}
  \label{algo:gram_iteration_conv}
  \begin{algorithmic}[1]
    \State \textbf{Inputs}: Filter \(\tK\), iterations \(N_\text{iter}\)
    \State \(\mD \gets \texttt{FFT2}(\tK)\) \hfill \Comment{Apply 2D FFT to the filter}
    \State \(r \gets 0_{n^2}\)  \Comment{Initialize rescaling factors}
    \For{$1 \ldots N_\text{iter}$}
    \For{$i = 1 \ldots n^2$ } \Comment{Parallel execution}
    \State \(r_i \gets 2 (r_i + \log\norm{\mD_i}_\frob)\)
    \State \(\mD_i \gets \mD_i / \norm{\mD_i}_\frob\) \Comment{Rescale to prevent overflow}
    \State \(\mD_i \gets \mD_i^* \mD_i\) \Comment{Gram iteration step}
    \EndFor
    \EndFor
    \State \(\sigma_1 \gets \max_i \{ \norm{\mD_i}_\frob^{2^{-N_\text{iter}}} \exp(2^{-N_\text{iter}} r_i) \}\)
    \State \textbf{return} \(\sigma_1\)
  \end{algorithmic}
\end{algorithm}
Using Proposition~\ref{prop:upper_bound_conv}, Algorithm~\ref{algo:gram_iteration_conv} computes the spectral norm of a circular convolutional layer represented by \(\tK\). The method is deterministic and precise due to the properties of Gram iteration and scales efficiently by handling \(n^2\) small matrices of size \(\cout \times \cin\) in parallel. For large input sizes, using a reduced spatial dimension \(n_0 \leq n\) can further reduce computation while maintaining an upper bound, as detailed in Appendix~\ref{section:approx_n0}. For the backward computation, we use the previous Proposition~\ref{prop:gradient_gram_iteration} on each bound derived from the $\mD_i$ for each $i$.
The 2D FFT step is quite efficient, with complexity \(O(n \log_2 n)\), making the method practical for small to moderate \(n\). Approximations between circular and zero padding are discussed in Appendix~\ref{section:approx_circ}.
In the following section, we present an application of $\GI$ to estimate directly the spectral norm of convolutional layers with zero padding.

\section{Spectral norm of zero padding convolutional layers}
In practical implementations, zero padding is commonly used to preserve the spatial dimensions of the input and output. It is the default padding in most deep learning frameworks and convolutional neural network (CNN) architectures.
For a convolutional layer with zero padding, the operation can be represented by a banded doubly block Toeplitz matrix \( \mT \), formed by concatenating structured blocks. This matrix, of size \( n^2 \cout \times n^2 \cin \), is too large for direct spectral norm computation.
Most approaches to spectral norm estimation in convolutional layers assume circular padding due to the convenient properties of the Fourier transform, which simplifies the computation. However, the zero-padding case is more challenging, as it does not share the same spectral properties.

To handle the large matrix $\mT$, we follow the convolutional operator approach of~\citet{wang2020orthogonal} and~\citet[Appendix~A.1]{prach2022almost}.
In particular,~\citet{prach2022almost} introduce a rescaling along the output channel dimension $j$ in order to obtain Lipschitz convolutional layers:
\begin{equation*}
  r_j \;=\; \sum_{i_1, k', l'}
  \left| \sum_{j=1}^{\cout} \tK_{j, i_1} \star \tK_{j, i_2} \right|_{k', l'} \, .
\end{equation*}
The rescaled kernel is then defined as
\[
  \tK_{\text{rescaled}} \;=\; \tK \star \tR,
  \qquad \tR \in \R^{1 \times 1 \times \cout \times \cout},
  \quad \tR_{1,1,j,j} = r_j.
\]
This rescaling can be interpreted as performing a single Gram iteration on the filter $\tK$.
We propose to generalize this idea to an arbitrary number of iterations, thereby establishing a new bound on the spectral norm of $\mT$.

\begin{theorem}[Bound on the spectral norm of zero-padded convolutional layers]\label{thm:bound_spectral_norm_toeplitz}
  For a zero-padded convolutional layer $\mT$ parametrized by a filter $\tK \in \mathbb{R}^{\cout \times \cin \times k \times k}$,
  let us note the Gram iterate for matrix $\mT$ as $\mT^{(t)}$.
  We define the Gram iterate for filter as $\tK^{(t+1)}_{i_1, i_2} = \sum_{j=1}^\cout \tK^{(t)}_{j, i_1} \star \tK^{(t)}_{j, i_2}$, with convolution done with maximal non-trivial zero padding, with $\tK^{(1)} = \tK$.
  \\
  For norm $||~\cdot~||_{\infty}$, we have the following bound:
  \begin{align*}
    \norm{\mT^{(t+1)}}_\infty^{2^{-{t}}}
    \leq
    \left( \max_{i_2} \sum_{i_1, k^\prime, l^\prime} \left| \sum_{j=1}^\cout   \tK^{(t)}_{j, i_1} \star \tK^{(t)}_{j, i_2} \right|_{k^\prime, l^\prime}
    \right)^{2^{-{t}}} \ .
  \end{align*}
  For norm $||~\cdot~||_\frob$, we have:
  \begin{align*}
    \norm{\mT^{(t+1)}}_\frob^{2^{-{t}}}
    \leq \left(
    k^2 \sum_{i_1, i_2, k^\prime, l^\prime} \left| \sum_{j=1}^\cout \tK^{(t)}_{j, i_1} \star \tK^{(t)}_{j, i_2} \right|_{k^\prime, l^\prime}^2
    \right)^{^{2^{-(t+1)}}} ,
  \end{align*}
  and $\sigma_1(\mT) \leq || \mT^{(t)} ||^{2^{-t}} \underset{t \to \infty}{\longrightarrow} \sigma_1(\mT)$ with super geometric convergence.
\end{theorem}
The maximum non-trivial padding size is the maximum padding size, $k-1$, where the convolution kernel still interacts with the original input regions, ensuring it does not interact exclusively with zeros from the padding.
Note that this bound also holds for the case of circular padding, see proof in Appendix.
Pairing with Equation~(\ref{eq:toeplitz_gram_conv_relation}), we have  Algo.~\ref{algo:gram_iteration_toep} for the choice of matrix norm $||~.~||_{\infty}$.

\begin{algorithm}[h]
  \caption{: norm2\_toep$(\tK, N_\text{iter})$}\label{algo:gram_iteration_toep}
  \begin{algorithmic}[1]
    \State~\textbf{Inputs} filter: $K$, number of iterations: $N_\text{iter}$
    \State~$r \gets 0$
    \State~\textbf{for} $1 \ldots N_\text{iter}$
    \State~\quad  $r \gets 2
      (r + \log\norm{\tK}_F)$
    \State~\quad  $\tK \gets \tK /\norm{\tK}_F ~ $
    \State~\quad  $\tK \gets \mathrm{conv2d}(\tK, \tK, \mathrm{padding}=\mathrm{kernel\_size}(\tK)-1)$
    \State~$\sigma_1 \gets \max_i \left\{ \left(\sum_{j, k_1, k_2} |\tK_{j, i, k_1, k_2}|\right)^{2^{-N_\text{iter}}} \exp{(2^{-N_\text{iter}} r)} \right\}$
    \State \textbf{return} $\sigma_1$
  \end{algorithmic}
\end{algorithm}
The bound may not be tight, as the use of absolute values in the summations discards potential cancellation effects between filter coefficients. This yields a conservative estimate of the spectral norm, although the Gram iteration ensures fast convergence of the bound.

The method is referred to as \texttt{norm2\_toep} because it provides an upper bound on the spectral norm of convolutional layers with zero-padding, which are classically represented as Toeplitz matrices. However, it also applies to circular convolutions, where the associated matrices are circulant.

Our proposed method leverages the same principles as power iteration \citep{ryu2019plug} but with significant advantages for convolutional layers. Unlike power iteration, it is deterministic and provides a quadratically convergent certified upper bound on the spectral norm at any iteration, whereas power iteration does not guarantee this property. Specifically adapted for convolutional layers, it utilizes efficient \texttt{conv2d} operations with GPU acceleration, making it highly practical for large-scale applications.

In the following, we study the relation between the spectral norm of convolutional layers with circular and zero padding.


\section{Bound approximations between input sizes and padding types}\label{section:approx}
This section presents theoretical and practical approximations of spectral-norm bounds for convolutional layers that remain accurate while reducing cost. Exact computation scales poorly with the spatial size $n$: with circular padding it requires evaluating $n^2$ spectral norms of the $\cin \times \cout$ matrices $\mD_i$, and with zero padding and large kernels the Gram-iteration approach (Alg.~\ref{algo:gram_iteration_toep}) is likewise expensive.

To address these challenges, we propose an approximation that extends the computationally efficient spectral norm bounds for circular padding to the zero-padding case. While circular padding, modeled by circulant matrices, facilitates spectral norm computations due to its connection to Fourier transforms, zero padding, modeled by Toeplitz matrices, is more commonly used in practice. Section~\ref{section:approx_circ} and Theorem~\ref{thm:bound_circ_toep} establish how the spectral norm bound computed under circular padding can approximate the bound for zero padding, leveraging the structure of circulant matrices to simplify computation.
In particular, for zero padding, spectral norm computations scale with the kernel size but remain independent of the input size, making this approximation particularly relevant for the large kernel size.
Another consideration arises when dealing with large input sizes. Circular padding bounds are computationally efficient with respect to kernel size but become expensive as the input size \( n \) increases. Section~\ref{section:approx_n0} and Theorem~\ref{thm:bound_approximation_for_lower_input_size} examine how spectral norm bounds computed for a smaller spatial dimension \( n_0 \leq n \) can serve as a reliable approximation for larger inputs, significantly reducing computational costs.

This section presents theoretical results and practical methods for approximating spectral norm bounds in these scenarios, ensuring efficient and scalable computations for convolutional layers.

\subsection{From circular to zero padding}
\label{section:approx_circ}
In CNN architectures, zero-padded convolutions are more common than circular ones. We want to use the circular convolution theory to obtain a bound on the spectral norm of zero padding convolution.

Considering the block diagonal matrix \( \mD_i \) built from the filter \( \tK \), we re-index \( (\mD_i)_{i = 1, \dots, n^2} \) as \( (\mD_{u,v})_{1 \leq u,v \leq n} \), where \( \mC \) denotes the associated matrix representing the circular convolutional layer.
Using the discrete Fourier transform (DFT) matrix \( \mU \in \mathbb{C}^{n \times n} \), the expression for \( \mD_{u,v} \) is given by:
\begin{align*}
  \mD_{u,v} & = \sum_{k_1=0}^{n-1} \sum_{k_2=0}^{n-1} \mU_{k_1, u} \ \tK_{:,:,k_1, k_2} \ \mU_{k_2, v}
            & = \sum_{k_1=0}^{k-1} \sum_{k_2=0}^{k-1} e^{-\frac{2\pi\ci k_1 u}{n}} \ e^{-\frac{2\pi\ci k_2 v}{n}} \ \tK_{:,:,k_1, k_2} \ ,
\end{align*}
where \( \tK \) is zero-padded.
Similarly, we define \( \mT \) as the matrix constructed from the filter \( \tK \), representing the zero-padded convolutional layer. We recall the bound on spectral norm  produced with Gram iteration
from Theorem~\ref{thm:gram_iteration_main_result}:
\begin{equation*}
  \sigma_1(\mC)
  = \max_{1 \leq u,v\leq n} \sigma_1(\mD_{u,v})
  \leq \max_{1 \leq u,v \leq n} \norm{{\mD}^{(t)}_{u,v}}_\frob^{2^{1-t}} \ .
\end{equation*}
The following theorem establishes a bound approximation between circular and zero padding convolutional layers.
\begin{theorem}[Bound approximation for zero padding]
  \label{thm:bound_circ_toep}
  For given \(t\) and kernel size \(k\), let
  \[
    n \ge 2^t\bigl\lfloor\tfrac{k}{2}\bigr\rfloor + 1,
    \quad
    \alpha = \frac{2^t\lfloor k/2\rfloor}{n}.
  \]
  Then the spectral norm of the zero-padded convolutional layer can be bounded by:
  \begin{equation*}
    \sigma_1(\mT) \leq \left(\frac{1}{1 - \alpha}\right)^{2^{-t}} \max_{1 \leq u,v \leq n} \norm{{\mD}^{(t)}_{u,v}}_\frob^{2^{1-t}} \ .
  \end{equation*}
\end{theorem}
See proof in Appendix~\ref{appendix:proof_bound_circ_toep}.
This result can be used when computing the bound on the spectral norm of convolution layers with circular padding is more convenient than zero padding, namely because circulant matrix theory allows the use of Fourier transform to ease computation. It can be crucial in the case of large kernel size $k$.

Orthogonal convolutional operators~\citep{li2019preventing,trockman2021orthogonalizing,singla2021skew,yu2022constructing} allow for spectral norm control, particularly in the case of circular padding.
\begin{proposition}[Contractant property of orthogonal convolutional layers with zero padding]
  \label{prop:orthogonal_circular_padding}
  For orthogonal convolutional layers with circular padding and filter $\tK$, we have:
  \begin{align*}
    \sigma_1(\mT) \leq \sigma_1(\mC) = 1 \ ,
  \end{align*}
  where \( \sigma_1(\cdot) \) denotes the spectral norm of the corresponding matrix.
\end{proposition}
See proof in Appendix~\ref{appendix:proof_orthogonal_circular_padding}.
This property extends to zero-padding convolutional layers by leveraging the relationship between their spectral norms. Specifically, contractive zero-padding convolutional layers can be derived from orthogonal circular padding layers, ensuring \( \sigma_1(\cdot) \leq 1 \). By applying orthogonalization techniques the stability and 1-Lipschitz constraint of circular padding convolutional layers can be transferred to the zero-padding setting, facilitating the construction of orthogonal neural networks in the standard zero-padding setting.

\subsection{Lower input size approximation}
\label{section:approx_n0}
We distinguish between the multichannel filter padded to match the input size \( n \), denoted as \( \tK \in \R^{\cout \times \cin \times n \times n} \), and the filter padded to a reduced input size \( n_0 \), where \( k \ll n_0 \leq n \), denoted as \( \tKarrow \in \R^{\cout \times \cin \times n_0 \times n_0} \).
In the same manner, we define \( (\mD_{u,v}^\downarrow)_{1 \leq u,v \leq n} \) as the block diagonal matrices constructed from \( \tKarrow \), with the corresponding circular convolution matrix denoted as \( \mC^\downarrow \).
A similar expression holds for \( \mD_{u,v}^\downarrow \):
\begin{align*}
  \mD_{u,v}^\downarrow = \sum_{k_1=0}^{k-1} \sum_{k_2=0}^{k-1} e^{-\frac{2\pi\ci k_1 u}{n_0}} \ e^{-\frac{2\pi\ci k_2 v}{n_0}}
  \tK_{:,:,k_1, k_2} \ .
\end{align*}
Similarly, the matrix \( \mTarrow \) is constructed from \( \tKarrow \), representing the zero-padded convolutional layer at spatial resolution \( n_0 \).

To consider a very large input spatial size $n$, we can use our bound by approximating the spatial size for $n_0 \leq n$. It means we pad the kernel $\tK$ to match the spatial dimension $n_0 \times n_0$, instead of $n \times n$. To compensate for the error committed by the sub-sampling approximation, we multiply the bound by a factor $\alpha$.
The work of~\citep{pfister2019bounding} analyzes the quality of approximation depending on $n_0$ and gives an expression for factor $\alpha$ to compensate and ensure to remain an upper-bound, as studied in~\citep{araujo2021lipschitz}.

We can estimate the spectral norm bound for a convolution defined by kernel $\tK$ and input size $n$ with the spectral norm bound for a lower input size $n_0 \leq n$.
\begin{theorem}[Bound approximation for lower input size]\label{thm:bound_approximation_for_lower_input_size}
  For given \(t\) and kernel size \(k\), let
  \[
    n_0 \ge 2^t\bigl\lfloor\tfrac{k}{2}\bigr\rfloor + 1,
    \quad
    \alpha = \frac{2^t\lfloor k/2\rfloor}{n_0}.
  \]
  the spectral norm of the circular convolutional layer can be bounded by:
  \begin{equation*}
    \sigma_1(\mC) \leq \left(\frac{1}{1 - \alpha}\right)^{2^{-t}}  \max_{1 \leq u,v \leq n_0} \norm{{\mD^\downarrow}^{(t)}_{u,v}}_\frob^{2^{1-t}} \ .
  \end{equation*}
  Same bound holds for zero padding convolutional layers:
  \begin{equation*}
    \sigma_1(\mT) \leq \left(\frac{1}{1 - \alpha}\right)^{2^{-t}}  \max_{1 \leq u,v \leq n_0} \norm{{\mD^\downarrow}^{(t)}_{u,v}}_\frob^{2^{1-t}}  \ .
  \end{equation*}
\end{theorem}

\begin{figure}[t]
  \centering
  \includegraphics[width=0.5\linewidth]{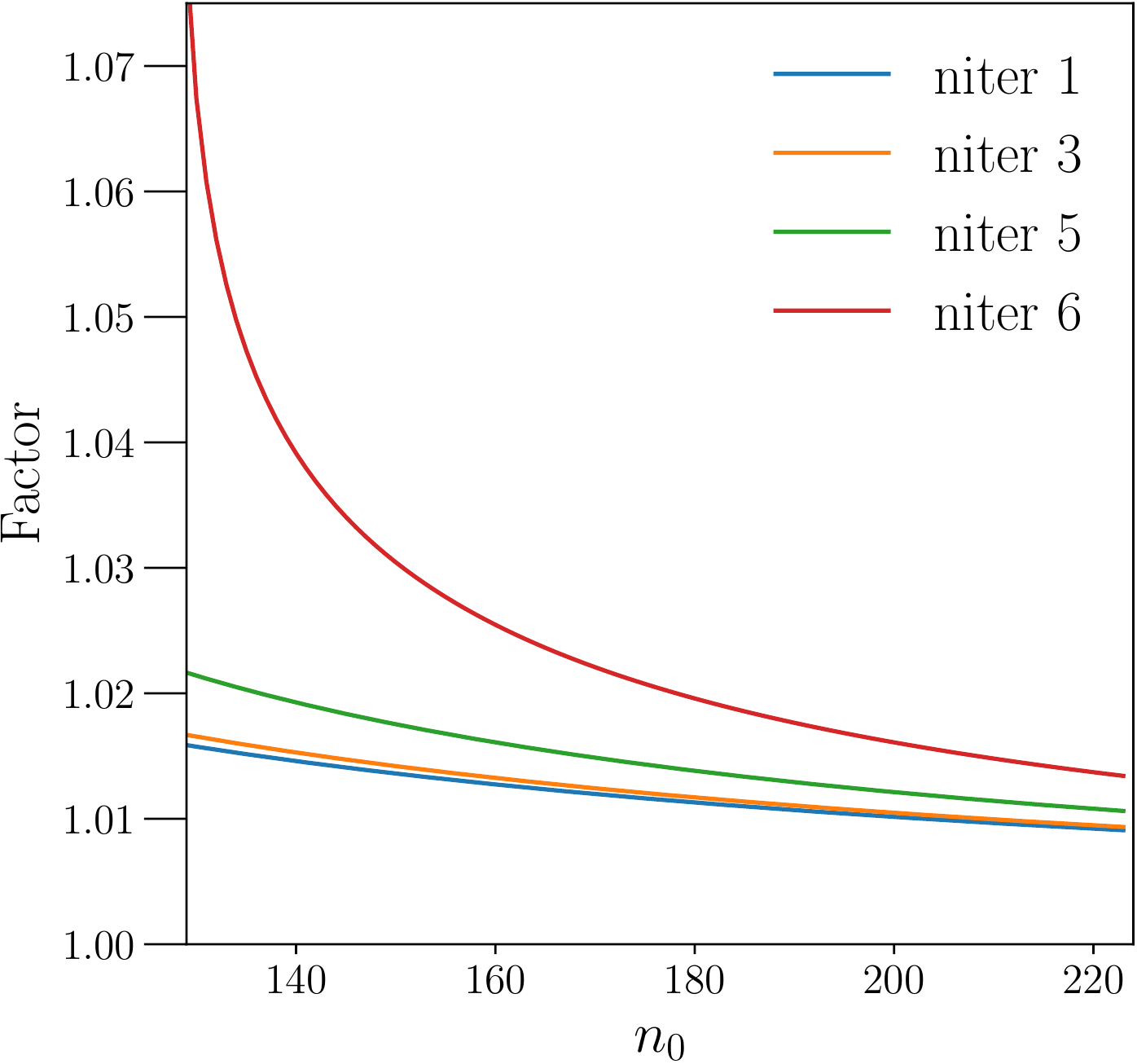}
  \caption{Evolution of bound factor $(1/(1 - \alpha))^{2^{-t}}$, described in Theorem~\ref{thm:bound_approximation_for_lower_input_size}, for input size $n=224$, kernel size $k=3$ and number of Gram iterations  $t \in \{1, 3, 5, 6 \}$. }
  \label{fig:alpha_factor_evolution}
\end{figure}
We illustrate the evolution of the factor term $\left(\frac{1}{1 - \alpha}\right)^{2^{-t}}$ in Figure~\ref{fig:alpha_factor_evolution} for different $n_0$ and for different number of Gram iteration $t$.
We see that the more the number of Gram iterations $t$ increases the larger the factor, however, the estimation of the spectral norm is also much more accurate (as the bound is tighter).
For a very large image size $n$, this theorem allows the computation of a tight bound on the spectral norm of a convolutional layer at a low computational cost w.r.t input spatial size.

\section{The case of strided convolutions}

Convolutional layers with stride $s > 1$ induce a downsampling effect that reduces the output resolution and modifies the associated linear operator. Let $\mW_{\tK, s}$ denote the operator corresponding to a convolution with kernel $\tK$ and stride $s$. When $s = k$ (i.e., non-overlapping convolutions), the spectral norm of the convolution is equal to that of the kernel:
\[
  \|\mW_{\tK, s}\|_2 = \|\tK\|_2.
\]
In general, the spectral norm of a strided convolution can be upper bounded as
\[
  \|\mW_{\tK, s}\|_2 \leq \|\text{DownSampling}_s\|_2 \cdot \|\mW_{\tK, 1}\|_2,
\]
where $\text{DownSampling}_s$ is the operator that selects one entry every $s$ positions. Since this operator is contractive, we have $\|\text{DownSampling}_s\|_2 \leq 1$, implying that stride reduces the spectral norm.

In practice, estimating the spectral norm of $\mW_{\tK, s}$ can be done directly by applying power iteration to the convolution operator with stride $s$, i.e., using \texttt{conv2d} with \texttt{stride=s}. For a more refined analysis, one can adapt the Gram iteration procedure to take into account the downsampling effect, as proposed by~\citet{senderovich2022practical}.

\section{Experiments}\label{section:experiments}
For research reproducibility, the code of experiments is available at
\url{https://github.com/blaisedelattre/lip4conv}.
Experiments were done on one NVIDIA RTX A6000 GPU.

\subsection{Spectral norm computation for dense layers}
\label{subsection:expe_comput_spectral_norm}

This experiment aims to compare the convergence behavior of Power iteration ($\PI$) and Gram iteration ($\GI$) in computing the spectral norm of dense layers, specifically focusing on matrices.

\begin{figure}[h]
  \centering
  \includegraphics[scale=0.55]{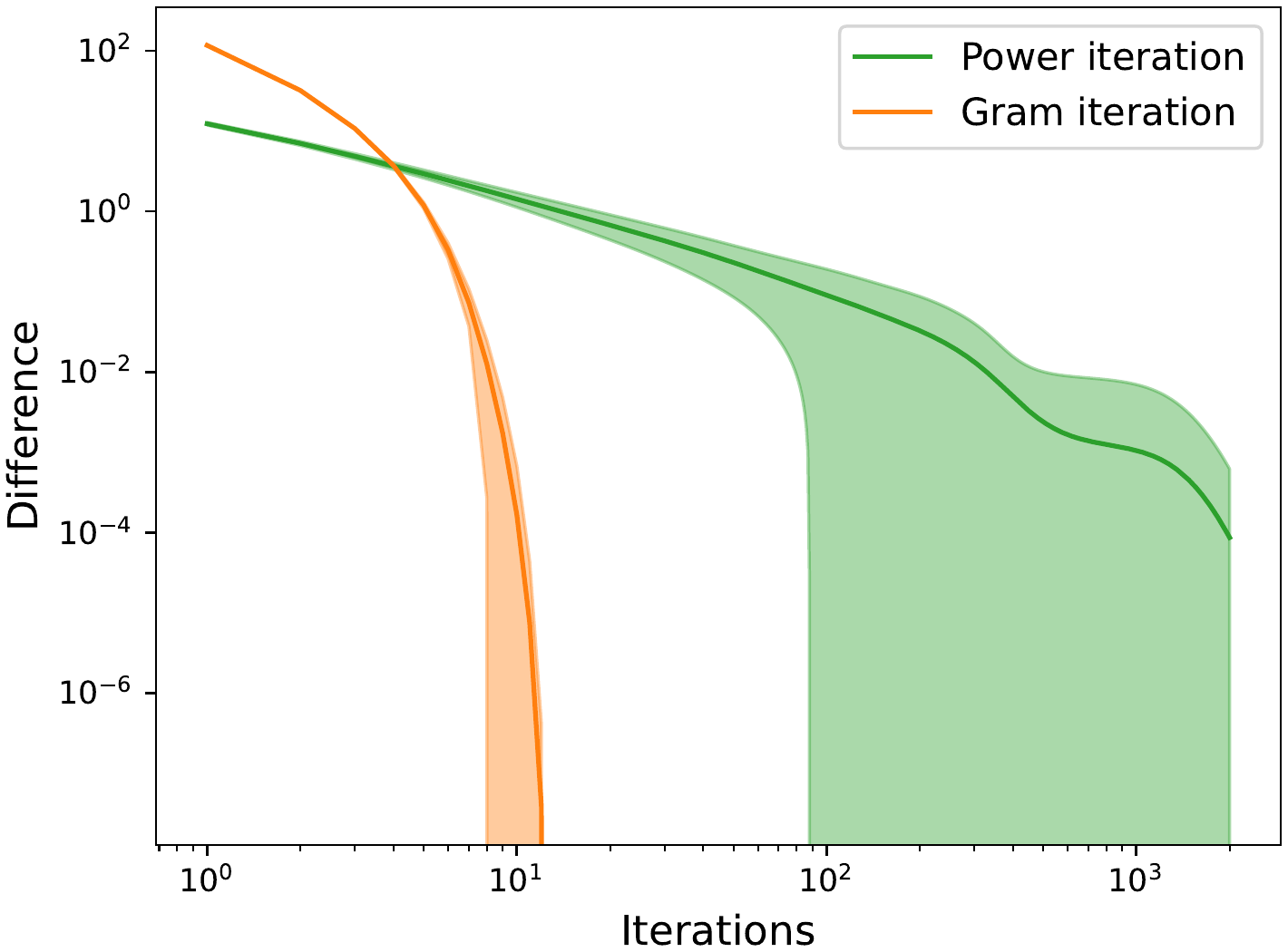}
  \caption{Convergence plot in log-log scale for spectral norm computation, comparing Power iteration and Gram iteration, one standard deviation shell is represented in a light color.
  Difference at each iteration is defined as $| {\sigma_1}_{\text{method}} - {\sigma_1}_{\text{ref}} |$.
  Gram iteration converges to numerical 0 in less than 12 iterations while Power iteration has not yet converged with 2,000  iterations and has a larger dispersion through runs.}
  \label{fig:convergence_plot_gram_vs_power_itreration}
\end{figure}

We compare $\PI$ and $\GI$ methods for spectral norms computation, and a set of $100$ random Gaussian matrices $2000 \times 1000$ is generated.
The reference value for the spectral norm of a matrix ${\sigma_1}_{\text{ref}}$ is obtained from PyTorch using SVD, and Algorithm~\ref{algo:gram_iteration_dense} is used for $\GI$.
Estimation error is defined as $| {\sigma_1}_{\text{method}} - {\sigma_1}_{\text{ref}} |$.
We observe in Figure~\ref{fig:convergence_plot_gram_vs_power_itreration} that PI needs more than 2,000 iterations to fully converge to numerical 0 and at minimum $90$ iterations. Runs have a much larger standard deviation in comparison to $\GI$. For that experiment, full convergence required up to 5,000 iterations for some realizations.
This highlights how PI convergence can be very slow from one run to another for different generated matrices.
Furthermore, in Figure~\ref{fig:convergence_plot_gram_vs_power_itreration_same_mat} in Appendix we show convergence of methods when a generated matrix is fixed: variance PI observed for PI is only due to its non-deterministic behavior, unlike GI which is deterministic.
This shows how the variance of PI is sensitive due to two causes: random generation of input matrices and the method in itself with random vector initialization at the start of the algorithm. On the contrary, $\GI$ converges under 15 iterations in every case.

\begin{figure}[h]
  \centering
  \includegraphics[scale=0.8]{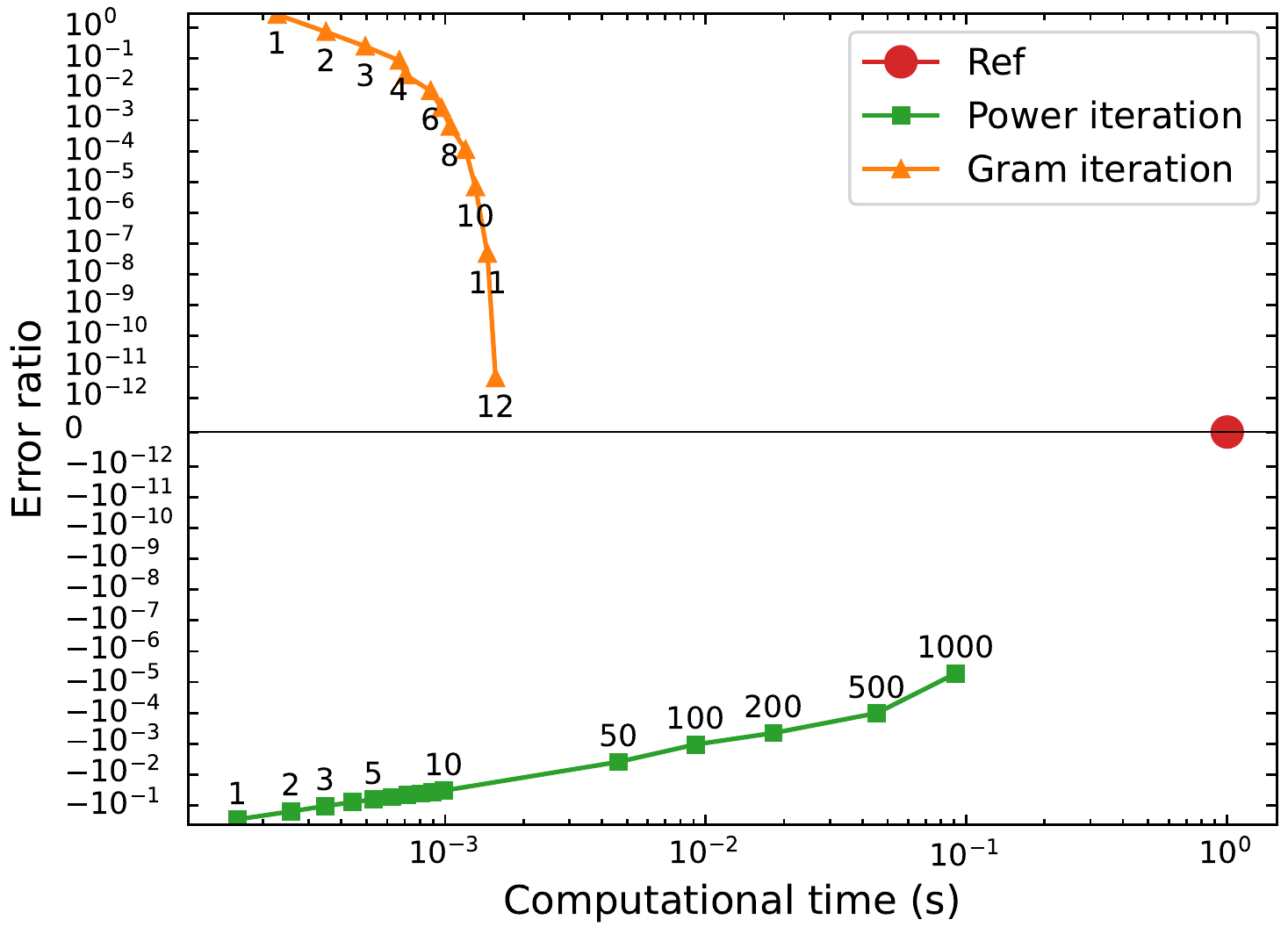}
  \caption{Error ratios over computational times of methods for spectral norm computation. Error ratio is defined as ${\sigma_1}_{\text{method}} / {\sigma_1}_{\text{ref}} - 1$, PyTorch $\SVD$ is taken as a reference. Points are annotated with the number of iterations. $\GI$ converges in $\pm 10^{-3}$s to numerical 0, while PI convergence is slower.}
  \label{fig:comparaison_methods_compute_spectral_norm_matrix}
\end{figure}

To capture the sign of the error, \ie if the current estimator is an upper bound or not, we define the error ratio as ${\sigma_1}_{\text{method}} / {\sigma_1}_{\text{ref}} - 1$, depicted in
Figure~\ref{fig:comparaison_methods_compute_spectral_norm_matrix} also reported in Table~\ref{tab:comparaison_methods_compute_spectral_norm_matrix}.
We observe that the error ratio for PI is negative at each iteration, meaning that PI estimation approaches the reference while being inferior to it, thus being a lower bound.
$\GI$ empirically enjoys a very fast convergence, with $10$ iterations: doing one more iteration divide the error ratio by approximately $1e2$ and two more iterations by $1e4$. In contrast, PI has a much slower convergence: passing from $10$ to $100$ iterations only divides the ratio by a factor of $10$.
This faster convergence rate from $\GI$ makes the method much faster in terms of computational time than PI and SVD for the same precision.
Those results motivate the use of $\GI$ for dense layers as well.

\begin{figure}[h]
  \centering
  \includegraphics[width=0.55\textwidth]{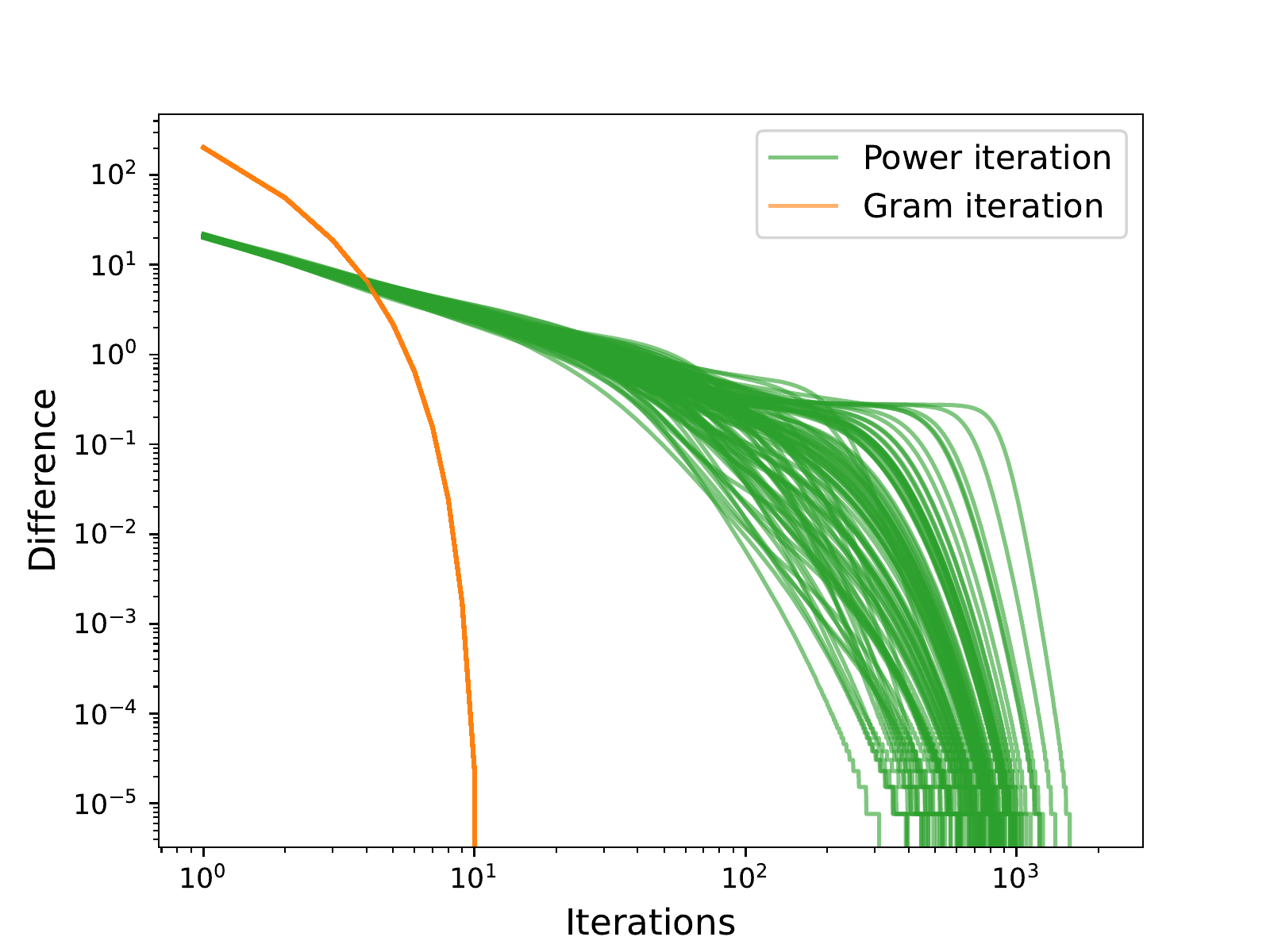}
  \caption{
  Convergence plot in log-log scale for spectral norm computation, comparing power iteration and Gram iteration, the same matrix is used here, each line corresponds to one run of Power iteration and Gram iteration.
  The difference is defined as
  $|{\sigma_1}_{\text{method}} - {\sigma_1}_{\text{ref}}|$ where reference is taken from PyTorch.
  We see that Power iteration has inherent randomness by design whereas Gram iteration is fully deterministic.
  }
  \label{fig:convergence_plot_gram_vs_power_itreration_same_mat}
\end{figure}

\begin{figure}[h]
  \centering
  \includegraphics[width=1.0\linewidth]{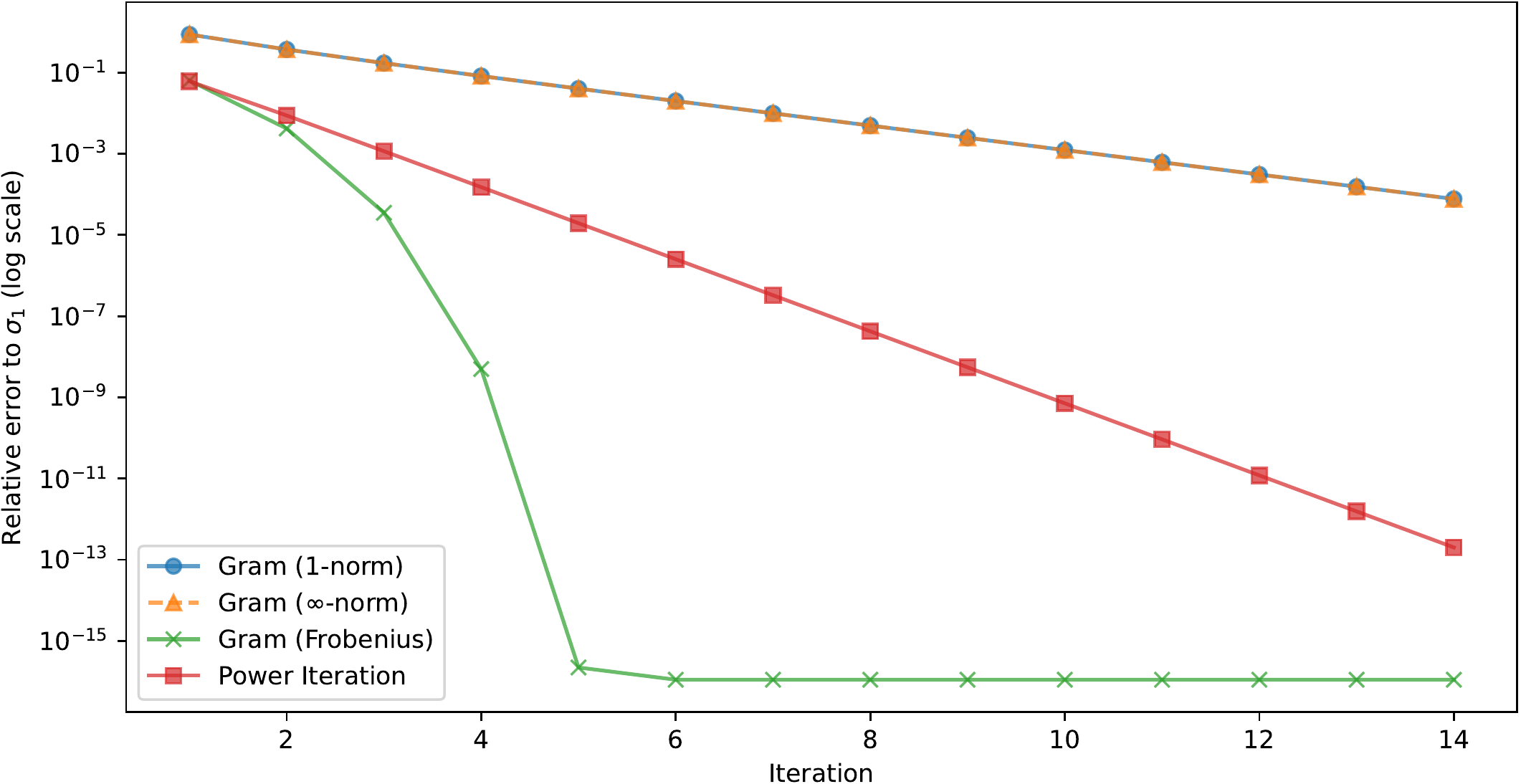}
  \caption{
    Comparison of Gram iteration with different final readouts (Frobenius, $\|\cdot\|_1$, $\|\cdot\|_\infty$)
    and Power Iteration (PI).
    Only the Frobenius readout exhibits \emph{supergeometric} convergence with local order tending to~2
    (near–quadratic appearance), consistent with $e_t=\Theta(q^{2^t}/2^t)$.
    Other readouts converge more slowly (effectively R–linear), while PI converges linearly at rate~$\sigma_2/\sigma_1$.
  }
  \label{fig:compare_gram_pi_inf_1_frob}
\end{figure}

\paragraph{Experiment: PI vs.\ GI with different readouts.}
We compare Gram iteration (GI) with different choices of final readout norm
against the classical Power Iteration (PI); see Fig.~\ref{fig:compare_gram_pi_inf_1_frob}.
The goal is to highlight how the choice of norm in the final evaluation of
$s_t=\|\mW^{(t)}\|^{2^{1-t}}$ impacts the asymptotic rate.

The test matrix is constructed as
\[
  \mW=\mU\,\mathrm{diag}(\sigma_1,\sigma_2,\ldots,\sigma_r)\mV^\top,
\]
where $\mU,\mV$ are independent Haar–distributed orthogonal matrices (obtained
from the QR factorization of Gaussian matrices). We fix $\sigma_1=1$,
$\sigma_2=\rho\in\{0.6,0.8\}$ to control the spectral gap, and let the remaining
singular values decrease geometrically. Unless otherwise stated we set $p=q=60$
and keep the random seed fixed to ensure reproducibility.

GI follows Algorithm~\ref{algo:gram_iteration}: at each step the iterate is
rescaled by the Frobenius norm to avoid overflow, while the final readout is
taken with $\|\cdot\|_\star\in\{\|\cdot\|_F,\|\cdot\|_1,\|\cdot\|_\infty\}$.
PI is run on $\mA=\mW^*\mW$, with iterates
$\vv_{k+1}=\mA\vv_k/\|\mA\vv_k\|_2$ and Rayleigh quotient estimate
$s_k^{\text{(PI)}}=\sqrt{\vv_k^\top \mA \vv_k}$.
We measure the relative error $e_t=|s_t-\sigma_1(\mW)|/\sigma_1(\mW)$ across
iterations, as well as the local order
$p_t=\log(e_{t+1})/\log(e_t)$. This allows us to diagnose whether the observed
convergence is linear, superlinear, or close to quadratic.

The results clearly separate the methods. With Frobenius readout, GI achieves
supergeometric decay $e_t=\Theta(q^{2^t}/2^t)$ and local order $p_t\uparrow 2$,
giving an empirical convergence nearly indistinguishable from quadratic (although
not R–quadratic in theory). In contrast, GI with $\|\cdot\|_1$ or
$\|\cdot\|_\infty$ readout converges much more slowly, consistently with an
R–linear bound. PI recovers its classical behavior, converging linearly with rate
$q=\sigma_2/\sigma_1$.

\subsection{Estimation for circular-padded convolutions}
\paragraph{Estimation on simulated convolutional layers}
\label{subsection:expe_estim_sim_conv_cout}
\begin{figure}[!ht]
  \centering
  \includegraphics[scale=0.55]{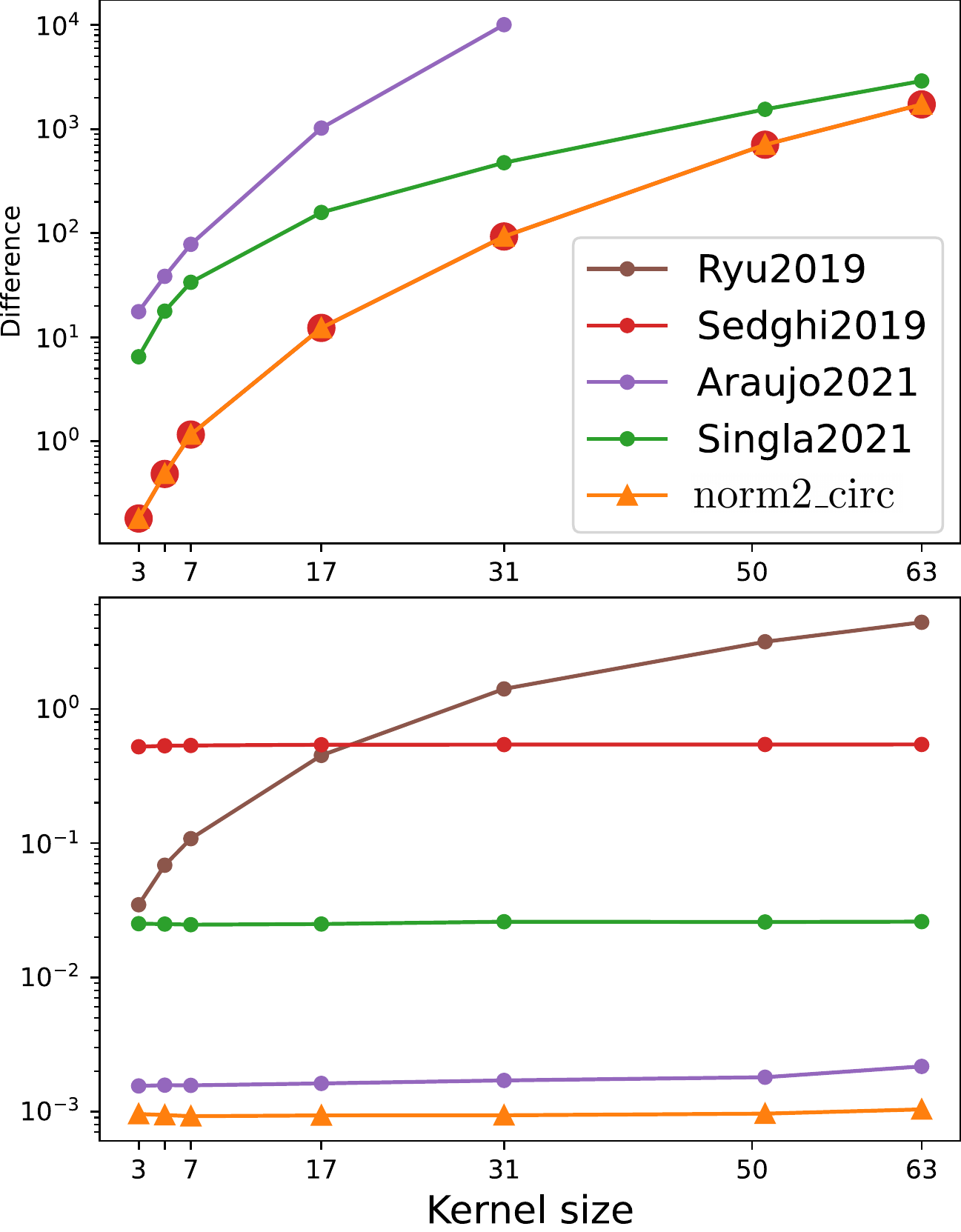}
  \caption{
  Effect of varying kernel size $k$ with $\cin=16$, $\cout=16$, and $n=252$. The difference is expressed as ${\sigma_1}_{\text{method}} - {\sigma_1}_{\text{Ryu2019}}$. As $k$ increases, the Lipschitz difference between circular and zero padding becomes more pronounced. Our method outperforms others in speed and precision, matching \citet{sedghi2019singular}.
  On the y-axis of the first plot,
  the difference is expressed as ${\sigma_1}_{\text{method}} - {\sigma_1}_{\text{Ryu2019}}$.
  On the second plot, the computational time is expressed in seconds.
  }
  \label{fig:times_estimation_error_vary_kernel_size}
\end{figure}
\begin{figure}[!ht]
  \centering
  \includegraphics[scale=0.55]{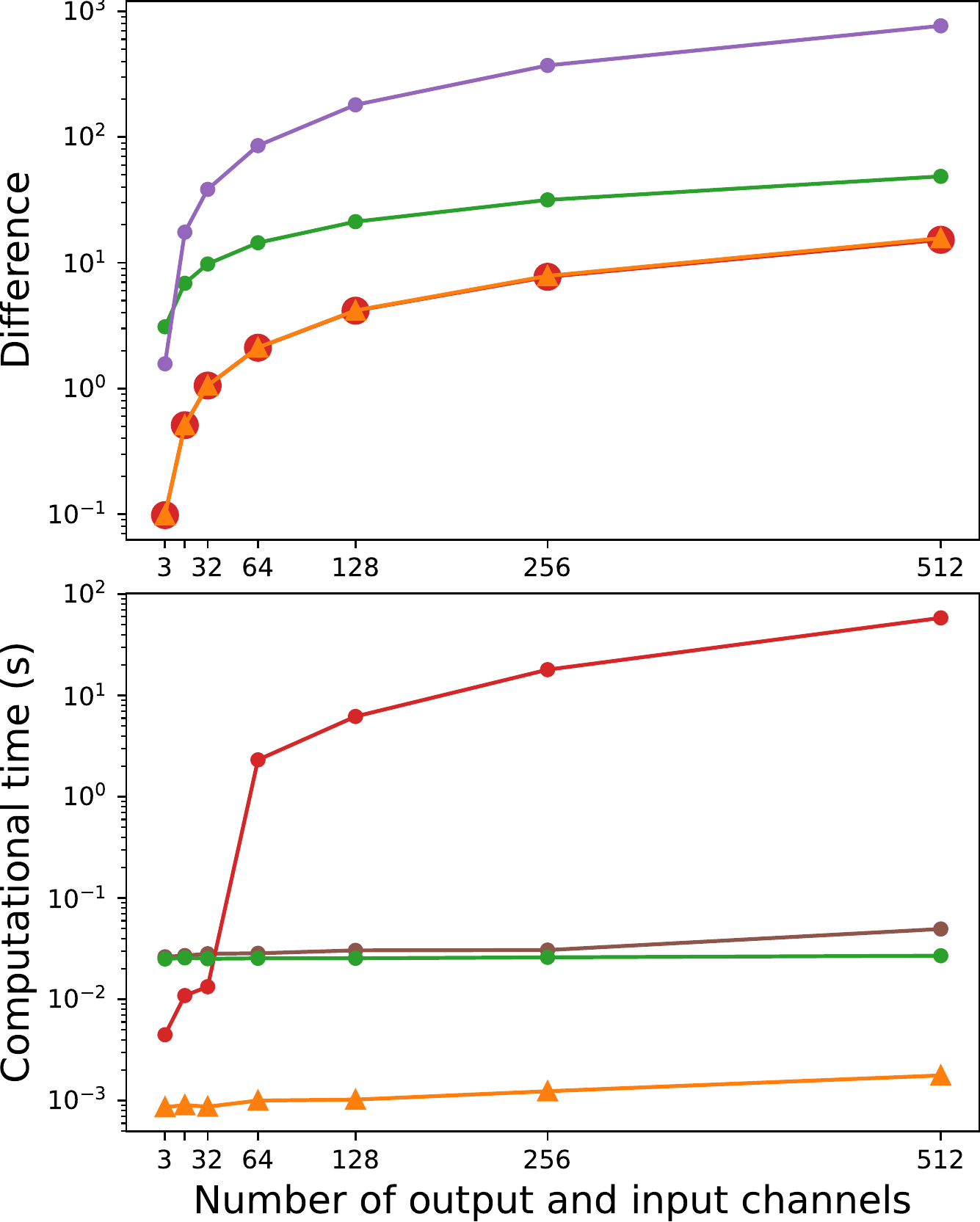}
  \caption{
  Effect of varying the number of channels $\cin$ and $\cout$ with fixed input size $n=32$ and kernel size $k=3$. The difference is expressed as ${\sigma_1}_{\text{method}} - {\sigma_1}_{\text{Ryu2019}}$. Our method achieves the best balance between speed and precision, matching the values from \citet{sedghi2019singular}, while \citet{ryu2019plug} scales strongly with input and kernel size.
  On the y-axis of the first plot,
  the difference is expressed as ${\sigma_1}_{\text{method}} - {\sigma_1}_{\text{Ryu2019}}$.
  On the second plot, the computational time is expressed in seconds.
  }
  \label{fig:times_estimation_error_vary_cin_cout}
\end{figure}
\begin{figure}[!ht]
  \centering
  \includegraphics[scale=0.55]{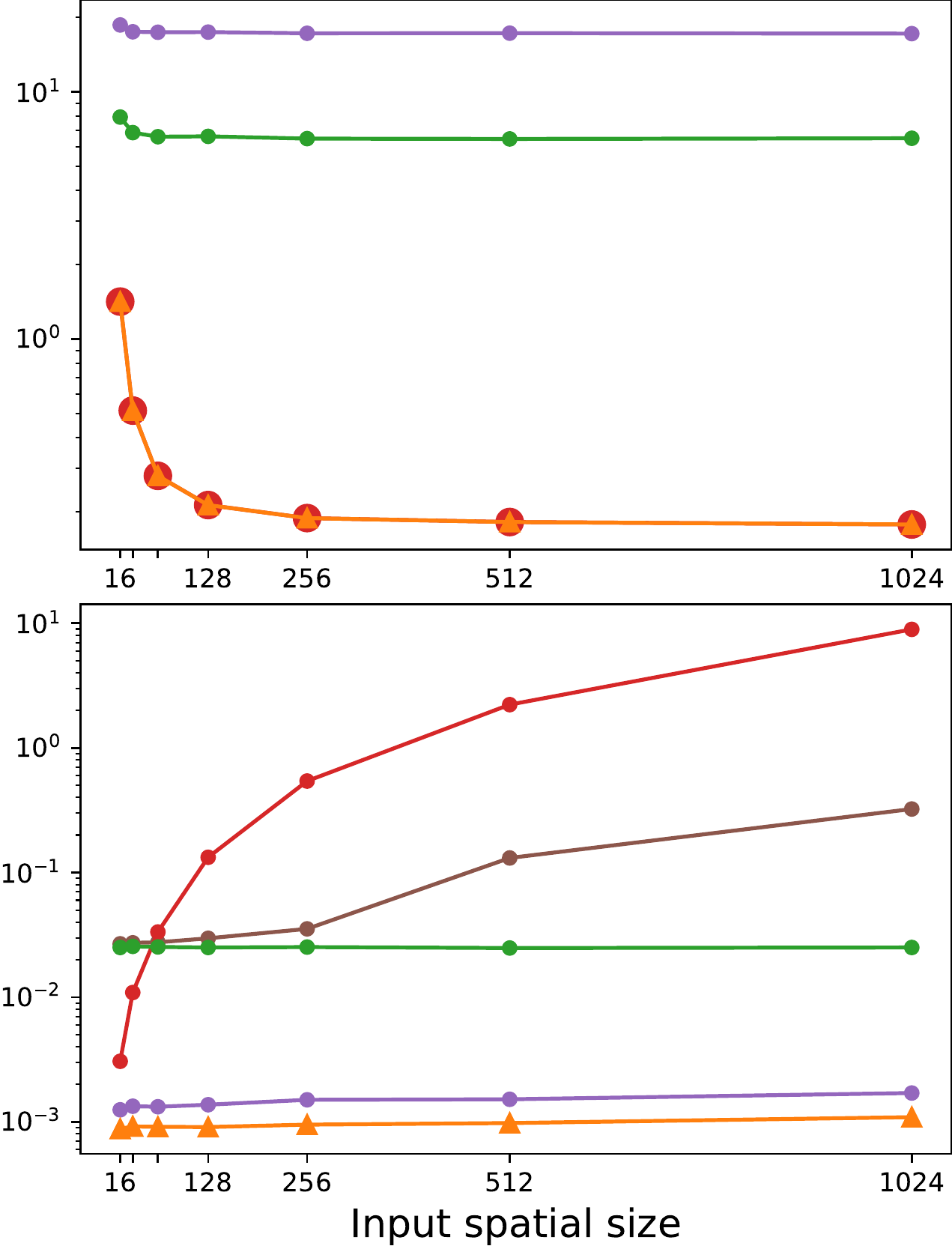}
  \caption{
  Effect of varying input size $n$, with $\cin=16$, $\cout=16$, and $k=3$. The difference is expressed as ${\sigma_1}_{\text{method}} - {\sigma_1}_{\text{Ryu2019}}$. Our method is efficient and precise, comparable to \citet{sedghi2019singular}, while \citet{ryu2019plug} exhibits high computational costs for larger $n$.
  On the y-axis of the first plot,
  the difference is expressed as ${\sigma_1}_{\text{method}} - {\sigma_1}_{\text{Ryu2019}}$.
  On the second plot, the computational time is expressed in seconds.
  }
  \label{fig:times_estimation_error_vary_input_size}
\end{figure}
This experiment evaluates the performance of various Lipschitz-bound methods in terms of precision and computational time. We consider convolutional layers with constant padding and vary three parameters: number of input and output channels (Figure~\ref{fig:times_estimation_error_vary_cin_cout}), input size $n$ (Figure~\ref{fig:times_estimation_error_vary_input_size}), and kernel size $k$ (Figure~\ref{fig:times_estimation_error_vary_kernel_size}). For each experiment, random kernels are sampled from Gaussian and uniform distributions, repeated 50 times, with results averaged for both errors (\({\sigma_1}_{\text{method}} - {\sigma_1}_{\text{Ryu2019}}\)) and computational time.

The methods compared include \citet{ryu2019plug} (100 iterations for precise estimates), \citet{singla2021fantastic} (50 iterations as per their code), \citet{araujo2021lipschitz} (50 samples), and our Algorithm~\ref{algo:gram_iteration_conv} (5 iterations). Our method achieves the same spectral bound as \citet{sedghi2019singular} under the circular convolution hypothesis, while being the fastest among all methods.

We observe that \citet{ryu2019plug}'s computational cost scales significantly with $n$ and $k$, whereas other methods based on the circular convolution hypothesis have near-constant computational times. As kernel size $k$ increases, the difference between circular and zero padding becomes more significant. These results highlight the scalability and precision of our approach for varying convolutional layer parameters.


\paragraph{Estimation on real convolutional layers}

\begin{figure}[h]
  \centering
  \includegraphics[scale=0.7]{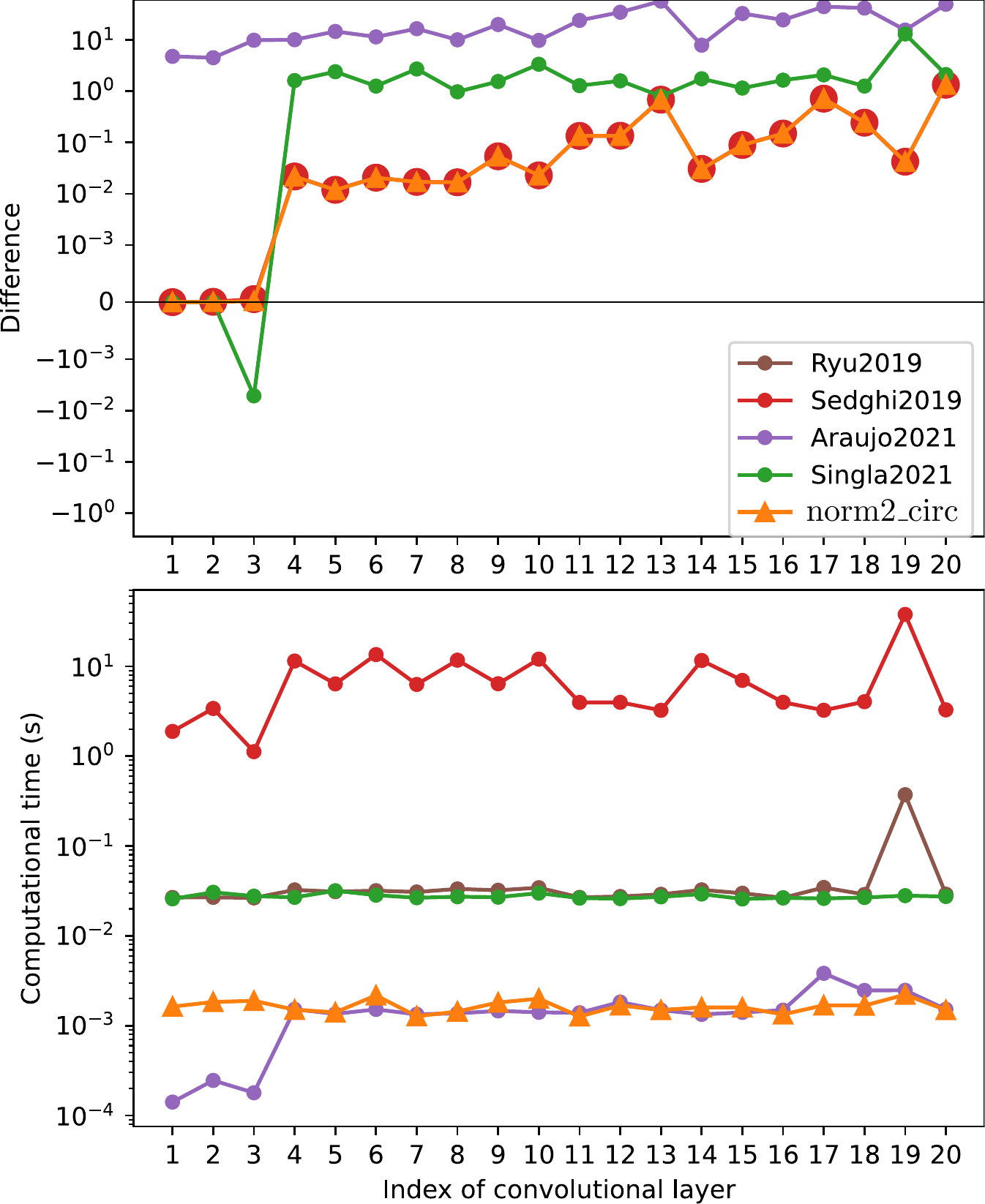}
  \caption{
    Comparison of spectral norm bounds for all convolutional layers of a pre-trained ResNet18. The reference value is obtained using the method of~\citet{ryu2019plug}. Each convolutional layer in ResNet18 has varying numbers of input and output channels, kernel sizes, and input spatial dimensions. This experiment evaluates the bounds on real kernel filters, demonstrating that our method achieves the best precision while maintaining comparable computational times to~\citet{araujo2021lipschitz}.
  }
  \label{fig:resnet18_convlip}
\end{figure}

In this experiment, we use the method of~\citet{ryu2019plug} with 100 iterations (to ensure convergence) as the reference for spectral norm computation. Since most convolutional layers in CNNs use constant zero padding, we also compare with~\citet{sedghi2019singular}, which provides the exact spectral norm under the circular padding assumption.

Inspired by~\citet{singla2021fantastic}, we estimate the spectral norm for each convolutional layer of a ResNet18~\citep{he2016deep} pre-trained on the ImageNet-1k dataset. Figure~\ref{fig:resnet18_convlip} shows the difference between the reference and the spectral norm bounds for various methods across all convolutional layers.

Our method produces bounds very close to the exact values given by~\citet{sedghi2019singular}, while consistently remaining above the reference value of~\citet{ryu2019plug}, ensuring a certified upper bound. In contrast, we observe that for filter index 3, the method of~\citet{singla2021fantastic} reports a value of $2.03$, which is below the reference value of $2.05$. This highlights the limitations of using power iteration (PI) in the pipeline to obtain a reliable upper bound. Detailed numerical results are provided in Table~\ref{tab:lipschitz_resnet18}.

\subsection{Estimation for zero-padded convolutions}
\begin{figure}[h]
  \centering
  \includegraphics[width=0.7\linewidth]{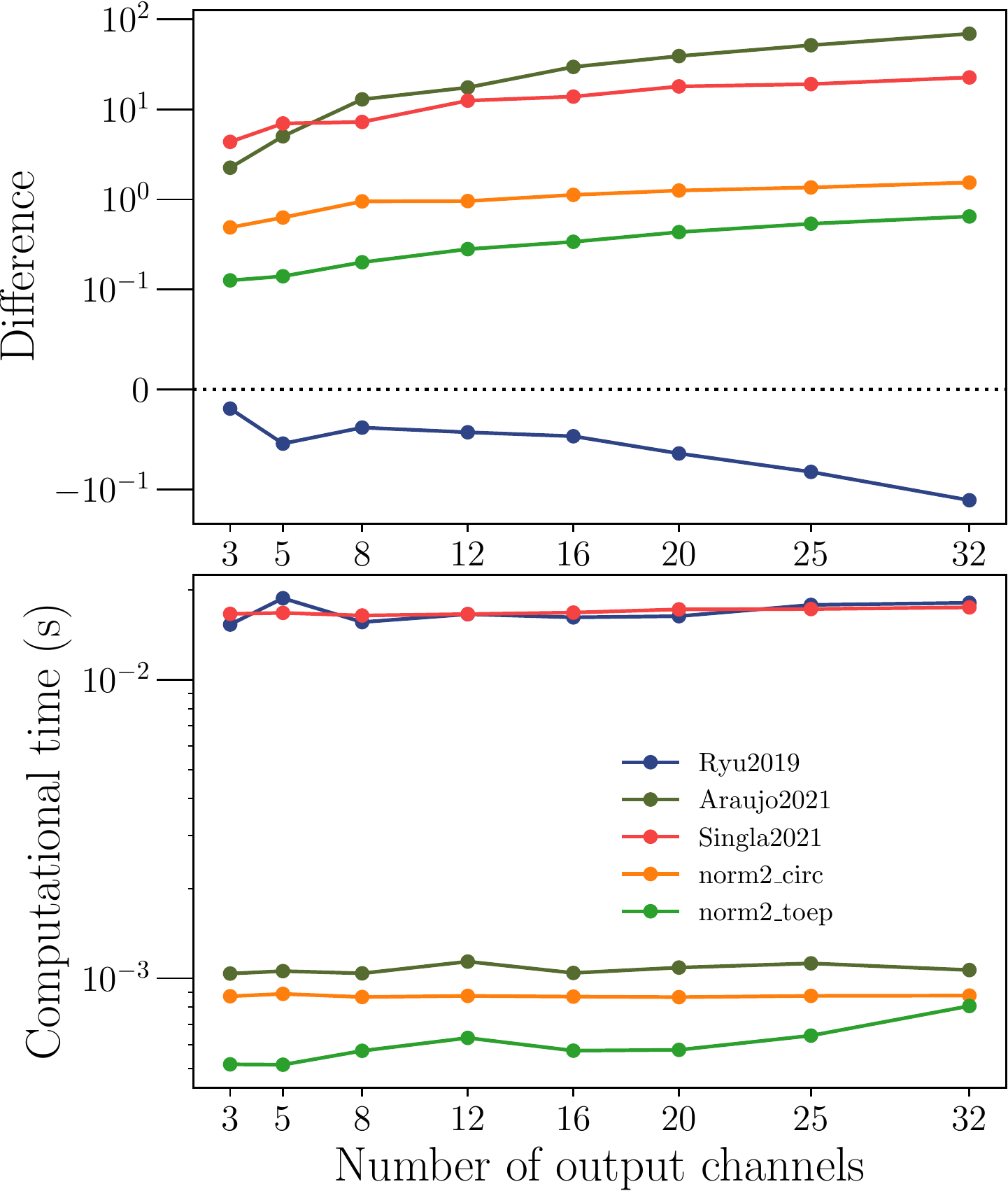}
  \caption{
    Estimation error and computational time for spectral norm computation of zeros-padded convolutional layers with varying numbers of input and output channels ($\cin, \cout$). Kernel size is $k=3$ and input size is $n=32$.
    The reference value for the spectral norm is taken as ground truth by computing the spectral norm of the corresponding Toeplitz matrix with NumPy's $\mathrm{matrix\_norm}$ function. The results compare different state-of-the-art methods.
  }
  \label{fig:accuracy_time_conv_spectral_norm}
\end{figure}

This experiment evaluates the accuracy and computational cost of spectral norm estimation methods for zeros-padded convolutional layers. While the case of circular padding has been addressed in the previous subsection, it does not provide theoretical guarantees for zeros-padded convolutional layers.

We exclude the method of \citet{sedghi2019singular} from this comparison because our Gram iteration-based circulant method ($\mathrm{norm2\_circ}$) achieves the same reference values for circular padding in significantly less computational time. For ground truth, we generate random filters $\tK$ sampled from a Gaussian distribution and construct the corresponding doubly banded Toeplitz matrix, as described in Section~\ref{section:circulant_and_toeplitz_matrices}. The spectral norm of this matrix is then computed using NumPy's $\mathrm{matrix\_norm}$ function.

For Gram iteration-based methods, we use six iterations, while the power iteration method of \citet{ryu2019plug} is run with 100 iterations to ensure convergence. We limit our tests to $1 \leq \cin, \cout \leq 32$ to avoid memory constraints associated with the large size of the Toeplitz matrix. The circulant approach ($\mathrm{norm2\_circ}$) incorporates a corrective factor $\alpha$, ensuring an upper bound on the spectral norm, as demonstrated in Theorem~\ref{thm:bound_circ_toep}, to make it applicable for zeros-padded convolutional layers.

Figure~\ref{fig:accuracy_time_conv_spectral_norm} shows that the power iteration method fails to guarantee an upper bound on the spectral norm. In particular, the method of \citet{ryu2019plug} consistently underestimates the true spectral norm. By contrast, our $\mathrm{norm\_toep}$ method delivers the best accuracy with a low computational cost, demonstrating its effectiveness for zeros-padded convolutional layers.

\subsection{Memory impact of gradient computation}

\paragraph{Memory cost of \texttt{norm2\_circ}.}
For the case of $\mathrm{norm2\_circ}$, naively applying automatic differentiation to estimate the gradient of the spectral norm bound via Gram iteration leads to a high memory footprint, particularly when FFT-based convolution is used. This is because autodiff stores all intermediate activation maps for the backward pass, which becomes prohibitive when regularizing all convolutional layers of deep networks. Instead, we derive the gradient explicitly (using Proposition~\ref{prop:gradient_gram_iteration}), avoiding backpropagation through the full iterative Gram procedure.

Table~\ref{tab:memory_impact_grad} compares the peak GPU memory usage when using autodiff versus our explicit formulation. On high-resolution inputs ($224 \times 224$), the explicit method reduces memory usage by more than 2× on ResNet18/34 and remains feasible on ResNet50, whereas autodiff exceeds available memory.

\begin{table}[h]
  \centering
  \caption{Peak GPU memory (in MB) during training with Gram regularization on all convolutional layers ($t=6$ iterations).}
  \begin{tabular}{lcc}
    \toprule
    \textbf{Model} & \textbf{Autodiff} & \textbf{Explicit Gradient} \\
    \midrule
    ResNet18       & 9770              & \textbf{4322}              \\
    ResNet34       & 17152             & \textbf{7088}              \\
    ResNet50       & $>42000$          & \textbf{27535}             \\
    \bottomrule
  \end{tabular}
  \label{tab:memory_impact_grad}
\end{table}

\paragraph{Memory cost of \texttt{norm2\_toep}.}
The \texttt{norm2\_toep} method estimates the spectral norm of a convolutional filter via recursive Gram iterations, where each iteration involves a self-convolution of the kernel. To preserve spatial alignment, zero-padding of size $k - 1$ is applied at every iteration, where $k$ is the initial kernel size. This results in a progressive expansion of the kernel’s spatial support. After $t$ Gram iterations, the effective kernel size becomes:
\[
  s_t = (k - 1)\,t + 1 \, .
\]
Hence, the memory required to store the filter grows quadratically with the number of iterations:
\[
  \#\text{elements}_t = s_t^2 = \left((k - 1)\,t + 1\right)^2.
\]
This quadratic growth imposes practical constraints, especially for large $t$ or wide filters. In practice, moderate values such as $t = 6$ remain tractable.
Nevertheless, the overall memory usage of \texttt{norm2\_toep} is typically lower if convolutional filters are usually small (e.g., $k=3$); second, fewer iterations are needed to obtain a reliable upper bound.

\section{Conclusion}
In this chapter, we introduced novel methods based upon Gram iteration
for spectral norm estimation that are deterministic, near quadratically convergent (super geometric in theory), differentiable, and consistently provide certified upper bounds. These methods are designed for three specific cases:

\begin{itemize}
  \item Dense layers represented by dense matrices.
  \item Circularly padded convolutional layers modeled as concatenations of doubly block circulant matrices.
  \item Zeros-padded convolutional layers modeled as concatenations of doubly block Toeplitz matrices.
\end{itemize}
These methods leverage the structural properties of the matrices involved, ensuring computational efficiency while maintaining theoretical guarantees.
They can be used to have a finer estimate of the spectral norm of linear layers in neural networks. The code is available at this link \href{https://github.com/blaisedelattre/lip4conv}{\texttt{lip4conv}}.\\
The proposed approaches are relevant in domains where grounded upper bounds are essential, primarily in robustness certification, stability analysis, and adversarial defense. Additionally, they play a crucial role in normalizing flow architectures, where layer inversion is required~\citep{behrmann2019invertible} and ensuring a tight bound on the Lipschitz constant is critical for preserving invertibility and numerical stability. \\
As differentiable techniques, they can be seamlessly integrated into neural network training pipelines, enabling their use for regularization. This is particularly useful in enforcing Lipschitz constraints or improving the robustness and stability of models. \\
Furthermore, we analyze the relationship between circular and zero padding, providing a theoretical approximation that allows spectral norm estimation under zero padding using results from circular padding. These findings enhance the understanding of convolutional layers’ spectral properties. \\
Recently, the work of~\cite{boissin2025adaptive} and the package \href{https://github.com/thib-s/orthogonium}{\texttt{orthogonium}} have expanded the design of orthogonal convolutional layers, making them more competitive in deep learning architectures. Our theoretical results can be leveraged to provide stronger guarantees on the spectral norm of these layers, ensuring more reliable control over their Lipschitz properties. Furthermore, our methods facilitate the extension of spectral norm control from circular padding to the more commonly used zero-padding layers, broadening their practical applicability.

In the next chapter, we will apply these spectral norm estimation techniques to enforce Lipschitz constraints in neural networks, demonstrating their impact on robustness, generalization, and stability in deep learning.

%% file: content/chapter-lipschitz_computation.tex
%

\setcounter{question}{0}
\chapter{Lipschitz networks}\label{chap:lipschitz_computation}



\minitoc%

Deep learning models often exhibit sensitivity to input perturbations~\citep{szegedy2013intriguing}, making stability a crucial consideration for both theoretical analysis and practical deployment. Lipschitz continuity provides a fundamental framework for quantifying and controlling this stability, playing a key role in generalization~\citep{bartlett2017spectrally} and adversarial robustness~\citep{tsuzuku2018lipschitz}. However, designing Lipschitz networks remains a significant challenge, as their Lipschitz constant is difficult to compute and constrain, particularly in deep architectures.
A major gap persists between the performance of state-of-the-art architectures and the theoretical guarantees offered by Lipschitz networks. While Lipschitz continuity has been leveraged to improve robustness and certified guarantees, existing approaches often impose strong constraints on network expressivity, leading to a trade-off between stability and performance. This calls for improved methods to estimate, regularize, and construct Lipschitz networks that balance theoretical guarantees with practical effectiveness.

This chapter presents our contributions to the study of Lipschitz continuity in neural networks and introduces theoretical and practical tools that will be leveraged in subsequent Chapter~\ref{chapter:robustness} and Chapter~\ref{chap:regularization}.
We begin by exploring methods for estimating the global Lipschitz constant of convolutional neural networks using Product Upper Bound (PUB), which provides an efficient approach to computing this Lipschitz bound.
We then introduce novel regularization techniques designed to directly control the Lipschitz constant of convolutional layers during training. To complement these methods, we propose approaches for constructing neural network layers that are inherently Lipschitz continuous. Specifically, we develop a new rescaling method \emph{spectral rescaling} tailored for Lipschitz networks.
Finally, we present new theoretical results on the regularity of the Weierstrass transform, highlighting its applications in designing Lipschitz networks.

\section{Lipschitz constant estimation with PUB}
\label{section:lip_cnn}

Building on the work presented in the chapter on spectral norm computation~\ref{chapter:spectral_norm_estimation} for neural network layers, we aim to refine the Product Upper Bound ($\PUB$) approximation for neural networks.
By leveraging precise spectral norm calculations, using Gram iteration introduced in Chapter~\ref{chapter:spectral_norm_estimation} (Equation~\ref{eq:gram_iterate}),
we demonstrate improvements over state-of-the-art methods for estimating the Lipschitz constant through $\PUB$, particularly for convolutional networks.
For our experiments, we use the method proposed by~\citet{ryu2019plug} with $100$ iterations to ensure convergence as a reference value. This serves as a benchmark for comparison, given its established accuracy. It is worth noting that most convolutional layers in CNNs employ constant zero padding, whereas~\citet{sedghi2019singular} provides an exact spectral norm computation specifically for convolutional layers with circular padding. To compute the $\PUB$ for a CNN or ResNet architecture we follow the rule stated in the Appendix~\ref{app:sec:pub_approximation}.
\begin{table*}[t]
  \caption{Comparison of networks bound Lipschitz ratio with standard deviation for several CNNs, reference for computing the ratio is the bound given by~\citet{ryu2019plug} method. Overall Lipschitz constant bound $\PUB_{\text{method}}$ is estimated for each method. Ratio of network Lipschitz bound is defined as $\PUB_{\text{method}}(f) / \PUB_{\text{Ryu2019}}(f)$. Results are averaged over 100 runs.
    Ratio standard deviations of ours,~\citet{sedghi2019singular} are induced by ~\citet{ryu2019plug} method. We give the same ratio as~\citet{sedghi2019singular} in a much lower time.
  }
  \vspace{0.1cm}
  \centering
  \vspace{0.1cm}
  \begin{tabular}
    {
      l
      l
      l
      l
      l
    }
    \toprule
    \multirow{2}[2]{*}{\textbf{Model}} & \multicolumn{4}{c}{\textbf{Ratio of Network Lipschitz Bound  (Total Running Time (s))}}                                                                                                                    \\
    \cmidrule{2-5}
                                       & \textbf{Ours}                                                                           & ~\citeauthor{singla2021fantastic}                  & \citeauthor{sedghi2019singular} & ~\citeauthor{ryu2019plug} \\
    \midrule
    VGG16                              & $1.14 \pm 0.02 ~(0.100)$                                                                & $23.90 \pm 6.80 ~(0.47)$                           & $1.14 \pm 0.020 ~(525)$         & $(0.64)$                  \\
    VGG19                              & $1.16 \pm 0.005 ~(0.110)$                                                               & $30.63 \pm 0.30 ~(0.53)$                           & $1.16 \pm 0.005 ~(639)$         & $(0.71)$                  \\
    ResNet18                           & $1.47 \pm 0.007 ~(0.039)$                                                               & $87.93 \pm 0.88 ~(0.50)$                           & $1.47 \pm 0.007 ~(185)$         & $(0.71)$                  \\
    ResNet34                           & $1.82 \pm 0.35 ~(0.060)$                                                                & $4982 \pm 4894 ~(0.88)$                            & $2.15 \pm 0.35 ~(237)$          & $(1.16)$                  \\
    ResNet50                           & $1.68 \pm 0.35 ~(0.100)$                                                                & $3338 \pm 4622 ~(1.05)$                            & $1.67 \pm 0.35 ~(377)$          & $(1.49)$                  \\
    ResNet101                          & $1.74 \pm 0.32 ~(0.173)$                                                                & $4026 \pm 4178 ~(1.50)$                            & $1.74\pm 0.32 ~(551)$           & $(2.20)$                  \\
    ResNet152                          & $1.92 \pm 0.46 ~(0.260)$                                                                & $8.39\mathrm{e}{+4} \pm 1.6\mathrm{e}{+5} ~(2.05)$ & $1.92 \pm 0.460 ~(725)$         & $(3.01)$                  \\
    \bottomrule
  \end{tabular}
  \label{tab:lipschitz_product_ratio_ryu}
\end{table*}
The overall Lipschitz constant of the network is assessed here using $\PUB$. We consider several CNNs pre-trained on the ImageNet-1k dataset: VGG~\citep{simonyan2014vgg} and ResNet~\citep{he2016deep} of different sizes.
Using rules from Appendix~\ref{section:lip_cnn} to compute the Lipschitz constant of each individual layer in CNNs, we compute bound $\PUB(f)$ on the overall Lipschitz constant of the network.
Then we compare this produced bound for each method to the one obtained by~\citep{ryu2019plug} by considering the ratio: $\PUB_{\text{method}}(f) / \PUB_{\text{Ryu2019}}(f)$.

We take $100$ iterations for~\citet{ryu2019plug} to have a precise reference, $50$ iterations for~\citep{singla2021fantastic}, $20$ samples for~\citep{araujo2021lipschitz} and $7$ iterations for our method, as the task requires increased precision. Results are reported in Table~\ref{tab:lipschitz_product_ratio_ryu}, we see that our method gives results similar to~\citep{sedghi2019singular} and overall network Lipschitz bound is tighter than~\citet{singla2021fantastic} and~\citet{araujo2021lipschitz} in comparison to~\citet{ryu2019plug}. This experience illustrates that small errors in the estimation of a single Lipschitz constant layer can lead to major errors on the Lipschitz bound of the overall network and that acute precision is crucial. Moreover, methods using PI such as~\citep{singla2021fantastic} have a huge standard deviation in this task which is problematic for deep networks whereas ours and~\citet{sedghi2019singular} have standard deviations that remain small. Our method offers the same precision quality as~\citet{sedghi2019singular} but significantly faster.

These results demonstrate that precise spectral norm calculations for individual layers are critical for obtaining tighter Lipschitz bounds. This precision not only improves robustness guarantees but also reduces variability, making the bounds more reliable for deep networks.
However, the $\PUB$ is still a loose bound, especially when we compare it to empirical local Lipschitz computed with gradient ascent.

\section{Controlling the Lipschitz constant of convolutional layers through regularization}
\label{section:reg_conv_layers}
\begin{figure}[h]
  \centering
  \includegraphics[width=0.75\textwidth]{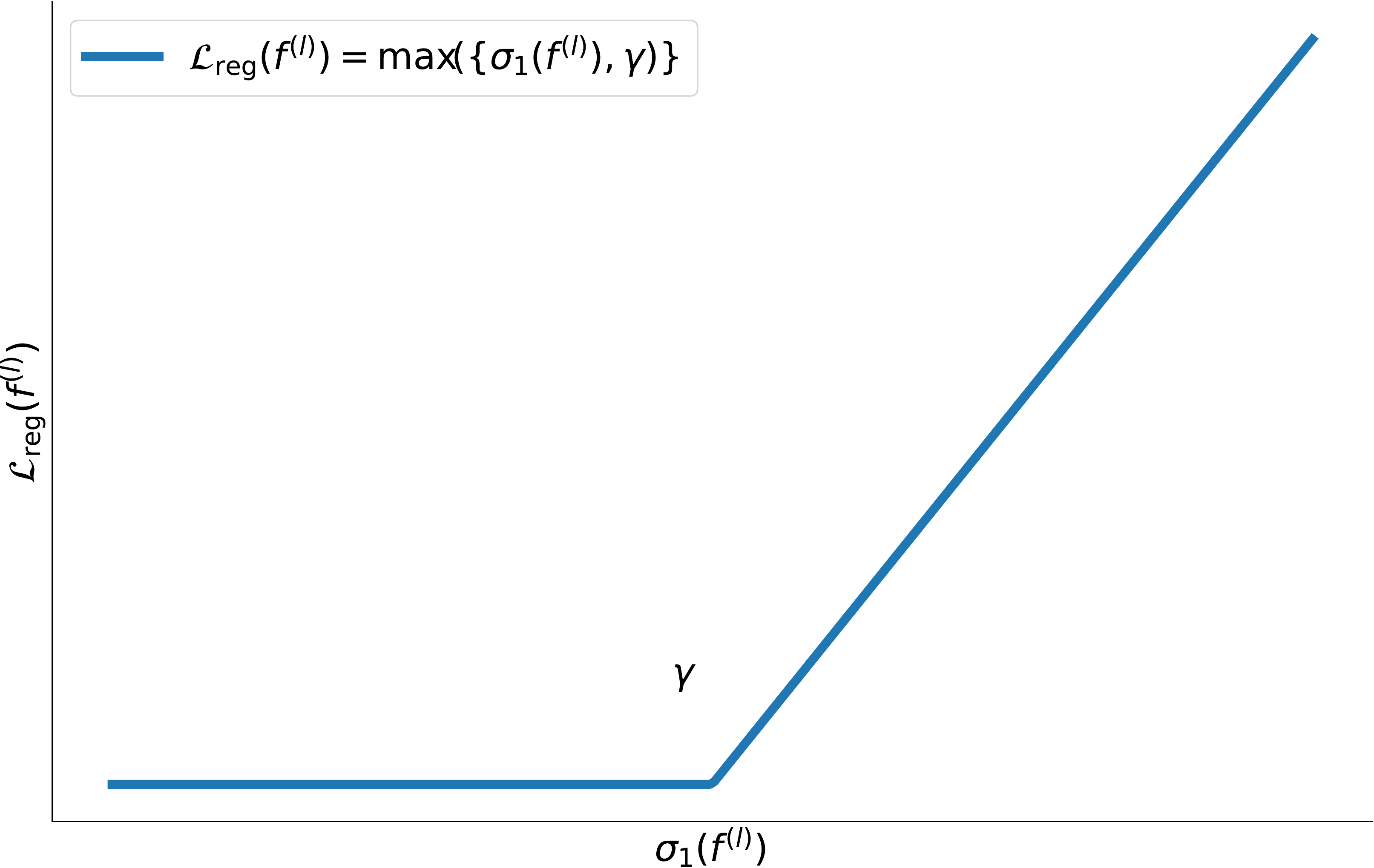}
  \caption{Illustration of the loss $\mathcal{L_{\text{reg}}}$ in function of the largest singular value ${\sigma_1}_\text{method}(f^{(l)})$.}
  \label{fig:loss_reg_lip}
\end{figure}

Regularization of spectral norms of convolutional layers has been studied in previous works~\citep{yoshida2017spectral, miyato2018spectral, gouk2021regularisation}.
\citet{wang2020orthogonal} regularized singular values by enforcing orthogonalization, offering a computationally cheap alternative. \citet{el_mehdi2022existence} further studied this orthogonalization technique.
The goal of this experiment is to assess how different bounds control the Lipschitz constant $\Lip(f^{(l)})$ of each convolution layer along the training process. To perform this control, a target Lipschitz constant $\gamma$ is set for all convolutions. The loss optimized during training becomes:
$\mathcal{L_{\text{train}}} + \mu_{\text{reg}} \sum_{l=1}^\nblayers \mathcal{L_{\text{reg}}}(f^{(l)})$, with
\begin{equation}
  \mathcal{L_{\text{reg}}}(f^{(l)})
  = \max\!\bigl\{ {\sigma_1}_\text{method}(f^{(l)}), \, \gamma \bigr\}
\end{equation}
This regularization loss ensures that the Lipschitz constant of the convolutional layer remains close to the target value~$\gamma$, by penalizing the maximum between the spectral norm and~$\gamma$. The function $x \mapsto \max\{x, \gamma\}$ is continuous but not differentiable at~$x=\gamma$; however, it admits well-defined subgradients.
See Figure~\ref{fig:loss_reg_lip} for an illustration of this regularization loss.
We use ResNet18 architecture~\citep{he2016deep}, trained on the CIFAR-10 dataset for 200 epochs, and with a batch size of 256. We use SGD with a momentum of $0.9$ and an initial learning rate of $0.1$ with a cosine annealing schedule.
The baseline is trained without regularization.
We compare different types of regularization: Lipschitz regularization using various methods, all with $\mu_{\text{reg}} = 10^{-1}$, and standard weight decay (WD) applied to the convolutional filters with $\mu_{\text{reg}} = 5 \times 10^{-3}$. This weight decay coefficient was chosen to ensure that the spectral norms of the layers remain approximately bounded around the target value, thus serving as a baseline for comparison.
We only compare methods under the circular padding hypothesis and use~\citep{sedghi2019singular} as the reference since it provides an exact value. We take $6$ iterations for Gram iteration in our method, $10$ for power iteration in~\citep{singla2021fantastic}, and $10$ samples in~\citep{araujo2021lipschitz} to have similar training times.
For our method $\mathrm{norm2\_circ}$, see Algorithm~\ref{algo:gram_iteration_conv}, we implement an explicit backward differentiation using Equation~\ref{eq:gram_iteration_bound_gradient} to speed up the gradient computation and reduce the memory footprint.
%

\begin{figure}[h]
  \centering
  \includegraphics[width=0.8\textwidth]{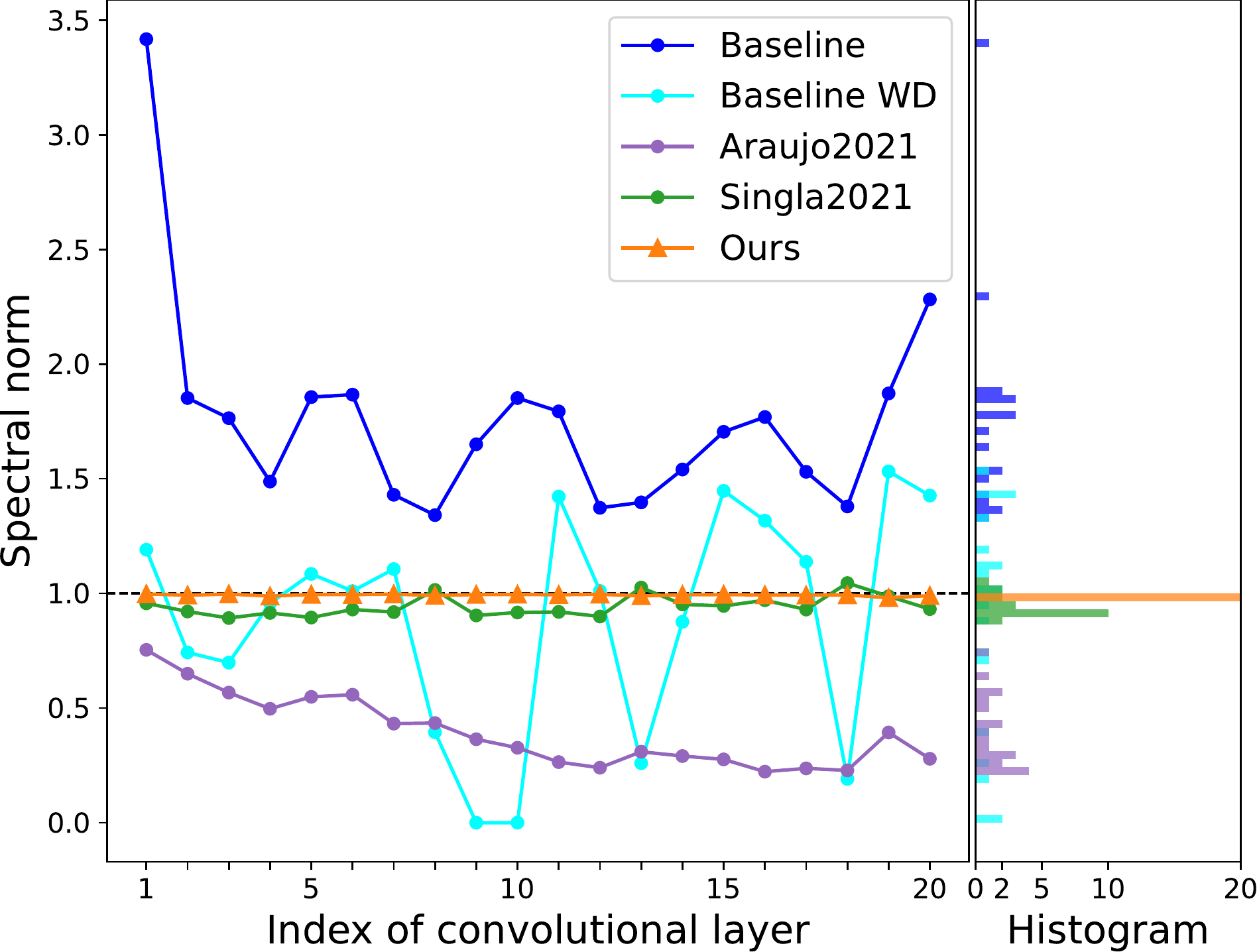}
  \caption{Plot and histogram of spectral norms for each convolution layer in ResNet18 at the end of training on CIFAR-10, for different regularization methods. We observe that the histogram of spectral norms regularized with our method is all at the target Lipschitz constant, meanwhile, other methods' histograms are scattered. }
  \label{fig:histograms_convolution_resnet_training_for_bounds}
\end{figure}

\begin{figure}[h]
  \includegraphics[width=0.95\textwidth, height=0.3\textwidth]{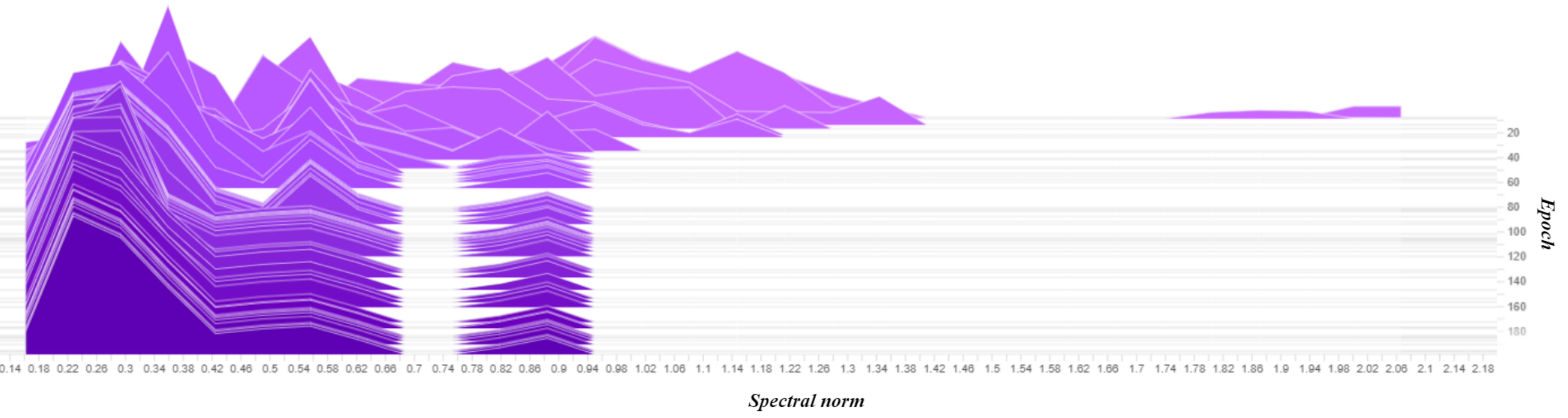}
  \caption{Spectral norm histograms in the function of epoch for  training on CIFAR10 with regularization using~\citet{araujo2021lipschitz}.}
  \label{fig:spectral_norm_over_epochs_reg_resnet_training_for_bounds1}
  \includegraphics[width=0.7\textwidth, height=0.3\textwidth]{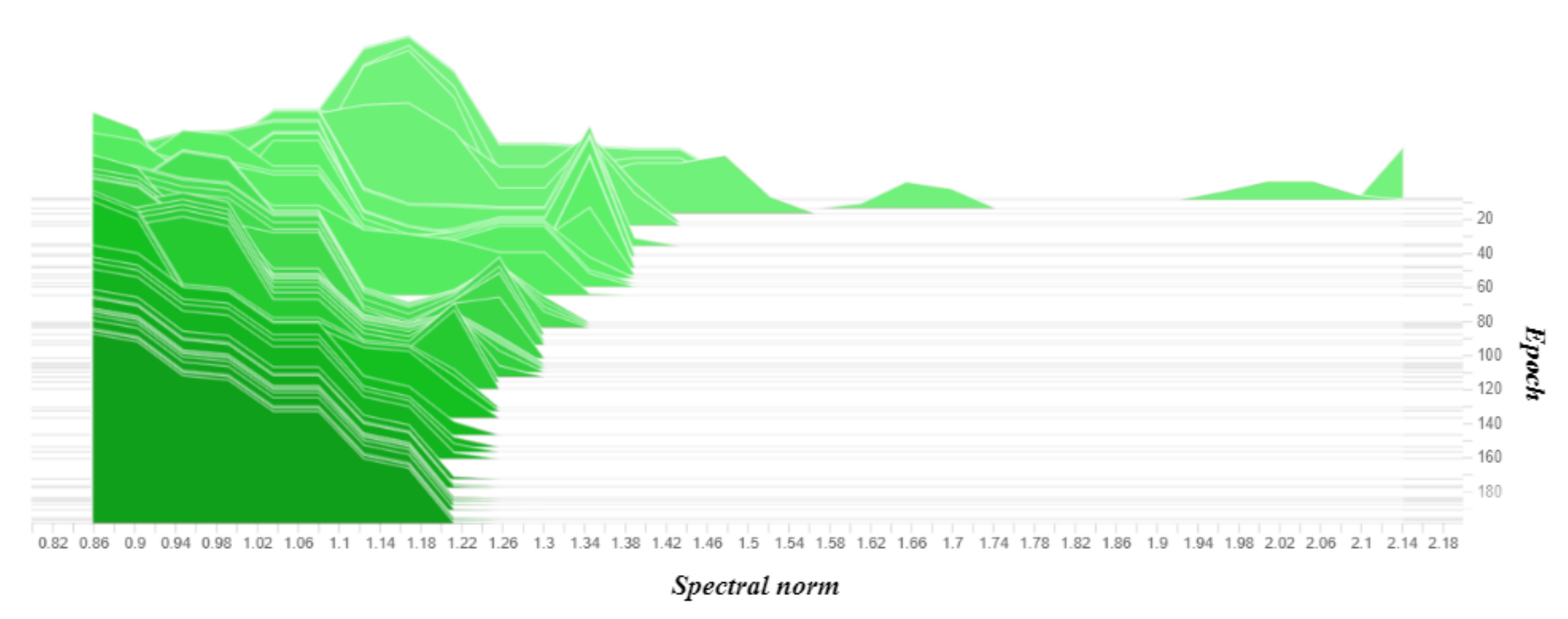}
  \centering
  \caption{Spectral norm histograms in the function of epoch for  training on CIFAR10 with regularization using~\citet{singla2021fantastic}.}
  \label{fig:spectral_norm_over_epochs_reg_resnet_training_for_bounds2}
  \includegraphics[width=0.7\textwidth, height=0.3\textwidth]{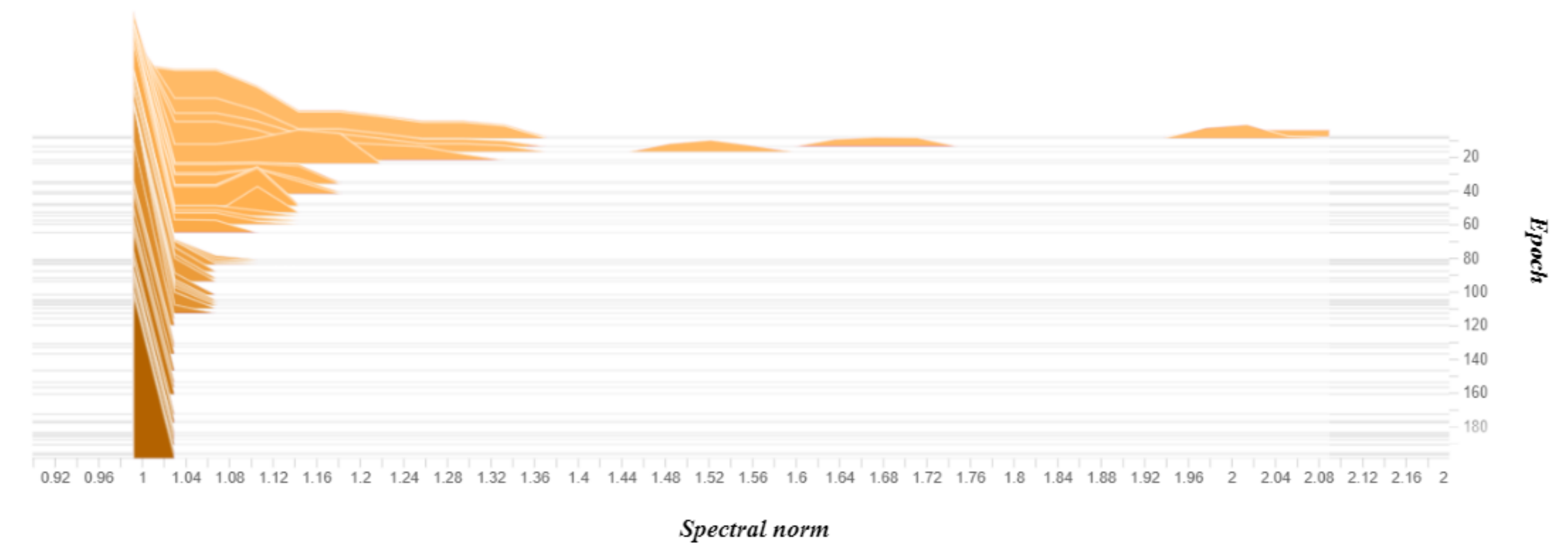}
  \caption{Spectral norm histograms in the function of epoch for  training on CIFAR10 with regularization using our method.}
  \label{fig:spectral_norm_over_epochs_reg_resnet_training_for_bounds3}
\end{figure}

Figure~\ref{fig:histograms_convolution_resnet_training_for_bounds} presents results on Lipschitz constant control of convolutional layers for ResNet18. The goal here is to have a Dirac distribution at target value $\gamma=1$, \ie we want each convolutional layer to have a spectral norm of $1$.
We observe that the more precise a method is, the more well-controlled the resulting distribution of the convolutional layers' spectral norm is after training.
Our regularization method matches each layer's target spectral norm of $1$.
Weight decay gives gross estimations and~\citep{araujo2021lipschitz} overestimates spectral norms and thus results in over-constrained Lipschitz constant for some layers.~\citep{singla2021fantastic} produces a maximum spectral of $1.1$ norm which is above target $\gamma$ but overall Lipschitz constant is centered around the target with dispersion.
This suggests that the convolution’s spectral norm is only loosely constrained, and that the differentiation of our bound behaves as expected.
Detailed spectral norm histograms over epochs for each Lipschitz regularization method are presented in Figure~\ref{fig:spectral_norm_over_epochs_reg_resnet_training_for_bounds1, fig:spectral_norm_over_epochs_reg_resnet_training_for_bounds2, fig:spectral_norm_over_epochs_reg_resnet_training_for_bounds3}, showing that all spectral norms are below the target Lipschitz constant at epoch $120$ for our method and accounts for fast convergence in Lipschitz regularization in comparison to other methods.

We provide a study on the impact of Lipschitz regularization  on the generalization performance of the network in Section~\ref{section:spectral_norm_regularization}, using the same experimental setup as in this section.

\section{Local Lipschitz property of neural networks}
\label{section:local_lipschitz}

Neural networks used are theoretically Lipschitz continuous, but estimating a meaningful Lipschitz constant is challenging.
A common approach to obtain a cheap upper bound is to use the $\PUB(f)$. While this bound is easy to compute, it can be extremely loose for deeper networks, as the actual local Lipschitz behavior may be much smaller than this product.
$1$-Lipschitz networks control the overall Lipschitz constant by design, using a product of upper bounds across layers and making it close to true Lipschitz. However, this global upper bound can be arbitrarily loose compared to the actual local Lipschitz constant around a given input point \(\vx\), particularly for standard architectures like ResNet-110.
\begin{table}[h!]
  \centering
  \caption{Comparison of local Lipschitz constant estimation (Local Lip.) and product upper bound (PUB) for different models and noise levels (\( \sigma \)) used during training.}
  \begin{tabular}{@{}lcc@{}}
    \toprule
    \textbf{Model}                   & \textbf{Local Lipschitz} & \textbf{PUB}              \\ \midrule
    LiResNet (\( \sigma = 0.00 \))   & 22                       & 37                        \\
    LiResNet (\( \sigma = 0.12 \))   & 11                       & 13                        \\
    LiResNet (\( \sigma = 0.25 \))   & 7.9                      & 9                         \\
    LiResNet (\( \sigma = 0.50 \))   & 4.8                      & 6                         \\ \midrule
    ResNet-110 (\( \sigma = 0.00 \)) & 235                      & \( 2.34 \times 10^{10} \) \\
    ResNet-110 (\( \sigma = 0.12 \)) & 27                       & \( 1.03 \times 10^{12} \) \\
    ResNet-110 (\( \sigma = 0.25 \)) & 25                       & \( 3.19 \times 10^{12} \) \\
    ResNet-110 (\( \sigma = 0.50 \)) & 19                       & \( 9.71 \times 10^{9} \)  \\
    ResNet-110 (\( \sigma = 1.0 \))  & 3.8                      & \( 1.32 \times 10^{11} \) \\
    \bottomrule
  \end{tabular}
  \label{tab:lip_comparison}
\end{table}
Table~\ref{tab:lip_comparison} illustrates the significant discrepancy between local Lipschitz estimates given by empirical Lipschitz estimation (Section~\ref{sec:local_lipschitzness}),
and the $\mathrm{PUB}$, especially for ResNet-110, where the global bound becomes impractically loose. This gap underscores the limitations of relying on PUB for certification, motivating the need for research focused on deriving certified bounds for local Lipschitz constants.
%
%
To elucidate why the empirical certified radius often exceeds the theoretical one, we estimate the local Lipschitz constant \(L(\vx, \mathcal{B})\) in a neighborhood of \(x\) of radius \(\epsilon = 0.15\) called $\mathcal{B}$, as detailed in Equation~\ref{eq:empirical_lipschitz}.
However, because these local Lipschitz estimates do not constitute rigorous global upper bounds, they cannot be directly employed for formal certification.

Training with Gaussian noise injection helps to reduce the empirical local Lipschitz constant, making the network smoother and more robust.
For example, increasing the noise level \(\sigma\) significantly decreases the empirical local Lipschitz constant, as shown in Table~\ref{tab:lip_comparison}.
This local reduction in sensitivity can be understood through the lens of function smoothing: as shown by \citet{cohen2019certified}, training with input noise effectively trains a smoothed version of the network, obtained by the Weierstrass transform \(\tilde{f}\) of the original function \(f\). Indeed, by Jensen's inequality,
\[
  \loss \bigl(\tilde{f}(\vx), y\bigr)
  \;\leq\;
  \E \!\bigl(\loss\bigl(f(\vx + \delta), y\bigr)\bigr),
\]
illustrating that minimizing the expected noisy loss indirectly minimizes the loss of the smoothed function.
Moreover, explicit bounds on the Lipschitz constant of \(\tilde{f}\) are known, showing that smoothing inherently reduces sensitivity to input perturbations.

These results highlight two complementary approaches:
(1) designing 1-Lipschitz blocks using the Product Upper Bound to explicitly enforce global control of the Lipschitz constant,
and (2) leveraging the Weierstrass transform by training with noise injection to induce smoother, more robust networks.
Together, these strategies provide a principled foundation for enhancing stability and robustness in neural networks.

\begin{figure}[h]
  \centering
  \includegraphics[width=1.0\textwidth]{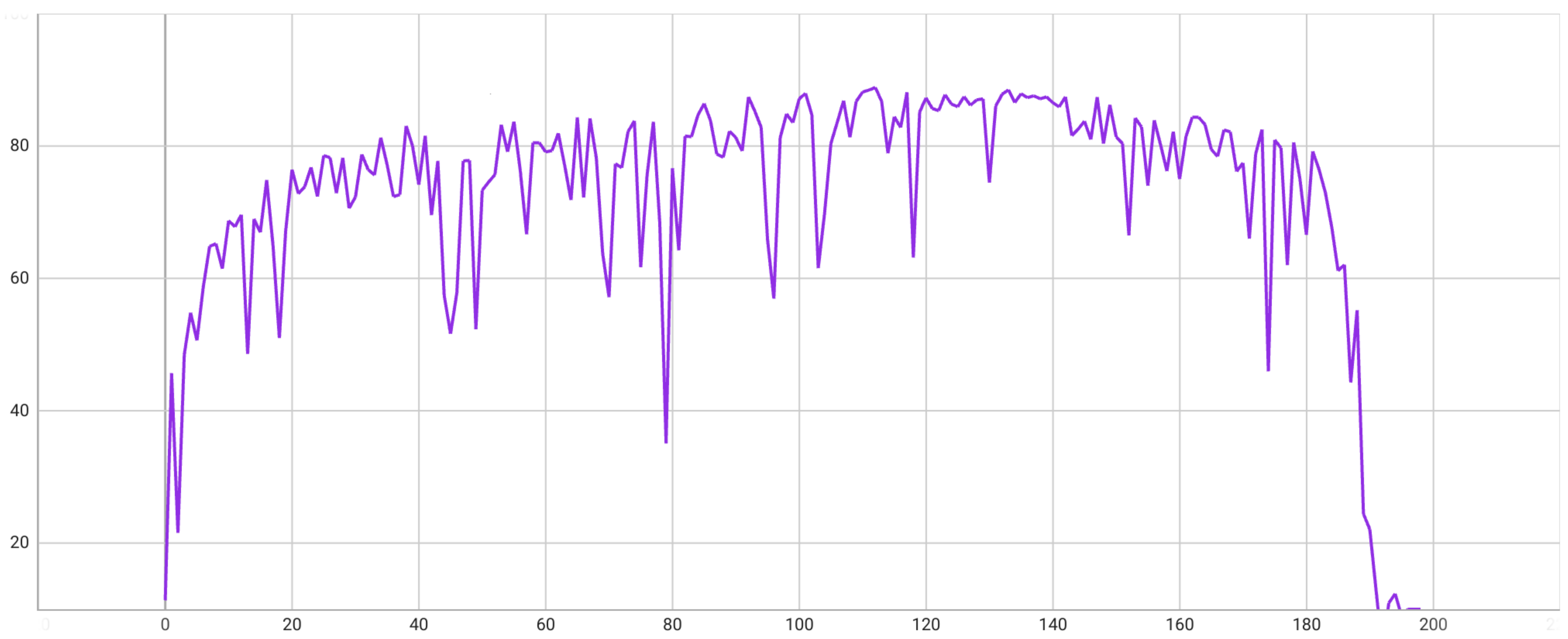}
  \caption{Classification accuracies (y-axis) for every epcoh (x-axis) for ResNet-18 models trained with Lipschitz regularization applied to all convolutional layers and normalization layers. Training is very unstable.}
  \label{fig:accuracy_reg_bn_and_all_linear_layers}
\end{figure}
\section{Training stability with Lipschitz regularization}
In this section, we investigate the use of Lipschitz regularization to enhance training stability in deep neural networks. We explore different regularization strategies, including Lipschitz regularization applied to all layers and we try to reduce the number of normalization layers such as batch normalization~\citep{ioffe2015batch}.

\begin{figure}[h]
  \centering
  \includegraphics[width=1.0\textwidth]{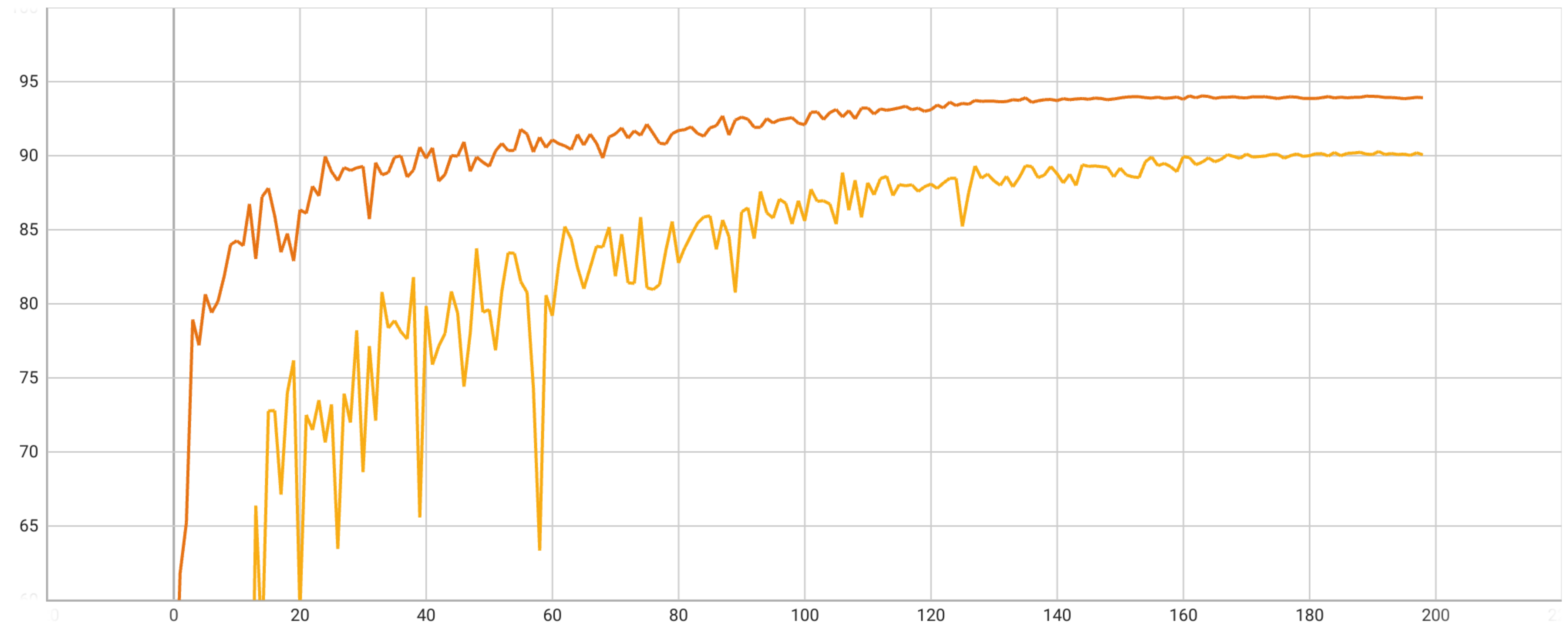}
  \caption{Comparison of classification accuracies (y-axis) w.r.t epochs (x-axis) for ResNet-18 models trained with Lipschitz regularization (yellow) applied to all layers (All) and a single Batch Normalization layer (OneBatchNorm) vs baseline (orange).}
  \label{fig:accuracies_baseline_vs_one_bn}
\end{figure}

\begin{figure}[h]
  \centering
  \includegraphics[width=1.0\textwidth]{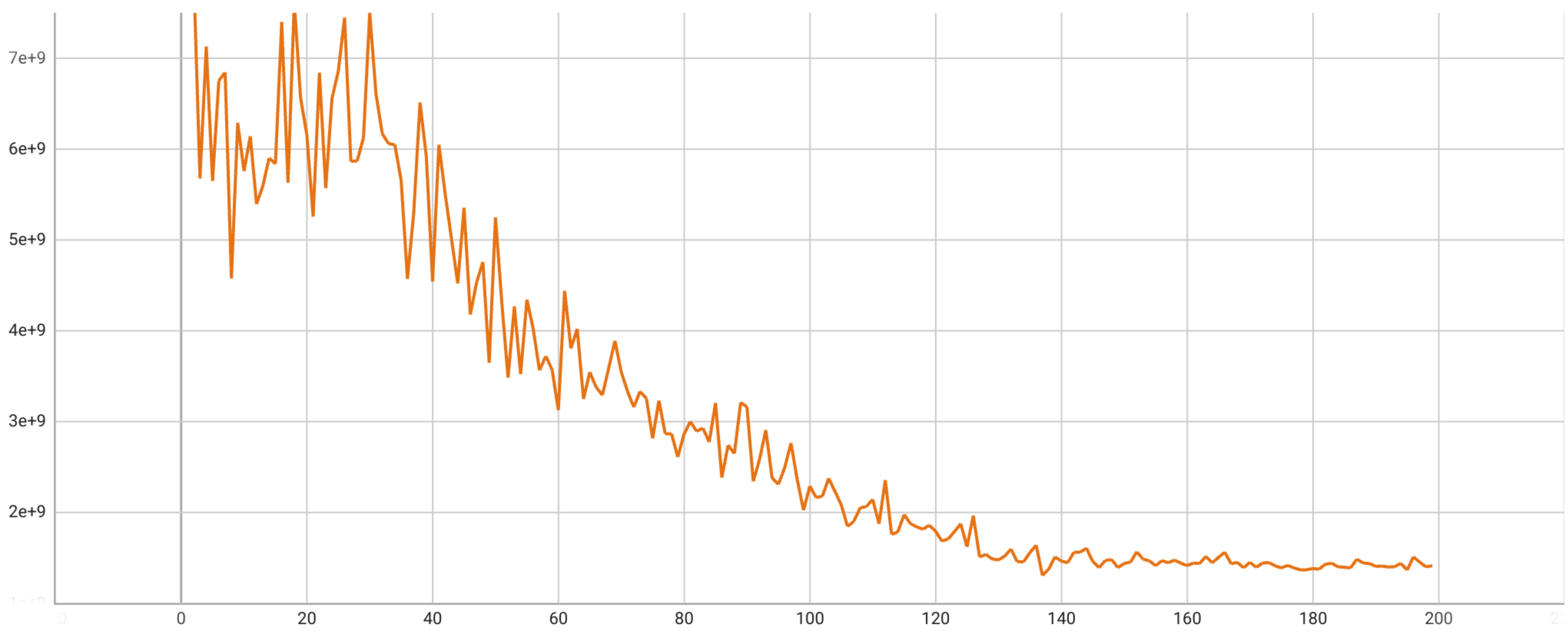}
  \caption{Evolution of the $\PUB$ (y-axis) over training epochs (x-axis) for ResNet-18 models with Lipschitz regularization applied to convolutional/dense layers and batch normalization after each convolutional layer (Baseline).}
  \label{fig:lip_over_epoch_baseline}
\end{figure}
\begin{figure}[h]
  \centering
  \includegraphics[width=1.0\textwidth]{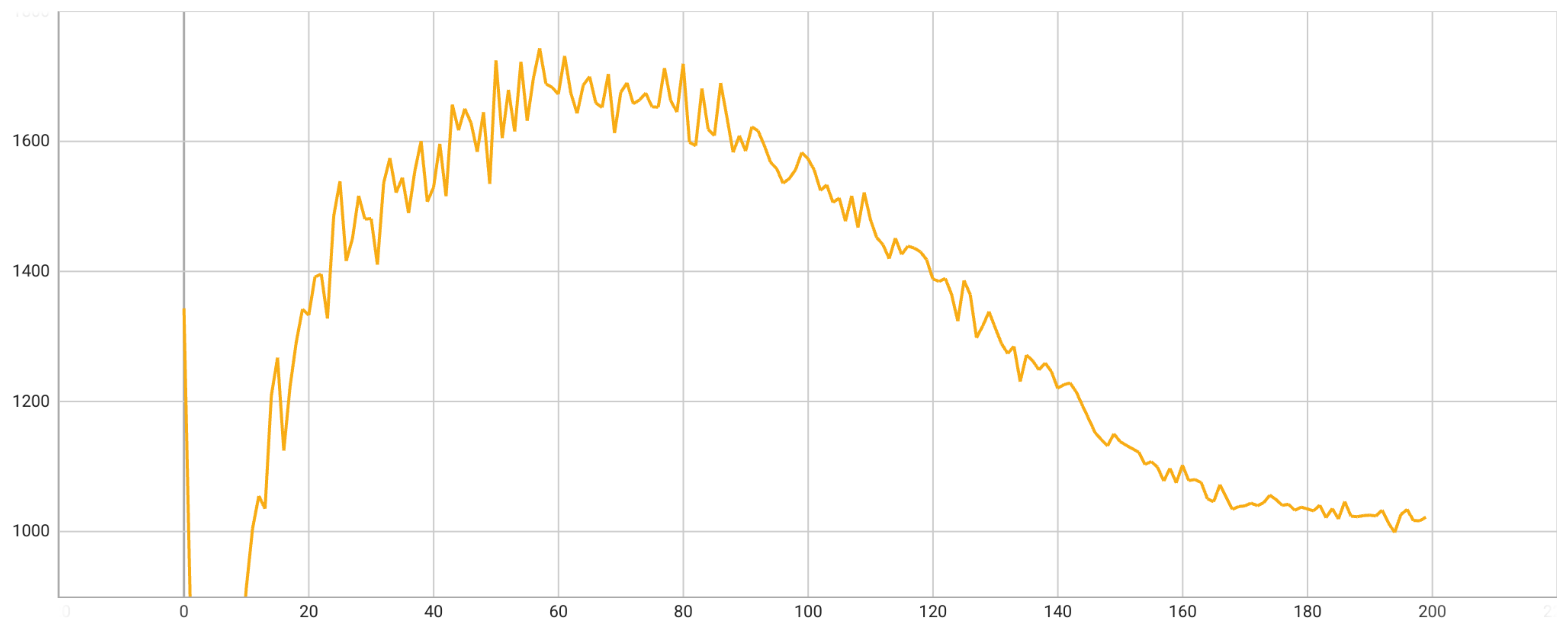}
  \caption{Evolution of the $\PUB$ (y-axis) over training epochs (x-axis) for ResNet-18 models with Lipschitz regularization applied to convolutional/dense layers and a single batch normalization layer after the last layer (OneBatchNorm).}
  \label{fig:lip_over_epoch_bn}
\end{figure}

\textbf{Training with Lipschitz-Regularized layers and batch normalization}
We train ResNet-18 models with Lipschitz regularization applied to all convolutional, dense, and batch normalization layers, ensuring Lipschitzness. However, training under these constraints proves highly unstable, as shown in Figure~\ref{fig:accuracy_reg_bn_and_all_linear_layers}. Batch normalization normalizes activations using statistics computed over the batch, which inherently introduces scale variations. When constrained by Lipschitz regularization, these fluctuations can become unpredictable, potentially causing unstable optimization dynamics. In the following, we try to remove batch normalization layers and replace them with Lipschitz regularization on linear layers.


\textbf{Training with Lipschitz-Regularized layers and reduced batch normalization}
Training deep neural networks without batch normalization or other normalization techniques is often unstable due to vanishing and exploding gradients~\citep{ioffe2015batch, ba2016layer}. As depth increases, activations and gradients can become poorly scaled, leading to slow convergence or divergence. Batch normalization mitigates these issues by normalizing activations across mini-batches, improving gradient flow, and enabling higher learning rates~\citep{santurkar2018does}. However, it also introduces a dependence on batch size, which can be problematic in scenarios such as training on memory-limited edge devices, distributed training with asynchronous updates, or when batch statistics become unreliable due to domain shifts or small dataset sizes.
Most CNN architectures incorporate batch normalization after each convolutional layer~\citep{he2016deep}. In such cases, Lipschitz regularization provides a promising alternative, as it stabilizes gradients without relying on batch-dependent normalization. By explicitly constraining the Lipschitz constant of each layer, this regularization method prevents gradients from exploding.

We investigate the impact of Lipschitz regularization on ResNet-18 models under two different normalization strategies: a standard batch normalization setup (Baseline) and a configuration with a single batch normalization layer (OneBatchNorm).
In the (Baseline setup), Lipschitz regularization is applied to all convolutional, dense, and batch normalization layers, following the standard ResNet-18 architecture where batch normalization is placed after each convolutional layer. This enforces strict Lipschitz constraints but leads to significant instability.
In contrast, (OneBatchNorm) enforces Lipschitz regularization  on the convolutional and dense layers, and structurally replaces all intermediate batch normalization layers by a single batch normalization layer placed just before the final dense (logit) layer. This design aims to preserve the global Lipschitz constraints while still benefiting from normalization, but avoids the instability that multiple batch normalization layers can introduce under Lipschitz control.
Figure~\ref{fig:accuracies_baseline_vs_one_bn} compares classification accuracies, showing that (OneBatchNorm) incurs an accuracy drop of approximately 3\% compared to (Baseline). However, as shown in Figures~\ref{fig:lip_over_epoch_baseline} and \ref{fig:lip_over_epoch_bn}, the evolution of the Lipschitz upper bound (\(\PUB\)) differs significantly.
In (OneBatchNorm), \(\PUB\) stabilizes around \( \approx 1000 \). In (Baseline), \(\PUB\) grows uncontrollably, reaching \( \approx 10^9 \), indicating severe instability.

While (OneBatchNorm) reduces instability, the accuracy drop and the difficulty in obtaining a tight Lipschitz constant suggest the need for alternative architectures with better gradient norm preservation.
Although Lipschitz constraints prevent gradient explosion, they do not inherently address gradient vanishing. This experiment highlights the importance of balancing Lipschitz regularization with architectural choices to ensure stable training and accurate control of the Lipschitz constant.

\section{Rescaling layers for Lipschitz networks}
\label{section:by_design_lipschitz_layers}
Unlike the previous sections, which focus on controlling the Lipschitz constant through explicit regularization during training, this section addresses a complementary strategy that enforces Lipschitz constraints directly by design.
We build upon existing Lipschitz layers such as the Spectral Lipschitz Layer (SLL), Convex Potential Layer (CPL), Almost Orthogonal Layer (AOL), and Spectral Normalization (SN) by introducing a new approach: spectral rescaling (SR). Spectral Normalization, as introduced by \citet{miyato2018spectral}, divides all singular values of a weight matrix by its largest singular value, ensuring the spectral norm is exactly one. However, this approach squashes smaller singular values towards zero, resulting in an ill-conditioned matrix.
The Almost Orthogonal Layer (AOL), proposed by \citet{prach2022almost}, mitigates this issue by applying a diagonal rescaling matrix \(\mR\) such that the spectral norm of \(\mW \mR\) is bounded by one. While this improves the condition number compared to Spectral Normalization, the rescaling is often overly conservative, with the spectral norm \(\|\mW \mR\|_2\) significantly lower than one.

To address these limitations, we propose spectral rescaling (SR), a tight rescaling method that interpolates between Spectral Normalization and AOL. Our method converges to a spectral norm of one while maintaining a good condition number, offering improved stability and precision. The rescaling is achieved by generalizing the Gram matrix \(\mW^\top \mW\) to its \(t\)-th iterate, providing a more flexible and effective approach to constructing Lipschitz layers.
We note the $t$-th Gram iterate matrix, $\mW^{(t +1)} = {\mW^{(t)}}^\star \mW^{(t)}$  and $\mW^{(0)}=\mW$, is computed using algorithms described in Chapter~\ref{chapter:spectral_norm_estimation}.
\begin{theorem}[Spectral rescaling for matrices]
  \label{thm:improve_aol}
  For any $\mW \in \mathbb{R}^{n \times d}$, integer $t \geq 1$,
  define $\mW^{(t+1)} = {\mW^{(t)}}^\top \mW^{(t)}$, with $\mW^{(0)} = \mW$.
  We define the spectral rescaling
  $\mR^{(t)}$ as the diagonal matrix with
  $\mR^{(t)}_{ii} = \left( \sum_j \left| \mW^{(t+1)} \right|_{ij} \right)^{-2^{-(t+1)}}$ if the expression in the brackets is non zero, or $\mR_{ii} = 0 $ otherwise.

  Then  $\sigma_1(\mW \mR^{(t)}) \underset{t \to +\infty}{\longrightarrow} 1$, with super geometric convergence  and the iterates upper bound the limit $\sigma_1(\mW \mR^{(t)}) \leq 1$.
\end{theorem}
See proof in Appendix~\ref{app:sec:proof_improve_aol}.
Using previous Theorem~\ref{thm:improve_aol} and Theorem~1 of \citep{araujo2023a}, we can design the following two $1$-Lipschitz layers.
%

\begin{corollary}[Spectral rescaling for $1$-Lipschitz layers]
  \label{corol:layers_SR}
  Under the notations of Theorem~\ref{thm:improve_aol}, let \(\mQ = \mathrm{diag}(q_i)\) with \(q_i > 0\), and fix \(t \geq 0\). Define the spectral rescaling matrix \(\mR^{(t)}\) by
  \begin{align*}
    \mR^{(t)}_{ii} \;=\;
    \begin{cases}
      \Big( \sum\nolimits_j \, \big|\mW^{(t)}\big|_{ij} \,\tfrac{q_i}{q_j} \Big)^{-2^{-t}}, &
      \text{if the sum is nonzero},                                                                             \\[6pt]
      0,                                                                                    & \text{otherwise}.
    \end{cases}
  \end{align*}
  Then, for any \(t \geq 0\), the following mappings are \(1\)-Lipschitz:
  \begin{itemize}
    \item the affine layer \(\vx \mapsto \mW \mR^{(t)} \vx + \vb\),
    \item the residual layer \(\vx \mapsto \vx - 2 \mW (\mR^{(t)})^2 \, \nonlin(\mW^\top \vx + \vb)\),
  \end{itemize}
  where \(\vb \in \R^d\) is a bias vector, \(\mW\) is the weight matrix, and \(\nonlin\) is any \(1\)-Lipschitz activation (e.g., \(\relu\), \(\mathrm{tanh}\), \(\mathrm{sigmoid}\)).
\end{corollary}

Spectral rescaling is differentiable so training is possible through the rescaled layer as a whole.
The introduction of the diagonal matrix $\mQ = \mathrm{diag}(q_i)$ adds flexibility to the spectral rescaling by enabling anisotropic control over the activation dimensions. When $\mQ$ is learned jointly with the model parameters, it enhances the expressivity of the rescaling operation while preserving $1$-Lipschitzness through the construction in Corollary~\ref{corol:layers_SR}. Each learned rescaling coefficient $q_i > 0$ can adaptively emphasize or attenuate certain directions, allowing the network to better align with the data geometry under the Lipschitz constraint.
Previous Corollary~\ref{corol:layers_SR} can be applied to convolutional layers using Algorithm~\ref{algo:gram_iteration_toep} to compute the rescaling.

Note that for $t=1$, SR is equivalent to $\AOL$ \citep{prach2022almost} (see Equation~\ref{eq:DAOL}).
As $t \to \infty$ the SR is more similar to SN,
indeed the stable rank ${\norm{ \mW^{(t)}}_{\frob}}^2 / \|\mW^{(t)}\|_2^2$ of the Gram iterate matrix $\mW^{(t)}$,
is closer and closer to one.
The Gram iterate for large iteration number $t$ is numerically a rank one matrix and the rescaling is equivalent to spectral normalization. Therefore, our method is an interpolation between the AOL rescaling and the $\SN$ rescaling, the number of iterations $t$ is the cursor which allows us to select at which degree the rescaling is more $\AOL$ or $\SN$.
Note that $\SN$ used on the $\SLL$ block of \citep{araujo2023a} retrieves the $\CPL$ block of \citep{meunier2022dynamical}.

\begin{figure}
  \centering
  \includegraphics[width=0.8\textwidth]{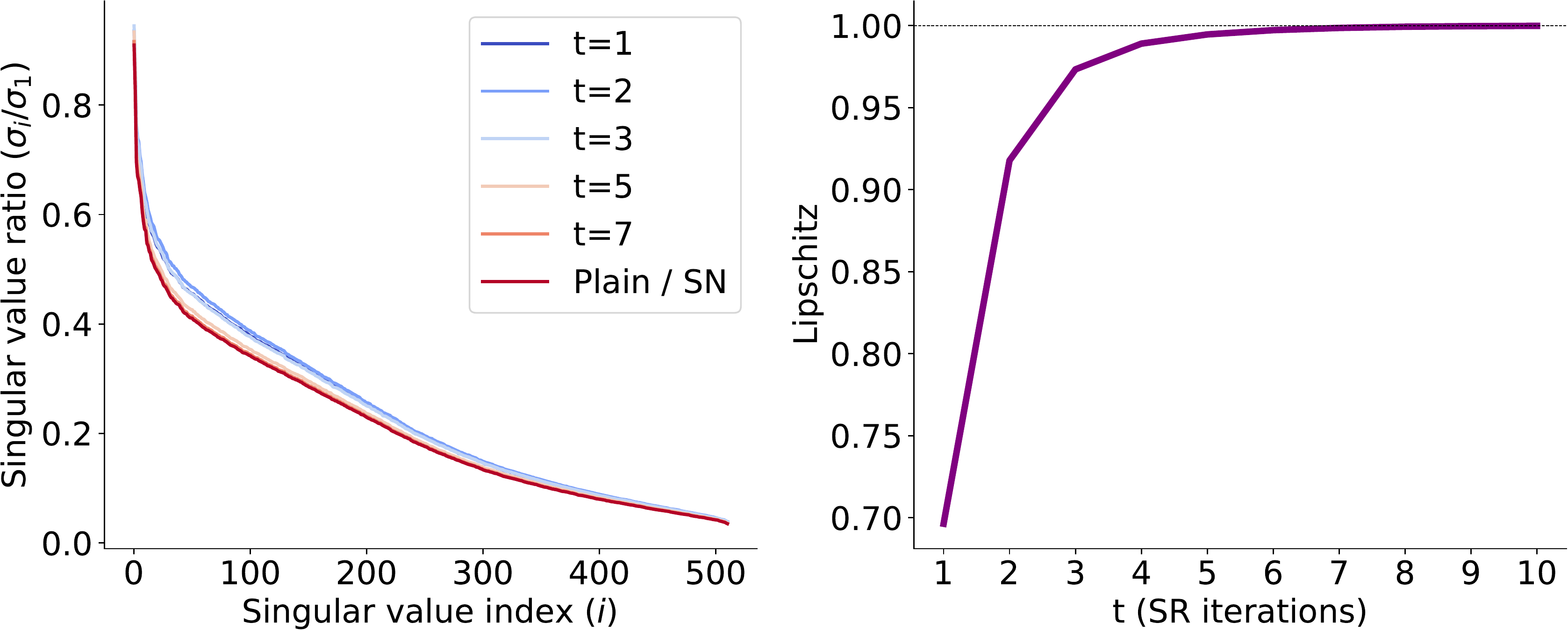}
  \caption{On the left, the ratio of \(\frac{\sigma_i}{\sigma_1}\) for a dense layer from ResNet18 ($512 \times 1000$) trained on the ImageNet-1k dataset with different rescaling methods.
    On the right, the convergence of SR to a 1-Lipschitz layer is depicted.}
  \label{fig:condition_number_sr}
\end{figure}

One main advantage of our method over $\SN$ done with power iteration is that it provides a guarantee that the layer is $1$-Lipschitz as the rescaling is upper bounding the singular values.
As we are rescaling, the Lipschitz constant is always lower than $1$, see Figure~\ref{fig:condition_number_sr} the right figure depicts the convergence of SR to a Lipschitz constant of 1 as iteration increases.
We see empirically that power iteration (with $100$ iterations) can be a loose lower bound on spectral norm as depicted in Figure~\ref{fig:accuracy_time_conv_spectral_norm}. Moreover, $\SR$ is deterministic and converges super geometrically, whereas SN with power iteration is random and converges linearly.

We observe that the ratio of singular values \(\frac{\sigma_1}{\sigma_i}\) is systematically lower for SR compared to \(\AOL\) and \(\SN\) (Figure~\ref{fig:condition_number_sr}, left).
This suggests that the singular values are more evenly distributed, meaning that transformations induced by SR are more isotropic than those of \(\AOL\) and \(\SN\). This effect reduces average directional distortions in feature space, potentially leading to more stable representations.

\section{Results on regularity of the Weierstrass transform}
In this section, we present new results on the regularity properties of the Weierstrass transform when the underlying network is Lipschitz continuous. This is particularly relevant for networks designed to be Lipschitz by construction where we would perform on top Weierstrass transform to smooth the network. Also, as we see in the previous section~\ref{section:local_lipschitz} the Lipschitz constant of vanilla networks such as ResNet is overestimated by the product of upper bounds and the true Lipschitz constant is probably much smaller. Incorporating this piece of information can help to derive tighter bounds on the Lipschitz constant of the smoothed network.

In the Lemma~\ref{lemma:lip_weierstrass_transform} of \citet{salman2019provably}, the Lipschitz constant of the Weierstrass transform is bounded by the Lipschitz constant of the smoothed function $f$ and the standard deviation $\sigma$ of the Gaussian kernel used for smoothing.
Lemma~\ref{lemma:lip_weierstrass_transform_lipschitz} states that the Lipschitz constant of the Weierstrass transform
$\widetilde{f} = f * \phi_{\sigma}$
is bounded by the Lipschitz constant  function $f$. Note that in each case the two quantities $\sigma$ and $\Lip(f)$ do not interact in the bound.
We derive enhanced bounds on the Lipschitz constant of  $\widetilde{f}$ with the additional assumption that $f$ is Lipschitz continuous. Using the same notation and assumption as previous lemmas we have the following result:
\begin{theorem}[Lipschitz bound for the Weierstrass transform with based Lipschitz continuity]
  \label{thm:thm_bound_lip_sigma_element_wise}
  Suppose that $f : \R^d \to [0, 1]$
  is Lipschitz continuous, then
  \begin{align}
    \label{eq:bound_lip_sigma_element_wise}
    \Lip({\widetilde{f}})
    \leq \Lip({ f})\operatorname{erf}\left(\frac{1}{2^\frac{3}{2} \Lip({f}){\sigma}}\right) \ .
  \end{align}
\end{theorem}
See proof in Appendix~\ref{app:sec:proof_bound_lip_sigma_element_wise}.
We can also derive the following bound which depends only on $\sigma$ and $\Lip(f)$ independently:
\begin{corollary}(Lipschitz Bounds on the Weierstrass transform)
  \label{corol:bound_lip_sigma}
  Suppose that $f : \R^d \to [0, 1]$ is Lipschitz continuous, then
  \begin{align}
    \label{eq:corol_bound_lip_ftilde_sigma}
    \Lip({\widetilde{f}})
    \leq \frac{1}{\sqrt{2\pi\sigma^2}} \
  \end{align}
  and also
  \begin{align}
    \label{eq:corol_bound_lip_ftilde_lip}
    \Lip({\widetilde{f}})
    \leq \Lip(f) \ .
  \end{align}
\end{corollary}
The smoothed function $\widetilde{f}$ is at least as regular as the original function $f$, also it is noteworthy that Equation~\ref{eq:corol_bound_lip_ftilde_sigma} enhances the bound
on \(\Lip(\widetilde{f})\) originally derived in Lemma~\ref{lemma:lip_weierstrass_transform} of~\citet{salman2019provably} by a factor of 2.
This refinement on the bound was possible by supposing Lipschitz continuity on the function $f$. Note that its Lipschitz constant can be arbitrarily high, so this assumption is quite light: the Lipschitz constant does not play into the derived bound.
In the proof we demonstrate that the bound is optimal as it exists a function $f$ such that the bound is an equality.
These improved bounds can be seamlessly incorporated into subsequent works, such as \citep{pautov2022smoothed, franco2023diffusion, chen2024diffusion}.

We observe that the Weierstrass transform and Lipschitz continuity exhibit a synergistic effect on the Lipschitz constant of the smoothed function \(\tilde{f}\).
In particular, there exists an intermediate regime, characterized by a specific relationship between \(\sigma\) and \(\Lip(f)\), where these effects combine to reduce the Lipschitz constant of \(\tilde{f}\) more than either mechanism would individually.
To formalize this, we define the gap function
\[
  \Delta(\sigma)
  =
  \min\bigl\{\Lip(f), \,\frac{1}{\sqrt{2\pi}\,\sigma}\bigr\}
  \;-\;
  \Lip(f) \,\mathrm{erf} \Bigl(\frac{1}{2^{3/2} \Lip(f) \sigma}\Bigr),
\]
which quantifies the improvement provided by considering the refined Weierstrass bound over the independent Lipschitz and smoothing bounds.

\begin{proposition}
  \label{prop:sigma_star_lip}
  Let \(f : \R^d \to [0, 1]\) be a Lipschitz continuous function.
  The optimal smoothing parameter that maximizes this gap, i.e.
  \[
    \sigma^*
    = \argmax_{\sigma > 0} \Delta(\sigma),
  \]
  is given explicitly by
  \[
    \sigma^* = \frac{1}{\Lip(f) \sqrt{2\pi}}
    \quad \Longrightarrow \quad
    \Lip(\tilde f)
    \;\le\;
    \mathrm{erf}\bigl(\tfrac{\sqrt\pi}{2}\bigr)\,\Lip(f)
    \;\approx\;
    0.79\,\Lip(f).
  \]
\end{proposition}
See proof in Appendix~\ref{app:proof_sigma_star_lip}.
For this choice of \(\sigma^*\), we also have that the randomized smoothing bound exactly meets the deterministic Lipschitz bound, since
\[
  \frac{1}{\sqrt{2\pi}\,\sigma^*}
  = \Lip(f).
\]
This reveals a principled trade-off: by selecting \(\sigma\) in relation to the intrinsic Lipschitz constant \(\Lip(f)\), or by designing \(\Lip(f)\) to match a target smoothing scale, we maximize the synergistic reduction captured by \(\Delta(\sigma)\).
Consequently, this combination can reduce the certified Lipschitz bound by up to \(21\%\).
This provides a practical guideline: given either a desired smoothing level or an inherent Lipschitz constraint, one can tune the complementary parameter to maximize the benefits of Lipschitz control and Weierstrass smoothing captured through \(\Delta(\sigma)\).

To illustrate the impact of this result, we consider the example of the smoothed variant of binary probit regression where Gaussian noise is injected into the input during inference. This technique enhances the robustness of the model by averaging predictions over perturbations of the input, mitigating sensitivity to adversarial noise.
Logistic regression is a linear model for binary classification that predicts the probability of an input \( \vx \in \mathbb{R}^d \) belonging to the positive class. It models the relationship between the input features and the output class using the sigmoid function. The prediction is given by:
\[
  f(\vx) = \frac{1}{1 + e^{- (\vw^\top \vx + b) }},
\] where \( \vw \in \mathbb{R}^{d} \) is the weight vector and \( b \in \mathbb{R} \) is the bias. This output represents the probability of the positive class \( y = 1 \), while \( 1 - f(\vx) \) corresponds to the probability of the negative class \( y = 0 \).
The sigmoid function \( s(z) = \frac{1}{1 + e^{-z}} \) ensures that the output is bounded between 0 and 1, making it interpretable as a probability. Logistic regression is typically trained by minimizing the cross-entropy loss, which measures the difference between the predicted probabilities and the ground truth labels.

\begin{example}[Example of smoothed binary probit regression]
  \label{example:smoothed_linear_logistic_regression}
  Consider a linear function \( x \mapsto \vw^\top \vx + b \) with \( \vw \in \mathbb{R}^{d} \) and \( b \in \mathbb{R} \).
  The smoothed probit regression model is defined as:
  \[
    \tilde{f}(\vx) = \E_{\delta \sim \mathcal{N}(0, \sigma^2 \mI)}[s(\vw^\top (\vx + \delta) + b)],
  \]
  where \( s(z) \) is an approximation of the sigmoid function, expressed as:
  \[
    s(z) = \Phi(\lambda z),
  \]
  with \( \Phi \) representing the Gaussian cumulative distribution function. Choosing \( \lambda = \sqrt{\frac{\pi}{9}} \) or \( \lambda = \sqrt{\frac{\pi}{5.35}} \) provides a close approximation to the standard sigmoid \( s(z) = \frac{1}{1 + e^{-z}} \). These values minimize approximation error, ensuring that the smoothed sigmoid closely matches the standard sigmoid across various inputs.

  The closed-form solution for the smoothed output, following~[Eq. 4.152]\citet{bishop2006pattern}, is:
  \[
    \tilde{f}(\vx) = \Phi\left(\frac{\lambda (\vw^\top \vx + b)}
      {(1 + \lambda^2 \sigma^2 \|\vw\|^2)^{\frac{1}{2}}}
    \right).
  \]
  This formulation demonstrates how smoothing with Gaussian noise yields an analytical solution, providing robustness against perturbations and adversarial inputs.
\end{example}
\begin{figure}
  \centering
  \includegraphics[width=0.7\textwidth]{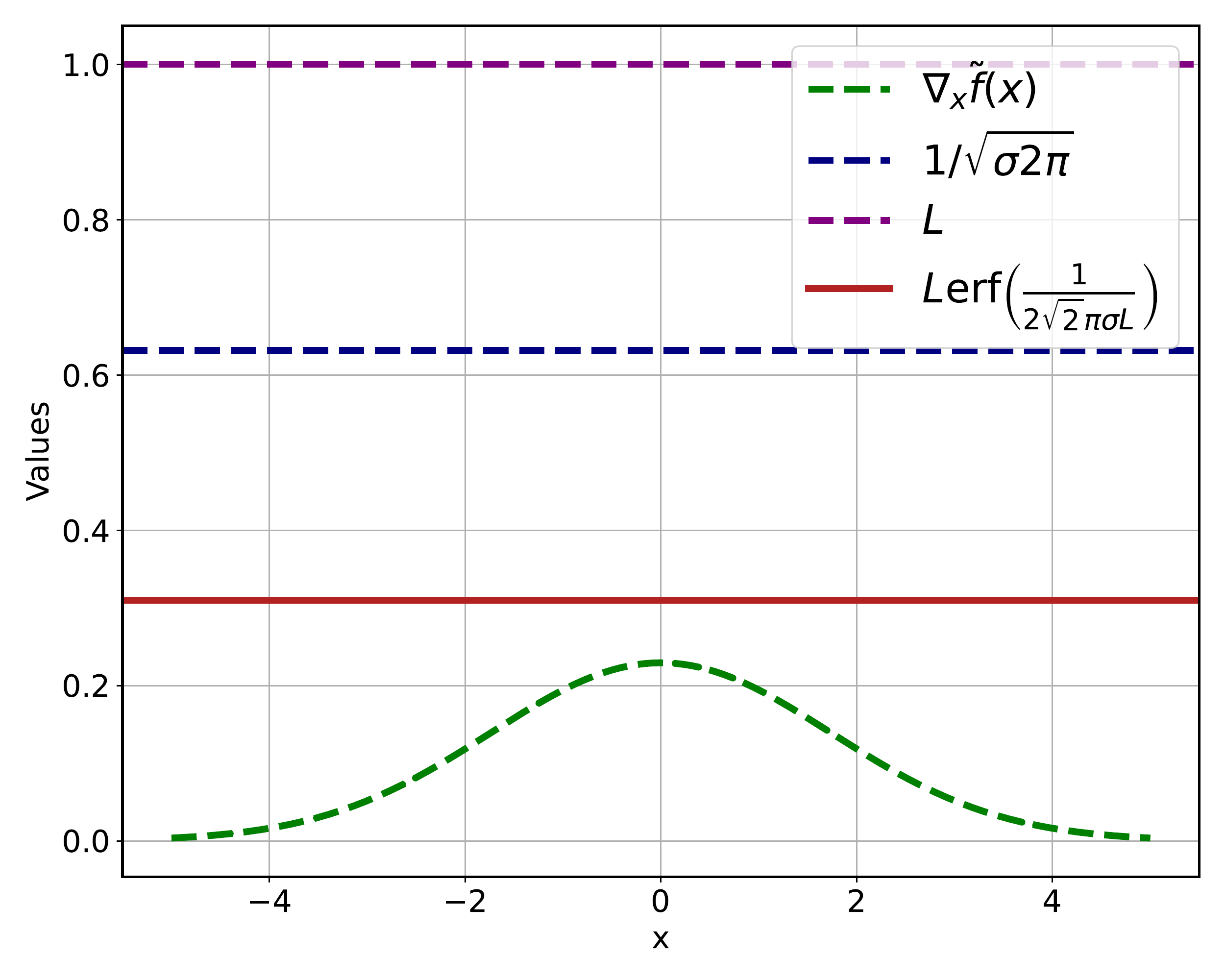}
  \caption{Plot of the gradient and bounds of function $\widetilde{f}_k$ for a smoothed probit regression model with $d=1$ and $\sigma = 0.2$ and $\Lip(f) = 1$.}
  \label{fig:smoothed_linear_regression}
\end{figure}
Figure~\ref{fig:smoothed_linear_regression} shows the gradient and bounds of the smoothed function $\widetilde{f}_k$ for a smoothed probit regression model, see Example~\ref{example:smoothed_linear_logistic_regression}, with $\Lip(f) = 1$ and associated $\sigma^\star = \frac{1}{\sqrt{2 \pi}}$ .
The bounds are computed using the Lipschitz constant of the original function $f$ and the RS one with standard deviation $\sigma$ of the Gaussian noise. We observe that our bound $\Lip({\widetilde{f}})$ is tighter around the gradient of the smoothed function, demonstrating the effectiveness of the bound using the Lipschitz constant mixed with smoothing standard deviation in controlling the regularity of the smoothed function $\tilde{f}$.

We derive a similar result for the Lemma~\ref{lemma:lip_quantile_smoothed_classifier} of \citep{salman2019provably} by incorporating Lipschitz continuity assumptions on the network. This result provides a tighter bound on the Lipschitz constant of the quantile-composed smoothed network.
The following theorem provides a key result for deriving certified radii under the assumption of Lipschitz continuity on the function \( f \) for the randomized smoothing framework.
\begin{theorem}[Weierstrass transform with local Lipschitz continuity]
  \label{thm:lip_quantile_smoothed_classifier_ours}
  The function \( \quant \circ \tilde{f} \) is locally Lipschitz continuous within the ball
  \( B(\vx, \budget) = \{ \vx^\prime \in \mathbb{R}^d : \lVert \vx^\prime - \vx \rVert_2 \leq \budget \} \), with Lipschitz constant:
  \begin{align*}
    \Lip\left( \quant \circ \tilde{f}, B(\vx, \budget) \right)
    \leq \Lip(f)
    \sup_{\vx^\prime \in B(\vx, \budget)}
    \left\{
    \frac
    {
      \Phi_\sigma\left(
      s_0(\vx^\prime)
      + \dfrac{1}{\Lip(f)}
      \right)
      - \Phi_\sigma\left(s_0(\vx^\prime)\right)
    }
    {
      \phi(\quant(\tilde{f}(\vx^\prime)))
    }
    \right\} \
  \end{align*}
  where \( \Phi_\sigma \) is the cumulative distribution function of the normal distribution with standard deviation \( \sigma \), $\phi$ is the standard Gaussian density function,
  and \( s_0(\vx^\prime) \) is determined
  by solving:
  \[
    \tilde{f}(\vx^\prime) = 1 - \Lip(f) \int_{s_0(\vx^\prime)}^{s_0(\vx^\prime) + \frac{1}{\Lip(f)}} \Phi_\sigma(s) \, ds \ .
  \]
\end{theorem}
See proof in Appendix~\ref{app:sec:proof_lip_quantile_smoothed_classifier_ours}.
Incorporating the Lipschitz constant of $f$ gives a local Lipschitz constant for the quantile-composed smoothed classifier. This result provides a tighter bound on the Lipschitz constant of the quantile-composed smoothed network.
Here the choice of an optimal $\sigma^\star$ is more complex than in the previous case, as it depends on the Lipschitz constant is local and it depends on $\vx$ and radius $\epsilon$.
\begin{figure}[h]
  \centering
  \includegraphics[width=0.7\textwidth]{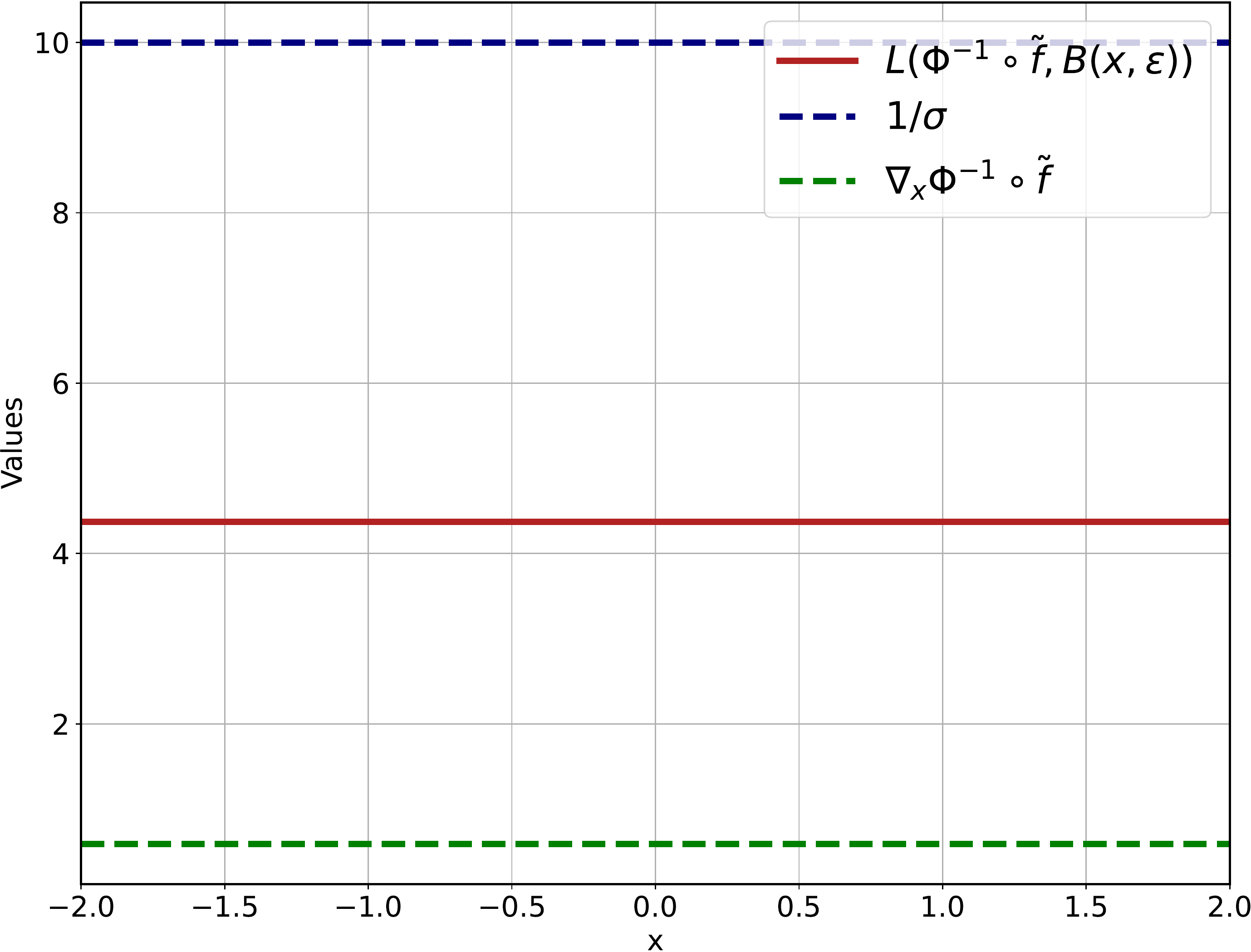}
  \caption{Plot of the gradient and bounds of function $\quant \circ \widetilde{f}_k$ for a smoothed probit regression model with $d=1$ and $\sigma = 0.1$ and $\Lip(f) = 1$.}
  \label{fig:smoothed_linear_regression_phi_inverse}
\end{figure}
Same as before we can compute the expectation under Gaussian noise injected over input exactly for a probit regression model, see Example~\ref{example:smoothed_linear_logistic_regression}.
Figure~\ref{fig:smoothed_linear_regression_phi_inverse} shows the gradient and bounds of the smoothed function $\quant \circ \widetilde{f}_k$ for a smoothed probit regression model with $d=1$, $\Lip(f) = 1$ and associated $\sigma = 0.1$.

\section{Conclusion}
This chapter presented an in-depth exploration of Lipschitz networks, focusing on methods to estimate, regularize, and design neural architectures with controlled Lipschitz constants. We introduced the Product Upper Bound ($\PUB$) as an efficient technique for estimating the Lipschitz constant of convolutional networks, demonstrating its accuracy and scalability.

To further constrain the Lipschitz constant during training, we investigated regularization strategies applied to convolutional layers, showing that precise spectral control enhances network stability and reduces variability. We also proposed spectral rescaling (SR), a novel method that interpolates between existing normalization techniques, providing tighter bounds and better conditioning for deep networks.

Additionally, we explored the theoretical properties of the Weierstrass transform, highlighting its role in smoothing neural networks. We examined the interaction between Lipschitz continuity and the Weierstrass transform, demonstrating how their combined effect can lead to tighter bounds on the Lipschitz constant of smoothed networks.

While our work has focused on estimating, regularizing, and designing Lipschitz networks, another perspective highlights the critical role of loss functions in shaping the accuracy-robustness trade-off~\citep{bethune2022pay}. This suggests that beyond architectural constraints, optimization choices and loss design play a fundamental role in fully leveraging the benefits of Lipschitz-constrained models.
A deeper understanding of these interactions could bridge the gap between theoretical guarantees and practical performance, paving the way for models that are both provably robust and highly expressive.
The techniques developed in this chapter contribute to this broader effort by addressing key aspects of network stability and expressivity, with direct implications for robustness against adversarial perturbations.
In the next chapter, we apply these tools developed in the current chapter to certified robustness, demonstrating how bounding the Lipschitz constant provides formal guarantees on adversarial resilience and connects theoretical stability measures to practical security concerns in deep learning.

%% file: content/chapter-robustness.tex
%
\chapter{Robustness through Lipschitz networks}\label{chapter:robustness}


\minitoc%

Deep learning models are highly susceptible to adversarial attacks, where imperceptible perturbations to input data can drastically alter model predictions~\citep{szegedy2013intriguing}. This vulnerability raises significant concerns for the deployment of neural networks in safety-critical applications such as autonomous driving, medical diagnosis, and security-sensitive systems. Addressing these concerns requires robust models capable of certifying their predictions against adversarial perturbations~\citep{katz2017reluplex}.
Lipschitz continuity provides a theoretical foundation for quantifying and enforcing robustness by controlling how much a model's output can change in response to input variations. Constraining the Lipschitz constant has been shown to enhance robustness against adversarial perturbations~\citep{tsuzuku2018lipschitz}. However, enforcing these constraints in deep architectures remains challenging due to computational difficulties in estimating and maintaining a tight Lipschitz bound.
Certified robustness techniques, such as randomized smoothing, offer a probabilistic guarantee against adversarial attacks by constructing smoothed classifiers that are inherently more stable. However, a critical limitation of current certification methods is that the obtained certified robustness is significantly lower than the empirical robustness observed under adversarial attacks such as projected gradient descent (PGD)~\citep{croce2021robustbench, cohen2019certified}. This gap between theoretical guarantees and empirical observations limits the practical effectiveness of certification techniques and raises questions about their applicability in real-world scenarios.

In this chapter, we evaluate the certified robustness of Lipschitz networks under both deterministic and non-deterministic settings. We first assess robustness using Spectrally Rescaled (SR) layers, then extend the analysis to randomized smoothing with the Weierstrass transform, deriving new certified radius bounds incorporating Lipschitz constraints. This analysis aims to bridge the gap between theoretical robustness guarantees and practical improvements, highlighting the interplay between Lipschitz regularization and smoothing techniques.
\section{Spectrally rescaled layers for certified robustness}
%
Lipschitz networks improve robustness by explicitly constraining the Lipschitz constant of each layer, ensuring that the overall Lipschitz bound remains controlled through the product upper bound (\(\PUB\)).
A notable property of Lipschitz networks is their stability during training: a low Lipschitz constant mitigates gradient explosion and stabilizes optimization dynamics~\citep{kodali2018on}.
Indeed, unlike standard networks, Lipschitz architectures inherently reduce the need for batch normalization, layer normalization, or gradient clipping, as their controlled gradients naturally promote stable convergence.
However, enforcing a Lipschitz constraint can also introduce challenges: the contractive nature of such layers may lead to vanishing gradients, thereby limiting the effective depth of trainable networks.
An important property to counteract this issue is norm preservation, which helps mitigate gradient attenuation and is crucial for training deep Lipschitz networks~\citep{anil2019sorting}.

Residual connections are essential for addressing this issue, as they facilitate stable gradient propagation and mitigate vanishing gradients. As demonstrated by~\citet{he2016deep} with ResNet, residual layers enable the training of deeper networks by preserving gradient flow across layers. This property is particularly valuable for Lipschitz networks, where the controlled weight magnitudes further ensure stability. Additionally, residual architectures have been shown to boost performance, as highlighted by~\citet{araujo2023a}.

In this work, we employ $\SLL$ layers which is a Lipschitz residual layer, which requires rescaling to enforce the \(1\)-Lipschitz constraint. The baseline method utilizes Almost Orthogonal Layers ($\AOL$)~\cite{prach2022almost} for rescaling, while we introduce Spectral Rescaling ($\SR$) to improve both robustness and accuracy.
As detailed in Corollary~\ref{corol:layers_SR}, the rescaling matrix \( \mR \) is defined by:
\[
    \mR^{(t)}_{ii} = \left( \sum_j \left| \mW^{(t)} \right|_{ij} \frac{\vq_i}{\vq_j} \right)^{-2^{-t}} \quad \text{if the sum is non-zero}, \quad \text{otherwise} \quad \mR^{(t)}_{ii} = 0.
\]
Using the Corollary~\ref{corol:layers_SR}, for any layer \( 1 \leq l \leq \nblayers \), the following residual structure is \(1\)-Lipschitz:
\[
    f^{(l)}(\vx) = \vx - 2 \mW {(\mR^{(t)})}^{2} \nonlin(\mW^\top \vx + \vb),
\]
where \( \mW \) is the weight matrix, \( \vb \) the bias vector, and \( \nonlin = \relu \) is the activation function.

As each layer for $1 \leq l \leq \nblayers$ is $1$-Lipschitz, the overall network is also $1$-Lipschitz, thus
\[
    \Lip(f) = \prod_{l=1}^{\nblayers} \Lip(f^{(l)}) = 1 \ .
\]
We can then use the certified radius bound from~\citet{tsuzuku2018lipschitz}, see Equation~\ref{eq:radius_tsuzuku}, to compute the certified robustness of the network and the certified accuracy on the validation set.

\paragraph{Experimental setup and results.}
We evaluate the robustness of Lipschitz networks designed with $\SLL$ layers by comparing AOL and SR rescaling methods. Following the experimental setup of \citet{araujo2023a}, we conduct experiments on CIFAR-10 and CIFAR-100 datasets.
Training is performed over $200$ epochs with a batch size of $256$, using standard data augmentation techniques. The networks are optimized with
Adam~\citep{kingma2017adam}, with a learning rate of 0.01 and parameters $\beta_1 = 0.5$ and $\beta_2 = 0.9$, without weight decay. A piecewise triangular learning rate scheduler is applied to adjust the learning rate during training.
The loss function follows the CrossEntropy formulation used by \citet{prach2022almost}, with a temperature of $0.25$ and an offset value of \( \frac{3}{2} \sqrt{2} \). This ensures consistent evaluation across architectures and datasets, providing a reliable comparison between rescaling methods. To evaluate the scalability of $\SLL$ $\SR$ based Lipschitz networks, we consider two architectures: small and medium. The architectures are defined as: small (S): 20 convolutional layers, 7 dense layers, and medium (M): 30 convolutional layers, 10 dense layers.

\paragraph{Certified performance.}
Tables~\ref{tab:evaluation_results_cifar10} and \ref{tab:evaluation_results_cifar100} summarize natural and certified accuracies (see Definition~\ref{def:certified_accuracy}) across perturbation levels \( \epsilon \), averaged over three training runs. $\SR$ consistently outperforms $\AOL$ across all architectures and datasets, showcasing enhanced scalability and robustness. The tighter Lipschitz bounds achieved through $\SR$ layers improve network stability and certifiability.

\begin{table}[h]
    \centering
    \caption{Certified accuracy on CIFAR-10 for different perturbation levels $\epsilon$, comparing AOL and SR rescaling methods for small (S) and medium (M) architectures.
        Mean values are reported over 3 runs, standard deviation is omitted for brevity and is bounded by $0.12$ for all configurations.
    }
    \label{tab:evaluation_results_cifar10}
    \begin{tabular}{lcccccc}
        \toprule
        \multirow{2}{*}{\textbf{Rescaling}} & \textbf{Accuracy} & \multicolumn{4}{c}{\textbf{Certified accuracy ($\epsilon$)}}                                                    \\
        \cmidrule(lr){3-6}
                                            &                   & $0.141$                                                      & $0.283$        & $0.423$        & $1$            \\
        \midrule
        \text{AOL (S)}                      & 71.06             & 62.78                                                        & 53.67          & 45.37          & 19.18          \\
        \text{SR (S)}                       & \textbf{72.44}    & \textbf{63.49}                                               & \textbf{54.66} & \textbf{46.01} & \textbf{19.62} \\
        \midrule
        \text{AOL (M)}                      & 72.41             & 63.72                                                        & 54.48          & 46.38          & 19.92          \\
        \text{SR (M)}                       & \textbf{73.38}    & \textbf{64.64}                                               & \textbf{55.43} & \textbf{46.79} & \textbf{20.18} \\
        \bottomrule
    \end{tabular}
\end{table}

\begin{table}[h]
    \centering
    \caption{Certified accuracy on CIFAR-100 for different perturbation levels $\epsilon$, comparing AOL and SR rescaling methods for small (S) and medium (M) architectures.
        Mean values are reported over 3 runs, standard deviation is omitted for brevity and is bounded by $0.13$ for all configurations.
    }
    \label{tab:evaluation_results_cifar100}
    \begin{tabular}{lcccccc}
        \toprule
        \multirow{2}{*}{\textbf{Rescaling}} & \textbf{Accuracy} & \multicolumn{4}{c}{\textbf{Certified accuracy ($\epsilon$)}}                                                    \\
        \cmidrule(lr){3-6}
                                            &                   & $0.141$                                                      & $0.283$        & $0.423$        & $1$            \\
        \midrule
        \text{AOL (S)}                      & 46.05             & 35.13                                                        & 26.72          & 20.49          & 0.073          \\
        \text{SR (S)}                       & \textbf{46.62}    & \textbf{35.46}                                               & \textbf{27.42} & \textbf{21.14} & \textbf{0.075} \\
        \midrule
        \text{AOL (M)}                      & 46.50             & 35.81                                                        & 27.39          & 21.21          & 7.78           \\
        \text{SR (M)}                       & \textbf{47.50}    & \textbf{36.50}                                               & \textbf{28.54} & \textbf{21.89} & \textbf{8.13}  \\
        \bottomrule
    \end{tabular}
\end{table}

Spectral rescaling consistently enhances certified robustness and network stability across datasets and architectures, offering a scalable solution for certifiably robust networks.

\section{Certified robustness with randomized smoothing of Lipschitz networks}
\label{ssec:expe_rs_new_bound}

In the background sections,
we discussed the two main approaches for certifying robustness: deterministic, see Section~\ref{sec:lipschitz_networks_certified_robustness}, and probabilistic, see Section~\ref{sec:randomized_smoothing}. The deterministic approach, as proposed by~\citet{tsuzuku2018lipschitz}, relies on a Lipschitz network to provide a deterministic certificate. This method is computationally efficient but may be overly conservative, when the Lipschitz constant is bounded by the $\PUB$,
leading to a potential drop in performance.
The probabilistic approach, as proposed by~\citet{cohen2019certified}, smooths the soft classifier $F = \tau \circ f$ (see Equation~\ref{eq:soft_classifier}),
to create a soft smoothed classifier $\Ftilde$, which is estimated with Monte-Carlo integration to provide a probabilistic certificate with risk $\alpha$.
This method is more computationally expensive but offers a more accurate and flexible certification as it accounts for the architecture as a whole.

Recall that $F = \tau \circ f$, where $f$ is the Lipschitz network, $\tau$ maps to the simplex, and $F_k : \R^d \to [0,1]$ denotes the $k$-th component for $c = |\mathcal{Y}|$ classes. The smoothed classifier $\Ftilde$ is defined by:
\begin{align}
    \label{eq:smoothed_classifier}
    \Ftilde_k(\vx) = \int_{\R^d} F_k(\vx + \mathbf{\delta}) \phi_{\sigma}(\mathbf{\delta}) d\mathbf{\delta} ,
\end{align}
where \( \phi_{\sigma}(\mathbf{\delta}) = \frac{1}{(2\pi\sigma^2)^{d/2}} e^{-\|\mathbf{\delta}\|^2/(2\sigma^2)} \) is the Gaussian kernel with standard deviation \( \sigma \).

\subsection{Lipschitz interpretation of randomized smoothing}
Randomized smoothing, as described by \citet{salman2019provably}, can be interpreted through the Weierstrass transform, where smoothing is achieved by convolving the classifier output with a Gaussian kernel:
\[
    \Ftilde (\vx) = \mathbb{E}_{\mathbf{\delta} \sim \mathcal{N}(0, \sigma^2 \mI)} \left[ F(\vx + \mathbf{\delta}) \right]
    = \left(F * \phi_{\sigma}\right)(\vx) .
\]
This highlights the relationship between randomized smoothing and Lipschitz continuity, particularly the smoothness of the element-wise score from classifier \( \Ftilde_k\), for $1 \leq k \leq c$.
Leveraging this connection, we can derive two kinds of certified radius for $\Ftilde$. %
\paragraph{First certified radius bound.}
Using this interpretation, developed in~\cite{salman2019provably},
we can use the expression from Section~\ref{sec:lipschitz_networks_certified_robustness}.
To compute the certified radius, we first apply the coordinate-wise bound \( \Rcoord \) from Equation~\ref{eq:rcoord_bound}.
By plugging in the Lipschitz constant of the smoothed classifier, derived from Lemma~\ref{lemma:lip_weierstrass_transform}, from~\citet{salman2019provably}, for all $1 \leq k \leq c$:
$$\Lip(\Ftilde_k) \leq \sqrt{\frac{2}{\pi \sigma^2}} \ ,$$
we get the following certified radius bound:
\begin{align*}
    \Rcoord(\Ftilde, \vx, y)  \leq \frac{\sqrt{\pi \sigma^2} }{2 \sqrt{2}}\margin(\Ftilde(\vx), y) \ .
\end{align*}
To obtain a better bound, we can also use the refined bound derived in Corollary~\ref{corol:bound_lip_sigma},
$$\Lip({\widetilde{F}_k}) \leq \frac{1}{\sqrt{2\pi\sigma^2}} \ ,$$
which gives a tighter bound for the Lipschitz constant of the smoothed classifier:
\begin{align}
    \label{eq:radius_rs_ours}
    \Rcoord(\Ftilde, \vx, y)  \leq \sqrt{\frac{\pi\sigma^2}{2}} \margin(\Ftilde(\vx), y) \ .
\end{align} which is larger by a factor $2$.

To derive Lipschitzness for the smoothed classifier \(\Ftilde\), the outputs of \(F\) must be bounded. Typically, for a softmax classifier, these outputs lie in the interval \([0, 1]\). Consequently, the margin \(\margin(\Ftilde(\vx), y)\) is bounded by 1.
This results in a maximum certified radius of
\begin{align}
    \label{eq:max_radius_rs}
    \Rcoord(\Ftilde, \vx, y) \leq \sqrt{\frac{\pi \sigma^2}{2}},
\end{align} which is considerably lower than the empirical robustness upper bound obtained through adaptive attacks, as reported by \citet{chen2024robust}.
This certified radius is derived from the "weak law" of randomized smoothing, which assumes that the maximum Lipschitz condition holds along the entire perturbation path~\citep{chen2024diffusion}.
In contrast, the "strong law" of randomized smoothing,
as presented in Lemma~\ref{lemma:lip_quantile_smoothed_classifier}~\citep{salman2019provably}, allows for non-constant Lipschitzness for $\Ftilde$. This leads to a tighter and more accurate robust radius, with the potential for the upper bound of the certified radius to be unbounded, as it makes intervene $\quant \circ \Ftilde$ which is not bounded.
This observation motivates the exploration of an alternative certified radius based on the Lipschitz constant of the composition \(\Lip(\quant \circ \Ftilde)\).
\paragraph{Second certified radius bound.}
Same as for the first radius, we can plug the Lipschitz constant of the smoothed classifier, derived from Lemma~\ref{lemma:lip_quantile_smoothed_classifier} in $\Rcoord$.
%
\begin{align*}
    \Rcoord(\quant \circ \Ftilde, \vx, y) = \frac{\margin(\quant \circ \Ftilde(\vx), y)}{2 \Lip(\quant \circ \Ftilde_k)} \ .
\end{align*} %
Rewriting it, with $\Lip(\quant \circ \Ftilde_k) = \frac{1}{\sigma}$, we get the multi-class certified radius bound:
\begin{align}
    \label{eq:radius_rs_multiclass}
    \Rmult(\Ftilde(\vx), y) = \frac{\sigma}{2} \left( \Phi^{-1}( \Ftilde_{y} (\vx)) - \max_{k \neq y} \Phi^{-1}(\Ftilde_{k} (\vx)) \right) \ ,
\end{align}
which is similar to the radius derived in~\citet{cohen2019certified}, see Theorem~\ref{thm:certified_radius_rs_cohen}, for the choice $\underline{\vp}_{i_1} = \vp_{i_1}$ and $\overline{\vp}_{i_2} = \vp_{i_2}$, with $\vp = \Ftilde(\vx)$.
%

\subsection{Randomized smoothing on Lipschitz networks}
In the background section, we derive new Lipschitz bound on the smoothed classifiers $\Ftilde$ and $\quant \circ \Ftilde$,
in particular in the case where underlying classifier $F = f \circ \tau$ is Lipschitz by design, thus and
$\Lip(\Ftilde)$ is known. This requires that $\tau$ to be a mapping on $[0, 1]^c$ such as $\tau = \mathrm{softmax}$ or $\tau(x) = \frac{1}{1 + e^{-x}}$ for instance.
Following the same interpretation of the previous section we can examine the cross-effect of randomized smoothing and Lipschitz continuity on the Lipschitz constant of the smoothed classifier $\Ftilde$.

\paragraph{First certified radius bound.}
As in the previous section,  using the Lipschitz constant of the smoothed classifier element-wise,
from Theorem~\ref{thm:thm_bound_lip_sigma_element_wise},
for all $1 \leq k \leq c$:
$$
    \Lip({\widetilde{F}_k})
    \leq \Lip({ F}_k)\operatorname{erf}\left(\frac{1}{2^\frac{3}{2} \Lip({F}_k){\sigma}}\right) \ .
$$
We can derive the certified radius for $\Ftilde$:
\begin{align}
    \label{eq:radius_coord_and_randomized_smoothing}
    \Rcoord(\Ftilde, \vx, y)  \geq
    \underbrace{\operatorname{erf}\left(\frac{1}{2^\frac{3}{2} \Lip({F}_k){\sigma}}\right)^{-1}}_{\textcolor{blue}{\text{randomized smoothing}}}
    \underbrace{\frac{\margin(\Ftilde(\vx), y)}{2 \Lip({ F}_k)}}_{\textcolor{red}{\text{Lipschitz deterministic}}} \ .
\end{align}
We focus on an intermediate regime defined by a specific \(\sigma\) and \(\Lip(F)\), where these effects interact in a manner that is mutually beneficial, exceeding the individual impacts of randomized smoothing or Lipschitz continuity alone.
Using the Proposition~\ref{prop:sigma_star_lip}, optimal \(\sigma^\star\) is given by
$$
    \sigma^\star = \frac{1}{\sqrt{\pi} \Lip(F_k)} \ .
$$
For this choice, we obtain a certificate, see Equation.~\ref{eq:radius_coord_and_randomized_smoothing}, for the smoothed classifier $\Ftilde$ that is up to  \(26\%\) larger than the maximum certification given by randomized smoothing, see Equation~\ref{eq:radius_rs_ours}, or Lipschitz continuity alone, see Equation~\ref{eq:rcoord_bound}.
In the following we explain why this is the case.
We assume the same margins \(m = \mathrm{M}(F(x),y) = \mathrm{M}(\tilde{F}(x),y)\) for the deterministic network and smoothed network.
In the deterministic case, the certified radius is given by \(\mathrm{R}_{\text{det}} = \frac{m}{L}\).
For randomized smoothing with an optimal choice
\(\sigma^\star = \frac{1}{L\sqrt{2\pi}}\), we have
\[\mathrm{Lip}(\tilde F) \le L\,\mathrm{erf}\!\bigl(\tfrac{\sqrt\pi}{2}\bigr),\]
which leads to a certified radius
\[\mathrm{R}_{\text{rs}} = \frac{m}{\mathrm{Lip}(\tilde f)}
    \ge
    \frac{1}{\mathrm{erf}(\tfrac{\sqrt\pi}{2})}\,\frac{m}{L}.\]
This implies a maximum relative gain over the deterministic certificate of
\[\frac{\mathrm{R}_{\text{rs}}}{\mathrm{R}_{\text{det}}}
    \ge
    \frac{1}{\mathrm{erf}(\tfrac{\sqrt\pi}{2})}
    \approx
    \frac{1}{0.79}
    \approx
    1.27,\]
which corresponds to a certified radius about \(27\%\) larger.

Those bounds on the certified radius are using $\Rcoord$,  so they don't take into account the overall Lipschitz constant $\Lip(\Ftilde)$ and the radius $\Rglobal$ from~\citet{tsuzuku2018lipschitz}.
To derive a bound for $\Rglobal$, we can leverage the fact that $F$ takes values in the probability simplex $\simplex$, to derive a Lipschitz bound for the smoothed classifier $\Ftilde$.

We can adapt the Theorem~\ref{thm:thm_bound_lip_sigma_element_wise} to the case where $\tau$ is a mapping on the probability simplex $\simplex$.
\begin{corollary}[Global Lipschitz constant of the smoothed classifier]
    \label{cor:lip_smoothed_classifier}
    Let $f: \R^d \rightarrow \R^c$ be a Lipschitz function, $\tau : \R^c \rightarrow \simplex$ be a $1$-Lipschitz mapping on the simplex and the soft classifier
    $F = \tau \circ f$.
    Then the \emph{soft smoothed classifier} $\tilde{F}$, is Lipschitz continuous with:
    \begin{align*}
        \Lip({\widetilde{F}})
        \leq \Lip({ F})\operatorname{erf}\left(\frac{1}{2 \Lip({F}){\sigma}}\right) \leq \min\left\{ \frac{1}{\sqrt{\pi\sigma^2}}, \Lip({F})  \right\} \ .
    \end{align*}
\end{corollary}
Using this corollary we get the following certified radius bound for the smoothed classifier $\Ftilde$:
\begin{align}
    \label{eq:radius_global_and_randomized_smoothing}
    \Rglobal(\Ftilde, \vx, y)  \geq
    \underbrace{\operatorname{erf}\left(\frac{1}{2 \Lip({F}){\sigma}}\right)^{-1}}_{\textcolor{blue}{\text{randomized smoothing}}}
    \underbrace{\frac{\margin(\Ftilde(\vx), y)}{\sqrt{2} \Lip({ F})}}_{\textcolor{red}{\text{Lipschitz deterministic}}} \ .
\end{align}
\paragraph{Second certified radius bound.}
For the second radius, which uses the $\quant \circ \Ftilde$ we can use the Theorem~\ref{thm:lip_quantile_smoothed_classifier_ours} to derive a bound for the Lipschitz constant of $\quant \circ \Ftilde$. The difficulty is that we have a local Lipschitz constant, given by
\begin{align*}
    \Lip\left( \quant \circ \tilde{F}_k, B(\vx, \budget) \right)
    \leq \Lip(F_k)
    \sup_{\vx^\prime \in B(\vx, \budget)}
    \left\{
    \frac
    {
        \Phi_\sigma\left(
        s_0(\vx^\prime)
        + \dfrac{1}{\Lip(F_k)}
        \right)
        - \Phi_\sigma\left(s_0(\vx^\prime)\right)
    }
    {
        \phi(\quant(\tilde{F_k}(\vx^\prime)))
    }
    \right\} \ ,
\end{align*}

and \( s_0(\vx^\prime) \) is determined
by solving:
\[
    \tilde{F}_k(\vx^\prime) = 1 - \Lip(F_k) \int_{s_0(\vx^\prime)}^{s_0(\vx^\prime) + \frac{1}{\Lip(F_k)}} \Phi_\sigma(s) \, ds \ .
\]

Thus the certified radius $\Rcoord$ is also local, let say certification is required for a budget $\budget$, and we have to test if the certified radius  is superior to $\budget$:

\begin{align}
    \label{eq:radius_rs_multiclass_lip_local}
    \Rmultlip(\quant \circ \Ftilde, \vx, y) \geq
    \frac{1}{2}
    \left(
    \frac{\Ftilde_y(\vx)}{\Lip(\quant \circ \Ftilde_k, B(\vx, \budget))} - \max_{k \neq y} \frac{\Ftilde_k(\vx)}{\Lip(\quant \circ \Ftilde_k, B(\vx, \budget))}
    \right)
    \  \geq \budget \ .
\end{align}

A summary of the bounds on the Lipschitz constant of the smoothed classifier is presented in Table~\ref{tab:lipschitz_bounds_ftilde}.
\begin{table}[h]
    \centering
    \caption{Summary of bounds on the Lipschitz constant of the smoothed classifier. In this table $b=s_0(\vx^\prime)
        $ and $a=b
            + \dfrac{1}{\Lip(F_k)}$}
    \label{tab:lipschitz_bounds_ftilde}
    \resizebox{\textwidth}{!}{%
        \renewcommand{\arraystretch}{2}
        \begin{tabular}{|c|c|c|}
            \hline
            \textbf{Case} & \textbf{Bound expression} & \textbf{Reference} \\
            \hline
            $\Lip(\Ftilde_k)$ element-wise
                          &
            $\displaystyle
                \Lip(\Ftilde_k)
                \leq \sqrt{\frac{2}{\pi \sigma^2}}
            $
                          &
            \citet{salman2019provably}, Lemma~\ref{lemma:lip_weierstrass_transform}
            \\
            \hline
            $\Lip(\Ftilde_k)$ element-wise (refined)
                          &
            $\displaystyle
                \Lip(\Ftilde_k)
                \leq \frac{1}{\sqrt{2\pi \sigma^2}}
            $
                          &
            Corollary~\ref{corol:bound_lip_sigma}
            \\
            \hline
            $\Lip(\Ftilde_k)$ for Lipschitz $F_k$
                          &
            $\displaystyle
                \Lip(\Ftilde_k)
                \leq \Lip(F_k) \, \operatorname{erf}\!\left(\frac{1}{2^\frac{3}{2} \Lip(F_k) \sigma}\right)
                \leq \min\!\left\{ \frac{1}{\sqrt{2 \pi \sigma^2}}, \Lip(F_k)\right\}
            $
                          &
            Theorem~\ref{thm:thm_bound_lip_sigma_element_wise}
            \\
            \hline
            $\Lip(\Ftilde)$ global
                          &
            $\displaystyle
                \Lip(\Ftilde)
                \leq \Lip(F) \, \operatorname{erf}\!\left(\frac{1}{2 \Lip(F) \sigma}\right)
                \leq \min\!\left\{ \frac{1}{\sqrt{\pi \sigma^2}}, \Lip(F)\right\}
            $
                          &
            Corollary~\ref{cor:lip_smoothed_classifier}
            \\
            \hline
            $\Lip(\quant \circ \Ftilde)$ global
                          &
            $\displaystyle
                \Lip(\quant \circ \Ftilde)
                \leq \frac{1}{\sigma}
            $
                          &
            Standard RS theory, \citet{cohen2019certified}
            \\
            \hline
            $\Lip(\quant \circ \Ftilde)$ local
                          &
            $\displaystyle
                \Lip(\quant \circ \Ftilde_k, B(\vx, \budget))
                \leq \Lip(F_k)
                \sup_{\vx' \in B(\vx, \budget)} \frac{\Phi_\sigma(a) - \Phi_\sigma(b)}{\phi(\quant(\Ftilde_k(\vx')))}
            $
                          &
            Thm~\ref{thm:lip_quantile_smoothed_classifier_ours}
            \\
            \hline
        \end{tabular}
    } 
\end{table}

\subsection{Generalized simplex mapping}
In classification tasks, model outputs are typically mapped onto the standard probability simplex \(\simplex \subset \mathbb{R}^c\), where entries are non-negative and sum to 1, forming a valid class distribution. In the randomized smoothing (RS) framework, the output has traditionally been treated as a probability distribution due to the probabilistic nature of the method. Certified radii are derived by analyzing the probability of class scores under Gaussian noise, relying on the assumption that network outputs represent valid probabilities. While this probabilistic perspective has shaped the development of RS, strict adherence to the simplex is not essential for deriving Lipschitz bounds or applying smoothing techniques like the Weierstrass transform. For smoothed networks to exhibit Lipschitz continuity, it suffices for the underlying function to be bounded, allowing greater flexibility in the choice of mappings while still enabling the derivation of meaningful certified radii.

Relaxing the mass constraint leads to the generalized scaled simplex \( \rsimplex \), where entries sum to an arbitrary value \( \mass > 0 \).
Adjusting \( \mass \) influences the scale of the distribution,
modifying model margins and Lipschitz properties,
thus providing a tool to balance robustness and stability.
Indeed for a simplex of mass \( \mass \), the margins are bounded by $\mass$, and this can unlock a new certified radius beyond the mass-$1$ simplex, see Equation~\ref{eq:max_radius_rs}.

Mappings such as \(\mathrm{softmax}\) are often employed to transform logits onto the standard simplex due to their smooth, 1-Lipschitz properties.
However, \(\mathrm{softmax}\) can compress margins by producing dense distributions. To address this, \cite{martins2016softmax} proposed \(\mathrm{sparsemax}\), which projects logits onto \( \simplex \) while encouraging sparsity, resulting in larger margins and preserving 1-Lipschitz continuity.
The $\mathrm{sparsemax}$ projection onto the simplex is defined as:
\[
    \mathrm{sparsemax}(\vz) = \argmin_{\vp \in \Delta^{c-1}} \norm{\vp - \vz}_2^2,
\]
where $\Delta^{c-1}$ denotes the probability simplex.
The projection onto the simplex performed by \textit{sparsemax} is a classical operation in convex optimization. The specific algorithm used to compute sparsemax is closely related to the algorithm of \citet{held1974validation}, and later formalized in modern form by \citet{condat2016fast} for fast projection onto the simplex or $\ell_1$-ball.

\paragraph{Why sparsemax encourages sparsity.}
This projection leads to sparsity because it sets all components $z_i$ below a certain threshold $\tau(\vz)$ to zero:
\[
    \mathrm{sparsemax}_i(\vz) = \max\left(z_i - \tau(\vz), 0\right),
\]
where the threshold $\tau(\vz)$ is computed such that $\sum_i \mathrm{sparsemax}_i(\vz) = 1$ and only the largest components of $\vz$ remain positive. Consequently, many entries are exactly zero, producing a sparse output vector.

Building on this, we propose a generalized \(\mathrm{sparsemax}\) that maps logits onto the \( \mass \)-simplex \( \rsimplex \), where the total mass \( \mass \) can differ from the conventional value of 1. This generalized projection onto the scaled simplex $\Delta_r^{c-1}$ has already been employed in prior work (see, e.g., \cite[Algo.~1]{condat2016fast}), and we adopt it here to control the sparsity and margin of the output. Lower values of \( \mass \) lead to sparser, more selective distributions, whereas higher values allow for larger margins between active classes.
\begin{algorithm}
    \caption{generalized\_sparsemax($\vz$, $\mass$)}
    \begin{algorithmic}[1]
        \State Sort \( \vz \) in decreasing order \( \vz_{1} \geq \dots \geq \vz_{c} \)
        \State Find \( \kappa(z) \) such that
        \[
            \kappa(\vz) = \max_{k = 1 \ldots c} \left\{ k \, \bigg| \, \mass + k \vz_{k} > \sum_{j \leq k} \vz_{j} \right\}
        \]
        \State Define
        \[
            \rho(\vz) = \frac{\left(\sum_{j \leq \kappa(\vz)} \vz_j\right) - \mass}{\kappa(\vz)}
        \]
        \State \textbf{return} \( \vp \) such that \( \vp_i = \max(\vz_i - \rho(\vz), 0) \)
    \end{algorithmic}
    \label{algo:sparsemax_generalized}
\end{algorithm}
The generalized $\mathrm{sparsemax}$  is a $1$-Lipschitz mapping towards $\Delta^{c-1}_\mass$, following the same proof as in \cite[Appendix~A.5]{laha2018controllable}. This mapping is described in Algorithm~\ref{algo:sparsemax_generalized}.
Recall that $F = \tau \circ f$ where $f$ is the Lipschitz network and $\tau$ is a generalized simplex mapping.
For $r_1 \leq r_2$, when most of the logit vectors $f(x + \delta_i)$ are bounded by $r_1$, the mapping $\tau^{r_2}$ to the simplex $\Delta^{c-1}_{r_2}$ with $1$-Lipschitz simplex mapping is not going to increase the margin associated to vectors  $\tau^{r_2}(f(x + \delta_i))$.
In this case, it is better to map to the simplex $\Delta^{c-1}_{r_1}$ of lower mass and enjoy a tighter Lipschitz constant on $\tilde{F}$ or $\Phi^{-1} \circ \tilde{F_k}$.
Conversely, when $r_2 \geq r_1$, one can benefit from larger margins on $\tau^{r_2}(f(x + \delta_i))$ in comparison to margins on $\tau^{r_1}(f(x + \delta_i))$.
\begin{figure}[t]
    \centering
    \includegraphics[width=0.6\textwidth]{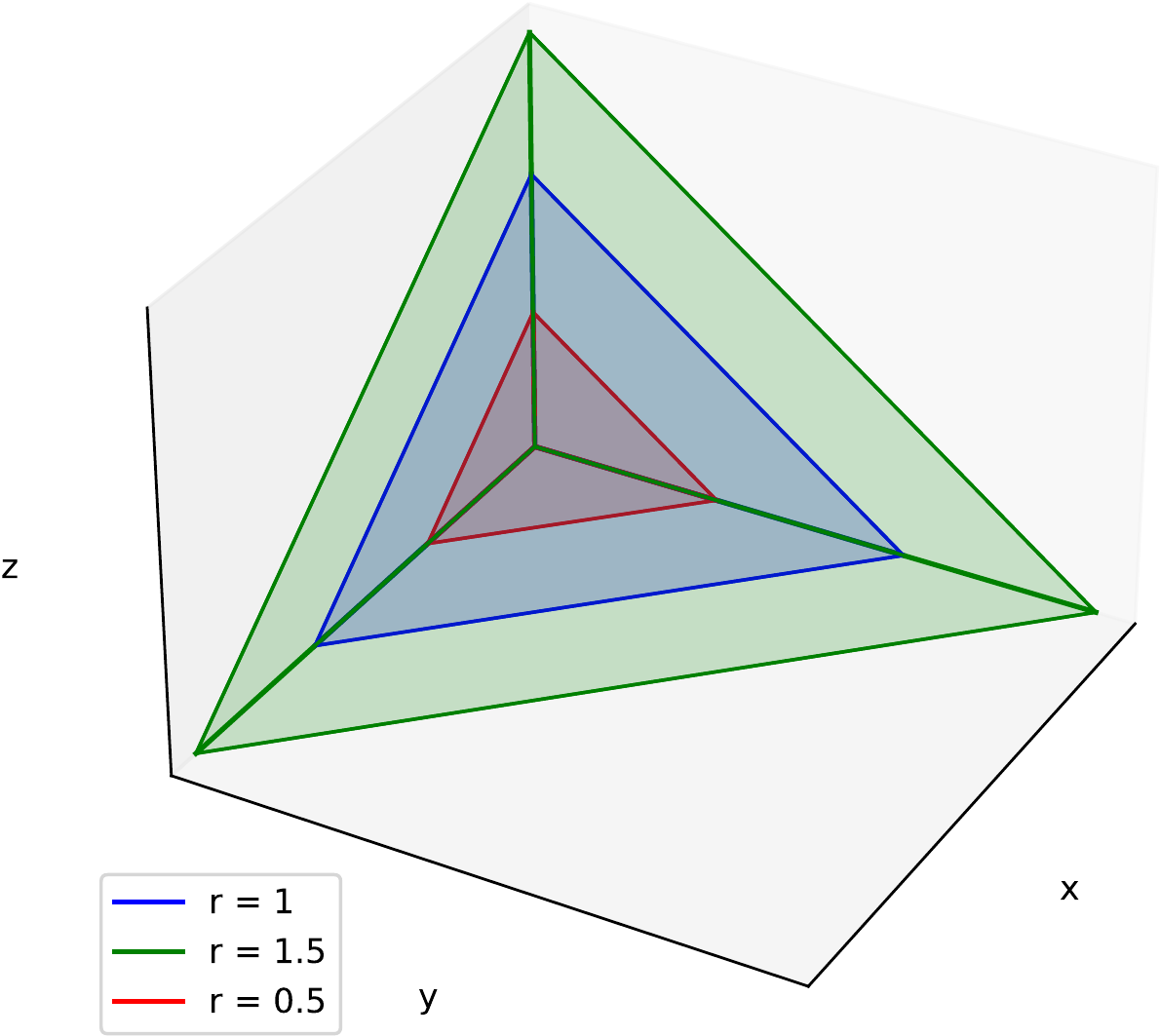}
    \caption{Illustration of simplexes of different mass $r \in \{0.5, 1, 1.5 \}$ for $c=3$ classes.}
    \label{fig:simplex}
\end{figure}
In Figure~\ref{fig:simplex}, we can see simplexes of different sizes, one point on a simplex gives a margin at most equal to the mass of the simplex.

Theorem results that establish Lipschitz bounds for functions \( F : \mathbb{R}^c \to \mathbb{R}^c \) can extend naturally to \( F : \mathbb{R}^d \to \rsimplex \), optimizing certified radii without restrictive simplex constraints. We get the following corollary for the Lipschitz constant of the \emph{smoothed classifier} $\tilde{F}$:
\begin{corollary}
    \label{cor:lip_smoothed_classifier_r}
    Let $F: \R^d \rightarrow \R^c$ be a $L$-Lipschitz function, $\tau : \R^c \rightarrow \rsimplex$ be a $1$-Lipschitz mapping on the $\mass$-simplex and the soft classifier
    $F = \tau \circ F$.
    Then the \emph{soft smoothed classifier} $\tilde{F}$, is Lipschitz continuous with:
    \begin{align}
        \Lip({\widetilde{F}})
        = \Lip({ F})\operatorname{erf}\left(\frac{\mass}{2 \Lip({F}){\sigma}}\right) \leq \min\left\{ \frac{\mass}{\sqrt{\pi\sigma^2}}, \Lip({F})  \right\} \ .
    \end{align}
\end{corollary}
This flexibility in tuning \( r \) makes the \( r \)-simplex and the generalized \(\mathrm{sparsemax}\) valuable tools, particularly for balancing the trade-off between variance and margin.
For instance, a smaller \( r \) reduces the Lipschitz constant further, enhancing robustness, while a larger \( r \) increases the margins between classes, see Corollary~\ref{cor:lip_smoothed_classifier_r}.
This approach extends traditional probabilistic frameworks in randomized smoothing by offering a spectrum of robustness profiles that go beyond the conventional mass-1 simplex.
We can also show that optimal $\sigma^\star$ is given by $\sigma^\star = \frac{\mass}{\sqrt{\pi} \Lip(f)}$.

\subsection{Experiments}
%
\begin{table}[t]
    \centering
    \caption{Certified accuracy on CIFAR-10 for different levels of perturbation $\budget$, for RS, Lipschitz deterministic, and ours.
        The risk is taken as $\alpha=1\mathrm{e-}3$ and the number of samples $n=10^4$.
    }
    \label{table:comparison_certified_acc_lip_vs_rs_vs_lip_and_rs_cifar10}
    \begin{tabular}{lccccccc}
        \toprule
        \multirow{2}[2]{*}{\textbf{Methods}} & \multicolumn{6}{c}{{\boldmath{}\textbf{Certified accuracy ($\budget$)}\unboldmath{}}} & \multirow{2}[2]{*}{\textbf{Average time (s)}}                                                                                     \\
        \cmidrule{2-7}
                                             & $0.14$                                                                                & $0.19$                                        & $0.25$         & $0.28$         & $0.42$        & $0.5$          &                \\
        \midrule
        \text{Lipschitz deterministic}
                                             & 40                                                                                    & 33.57                                         & 27.18          & 24.59          & 13.65         & 9.15           & \textbf{0.004} \\
        \text{Randomized smoothing}
                                             & 47.9                                                                                  & 31.99                                         & 28.17          & 27.86          & 6.42          & 0.0            & 0.9            \\
        \midrule
        \text{\textbf{RS with new bound}}
                                             & \textbf{52.56}                                                                        & \textbf{46.17}                                & \textbf{39.09} & \textbf{35.08} & \textbf{21.9} & \textbf{13.53} & 0.9            \\
        \bottomrule
    \end{tabular}
\end{table}
\textbf{Impact of Lipschitz constant on the first radius.}
To illustrate the gain of having a Lipschitz bound of the smoothed classifier $\Ftilde$ which includes information on the Lipschitz constant of base classifier $F$, simplex mass $\mass$ and variance $\sigma^2$, we compare certified accuracies on the same by design $5$-Lipschitz neural network backbone $\mathrm{Sandwhich ~ Small}$ from~\citep{wang2023direct}, trained with noise injection $\sigma=0.4$ and using the same certified robust radius $\Rglobal$ in Equation~\ref{eq:radius_tsuzuku}.

We choose for the smoothing variance $\sigma^* = \frac{r}{L(F)\sqrt{\pi}}$ as it maximizes the gain of our new bound, we use a Lipschitz constant of $5$ because $1$ is too restricitive for the network to be trained with this scale of noise injection.
Remark that the variance used for noise injection to train the $5$-Lipschitz classifier $F$ is close to $\sigma^*$ to mitigate a drop in performance on the smoothed classifier $\Ftilde$.
We consider the three procedures: Lipschitz deterministic using bound on $L(F)$, the RS using bound on $L(\tilde{F})$,
and our approach using new bound on $L(\tilde{F})$. For ours and RS, we fix $r=3$.
%
Results are displayed in Table~\ref{table:comparison_certified_acc_lip_vs_rs_vs_lip_and_rs_cifar10}.
We see that our procedure gives better-certified accuracies than RS and Lipschitz deterministic taken alone, indeed both methodologies provide the same Lipschitz constant for $\tilde{F}$ and $F$ respectively, whereas our method provides an inferior Lipschitz bound on $\Lip(\tilde{F})$.
Note that better results from the random procedure should not be directly construed as an intrinsic superiority over the deterministic one, as the element of randomness introduces variability that must be accounted for in the evaluation and large sampling computational cost. However, it gives a perspective on the performance of the theoretical Lipschitz smoothed classifier $\tilde{F}$.

\begin{figure}[t]
    \centering
    \includegraphics[width=0.95\linewidth]{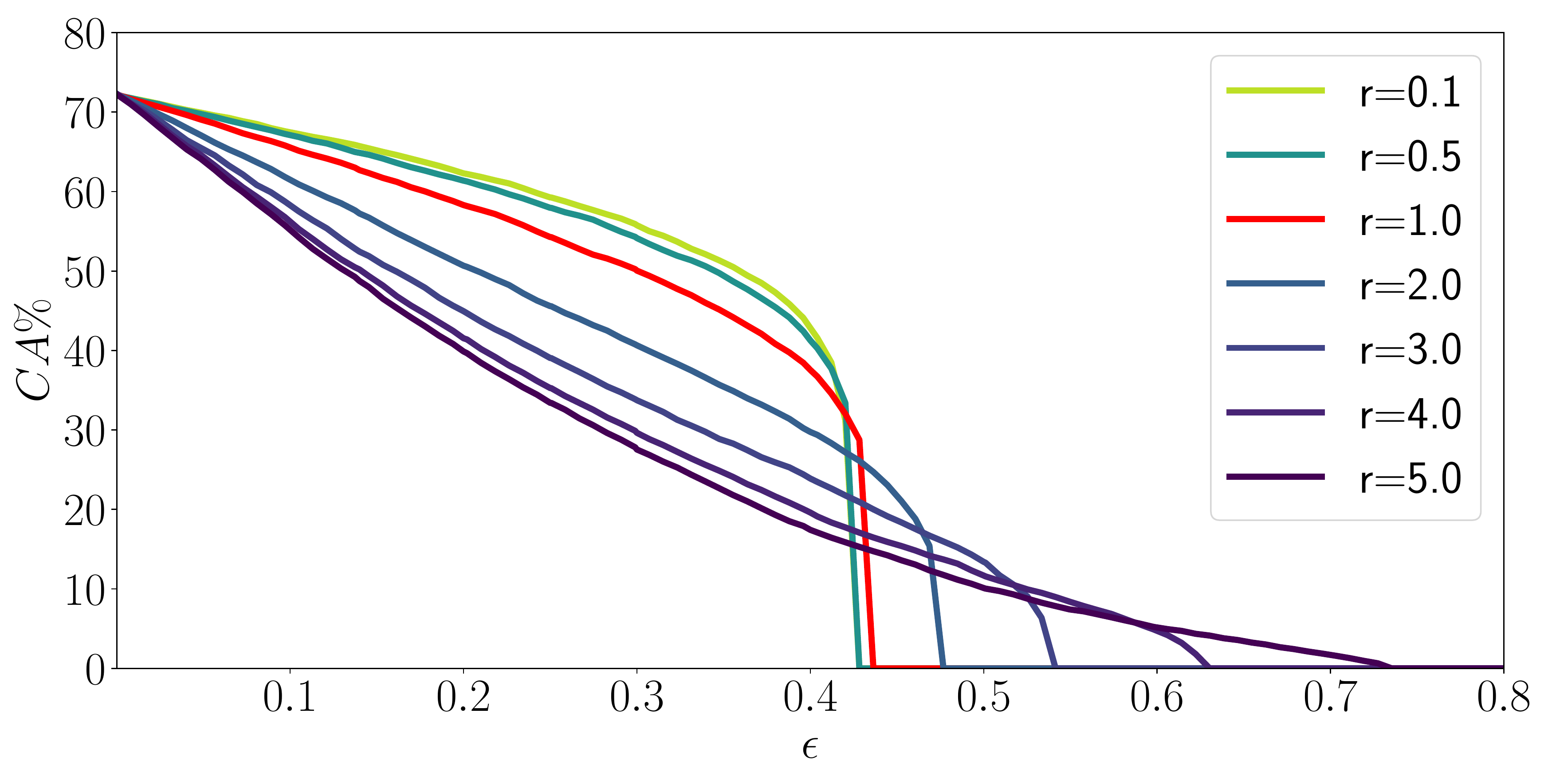}
    \caption{Certified accuracies ($CA$ in $\%$) with $\Rcoord$ in function of levels of perturbation $\mass$ on CIFAR-10, for different simplex mass $\mass$.
        Number of samples is $n=10^4$ and risk $\alpha = 1e\text{-}3$.
        The case $r=1$ in the color red corresponds to the regular RS probabilistic setting.
    }
    \label{fig:ca_for_different_r}
\end{figure}

\textbf{Impact of simplex mass on the first radius.}
We note that to reduce the Lipschitz constant, one can not only increase the smoothing noise \(\sigma\), as is traditionally done in randomized smoothing (RS) but also reduce \(\mass\), the total mass of the simplex. This introduces a new degree of freedom in the design of smoothed classifiers.
In Fig.~\ref{fig:ca_for_different_r}, we plot the evolution of certified accuracies for different simplex masses \(\mass\), using the same settings as described previously. It becomes evident that the classical RS setting, i.e. \(\mass = 1\), is merely one specific instance of a broader range of robustness profiles.
By varying the simplex mass \(\mass\), we can explore and unlock new, potentially larger certified radii, depending on the desired magnitude of the certification radius. This allows for more flexibility in tailoring the robustness of the classifier to meet specific application requirements.

Note that for $\epsilon = 0$ we recover the natural accuracy of the classifier, which is around $70\%$ for the considered network architecture on the CIFAR-10 dataset. It does not depend on the simplex mass $\mass$ as only the $\argmax$ operation is performed over the logits and margins are not meaningful in this case.
\begin{figure}[t]
    \centering
    \includegraphics[width=1\linewidth]{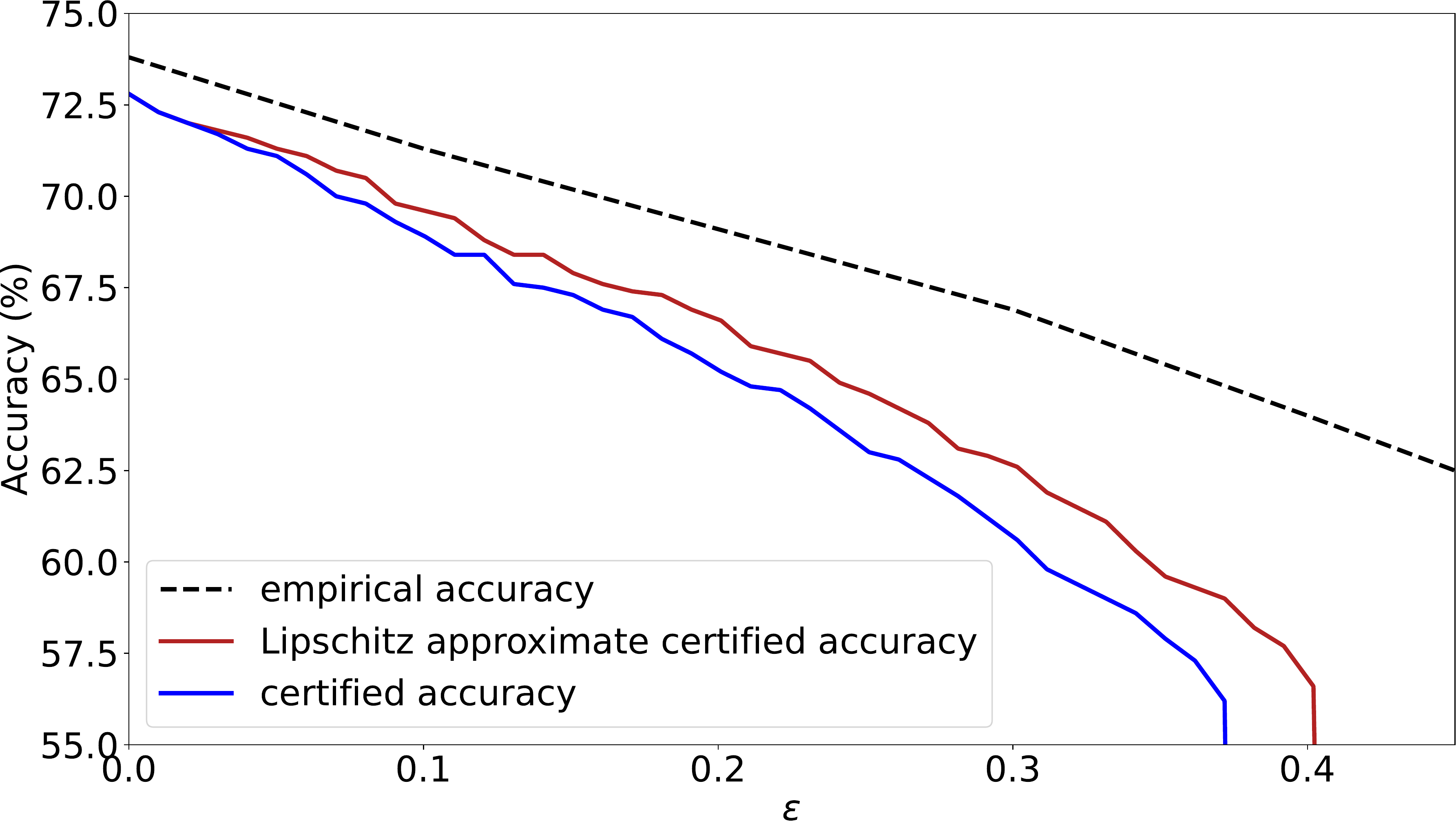}
    \caption{LiResNet trained with noise injection ($\sigma=0.5$), on CIFAR-10 dataset.
        Certified accuracy $R_{\mathrm{multi}}$ vs Lipschitz approximate certified accuracy $R_{\mathrm{multiLip}}$ vs empirical robust accuracy using projected gradient ascent. Randomized smoothing noise is taken as $\sigma=0.12$.}
    \label{fig:acc_certif_vs_empirical_vs_lip_liresnet}
\end{figure}

\textbf{Impact of Lipschitz constant on second radius.}
In the paper of~\citet{cohen2019certified} the authors point out that the produced certificates were too conservative compared to obtained empirical robustness. This could be due to the fact that the Lipschitz constant of the classifier $F$ is not taken into account in the certified radius.
To derive the certified radius \( R_{\mathrm{multiLip}} \), we acknowledge that this value is not exactly computable, as it requires evaluating the local Lipschitz constant through a certified procedure. However, to illustrate the potential gains of incorporating such Lipschitz constants into the certified radius under the "strong law" of randomized smoothing (RS), we employ Lipschitz gradient ascent locally (Section~\ref{sec:local_lipschitzness})  to estimate $\Lip(\quant \circ \Ftilde, B(0, \epsilon))$.

In this approximation, we utilize LiResNet~\citep{hu2023scaling}, a state-of-the-art Lipschitz network for certified robustness, trained on CIFAR-10. The resulting radius \( R_{\mathrm{multi}} \) underestimates the certified radius \( R_{\mathrm{multiLip}} \), demonstrating that the Lipschitz continuity assumption tightens the bound on the true certified radius.
We compared the standard certified robustness obtained from the classical RS procedure, which yields the radius \( R_{\mathrm{multi}} \), to our new Lipschitz-adjusted radius \( R_{\mathrm{multiLip}} \).
Both approaches apply the standard Bonferroni correction and the empirical Bernstein confidence bound~\citep{maurer2009empirical}.
For empirical evaluation, we conducted assessments using a projected gradient descent (PGD) \( \ell_2 \) attack~\citep{madry2018towards} with 40 iterations and a step size of 0.2. This experiment illustrates the potential improvements in certified robustness through the integration of the Lipschitz-adjusted radius.

As shown in Figure~\ref{fig:acc_certif_vs_empirical_vs_lip_liresnet}, the certified radius obtained with the Lipschitz continuity assumption, \( R_{\mathrm{multiLip}} \), more closely aligns with empirical robust accuracy (i.e., accuracy on the test set under the PGD \( \ell_2 \) attack).
This suggests that the inclusion of the Lipschitz assumption not only tightens the certified bound but also produces results that better reflect the actual robustness observed in practice, compared to the standard certified radius \( R_{\mathrm{multi}} \).
It is important to note that empirical robustness evaluations should ideally involve directly attacking the smoothed classifier, as demonstrated in~\citet{salman2019provably}, or utilize stronger adversarial attacks than PGD. Consequently, the empirical robustness reported here is likely slightly overestimated, as attacking the smoothed classifier is generally more challenging and often results in lower robustness values.

\section{Conclusion}

In this chapter, we explored the robustness properties of Lipschitz networks through deterministic and probabilistic approaches. We demonstrated that spectrally rescaled layers (\(\SR\)) enhance certified robustness by maintaining 1-Lipschitz constraints. This approach consistently outperformed baseline rescaling methods across multiple datasets and architectures.

We extended our analysis by integrating randomized smoothing (RS) with Lipschitz networks, highlighting the interaction between smoothing noise \(\sigma\) and the Lipschitz constant. By varying the simplex mass \(\mass\), we introduced a new degree of freedom in certified robustness, unlocking larger certified radii compared to traditional RS methods. This generalized smoothing approach enables the exploration of a broader range of robustness profiles, providing greater flexibility in network design.

Furthermore, we demonstrated the impact of local Lipschitz constants on the second certified radius \(R_{\mathrm{multiLip}}\). Through empirical evaluations on CIFAR-10, we showed that incorporating Lipschitz continuity assumptions aligns certified bounds more closely with empirical robust accuracy, outperforming classical RS techniques. This suggests a promising avenue for addressing the gap between certified radii in theory and those observed in practice. By refining the estimation of local Lipschitz constants, this approach holds the potential to further reduce the discrepancy, offering a pathway towards tighter and more realistic robustness guarantees.

Overall, our results indicate that Lipschitz-aware smoothing techniques can significantly enhance the robustness of neural networks.
The next chapter's work will focus on controlling the variance and risk in certification induced by the Monte-Carlo integration to estimate the smoothed classifier in randomized smoothing.

%% file: content/chapter-risk_management.tex
%
\chapter{Risk control and variance reduction in randomized smoothing}
\label{chap:risk_management}

\minitoc%

Randomized smoothing (RS) enhances adversarial robustness by averaging predictions over Gaussian-perturbed inputs, providing formal certification of classifier robustness.
%
A key challenge in RS is balancing the tradeoff between margin and variance. While maximizing classification margins improves robustness, it also increases variance, leading to overly conservative robustness estimates. Mono-class certified radii, \( R_{\text{mono}} \), often fail under high smoothing variance, as the top-class probability may fall below the certification threshold. Multi-class radii, \( R_{\text{multi}} \), mitigate this by considering the top two class probabilities, providing meaningful certification even when the top class confidence is low.

However, introducing \( R_{\text{multi}} \) necessitates controlling confidence intervals across all class probabilities, which scales with the number of classes. This can result in highly conservative bounds, particularly in datasets with many classes.
To address this, we propose the Class Partitioning Method (CPM), which groups less significant classes into meta-classes, reducing the number of confidence intervals that must be managed. This improves the standard confidence interval obtained with Bonferroni correction, lowering conservativeness and enabling tighter certified radii without increasing computational complexity.

Additionally, we introduce the Lipschitz-Variance-Margin (LVM) framework to reduce variance at the model level. By controlling the Lipschitz constant and applying temperature-scaled simplex mappings, LVM stabilizes smoothed classifiers, further enhancing certified accuracy across various perturbation levels.

\section{Certification procedures for randomized smoothing}


Randomized smoothing (RS) enhances adversarial robustness by averaging predictions over Gaussian perturbations of the input, producing a smoothed classifier.
\[
    \tilde{F}(\vx) = \mathbb{E}_{\mathbf{\delta} \sim \mathcal{N}(0, \sigma^2 \mI)} [ F(\vx + \mathbf{\delta}) ].
\]
Let \( \vp = \Ftilde(x) \), where \( \vp \in \Delta^{c-1} \) is the probability vector over \( c \) classes, $\vp = \left( \vp_1, \ldots, \vp_c \right)$, with \( \vp_i \) representing the probability of class \( i \).
RS provides a certified radius that quantifies robustness to adversarial perturbations, derived from the top two class probabilities \( \vp_{i_1} \) and \( \vp_{i_2} \).
For a risk level \( \alpha \), the certified radius is computed with a confidence interval for the probabilities \( \vp_{i_1} \) and \( \vp_{i_2} \) at level \( 1 - \alpha \):
\[
    \Rmult(\vp) = \frac{\sigma}{2} \left( \Phi^{-1}( \vp_{i_1}) - \Phi^{-1}(\vp_{i_2}) \right) \ .
\]
In practise, a simpler mono-class version is often used:
\(
\Rmono(\vp) = \sigma \Phi^{-1}(\vp_{i_1}) \leq \Rmult(\vp) \,
\)
%
as it requires only the top class probability \( \vp_{i_1} \) and is easier to compute.

\subsection{Advocating multi-class over mono-class certified radii}

\begin{figure}[h]
    \centering
    \includegraphics[width=0.5\textwidth]{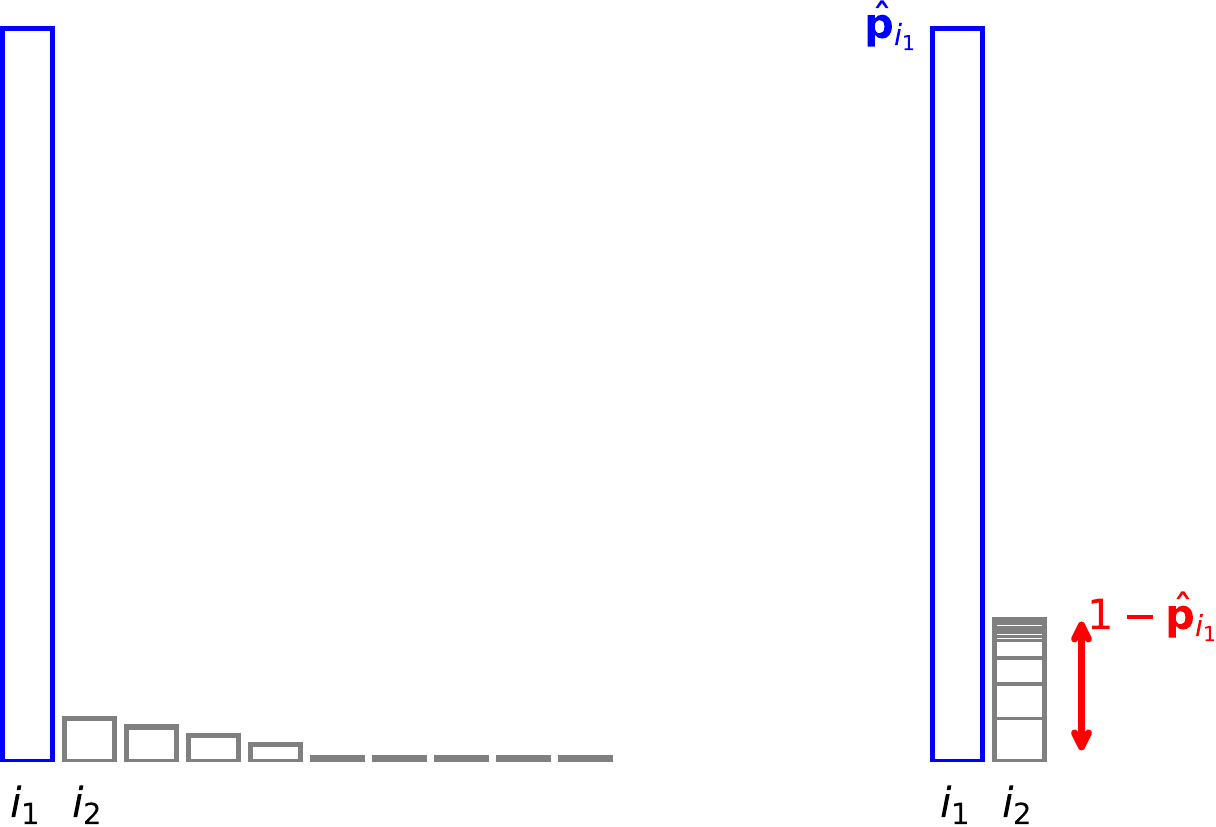}
    \caption{}
    \label{fig:summing_proba_rmono}
\end{figure}

When the top-class probability \( \vp_{i_1} \) is high, the mono-class certified radius \( \Rmono \) provides a meaningful robustness guarantee. This is illustrated in Figure~\ref{fig:summing_proba_rmono}, where the confidence in a single dominant class leads to a non-trivial certified region.
However, in binary classification settings—particularly when class \( i_1 \) is certified against a superclass that aggregates all remaining classes—the mono-class bound \( \Rmono \) becomes ineffective when \( \vp_{i_1} \leq \frac{1}{2} \), as highlighted by \citet{voracek2023improving, voracek2024treatment}.
\begin{figure}[h]
    \centering
    \includegraphics[width=0.5\textwidth]{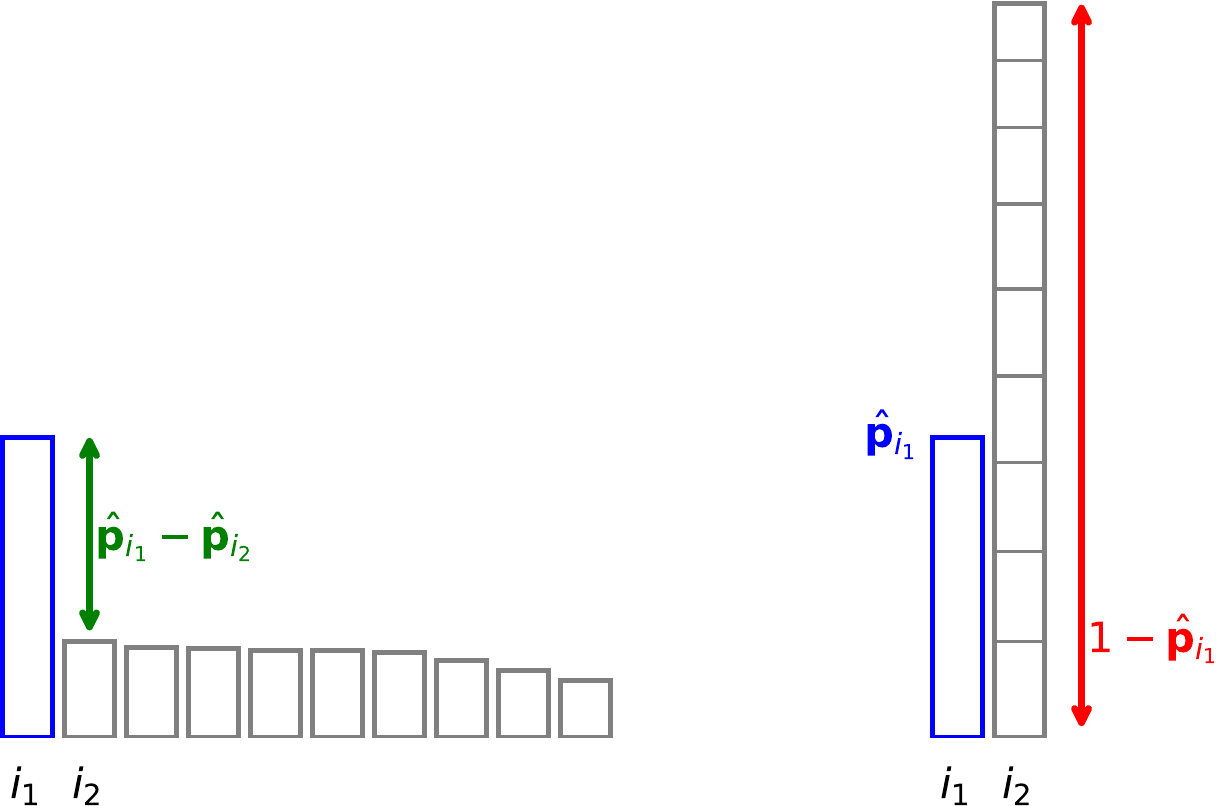}
    \caption{}
    \label{fig:summing_proba_flat}
\end{figure}
Figure~\ref{fig:summing_proba_flat} demonstrates this limitation: when the predicted top-class probability falls below \( \frac{1}{2} \), the mono-class radius drops to zero, indicating a complete lack of certified robustness.

To address this, the multi-class radius \( \Rmult \), which incorporates both \( \vp_{i_1} \) and the second-highest class probability \( \vp_{i_2} \), can yield a non-zero certification even when \( \vp_{i_1} < \frac{1}{2} \). Nonetheless, under high noise regimes (large variance \( \sigma^2 \)), this approach also shows limitations. The distribution of \( F(x + \delta) \) tends to flatten, approaching a uniform distribution. Consequently, the gap between \( \Rmult \) and \( \Rmono \) widens, as reported in Table~\ref{table:comparison_R2_vs_r2_for_different_sigmas} and previously observed by \citet{voracek2023improving}.
In this table $n$ is the number of samples used to estimate the certified radii, $\alpha$ is the risk level, and $\sigma$ is the smoothing variance.

\begin{table}[h!]
    \centering
    \caption{
        Comparison of two certified radii $\Rmult$ and $\Rmono$, and total variation distance to uniform distribution (TVU), for different values of \(\sigma\).
        We took a subset of images from the ImageNet test set with $n = 10^4$, $\alpha=0.001$.
        If the effect is not visible for $\sigma < 0.25$,
        we see that as the TVU decreases, the difference between the two radii increases as well.
    }
    \vspace{0.1cm}
    \begin{tabular}{cccccc}
        \toprule
        \(\sigma\) & TVU   & $\Rmult$      & $\Rmono$      & $\Rmult - \Rmono$ \\
        \midrule
        0.25       & 0.998 & \textbf{5.89} & \textbf{5.89} & 0.00              \\
        0.30       & 0.996 & \textbf{4.68} & 4.48          & 0.20              \\
        0.35       & 0.989 & \textbf{3.68} & 3.21          & 0.47              \\
        0.40       & 0.986 & \textbf{2.49} & 1.97          & 0.52              \\
        0.50       & 0.976 & \textbf{1.28} & 0.59          & 0.69              \\
        0.60       & 0.95  & \textbf{0.56} & 0.00          & 0.56              \\
        \bottomrule
    \end{tabular}
    \label{table:comparison_R2_vs_r2_for_different_sigmas}
\end{table}

\subsection{Confidence intervals for certified radii}

To estimate the certified radius, one must rely on approximations since the smoothed classifier cannot be computed directly. MC integration provides this approximation, the probabilities \( \vp \) are estimated via sampling:
\[
    \hat{\vp} = \frac{1}{n} \sum_{i=1}^n F(\vx + \delta_i),
\]
where \( \delta_i \sim \mathcal{N}(0, \sigma^2 \mI) \).
An \( (1 - \alpha) \)-exact coverage confidence interval is needed to control the risk $\alpha$ associated with the certified radius accurately.
\begin{definition}[Exact $\alpha$ coverage confidence interval]
    Let \( \xi \) be a random variable taking values in \( \mathbb{R} \).
    An \( (1 - \alpha) \)-exact coverage confidence interval for a parameter \(\theta\) is an interval \([\underline{\xi}, \overline{\xi}]\) such that the probability of \(\theta\) lying within the interval is greater than \(1 - \alpha\), i.e.,
    \[
        \mathbb{P}(\underline{\xi} \leq \theta \leq \overline{\xi}) \geq 1 - \alpha.
    \]
    Here, \(\alpha\) represents the risk, which is the probability that the interval does not contain the true value of \(\theta\).
\end{definition}
The inherent uncertainty is controlled by concentration inequalities or Clopper-Pearson confidence interval in the case of $\tau = \hardmax$ simplex mapping, they are used to derive. Commonly used methods to construct such intervals include empirical Bernstein bounds~\citep{maurer2009empirical}, the Hoeffding inequality~\citep{boucheron2013concentration} for continuous-valued simplex mapping, and Clopper-Pearson methods for discrete mapping (hardmax).
These approaches ensure the certified radius is computed with a controlled risk, providing reliable estimates for robustness certification.

\paragraph{Confidence validation for \( \Rmono \).} In the context of certification using \( \Rmono \), the radius is determined based on the index \( i_1 \), which is obtained from independent samples~\citep{cohen2019certified}. The certified radius is defined as:
\[
    \mu = \Rmono(\underline{\hat{\vp}_{i_1}}) \ ,
\]
where \( \underline{\hat{\vp}_{i_1}} \) represents the lower bound of the \( (1 - \alpha) \)-exact Clopper-Pearson confidence interval for the observed proportion \( \hat{\vp}_{i_1} \). The confidence test for this certification involves the following,
\( H_0 \): the probability of the certified radius being at least \( \mu \) is at least \( 1 - \alpha \), i.e.,
\[
    H_0: \mathbb{P}(\Rmono(\vp_{i_1}) \geq \mu) \geq 1 - \alpha \ .
\]
\( H_1 \): the probability of the certified radius being less than \( \mu \) is less than \( \alpha \), i.e.,
\[
    H_1: \mathbb{P}(\Rmono(\vp_{i_1}) < \mu) < \alpha \ .
\]
This framework helps to determine whether the certified radius provides a reliable measure of robustness under the specified risk level $\alpha$.

For the mono-class certification, the confidence interval is constructed for a single class, making it straightforward to manage the risk level.
However, constructing confidence intervals for \( \Rmult \) requires extra caution, as it involves managing several interdependent quantities.
Previous approaches~\citep{voracek2023improving, voracek2024treatment, scholten2023hierarchical, weber2023rab} introduce errors in how they allocate the risk budget \( \alpha \). This flawed allocation breaks the Bonferroni correction by assuming dependence between variables, as illustrated by a \href{https://github.com/blaisedelattre/bridging_the_gap_rs/blob/main/counter_example.py}{\textcolor{blue}{counter example}} discussed in the Appendix~\ref{app:sec:counterexample_bonferroni_correction}.

\paragraph{Confidence validation for \( \Rmult \).}
When certifying \( \Rmult \), we need to control the overall error rate across multiple comparisons, as we are dealing with several confidence intervals for different classes. The Bonferroni correction is used here to adjust the significance level, ensuring valid hypothesis testing even with dependent tests \citep{benjamini1995controlling, hochberg1987multiple, dunn1961multiple}.
\begin{theorem}[\cite{hochberg1987multiple}]
    \label{thm:bonferroni}
    Let \( H_1, H_2, \dots, H_m \) be a family of \( m \) null hypotheses with p-values \( \vp_i \), and let \( \alpha \) be the desired family-wise error rate. The family-wise error rate ($\mathrm{FWER}$) is defined as
    \[
        \mathrm{FWER} = \mathbb{P}\left(\bigcup_{i=1}^{m} \{ \text{reject } H_i \mid H_i \text{ is true}\}\right),
    \]
    representing the probability of making at least one Type I error among the multiple tests. The Bonferroni correction sets the individual significance level for each test to \( \frac{\alpha}{m} \), such that.
    \[
        (H_i) \text{ is rejected if } \vp_i \leq \frac{\alpha}{m} \ .
    \]
    This ensures control of the family-wise error rate at level \( \alpha \), even for dependent tests.
\end{theorem}
To handle the multiple comparisons in our problem, we choose \( \alpha' = \frac{\alpha}{c} \), where \( c \) is the number of classes, thereby adjusting each risk level to control the overall error rate. This adjustment allows us to  validate \( (1 - \alpha) \)-level confidence intervals for all classes simultaneously.
We define
\[
    \mu = \Rmult(\underline{\hat{\vp}_{I_1}}, \overline{\hat{\vp}_{I_2}}) = \frac{\sigma}{2} \left( \quant(\underline{\hat{\vp}_{I_1}})
    - \quant(\overline{\hat{\vp}_{I_2}}) \right),
\]
where \( I_1 \) is the index of the highest lower bound \( \underline{\hat{\vp}_i} \) among the \( (1 - \alpha') \)-level Clopper-Pearson confidence intervals, and \( I_2 \) is the index of the second-highest upper bound \( \overline{\hat{\vp}_i} \), with \( i \neq I_1 \).

By constructing these intervals at level \( 1 - \alpha' \), Theorem~\ref{thm:bonferroni} ensures that the probability of any interval failing to cover the true parameter \( \vp_i \) is at most \( \alpha \), thus validating our confidence validations. Similarly:
\begin{align*}
    H_0:\  & \mathbb{P}\left( \Rmult(\vp) \geq \mu \right) \geq 1 - \alpha \\
    H_1:\  & \mathbb{P}\left( \Rmult(\vp) < \mu \right) < \alpha \ .
\end{align*}
The choice of \( \alpha' = \frac{\alpha}{c} \) ensures correct coverage across multiple intervals. Using a different value, such as the often picked \( \alpha' = \frac{\alpha}{2} \)~\citep{voracek2023improving, voracek2024treatment, scholten2023hierarchical, weber2023rab}, would lead to incorrect coverage assumptions, as shown by a counterexample in the Appendix involving a simple multinomial distribution for \( \vp \), where dependence assumptions among the confidence intervals are invalid.

\subsection{Reducing computational overhead with class partitioning method (CPM)}

Despite advances in robustness certification, a significant gap persists between theoretical bounds and empirical robustness. This section introduces novel methods to reduce this gap, including an efficient procedure for constructing tighter confidence intervals using class partitioning and a new certificate based on Lipschitz continuity.
\begin{figure*}[h]
    \centering
    \includegraphics[width=1.0\textwidth]{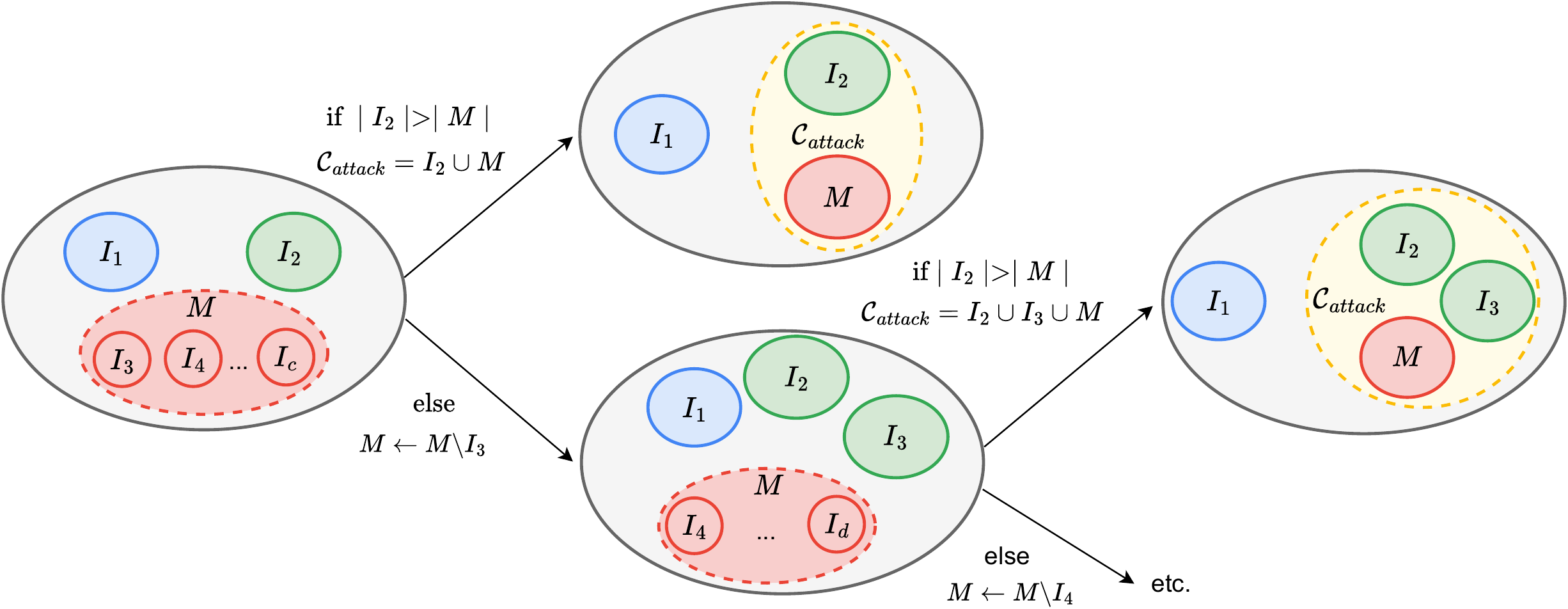}
    \caption{Illustration of the initial phase of the class partitioning method (CPM). The number of counts in class $I$ is noted $\mid I \mid$.}
    \label{fig:explanation_bucket_algo}
\end{figure*}
The robustness certification measure \( \Rmult \) performs better for smaller values of \( \hat{\vp}_{i_1} \), but it requires careful handling of confidence intervals and a conservative risk level \( \alpha' = \frac{\alpha}{c} \), where \( c \) is the number of classes. This conservative choice can be a limitation in cases with a large number of classes, such as ImageNet (\( c = 1000 \)), and a limited number of samples.

\begin{algorithm}
    \caption{Class partitioning method (CPM) for optimized Bonferroni correction}
    \label{algo:bucket_algo}
    \begin{algorithmic}[1]
        \Require Initial sample size \( n_0 \), estimation sample size \( n \), risk level \( \alpha \)
        \State \textbf{Initialize:} Class set \( \mathcal{C} = \{1, 2, \dotsc, c\} \)

        \Statex

        \State \textbf{Initial Sampling Phase:}

        \State Draw \( n_0 \) samples and compute the selection counts \( C_{\mathrm{select}}[i] \) for each class \( i \in \mathcal{C} \)

        \State \( I_1 \gets \arg\max_{i \in \mathcal{C}} C_{\mathrm{select}}[i] \) \Comment{Index of the class with the highest count}

        \State \( I_2 \gets \arg\max_{i \in \mathcal{C} \setminus \{I_1\}} C_{\mathrm{select}}[i] \) \Comment{Index of the class with the second-highest count}

        \State Initialize meta-class \( M \gets \mathcal{C} \setminus \{I_1, I_2\} \)

        \State Define \( \mathcal{C}_{\mathrm{attack}} \gets \{I_2, M\} \)

        \While{ \( \sum_{i \in M} C_{\mathrm{select}}[i] > C_{\mathrm{select}}[I_2] \) }
        \State \( k \gets \arg\max_{i \in M} C_{\mathrm{select}}[i] \)
        \State Remove class \( k \) from \( M \)
        \State Add class \( k \) to \( \mathcal{C}_{\mathrm{attack}} \)
        \EndWhile

        \State Let \( \mathcal{P} \) be the partitioning to form \( \mathcal{C}_{\mathrm{attack}} \)

        \Statex

        \State \textbf{Estimation Phase:}

        \State Draw \( n \) samples and compute the selection counts \( C_{\mathrm{estim}}[i] \) for each class \( i \in \mathcal{C} \)

        \State Apply partitioning \( \mathcal{P} \) to \( C_{\mathrm{estim}} \) to obtain bucket counts

        \State Compute \( \hat{\vp}_{I_1} = \dfrac{C_{\mathrm{estim}}[I_1]}{n} \)

        \For{ each bucket \( B \) in \( \mathcal{C}_{\mathrm{attack}} \) }
        \If{ \( B \) is a single class \( k \) }
        \State Compute \( \hat{\vp}_k = \dfrac{C_{\mathrm{estim}}[k]}{n} \)
        \Else \Comment{ \( B \) is meta-class \( M \) }
        \State Compute \( \hat{\vp}_M = \dfrac{\sum_{i \in M} C_{\mathrm{estim}}[i]}{n} \)
        \EndIf
        \EndFor

        \State Compute the total number of buckets \( c^\star = |\mathcal{C}_{\mathrm{attack}}| + 1 \)

        \State Compute the adjusted significance level \( \alpha' = \dfrac{\alpha}{c^\star} \)

        \State Compute the lower confidence bound \( \underline{\hat{\vp}}_{I_1} \) with risk \( \alpha' \)

        \For{ each bucket \( B \) in \( \mathcal{C}_{\mathrm{attack}} \) }
        \State Compute the upper confidence bound \( \overline{\hat{\vp}}_B \) with risk \( \alpha' \)
        \EndFor

        \State Let \( \overline{\hat{\vp}}_{\max} = \max_{B \in \mathcal{C}_{\mathrm{attack}}} \overline{\hat{\vp}}_B \)

        \Statex

        \State \textbf{Output:} Robustness radius \( R(\underline{\hat{\vp}}_{I_1}, \overline{\hat{\vp}}_{\max}) \)
    \end{algorithmic}
\end{algorithm}

To address the challenge of efficiently certifying robustness in multi-class classification, we propose a Class Partitioning Method (CPM) that integrates a structured partitioning strategy with multiple hypothesis testing using the Bonferroni correction, as detailed in Algorithm~\ref{algo:bucket_algo}. This method improves statistical power while reducing the computational cost of evaluating each class individually, making it particularly useful for large-scale settings such as ImageNet (\( c = 1000 \)). The CPM algorithm consists of two main phases: initial sampling and partitioning, followed by confidence-bound estimation.

\textbf{Initial sampling and class partitioning}
The algorithm begins by drawing an initial sample of size \( n_0 \) and computing the selection counts \( C_{\mathrm{select}}[i] \) for each class \( i \) in the set \( \mathcal{C} = \{1, 2, \dotsc, c\} \). The two classes with the highest selection counts are identified as \( I_1 \) the most probable class and \( I_2 \) the second most probable class.
The remaining classes are grouped into a meta-class \( M = \mathcal{C} \setminus \{I_1, I_2\} \). This setup reduces the number of direct comparisons by treating lower-probability classes as a single entity. The set of classes to attack is initialized as:
\( \mathcal{C}_{\mathrm{attack}} = \{I_2, M\}. \)
%
To refine the partitioning, whenever the total selection count of classes in \(M\) exceeds that of \(I_2\), the algorithm iteratively:
\begin{enumerate}[label=(\roman*)]
    \item removes from \(M\) the class \(k\) with the highest selection count, and
    \item adds it to \(\mathcal{C}_{\mathrm{attack}}\).
\end{enumerate}
This procedure repeats until the counts are balanced, yielding a partition \(\mathcal{P}\) that reduces the number of comparisons while preserving statistical power.

This strategy is visually illustrated in Figure~\ref{fig:explanation_bucket_algo}, where we show how the iterative reallocation of classes improves efficiency.

\textbf{Estimation phase and confidence bounds}
After partitioning, the algorithm draws an additional sample of size \( n \) and computes the selection counts \( C_{\mathrm{estim}}[i] \) for each class \( i \). Using the partitioning \( \mathcal{P} \), these counts are grouped into buckets, with empirical probabilities computed as:
\[ \hat{\vp}_{I_1} = \frac{C_{\mathrm{estim}}[I_1]}{n}, \quad \hat{\vp}_B = \frac{C_{\mathrm{estim}}[B]}{n}, \quad \forall B \in \mathcal{C}_{\mathrm{attack}}. \]

The total number of buckets is defined as:
\( c^\star = |\mathcal{C}_{\mathrm{attack}}| + 1. \)
To control the family-wise error rate, we apply the Bonferroni correction at the bucket level, adjusting the risk:
\( \alpha' = \frac{\alpha}{c^\star}. \)
This reduces conservativeness compared to the naive approach of \( \alpha' = \frac{\alpha}{c} \), resulting in tighter confidence intervals without sacrificing statistical guarantees.
Using the adjusted risk level, the algorithm computes the lower confidence bound \( \underline{\hat{\vp}}_{I_1} \) for \( I_1 \) and the upper confidence bounds \( \overline{\hat{\vp}}_B \) for each bucket \( B \in \mathcal{C}_{\mathrm{attack}} \).
The maximum upper bound across all attackable classes is then determined as:
\( \overline{\hat{\vp}}_{\max} = \max_{B \in \mathcal{C}_{\mathrm{attack}}} \overline{\hat{\vp}}_B. \)
Finally, the algorithm computes and outputs the robustness radius:
\( \mathrm{R}(\underline{\hat{\vp}}_{I_1}, \overline{\hat{\vp}}_{\max}), \)
which quantifies the classifier’s robustness against adversarial attacks.

Empirically, this partitioning approach significantly reduces the number of comparisons while improving the tightness of certified robustness bounds. By dynamically transitioning from \( \alpha' = \frac{\alpha}{c} \) to \( \alpha' = \frac{\alpha}{c^\star} \), the algorithm achieves more efficient robustness guarantees.

\subsection{Experiment}
\label{exp:exp_certif_imagenet_comparison_ic}
For all experiments, we use a subset of 500 images for ImageNet.
The risk level is set to \(\alpha = 0.001\), with \(n = 10^4\) samples and \(n_0 = 100\) used for class $I_1$ estimation. All experiments are conducted using a single A100 GPU.

\begin{figure*}
    \centering
    \includegraphics[width=1\linewidth]{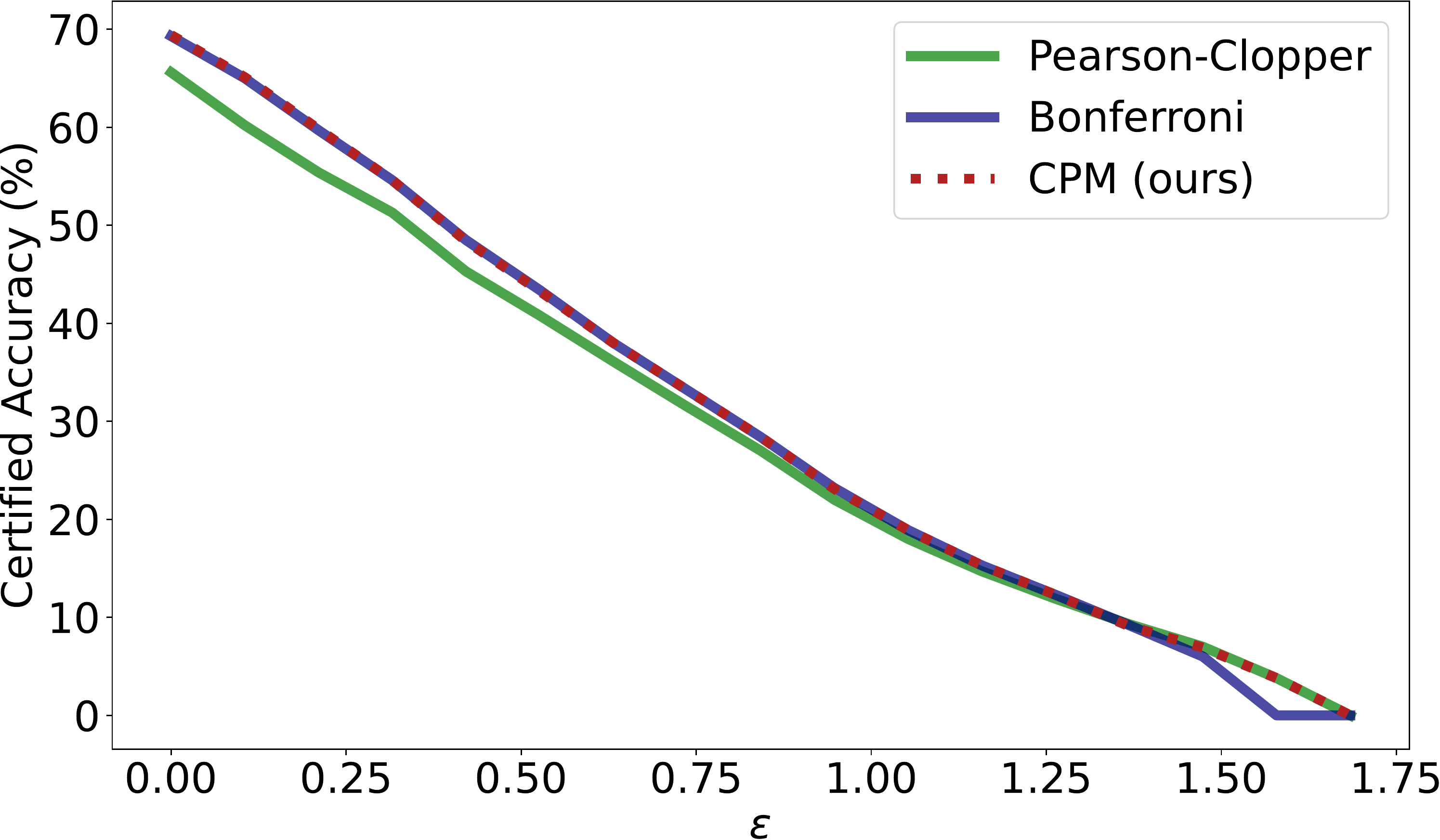}
    \caption{
        Comparison of various confidence interval methods for certified accuracy estimation with smoothing standard deviation $\sigma=0.5$ on the CIFAR-10 dataset, using ResNet-110 trained with noise injection ($\sigma=0.5$).
        The plot contrasts our CPM method with Bonferroni and Pearson-Clopper intervals across different perturbation levels $\epsilon$.}
    \label{fig:cifar10_sigma_0.5}
\end{figure*}
We evaluate our methods on the task of image classification using the CIFAR-10 dataset, employing a ResNet-110 architecture trained with noise injection (\(\sigma=0.5\)). The randomized smoothing procedure is applied by adding Gaussian noise to the inputs, providing probabilistic robustness guarantees. We report certified accuracy as a function of the level of \(\ell_2\) perturbation \(\epsilon\), which quantifies the size of adversarial perturbations against which the model is certified to be robust.
For certification, we use the certified radius \(\Rmult\) for CPM and the standard Bonferroni correction over \(c\) classes, while \(\Rmono\) is used for Pearson-Clopper intervals. Our CPM method typically produces an effective number of classes $c^\star$ between $2$ and $3$ providing a less conservative approach to certification. These gains are registered with no additional computational cost compared to the inference cost of the network.
As shown in Figure~\ref{fig:cifar10_sigma_0.5}, our CPM method achieves a balance between the strong performance of Bonferroni intervals at high perturbation levels and the robustness of Pearson-Clopper intervals at lower perturbations. This demonstrates its effectiveness in providing reliable certified accuracy across a wide range of \(\budget\) values.

\begin{figure*}[h]
    \centering
    \includegraphics[width=1\linewidth]{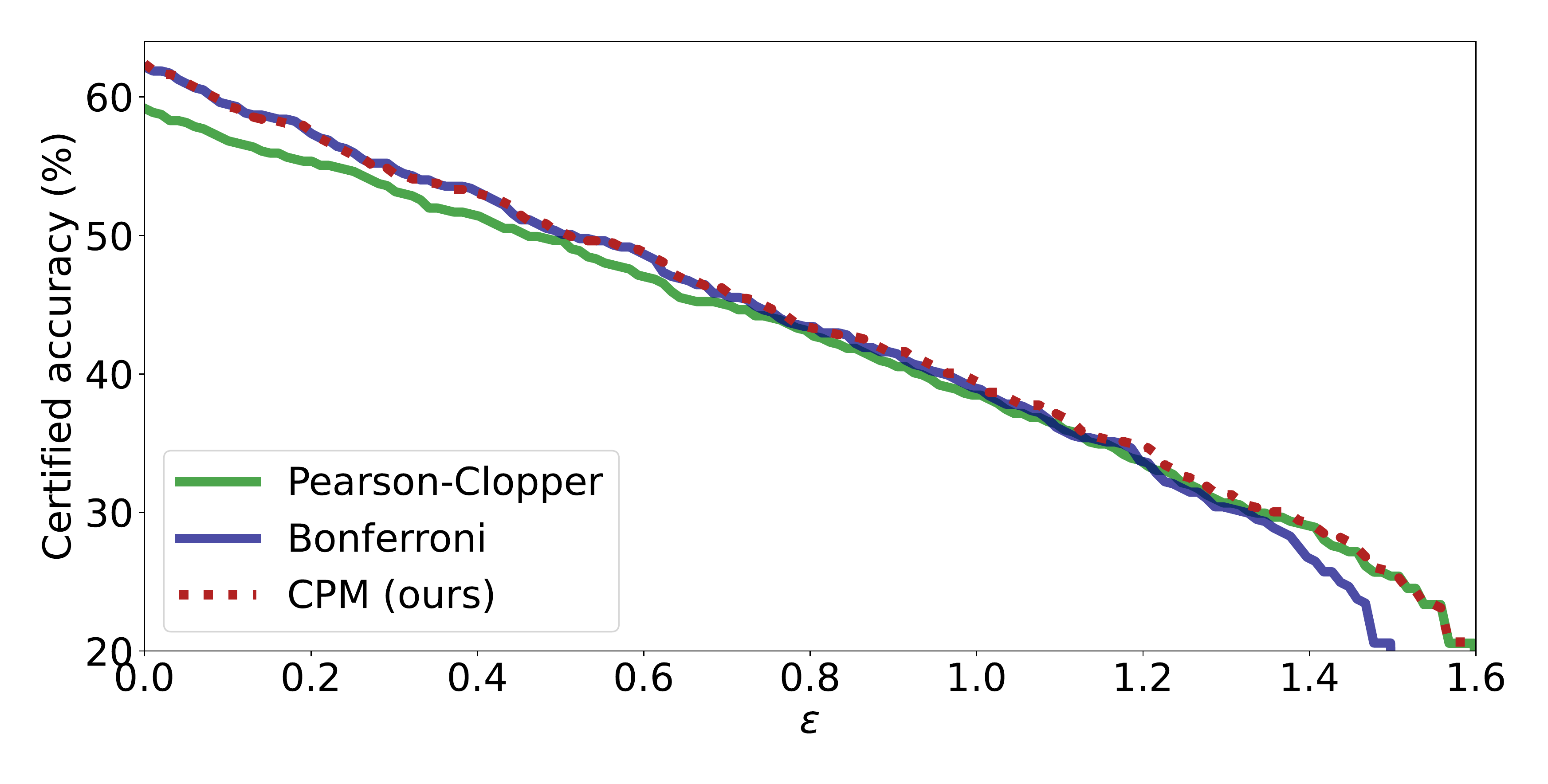}
    \caption{Comparison of various confidence interval methods for certified accuracy estimation with smoothing standard deviation $\sigma=0.5$ on the ImageNet dataset, using ResNet-50 trained with noise injection ($\sigma=0.5$). The plot contrasts our CPM method with Bonferroni and Pearson-Clopper intervals across different perturbation levels $\epsilon$.}
    \label{fig:ic^Heap_bf_pc-label}
\end{figure*}

To demonstrate the scalability of our method to a larger number of classes, we evaluate it on the ImageNet-1K dataset (1000 classes) using ResNet-50. As shown in Figure~\ref{fig:ic^Heap_bf_pc-label}, our CPM method provides tighter confidence intervals compared to the Bonferroni and Pearson-Clopper methods. Typically, CPM yields an effective number of classes, \( c^\star \), between 3 and 5. This improvement is particularly pronounced at both low and high perturbation levels, where CPM achieves a more accurate estimate of the certified accuracy. These results highlight the effectiveness of our approach in providing reliable robustness guarantees for large-scale classification tasks.

We provide additional results for different smoothing standard deviations in the Appendix~\ref{app:sec:additional_experiments_cpm}.
For high smoothing standard deviation $\sigma=1$ on CIFAR-10 CPM provides similar results to Pearson-Clopper while for ImageNet the gain remains significant.

We demonstrate the effectiveness of our CPM method in providing tighter confidence intervals for certification by reducing the number of classes in the design of the certified radius, however, the margin and the variance are not taken into account in this experiment. We will further investigate the impact of these factors in the next section.



\section{Lipschitz-variance-margin tradeoff}
\label{sec:lipschitz_variance_margin}

We introduce the \emph{Lipschitz-Variance-Margin Randomized Smoothing} (LVM-RS) procedure, illustrated in Figure~\ref{figure:mind_map}. This method is built upon the observation of a fundamental tradeoff between three core quantities underlying robustness certification: the Lipschitz constant of the network, the variance of Monte Carlo (MC) estimates, and the decision margin.
This tradeoff arises both in deterministic settings---such as Lipschitz-constrained networks---and in stochastic methods like randomized smoothing, where the classifier is a smoothed version of a base Lipschitz function. All such models share an intrinsic Lipschitz structure, but recognizing and formalizing this tradeoff reveals the underlying quantities that govern certified robustness.
We use this theoretical lens to design LVM-RS, a method that tackles the variance component directly. Specifically, we show that controlling the Lipschitz constant reduces the variance of the MC estimate, and we leverage Bernstein’s inequality to obtain tighter high-confidence bounds on the risk \( \alpha \). Additionally, we introduce a temperature-controlled simplex mapping to balance margin preservation and variance reduction. This regularized mapping mitigates the tradeoff, ensuring that robustness guarantees remain meaningful even under high noise levels.

\input{figures/mind_map}

\subsection{Lipschitz control for variance reduction}
\label{section:low_lip_low_variance}
Lipschitz continuity plays an important role in the sampling process, which is crucial for accurately estimating the smoothed classifier $\tilde{F}$. Specifically, by minimizing the local Lipschitz constant of a classifier $F$, one can reduce its variance. The following theorem illustrates this relationship for any $\sigma$.
\begin{theorem}[Gaussian Poincaré inequality \citep{boucheron2013concentration}]
    \label{prop:gaussian_poincarre_inequality}
    Let $Z = (Z_1, \dots, Z_n)$ represent a vector of i.i.d Gaussian random variables with variance $\sigma^2$. For any continuously differentiable function $h : \R^n \rightarrow \R$, the variance is given by:%
    \begin{align*}
        \V[h(Z)] \leq \sigma^2 \ \E\left[\norm{\nabla h(Z)}^2\right] \ .
    \end{align*}
\end{theorem}
We use the latter theorem to immediately derive:
\begin{corollary}
    \label{cor:gaussian_poincarre_inequality}
    With same hypothesis as Theorem~\ref{prop:gaussian_poincarre_inequality}, if $h$ exhibits Lipschitz continuity, we have that:
    \begin{align*}
        \V[h(Z)] \leq \sigma^2 \ \Lip(h)^2 \ .
    \end{align*}
\end{corollary}
Applying the above corollary to the classifiers $\{F_k\}$ which can be considered differentiable almost everywhere, it is evident that constraining the Lipschitz constant, $L(F_k)$, leads to a diminished variance for $F_k$.
This, in turn, results in a more precise estimation of $\E[F(\vx+\delta)]$, as captured by $\ \frac{1}{n} \sum_{i=1}^n F(\vx + \delta_i)$. Lowering the local Lipschitz constraints can significantly attenuate the variance and improve the certification results, but it can be too restrictive and cause a drop in performance.
To enforce low Lipschitz and reduce performance loss,~\cite{cohen2019certified} proposed the injection of Gaussian noise during training,~\cite{salman2019provably} introduced $\mathrm{SmoothAdv}$, which involves adversarial training of the smoothed classifier $\tilde{F}$, to reduce its local Lipschitz constant. The work of~\cite{pal2023understanding} studied how the noisy training on the \emph{sub-classifier} affects the performance and robustness of the \emph{smoothed classifier}.
Other noteworthy methods include those by \cite{salman2020denoised, carlini2023certified}, which combine a conventional classifier with a denoiser diffusion model, ensuring that the resulting architecture remains invariant to Gaussian noise, thereby giving Lipschitz continuity to the classifier and preserved performance.
\subsection{Statistical risk management for low variance}
\label{section:statistical_risk_management_in_rs}
To effectively leverage low-variance estimators in randomized smoothing, one must rely on confidence intervals or concentration inequalities where the variance plays a central role in risk control.

The Clopper–Pearson interval, tailored for binomial proportions, provides an exact \( \alpha \)-level confidence interval and is commonly used to compute lower and upper bounds \(\underline{\hat{\vp}}\) and \(\overline{\hat{\vp}}\), as in~\citet{cohen2019certified, carlini2023certified}. This interval naturally applies when the randomized classifier outputs are obtained via a \textit{hardmax} mapping, resulting in Bernoulli trials in Monte Carlo sampling.
By contrast, methods such as~\cite{lecuyer2019certified} and~\cite{levine2019certifiably}, which smooth scalar outputs in \([0,1]\) rather than discrete class labels, do not generate Bernoulli outcomes and thus cannot use the Clopper–Pearson interval. Instead, they rely on \emph{Hoeffding’s inequality}, which also guarantees an \( \alpha \)-level coverage but does not exploit the variance of the estimate.
Other inequalities such as the \emph{sub-Gaussian} bound, related to the Gaussian–Poincaré inequality~\citep{massart2007concentration}, incorporate the Lipschitz constant \(\Lip(F)\) of the classifier. However, this approach has practical limitations: computing \(\Lip(F)\) is NP-hard in general for neural networks, and worst-case Lipschitz bounds often drastically overestimate the true empirical variance.
To overcome these issues, we advocate for the use of the \emph{Empirical Bernstein inequality}, particularly when the empirical variance is low. This inequality yields tighter confidence intervals by explicitly incorporating observed variance, and still provides high-probability guarantees for the risk level \( \alpha \). While mentioned in~\cite{lecuyer2019certified}, we build on this idea and integrate it directly into our certification framework.
\begin{proposition}[Empirical Bernstein's inequality \citep{maurer2009empirical}]
    \label{prop:empirical_bernstein_inequality}
    Let \( Z_0, Z_1, \dots, Z_n \) be i.i.d random variables with values between 0 and 1. The risk level is denoted as \( \alpha  \in [0, 1] \).
    Then with probability at least  \( 1-\alpha \) in vector \( Z = (Z_1, \dots, Z_n) \), we have
    \begin{align}
        \label{eq:empirical_bernstein_inequality}
        \E [Z_0] - \frac{1}{n} \sum_{i=1}^n Z_i \leq \sqrt{\frac{2 S_n(Z) \log (2/\alpha)}{n}}
        + \frac{7 \log (2 / \alpha)}{3(n-1)}
        \ .
    \end{align}
    Here, \( S_n(Z) \) represents the sample variance \( \frac{1}{n(n-1)} \sum_{1 \leq i < j \leq n} (Z_i - Z_j)^2 \). Note that the bound is symmetric around $\E Z_0$.
\end{proposition}

In our setting, we apply the Empirical Bernstein inequality to the random variables \( Z_i = F_k(\vx + \delta_i) \). This formulation provides the flexibility to smooth a wide range of simplex mappings \( s \), beyond the traditional \(\mathrm{hardmax}\), thereby enabling better control of the tradeoff between margin and variance. Crucially, it allows us to construct exact confidence intervals for scalar outputs \( F_k(\vx + \delta) \), regardless of whether the distribution is discrete or continuous.
Also it allow to leverage the variance of the empirical distribution, which is particularly useful when the variance is low, as it leads to tighter bounds than those provided by Hoeffding's inequality. See Fig.~\ref{fig:comparison_concentration}, for a comparison between Hoeffding’s and Bernstein’s inequalities.
%
\subsection{Scaled simplex for high margin and low variance}
\label{section:r_simplex}
The $\mathrm{hardmax}$ mapping maximizes class margins by assigning all mass to the predicted class, yielding the largest possible margins on the simplex. However, $\mathrm{hardmax}$ is not Lipschitz continuous, and as shown in Example~\ref{ssec:example_var_argmax}, it can lead to significant variance amplification.
In contrast, the $\mathrm{softmax}$ mapping smooths class margins, reducing the difference between logits. While this compresses margins, $\mathrm{softmax}$ maintains $1$-Lipschitz continuity, preventing variance explosion.
To address the limitations of $\mathrm{softmax}$ and $\mathrm{hardmax}$,~\cite{martins2016softmax} introduced $\mathrm{sparsemax}$, a novel projection that is $1$-Lipschitz and yields margins larger than $\mathrm{softmax}$ but smaller than $\mathrm{hardmax}$.  This mapping promotes sparsity in the output distribution compared to $\mathrm{softmax}$, reducing variance through contraction, as formalized in Corollary~\ref{cor:gaussian_poincarre_inequality}.

To further refine the variance-margin trade-off, we introduce temperature scaling to simplex mappings. For a temperature \( T > 0 \), we define the scaled mapping $\tau^T: \mathbb{R}^c \to \Delta^{c-1}$ as:
\[
    \tau^{T}(\vz) = \tau\left(\frac{\vz}{T}\right).
\]
By adjusting the temperature \( T \), we interpolate between \( \mathrm{softmax} \) and \( \mathrm{hardmax} \), or between \( \mathrm{sparsemax} \) and \( \mathrm{hardmax} \). This temperature scaling allows for the selection of an optimal simplex mapping that balances variance reduction with margin maximization, addressing the variance-margin trade-off. The effect of temperature tuning is illustrated in Figure~\ref{fig:comparison_projection_simplex}.

\subsection{LVM-RS inference procedure}
\label{section:lvm_rs}

Given a trained neural network $f$, we choose a simplex mapping $\tau$ from a set of simplex maps $\mathcal{T}$ and a temperature $T \in [T_{\mathrm{lower}}, T_{\mathrm{upper}}]$,
defining an ensemble of smoothed soft classifiers $\{\tilde{F}^{\tau^T}\}$:
\begin{equation*}
    \tilde{F}^{\tau^T}(\mathbf{x})
    = \mathbb{E}_{\boldsymbol{\delta} \sim \mathcal{N}(0, \sigma^2 \mathbf{I})}
    \left[{\tau}_k^T(f(\mathbf{x} + \boldsymbol{\delta})) \right] \ .
\end{equation*}


We begin by sampling \( n_0 \) i.i.d. Gaussian perturbations \( \delta_i \sim \mathcal{N}(0, \sigma^2 I) \), and compute the smoothed output \( Z_i = F_k(\vx + \delta_i) \in [0,1] \) for each sample. We then form the empirical estimate:
\[
    \hat{\vp}^{\tau^T}_k := \frac{1}{n_0} \sum_{i=1}^{n_0} Z_i
    \quad\text{with variance}\quad
    \hat{v}^{\tau^T}_k := \frac{1}{n_0(n_0 - 1)} \sum_{i=1}^{n_0} (Z_i - \hat{\vp}^{\tau^T}_k)^2.
\]

Applying the empirical Bernstein inequality (Equation~\ref{eq:empirical_bernstein_inequality}) with risk level \( \alpha \), we obtain a high-probability lower bound \( \underline{\vp}^{\tau^T}_k \) and upper bound \( \overline{\vp}^{\tau^T}_k \) on the true mean:
\[
    \underline{\vp}^{\tau^T}_k = \hat{\vp}^{\tau^T}_k - \sqrt{ \frac{2 \hat{v}^{\tau^T}_k \log(2/\alpha)}{n_0} } - \frac{7 \log(2/\alpha)}{3(n_0 - 1)} ,
\]
\[
    \overline{\vp}^{\tau^T}_k = \hat{\vp}^{\tau^T}_k + \sqrt{ \frac{2 \hat{v}^{\tau^T}_k \log(2/\alpha)}{n_0} } + \frac{7 \log(2/\alpha)}{3(n_0 - 1)}.
\]

We denote by \( \underline{\vp}^{\tau^T}_{I_1} \) and \( \overline{\vp}^{\tau^T}_{I_2} \) the corrected probabilities associated respectively with the predicted class \( I_1 \) and the highest non-predicted class \( I_2 \). These corrected estimates are then used to compute the certified radius \( \Rmult(\underline{\vp}^{\tau^T}_{I_1}, \overline{\vp}^{\tau^T}_{I_2}) \), which accounts for estimation uncertainty at level \( \alpha \).

Finally, to maximize robustness, we perform a grid search over temperature and smoothing parameters and select the optimal pair:
\[
    (\tau^\star, T^\star) = \argmax_{\tau, T} \Rmult(\underline{\vp}^{\tau^T}_{I_1}, \overline{\vp}^{\tau^T}_{I_2}).
\]

Then, a sampling of size $n$ is performed and we evaluate an MC estimate of $\tilde{F}^{{\tau^\star}^{T^\star}}(\mathbf{x})$,
which gives $\hat{\mathbf{p}}^\star$ and associated risk-corrected $\underline{\mathbf{p}_{I_1}}^\star$ and  $\overline{\mathbf{p}_{I_2}}^\star$.
We return the prediction $I_1$ and associated certified radius $\Rmult(\underline{\mathbf{p}_{I_1}}^\star, \overline{\mathbf{p}_{I_2}}^\star)$.

\begin{algorithm}[htbp]
    \caption{LVM-RS ($f$, $\sigma$, $\mathbf{x}$, $n_0$, $n$)}
    \begin{algorithmic}[1]
        \State $\mathrm{samples}_{n_0} \gets \mathrm{sample\_scores}(f, \mathbf{x}, n_0, \sigma)$ \Comment{validation set, $n_0 \times c$}
        \State $\mathrm{samples}_{n} \gets \mathrm{sample\_scores}(f, \mathbf{x}, n, \sigma)$ \Comment{certification set, $n \times c$}
        \For{$T \in [T_{\mathrm{lower}}, T_{\mathrm{upper}}]$}
        \For{$\tau \in \mathcal{T}$}
        \State $\underline{\mathbf{p}_{I_1}}^{\tau^T} \gets \frac{1}{n_0} \sum_{i=1}^{n_0}
            \tau^T(\mathrm{samples}_{n_0}[i, :]) -
            \mathrm{shift}(S_{n_0}(\tau^T(\mathrm{samples}_{n_0})), \alpha, n_0)$ 
        \State $\overline{\mathbf{p}_{I_2}}^{\tau^T} \gets \frac{1}{n_0} \sum_{i=1}^{n_0}
            \tau^T(\mathrm{samples}_{n_0}[i, :]) +
            \mathrm{shift}(S_{n_0}(\tau^T(\mathrm{samples}_{n_0})), \alpha, n_0)$ 
        \EndFor
        \EndFor
        \State $(\tau^\star, T^\star) \gets \argmax_{\tau, T}
            \Rmult\left(\underline{\mathbf{p}_{I_1}}^{\tau^T}, \overline{\mathbf{p}_{I_2}}^{\tau^T}\right)$
        \State $\underline{\mathbf{p}_{I_1}}^\star \gets \frac{1}{n_0}
            \sum_{i=1}^{n_0} {\tau^\star}^{T^\star}(\mathrm{samples}_{n}[i, :]) +
            \mathrm{shift}(S_n(\tau_\star^{t^\star}(\mathrm{samples}_{n})), \alpha, n)$ \Comment{dim. $c$}
        \State $\overline{\mathbf{p}_{I_2}}^\star \gets \frac{1}{n_0}
            \sum_{i=1}^{n_0} {\tau^\star}^{T^\star}(\mathrm{samples}_{n}[i, :]) -
            \mathrm{shift}(S_n(\tau_\star^{t^\star}(\mathrm{samples}_{n})), \alpha, n)$ \Comment{dim. $c$}
        \State \Return prediction $\argmax_{k} \bar{\mathbf{p}}^\star_k$ and certified radius $\Rmult(\underline{\mathbf{p}_{I_1}}^\star, \overline{\mathbf{p}_{I_2}}^\star)$
    \end{algorithmic}
    \label{algo:lvmrs}
\end{algorithm}

\subsection{Experiment on certified accuracy with LVM-RS}
\label{ssec:expe_lvm_rs}

In this experiment, we empirically validate the efficacy of our proposed inference procedure presented in Algorithm~\ref{algo:lvmrs}. Our approach leverages the variance-margin trade-off to achieve state-of-the-art RS results. We use the following parameters: temperatures: $T_{\mathrm{lower}} = 0.01$, $T_{\mathrm{upper}} = 50$ (50 values) and implex maps: $\mathcal{T} = \{\mathrm{sparsemax}, \mathrm{softmax}, \mathrm{hardmax}\}$.

In this experiment, we empirically validate the efficacy of our proposed inference procedure presented in Algo.~\ref{algo:lvmrs}, highlighting its capability to improve randomized smoothing and achieve certified accuracy. Central to our approach is the leveraging of the variance-margin tradeoff, which as we demonstrate, yields state-of-the-art RS results. We further showcase how the procedure enhances the off-the-shelf state-of-the-art baseline model of \cite{carlini2023certified}, which utilizes a vision transformer coupled with a denoiser for randomized smoothing.
Note that the choice of simplex mapping and temperature depends on the input.
%
The baseline consists of the model of~\cite{carlini2023certified} which does smoothing of $\mathrm{hardmax}$ of \emph{base classifier} and uses the Pearson-Clopper confidence interval to control the risk $\alpha$.

To compare the baseline with our method, certified accuracies are computed with $\Rmult$ in the function of the level of perturbations $\epsilon$, for different noise levels $\sigma \in \{0.25, 0.5, 1\}$.
Results are presented in Figure~\ref{fig:certified_accuracy_for_different_eps_cifar10} for CIFAR-10 and in Figure~\ref{fig:certified_accuracy_for_different_eps_imagenet} for ImageNet. We see that our method increases results, especially in the case of high $\sigma$, in the case of $\sigma \in \{0.5, 1.0 \}$ the overall certified accuracy curve in the function of $\epsilon$ the maximum perturbation is lifted towards higher accuracies. Results are presented in Table~\ref{tab:best_results_cifar10} for CIFAR-10 and in Table~\ref{tab:best_results_imagenet} for ImageNet.
Computation was performed on GPU V100, reported average time is the computational cost of one input $\vx$ proceeds by RS and LVM-RS, we see that the computation gap between the two methods is narrow for CIFAR10 but is a bit wider for ImageNet.
Detailed results are presented in Appendix~\ref{appendix:expe_lvm_rs}.
\begin{table}[t]
    \centering
    \caption{Best certified accuracies across \( \sigma \in \{0.25, 0.5, 1.0\} \) for different levels of perturbation $\epsilon$, on CIFAR-10, for \( n=10^5 \) samples and risk \( \alpha=1\mathrm{e-}3 \).}
    \begin{tabular}{lcccccc}
        \toprule
        \multirow{2}[2]{*}{\textbf{Methods}} & \multicolumn{5}{c}{{\boldmath{}\textbf{Best certified accuracy (\( \varepsilon \))}\unboldmath{}}} & \multirow{2}[2]{*}{\textbf{Average time (s)}}                                                           \\
        \cmidrule{2-6}
                                             & 0.0                                                                                                & 0.25                                          & 0.5            & 0.75           & 1.0            &      \\
        \midrule
        {\citeauthor{carlini2023certified}}  & 86.72                                                                                              & 74.41                                         & 58.25          & 40.96          & 29.91          & 7.10 \\
        \textbf{LVM-RS (ours)}               & \textbf{88.49}                                                                                     & \textbf{76.21}                                & \textbf{60.22} & \textbf{43.76} & \textbf{32.35} & 7.11 \\
        \bottomrule
    \end{tabular}
    \label{tab:best_results_cifar10}
    \vspace{-0.2cm}
\end{table}
\begin{table}[t]
    \centering
    \caption{Best certified accuracies across $\sigma \in \{0.25, 0.5, 1.0\}$ for different levels of perturbation $\epsilon$, on ImageNet, for $n=10^4$ samples and risk $\alpha=1\mathrm{e-}3$.}
    \begin{tabular}{lccccccc}
        \toprule
        \multirow{2}[2]{*}{\textbf{Methods}} & \multicolumn{6}{c}{{\boldmath{}\textbf{Best certified accuracy ($\varepsilon$)}\unboldmath{}}} & \multirow{2}[2]{*}{\textbf{Average time (s)}}                                                                            \\
        \cmidrule{2-7}
                                             & 0.0                                                                                            & 0.5                                           & 1.0            & 1.5            & 2              & 3              &      \\
        \midrule
        {\citeauthor{carlini2023certified}}  & 79.88                                                                                          & 69.57                                         & 51.55          & \textbf{36.04} & 25.53          & 14.01          & 6.46 \\
        \textbf{LVM-RS (ours)}               & \textbf{80.66}                                                                                 & \textbf{69.84}                                & \textbf{53.85} & \textbf{36.04} & \textbf{27.43} & \textbf{14.31} & 7.03 \\
        \bottomrule
    \end{tabular}
    \label{tab:best_results_imagenet}
    \vspace{-0.2cm}
\end{table}
\section{Conclusion}
This chapter addressed risk management for randomized smoothing (RS) to achieve certified adversarial robustness. We demonstrated the superiority of multi-class certified radii \( R_{\text{multi}} \) over the mono-class approach \( R_{\text{mono}} \), particularly in high-variance scenarios where \( R_{\text{mono}} \) fails to provide meaningful certification. Our empirical analysis highlighted the increasing gap between \( R_{\text{multi}} \) and \( R_{\text{mono}} \) as variance grows.

To mitigate the computational overhead associated with multi-class certification, we introduced the Class Partitioning Method (CPM), which refines Bonferroni corrections by partitioning classes into relevant buckets. This method reduced conservativeness and enhanced certified accuracy without increasing computational complexity. Experiments on CIFAR-10 and ImageNet validated CPM's efficiency, showing notable improvements in certification accuracy at different noise levels.

We further addressed the variance-margin tradeoff by introducing the Lipschitz-Variance-Margin (LVM) tradeoff, which operates at the model level to minimize variance during Monte Carlo sampling. By leveraging scalable simplex mappings and temperature scaling, LVM improves the stability of smoothed classifiers. Our combined RS + LVM approach consistently outperformed baseline methods, achieving state-of-the-art results across perturbation budgets.

In the next chapter, we explore the impact of Lipschitz's constant regularization on neural network generalization.
Additionally, we extend randomized smoothing principles to the parameter space rather than the input space, aiming to enhance generalization further.

%% file: figures/mind_map.tex
\begin{figure*}[t]
  \centering
  \scalebox{0.65}{%
    \input{figures/figurev3}
  }%
  \caption{ 
\textbf{First}, \cite{tsuzuku2018lipschitz} proposes a deterministic certificate starting from a Lipschitz \emph{base subclassifier}, followed by margin calculation and radius binding. 
\textbf{Second}, \cite{cohen2019certified} introduces a \emph{base subclassifier} to create a \emph{smoothed subclassifier}. The risk factor $\alpha$ is then estimated using the Clopper-Pearson interval to provide a probabilistic certificate. 
\textbf{Third}, our method  (the \emph{Lipschitz-Variance-Margin Randomized Smoothing} or LVM-RS) extends a smoothed classifier constructed with a Lipschitz base classifier composed with a map which transforms logit to probability vector in simplex. The regularization of the Lipschitz constant is motivated by the Gaussian-Poincar\'e inequality in Theorem~\ref{prop:gaussian_poincarre_inequality}. The empirical variance is applied to the Empirical Bernstein inequality in Proposition~\ref{prop:empirical_bernstein_inequality} to accommodate for the risk factor $\alpha$, in the same flavor as in \cite{levine2019certifiably}. The pipeline also ends with a probabilistic certificate, 
similar to the methodology used in \cite{cohen2019certified}'s certified approach.}
  \label{figure:mind_map}
\end{figure*}
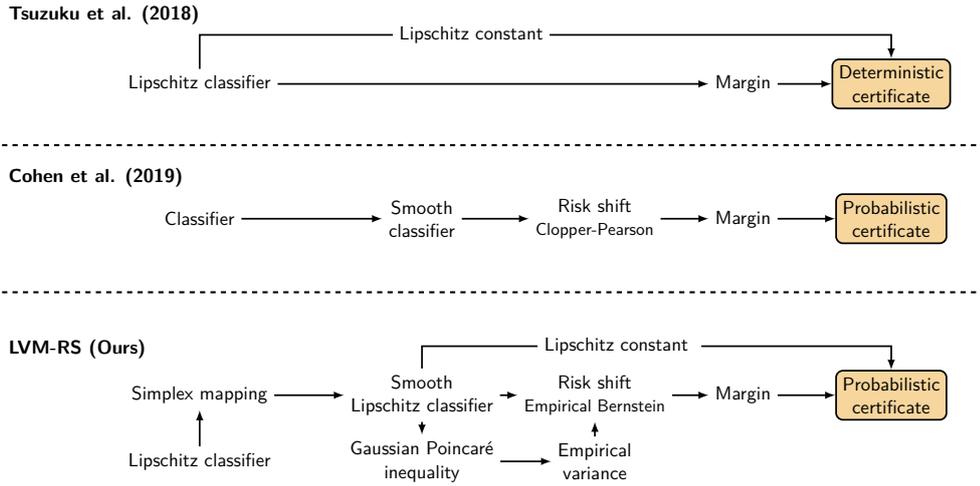

%% file: figures/figurev3.tex
\tikzset{%
  >={Latex[width=0mm,length=0mm]},
  base/.style = {fill=white, font=\sffamily},
  inputs0/.style={
    rounded corners,
    draw=black,
    line width=1pt,
    fill=burstblue!50,
  },
  inputs/.style={
    rounded corners,
    fill=none,
    draw=none,
    line width=1pt,
    minimum width=2cm,
    minimum height=1cm,
  },
  crops/.style={
    draw=black,
    rounded corners=0,
    line width=1pt,
    minimum width=0.75cm,
    minimum height=0.56cm
  },
  /.style={
  },
  rectangleinput/.style={
    rounded corners,
    fill=myorange,
    draw=black,
    line width=1pt,
    minimum width=3.5cm,
    minimum height=1cm
  },
  rectangle1/.style={
    rounded corners,
    fill=myorange,
    draw=black,
    line width=1pt,
    minimum width=2cm,
    minimum height=1cm
  },
  rectangle2/.style={
    rounded corners,
    fill=white,
    draw=black,
    dashed,
    line width=1pt,
    minimum width=2cm,
    minimum height=1cm
  },
  descr/.style={
    fill=white,
    inner sep=2.5pt
  },
  connector/.style={
    -latex,
    color=black,
    line width=1pt,
  },
  rectangle connector/.style={
    connector,
    to path={(\tikztostart) |- ++(3.1,0.25) \tikztonodes -- (\tikztotarget)},
    pos=0.5
  },
}
\def\shifta{+0.25}
\def\shiftb{-0.50}
\def\shiftc{-1.30}
\begin{tikzpicture}[every node/.style=base, align=center, scale=5]

  \draw[dashed,very thick] (+0., +0.15) -- (+4.0, +0.15);
  \draw[dashed,very thick] (+0., -0.45) -- (+4.0, -0.45);

  \node[fill=none, anchor=west] (ref1)   at (+0, 0.43+\shifta) {\textbf{Tsuzuku et al. (2018)}};
  \node[base] (lip_clas1) at (+0.8, +0.15+\shifta) {Lipschitz classifier};
  \node[base] (margin1)   at (+3.0, +0.15+\shifta) {Margin};
  \node[rectangle1] (certificate1) at (+3.6, +0.15+\shifta) {Deterministic \\ certificate};
  \draw[connector] (lip_clas1) -- (margin1);
  \draw[connector] (margin1) -- (certificate1);
  \draw[connector] (lip_clas1.north) -- (+0.8, 0.35+\shifta) -| (certificate1.north);
  \node[base,fill=white] (lipschitz) at (+1.9, 0.35+\shifta) {Lipschitz constant};

  \node[fill=none, anchor=west] (ref2)   at (+0, 0.52+\shiftb) {\textbf{Cohen et al. (2019)}};
  \node[base] (classifier)  at (+0.8, +0.35+\shiftb) {Classifier};
  \node[base] (smooth1)     at (+1.7, +0.35+\shiftb) {Smooth \\ classifier};
  \node[base] (risk_shift1) at (+2.4, +0.35+\shiftb) {Risk shift \\ {\small Clopper-Pearson}};
  \node[base] (margin2)     at (+3.0, +0.35+\shiftb) {Margin};
  \node[rectangle1] (certificate2) at (+3.6, +0.35+\shiftb) {Probabilistic \\ certificate};

  \draw[connector] (classifier) -- (smooth1);
  \draw[connector] (smooth1) -- (risk_shift1);
  \draw[connector] (risk_shift1) -- (margin2);
  \draw[connector] (margin2) -- (certificate2.west);

  \node[fill=none, anchor=west] (ref3)   at (+0, 0.62+\shiftc) {\textbf{LVM-RS (Ours)}};
  
  \node[base] (lip_clas2)     at (+0.8, 0.16+\shiftc) {Lipschitz classifier};
  \node[base] (simplex)       at (+0.8, 0.43+\shiftc) {Simplex mapping}; 
  \node[base] (smooth2)       at (+1.7, 0.43+\shiftc) {Smooth \\ Lipschitz classifier};
  \node[base] (emp_variance)  at (+2.4, 0.16+\shiftc) {Empirical \\ variance};
  \node[base] (risk_shift2)   at (+2.4, 0.43+\shiftc) {Risk shift \\ {\small Empirical Bernstein}};
  \node[base] (margin3)       at (+3.0, 0.43+\shiftc) {Margin};
  \node[base] (poincare)      at (+1.7, 0.16+\shiftc) {Gaussian Poincar\'e \\ inequality};
  \node[rectangle1] (certificate3) at (+3.6, 0.43+\shiftc) {Probabilistic \\ certificate};

  \draw[connector] (simplex) -- (smooth2);
  \draw[connector] (smooth2) -- (poincare);
  \draw[connector] (poincare) -- (emp_variance);
  \draw[connector] (emp_variance) -- (risk_shift2);
  \draw[connector] (smooth2) -- (risk_shift2);
  \draw[connector] (risk_shift2) -- (margin3);
  \draw[connector] (lip_clas2) -- (simplex);

  \draw[connector] (smooth2.north) -- (+1.7, 0.63+\shiftc) -- (3.6, 0.63+\shiftc) -- (certificate3.north);
  \draw[connector] (margin3) -- (certificate3.west);
  \node[base,fill=white] (lipschitz2) at (+2.5, 0.63+\shiftc) {Lipschitz constant };

\end{tikzpicture}%

%% file: content/chapter-regularization.tex
%
\chapter{Regularization to enhance generalization}
\label{chap:regularization}

\minitoc%

Generalization remains a fundamental challenge in deep learning, particularly as models grow in complexity and are trained on increasingly large and diverse datasets. Regularization techniques aim to improve generalization by constraining the learning process and guiding models toward solutions that perform well on unseen data. In this chapter, we explore two complementary approaches to regularization: Lipschitz regularization, which constrains the overall sensitivity of a network, and flatness-based regularization, which focuses on minimizing sharpness in the loss landscape.

To enforce Lipschitz regularization, we leverage the Gram iteration method developed in Chapter~\ref{chapter:spectral_norm_estimation}. This technique allows us to efficiently regularize the Lipschitz constants of individual layers, including convolutional layers, by estimating and constraining their spectral norms. By reducing the Lipschitz constant, we enhance the network's robustness and generalization performance.

For flatness-based regularization, we present a theoretical framework to quantify the reduction in the Hessian curvature of the loss landscape with Gaussian smoothing using the result on Weierstrass transform from Chapter~\ref{chap:lipschitz_computation}  and we link flat minima to Lipschitz continuity.
We introduce a novel regularization method, activation decay, which directly reduces the sharpness of the loss by smoothing the penultimate activations. This approach not only flattens the loss landscape but also improves generalization and robustness, providing a computationally efficient alternative to existing sharpness-aware techniques.

\section{Lipschitz-based regularization}

\label{section:spectral_norm_regularization}
The Lipschitz constant of a network plays a critical role in generalization
bounds \citep{bartlett2017spectrally}, as it quantifies the sensitivity of the model to variations in the input data, thereby linking generalization performance with robustness.
To regularize the Lipschitz constant, we apply spectral norm regularization on both the convolutional filters and dense layers, following the approach of~\citet{singla2021fantastic, araujo_lipschitz_2020}, as described in previous Section~\ref{section:reg_conv_layers}.
This technique is analogous to weight decay, which regularizes the Frobenius norm of the weights.

The regularization term added to the loss function is:
\begin{equation}
    \loss_{\text{reg}} = \mu_{\text{reg}} \sum_{l=1}^\nblayers \left\| \mW^{(l)} \right\|_2 \ .
\end{equation}
Regularizing each layer individually constrains the overall Lipschitz constant of the network via the product's upper bound.
The regularized training loss writes as $$\loss = \loss_{\mathrm{CE}} + \loss_{\text{reg}} .$$
Applying direct $\PUB$ as a regularization term is not ideal, as the product of Lipschitz bounds can become unstable and grow unbounded. One alternative is to employ the logarithm of the product, as proposed by~\cite{araujo_lipschitz_2020}. However, empirical results indicate that summing individual bounds during regularization yields the best performance.

To assess the impact of spectral norm regularization on classification accuracy and training time, we conducted experiments on CIFAR-10 using the ResNet-18 architecture~\citep{he2016deep}. The training setup follows the protocol described in Chapter~\ref{chap:lipschitz_computation}, with the following hyperparameters: 200 epochs, batch size 256, SGD with momentum 0.9, an initial learning rate of 0.1, and a cosine annealing schedule. The baseline model is trained without any regularization (\ie $\mu_{\text{reg}} = 0$).

We compare several spectral norm estimation methods from prior work~\citep{singla2021fantastic, araujo2021lipschitz, ryu2019plug} against our proposed estimator, denoted \texttt{norm2\_circ} (see Algorithm~\ref{algo:gram_iteration_conv}). Our method relies on explicit backward-mode differentiation (Equation~\ref{eq:gram_iteration_bound_gradient}) to improve both computational efficiency and memory usage.

\paragraph{Regularization terms.}
We apply two types of regularization to the convolutional layers:

\begin{itemize}
    \item \textbf{Spectral norm regularization:} we penalize the deviation of the spectral norm from a target threshold using
          \[
              \mathcal{L}_{\text{reg}} = \mu_{\text{reg}} \sum_{\ell}  \|\mW^{(\ell)}\|_2^2,
          \]
          where $\|\mW^{(\ell)}\|_2$ denotes the spectral norm of layer $\ell$, and $\mu_{\text{reg}} = 1\text{e-}1$.

    \item \textbf{Weight decay (WD) / Frobenius norm regularization:} we also apply a standard weight decay regularization
          \[
              \mathcal{L}_{\text{reg}} = \mu_{\text{wd}} \sum_{\ell} \|\mW^{(\ell)}\|_F^2,
          \]
          with $\mu_{\text{wd}} = 5\text{e-}3$.
\end{itemize}

\paragraph{On the choice of regularization coefficients.}
The coefficients $\mu_{\text{reg}}$ and $\mu_{\text{wd}}$ are not directly comparable and must be chosen separately. This is because the Frobenius norm scales with the square root of the number of weights in a layer, while the spectral norm corresponds to the largest singular value and is typically of smaller magnitude. Using the same coefficient for both would result in significantly unbalanced regularization strengths.

\paragraph{Additional details.}
We do not report results from~\citet{sedghi2019singular}, as their exact spectral norm computation is computationally prohibitive—requiring over 6750 seconds per epoch.
We also evaluate the effect of the Gram iteration count $t \in {3, 4, 5, 6}$ at the end of training. Higher values of $t$ reduce the standard deviation of the spectral norm estimate, with all reported training times corresponding to $t=6$. For statistical robustness, all experiments are repeated 10 times.

Results of experiments are reported in Table~\ref{tab:test_accuracies_method_reg_cifar10}.
Most of the bounds for Lipschitz regularization~\citep{ryu2019plug, araujo2021lipschitz, singla2021fantastic} tend to slightly degrade accuracy in comparison to baseline, contrary to our bound where a small accuracy gain can be observed: our Lipschitz regularization is not a trade-off on accuracy under this training configuration.
Furthermore, the low standard deviation of our accuracies suggests that our regularization stabilizes training contrary to all other approaches. The computational cost of our method is the lightest of all Lipschitz regularization approaches.

\begin{table}[h]
    \caption{This table shows test accuracies and training times for ResNet18 on CIFAR-10, repeated 10 times. Our method has a much narrower standard deviation, slightly better accuracy, and lower training time in comparison to other bounds for Lipschitz regularization. Here WD denotes weight decay.
    }
    \centering
    \footnotesize
    \begin{tabular}{lllllll}\toprule
                                           & \textbf{Test accuracy} (\%)  & \textbf{Time per epoch (s)} \\\cmidrule{1-3}
        {Baseline}                         & 93.32 $\pm$ 0.12             & 11.4                        \\
        {Baseline WD}                      & 93.30 $\pm$ 0.19             & 11.4                        \\
        {\citeauthor{ryu2019plug}}         & {93.27} $\pm$ 0.13           & 95.0                        \\
        {\citeauthor{sedghi2019singular}}  & {\ \ \ \ \ \ \ \ \ \ \xmark} & 6750                        \\
        {\citeauthor{araujo2021lipschitz}} & 90.66 $\pm$ 0.26             & 37.4                        \\
        {\citeauthor{singla2021fantastic}} & 93.29 $\pm$ 0.15             & 38.4
        \\\midrule
        Ours                               & 93.48 $\pm$ 0.08             & 31.9                        \\\midrule
    \end{tabular}
    \label{tab:test_accuracies_method_reg_cifar10}
\end{table}

To demonstrate the scalability of our method, a ResNet18 is trained on the ImageNet-1k dataset, processing $256 \times 256$ images.
ResNet18 baseline (with WD) is compared with two regularized ResNet18: one trained from scratch on 88 epochs, and one initialized on a pre-trained baseline and fined-tuned on 10 epochs. Trainings are repeated four times on 4 GPU V100.
The results of experiments are reported in
Table~\ref{tab:test_accuracies_method_reg_imagenet}.
This experiment shows that our bound remains efficient when dealing with large images. Moreover, it is not required to train a new ResNet from scratch: any trained ResNet can be fine-tuned on a few epochs to be regularized.

\begin{table}[h]
    \caption{This table shows test accuracies and training times for ResNet18 on ImageNet-1k, repeated 4 times.}
    \centering
    \footnotesize
    \begin{tabular}{lllllll}\toprule
                          & \textbf{Test accuracy} (\%) & \textbf{Time per epoch (s)} \\\cmidrule{1-3}
        {Baseline WD}     & 69.76                       & 746                         \\\midrule
        Ours              & 70.77 $\pm$ 0.12            & 782                         \\
        Ours (fine-tuned) & 70.50 $\pm$ 0.05            & 782                         \\\midrule
    \end{tabular}
    \label{tab:test_accuracies_method_reg_imagenet}
\end{table}
This experiment demonstrates that our method is efficient for large images and can be applied to pre-trained models, allowing for quick fine-tuning with regularization.

One limitation is that our method does not encompass the normalization layers, such as batch normalization~\citep{ioffe2015batch} or layer normalization~\citep{ba2016layer}, which are often used in deep networks. Penalizing the spectral norm of these layers is not straightforward, as they do not have a well-defined weight matrix, but are non linear transformations of the input.

In the following section, we explore a different approach to regularization, focusing on minimizing sharpness in the loss landscape to improve generalization. This approach is complementary to Lipschitz regularization and can be applied alongside it.

\section{Flatness regularization through activation decay}
\label{section:theoretical_results}

The nature of the minima in the loss landscape has often been cited as a key factor influencing generalization.
Models that converge to sharp minima tend to be highly sensitive to small perturbations, resulting in poor performance on unseen data. In contrast, models that converge to flatter minima are often associated with improved robustness and generalization~\citep{hochreiter1997flat}. Flatness intuitively corresponds to solutions that are less sensitive to small changes in inputs or parameters, making the model more robust to variations. While sharpness-based measures have been shown to correlate strongly with generalization under certain settings~\citep{jiang2020fantastic}, flatness is not a universal guarantee of better generalization. This is because sharpness measures can be sensitive to parameterization and re-scaling effects in deep, overparameterized networks~\citep{zhang2017rethinking}. Despite these nuances, regularization techniques aimed at minimizing sharpness or curvature remain effective heuristics for guiding models toward solutions that generalize well in practice.

A widely used regularization method for encouraging flatness is weight decay, or \( \ell_2 \) regularization on weights. This technique penalizes the magnitude of the model’s weights by adding a term proportional to the squared norm of the parameters to the loss function. Weight decay serves to constrain large parameter values, effectively smoothing the loss landscape and reducing the complexity of the learned model. By controlling the magnitude of the weights, weight decay helps the model find flatter minima, which is associated with improved generalization~\citep{krogh1992simple}. In addition to being computationally efficient, weight decay aligns with other regularization strategies, such as Lipschitz constraints, as explored in Section~\ref{section:spectral_norm_regularization}.

\subsection{Link between Lipschitz networks and flat minima}
\label{section:link_lipschitz_flat_minima}
We recall the definition of  a feed-forward neural network \( f \) as the composition of its \( \nblayers \) layers:
\(
f = f^{(\nblayers)} \circ f^{(\nblayers-1)} \circ \cdots \circ f^{(1)} \ .
\)
Given an input \( \vx \in \R^d \), we define the activations at each layer as:
\(
\vh^{(l)} = f^{(l)}(\vh^{(l-1)}) \ ,
\)
for \( l = 1, \dots, L \), where \( \vh^{(0)} = \vx \). Each layer applies a linear transformation with weights \( \mathbf{W}^{(l)} \), followed by a nonlinearity \( \nonlin^{(l)} \), typically \(\relu\) or \(\GELU\):
\[
    f^{(l)}(\vh^{(l-1)}) = \nonlin^{(l)}\left(\mathbf{W}^{(l)} \vh^{(l-1)}\right), \quad l = 1, \dots, L-1.
\]
The final layer applies no nonlinearity (\(\nonlin^{(L)}\) is the identity), producing:
\(
f^{(L)}(\vh^{(L-1)}) = \mathbf{W}^{(L)} \vh^{(L-1)}.
\)
For a given label \( \vy \), the loss is denoted by \( \mathcal{L}(\vh^{(L)}, \vy, \tens{\theta}) \), where
\(
\tens{\theta} = \mathrm{vec}\left( \{ \mathbf{W}^{(l)} \}_{l = 1, \dots, L} \right) \ .
\)

Penalizing the parameter norm \(\|\vtheta\|_2\) is a common regularization strategy that constrains the model's Lipschitz constant by reducing the product of the spectral norms of the layers. Specifically, the product upper bound of the network's Lipschitz constant can be expressed as: $$
    \PUB(f) = \prod_{l=1}^\nblayers \left\| \mW^{(l)} \right\|_2 \leq \prod_{l=1}^\nblayers \left\| \mW^{(l)} \right\|_\frob,
$$ where \(\|\mW^{(l)}\|_2\) and \(\|\mW^{(l)}\|_F\) denote the spectral and Frobenius norms of the weights in layer \(l\), respectively.

Both sharpness regularization and Lipschitz networks improve generalization by reducing the model’s sensitivity to perturbations. Sharpness regularization operates on the loss landscape to promote flatter minima, while Lipschitz regularization constrains the function’s overall behavior. Notably, Lipschitz constraints also encourage a regular loss landscape to some extent.

Sharp minima are usually characterised by a large spectral norm of the loss Hessian $\| \nabla^2_{\tens{\theta}} ~
    \mathcal{L}(\tens{\theta})
    \|_2$.
To make curvature control more tractable, we first analyze the structure of the Hessian. The following theorem provides a decomposition of the full Hessian norm $\| \nabla^2_{\tens{\theta}} ~
    \mathcal{L}(\tens{\theta}) \|_2$  into a product involving the sensitivity of the penultimate activations with respect to the parameters and the curvature of the loss with respect to these activations.
\begin{theorem}[Layerwise Hessian Norm Decomposition]
    \label{thm:bound_hessian_layers}
    Consider the empirical loss
    $\mathcal L(\theta)=\tfrac{1}{n}\sum_{i=1}^n \ell(\va^{(L)}_i(\theta),y_i)$,
    with $\ell(\cdot,y)$ twice differentiable in the logits and all activations $1$-Lipschitz.
    Suppose a minimizer $\tens{\theta}^\star$ satisfies \emph{logit stationarity},
    i.e.\ $\nabla_{\va^{(L)}} \ell(\va^{(L)}_i(\tens{\theta}^\star),y_i)=0$ for all $i$.
    Then the spectral norm of the Hessian at $\tens{\theta}^\star$ satisfies
    \begin{equation}
        \resizebox{\columnwidth}{!}{$
            \|\nabla^2_{\tens{\theta}}\mathcal{L}(\tens{\theta}^\star)\|_2
            \;\leq\;
            \underbrace{
                \left(\sum_{j=1}^{L-1}
                \Big\|\tfrac{\partial \vh^{(j)}}{\partial \tens{\theta}}(\tens{\theta}^\star)\Big\|_2
                \prod_{l=j+1}^{L-1} \|\mW^{(l)}\|_2
                \right)^2
            }_{\text{Sensitivity of $\vh^{(L-1)}$ to parameters}} \;
            \textcolor{red}{
            \underbrace{
            \|\nabla^2_{\vh^{(L-1)}} \mathcal{L}(\tens{\theta}^\star)\|_2
            }_{\text{Curvature w.r.t.\ activation } \vh^{(L-1)}}
            }
        $}
    \end{equation}
\end{theorem}

Find proof in Appendix~\ref{proof:thm:bound_hessian_layers}.
%
Given that over-parameterized networks typically reach near-zero training loss~\citep{zhang2017rethinking},
assuming logit stationarity at a minimizer $\tens{\theta}^\star$ is reasonable:
it holds in regression (MSE), in classification with label smoothing,
and asymptotically in cross-entropy with one-hot labels as logits diverge.
This theorem emphasizes the outsized influence of deeper layers’ spectral norms on curvature, highlighting their crucial role in shaping the loss landscape. Regularizing deeper layers is especially effective for controlling sharpness. Weight decay
or spectral norm regularization naturally accomplishes this by bounding the spectral norm through the Frobenius norm constraint.

This inequality highlights that the contributions from deeper layers to the Hessian norm bound are more significant, as the spectral norm \(\|\mW^{(l)}\|_2\) is accumulated
more times
for deeper layers. Consequently, the curvature of the loss landscape is strongly influenced by the Lipschitz constants of the deeper layers.

The term \(\left\| \nabla^2_{\vh^{(\nblayers-1)}} \loss(\vtheta) \right\|_2\) reflects the influence of the Hessian with respect to the penultimate activations. Combined with the accumulated Lipschitz constants, this term governs the overall smoothness of the loss landscape. Managing the spectral norms of intermediate layers \(\|\mW^{(l)}\|_2\) and the penultimate-layer Hessian offers a more tractable approach than directly constraining the Hessian of the loss function, which would require costly second-order computations. Specifically, for a model with $d$ parameters, the Hessian matrix has size $d \times d$,
requiring $O(d^2)$ storage and $O(d^3)$ computation. This makes direct computation infeasible for large networks~\cite{goodfellow2016deep}.

The above theorem demonstrates that Lipschitz networks encourage flatter loss landscapes, which is beneficial for generalization. Weight decay, by bounding the Frobenius norm \(\|\mW^{(l)}\|_F\), indirectly controls the spectral norm \(\|\mW^{(l)}\|_2\), providing a practical and computationally efficient regularization strategy. Section~\ref{section:spectral_norm_regularization} further explores spectral norm regularization as an efficient method for controlling curvature through Lipschitz regularization. However, weight decay does not control the term \(\left\| \nabla^2_{\vh^{(\nblayers-1)}} \loss(\vtheta) \right\|_2\), which motivates the need for additional techniques based on loss smoothing to reduce sharpness in the loss landscape, as discussed in the next section.
%
\subsection{Loss smoothing for flat minima}
\label{section:activation_decay}
\citet{bishop1995training} first established the connection between noise injection on inputs and deterministic regularization on parameters, demonstrating how noise injection can be cast as a form of regularization. In more recent work,~\citet{wei2020implicit,orvieto2022anticorrelated} further explored this connection by analyzing how specific forms of noise injection, such as anticorrelated noise and dropout~\citep{srivastava2014dropout}, can encourage flatter minima and improve generalization.
While noise injection introduces stochasticity during training, making models more robust to perturbations and guiding them toward flatter minima, those methods introduce randomness into the optimization process, which can lead to performance variability across different runs.

Another important method in the field of regularization is Sharpness-Aware Minimization (SAM)~\citep{foret2021sharpnessaware}, see Equation~\ref{eq:sam},
which directly targets sharpness by introducing a min-max optimization framework. SAM aims to minimize the worst-case sharpness in a neighborhood of the model’s parameters, encouraging convergence to flatter minima. While effective, SAM's computational cost and memory overhead are significantly higher than simpler methods like weight decay and noise injection. The perturbed loss evaluation requires multiple forward-backward.
~\citet{zhang2018local} show that batch normalization~\citep{ioffe2015batch} and residual connections~\citep{he2016deep} enhance backpropagation by improving the local Hessian's spectrum leading to better gradient flow.
Beyond simply smoothing the loss landscape, methods like SAM  provide robustness to label noise by guiding models to converge to flatter minima, where the sensitivity to noisy labels is reduced~\citep{baek2024why}.

Inspired by the works of~\citet{bishop1995training} and~\citet{orvieto2022anticorrelated}, we propose a novel deterministic noise-based regularization that operates on activations rather than weights, with the same low computational cost. This approach reduces sharpness in the loss landscape while maintaining computational efficiency.
In the regime near a minimum, where the loss gradient's norm $\| \nabla_{\vtheta} \loss(\tens{\vtheta}) \|_2 $ is small, understanding the impact of smoothing techniques like Gaussian noise on the curvature of the loss is crucial for generalization. The following corollary quantifies the reduction in the spectral norm of the Hessian—an indicator of sharpness—when Gaussian noise is applied to the parameters near a local minimum.
\begin{corollary}[Dimension-free bound on Hessian norm of Gaussian smoothed loss]
    \label{corol:bound_hessian_smoothed}
    Let \( \mathcal{L}: \mathbb{R}^d \to \mathbb{R} \) be  twice differentiable. Suppose that the gradient \( \nabla_{\theta} \mathcal{L} \) is $H$-Lipschitz continuous,
    and that for some $\epsilon > 0$, \( \| \nabla_{\theta} \mathcal{L}(\tens{\theta}) \|_2 \leq \epsilon \).
    For $\mathrm{\Delta} \sim \mathcal{N}(0, \sigma^2 I)$, the spectral norm of the Hessian of the Gaussian smoothed loss is bounded by:
    \begin{equation}
        \| \nabla^2_{\tens{\theta}} ~
        \mathcal{L}^\sigma(\tens{\theta})
        \|_2 \leq
        \textcolor{red}{
            \underbrace{{H}}_{\text{base curvature}}
        }
        ~
        \textcolor{blue}{
            \underbrace{
                \operatorname{erf}\left( \frac{\epsilon}{\sqrt{2} H \sigma} \right)
            }_{\text{smoothing term} \; < \; 1}
        }
        \ .
    \end{equation}
\end{corollary}
The proof (see Appendix~\ref{proof:corol:bound_hessian_smoothed})
of this corollary adapts Theorem~\ref{thm:thm_bound_lip_sigma_element_wise} which quantifies the regularity of Weierstrass transform when underlying transformed function is already regular, previously applied in randomized smoothing with Gaussian noise applied to the inputs to obtain Lipschitz continuity.
Here Gaussian noise is applied to the network's parameters $\vtheta$ and it translates to increased flatness of the loss landscape.
The work of~\citet{nesterov2017random} was the first to derive bounds on the regularity of the Gaussian smoothed loss.  Unlike earlier bounds, which depend on the dimensionality \(d\), this result is dimension-free, making it more suitable for high-dimensional deep networks.
The result provides a quantitative measure of how Gaussian noise injection smooths the loss landscape, reducing sharpness and promoting generalization. Notably, while~\citet{orvieto2022anticorrelated} explored the connection between noise injection and the trace of the Hessian, our bound focuses on the Hessian’s spectral norm, offering a complementary perspective on sharpness reduction.

\begin{figure}[h]
    \centering
    \includegraphics[width=0.5\linewidth]{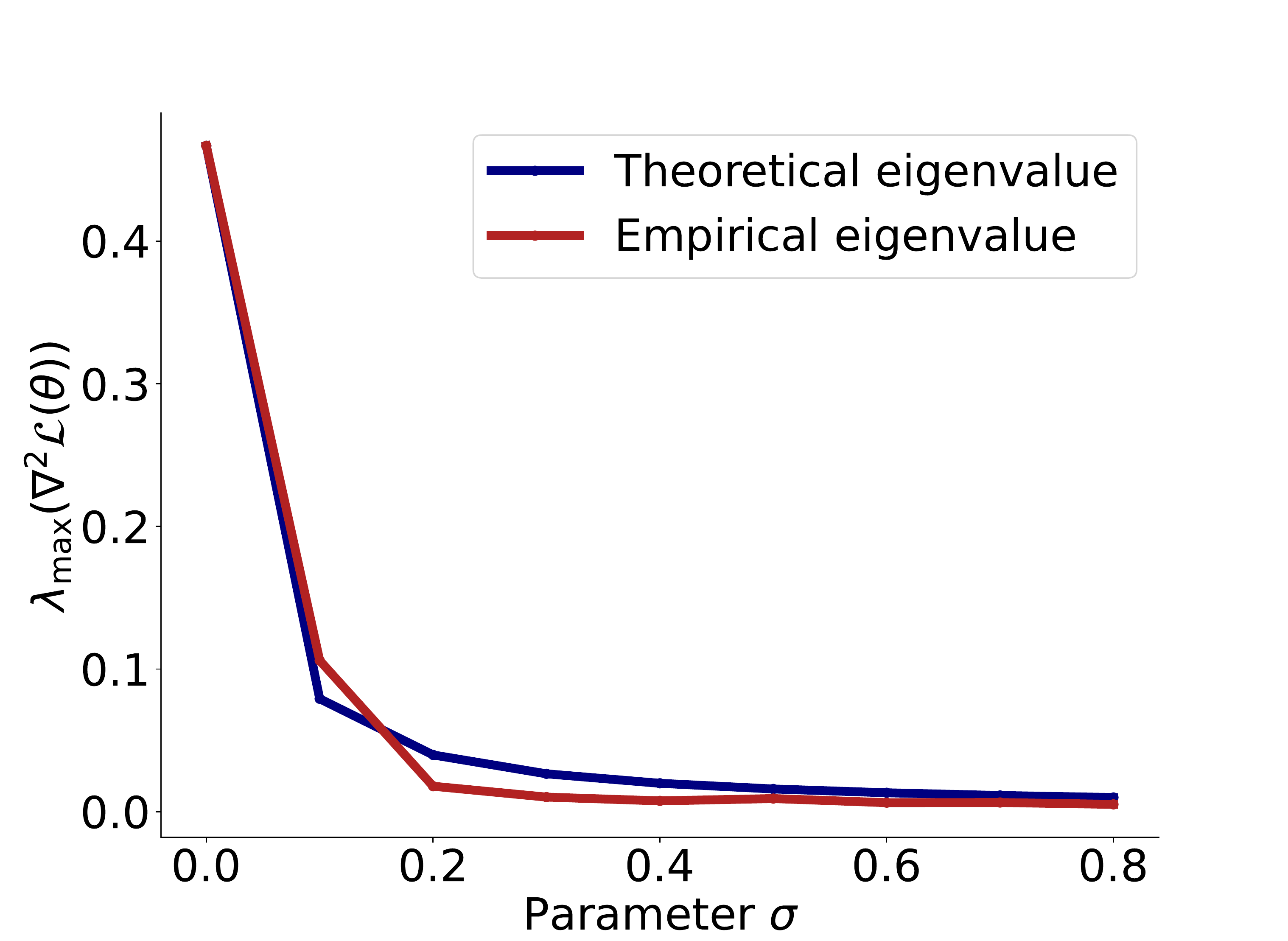}
    \caption{Comparison between the theoretical bound on the largest eigenvalue of the Hessian given by Corollary~\ref{corol:bound_hessian_smoothed} and the empirical value computed with PyHessian with a relative tolerance of $1\mathrm{e-}3$, on a ResNet-56 model trained on CIFAR-10 for 300 epochs.}
    \label{fig:theoritical_bound_vs_empirical}
\end{figure}
\paragraph{Empirical validation of regularization effects}
This experiment empirically evaluates how closely the theoretical bound from Corollary~\ref{corol:bound_hessian_smoothed} aligns with observed Hessian norms.
Several runs of training on CIFAR-10 are performed with the ResNet-56 model and different parameters $\sigma$ for the AD.
We use the PyHessian library to compute the Hessian operator norm~\citep{yao2020pyhessian}.
Then, we compute the largest eigenvalue of the Hessian on the final layers of the network to assess the sharpness reduction predicted by Gaussian smoothing.
Here Hessian is computed only w.r.t to $\mathbf{W}^{(\nblayers)}$ of the final layer. We estimate $\epsilon$ by averaging the gradient's norm near the minimum at the end of the training.
The Hessian eigenvalue is computed on the training set at the end of the training after 300 epochs. \\
We see in Figure~\ref{fig:theoritical_bound_vs_empirical},
that the theoretical bound gives the correct trend of the evolution of the curvature of the Hessian, the remaining mismatch might come from stochasticity in the Hessian eigenvalue computation, the Jensen gap, or another factor.

\subsection{From Noise to Regularization: Activation Decay}

Building on the curvature decomposition in Theorem~\ref{thm:bound_hessian_layers} and the smoothing guarantees from Corollary~\ref{corol:bound_hessian_smoothed}, we now derive a simple and efficient regularization method. The key idea is to contract the activation-level curvature term without injecting noise during training. To this end, we analytically translate the effect of Gaussian smoothing into a deterministic penalty, resulting in a method we call \emph{activation decay}.

Rather than adding noise as in dropout, we consider a Gaussian perturbation \( {\Delta} \sim \mathcal{N}(0, \sigma^2 \mI) \) applied solely to the final-layer weights \( \mW^{(L)} \). For fixed penultimate activations \( \vh^{(L-1)} \), this yields a smoothed loss:
\[
    \mathcal{L}^\sigma (\mW^{(L)} \vh^{(L-1)}, \vy) =
    \mathbb{E}_{{\Delta}} \left[
    \mathcal{L} \big((\mW^{(L)} + {\Delta}) \vh^{(L-1)}, \vy \big)
    \right].
\]
This smoothing strategy, closely related to softmax over Gaussian distributions~\citep{lu2021meanfield} and final-layer isolation techniques~\citep{newman2021train}, remains fully differentiable and admits a closed-form upper bound for standard losses such as cross-entropy.

\begin{theorem}[Smoothed Cross-Entropy Loss]
    \label{thm:smoothed_loss}
    Let $\mathcal{L}_{\mathrm{CE}}$  be the cross-entropy loss, \(\vh^{(L-1)} \in \mathbb{R}^d\) be an input from the penultimate layer, \(\vy \in \mathbb{R}^c\) a one-hot label vector, and \(\mW^{(L)} \in \mathbb{R}^{c \times d}\) a weight matrix. Then the following bound holds:
    \begin{align}
        \mathcal{L}_{\mathrm{CE}}^\sigma(\mW^{(L)} \vh^{(L-1)}, \vy) \leq \textcolor{red}{
        \underbrace{
        \mathcal{L}_{\mathrm{CE}}(\mW^{(L)} \vh^{(L-1)}, \vy)
        }_{\text{base loss}}
        }
        +
        \textcolor{blue}{
            \underbrace
            {\frac{\sigma^2}{2} \|\vh^{(L-1)}\|_2^2}_{\text{smoothing term}}
        }
        \ .
    \end{align}
\end{theorem}

Minimizing this upper bound\footnote{See Appendix~\ref{sec:mc_vs_ad} for empirical validation of the tightness of this bound.}
is equivalent to augmenting the original loss with an \( \ell_2 \) penalty on \( \vh^{(L-1)} \). We refer to this regularizer as \emph{activation decay} (AD), by analogy with weight decay. Unlike dropout or other stochastic methods, AD is deterministic, low-variance, and introduces no computational overhead during training.

The resulting penalty \( \|\vh^{(L-1)}\|_2^2 \) directly targets the curvature term \( \nabla^2_{\vh^{(L-1)}} \mathcal{L} \) identified in Theorem~\ref{thm:bound_hessian_layers}, acting as a surrogate for Gaussian smoothing at the classifier level. This construction is rigorously justified by Corollary~\ref{thm:bound_hessian_smoothed}, which proves that smoothing contracts the Hessian’s spectral norm in a dimension-independent way. Meanwhile, the parameter-sensitivity component of curvature can be independently controlled via standard techniques such as weight decay or spectral norm regularization. In this view, activation decay and weight regularization act complementarily, each addressing a distinct factor in the Hessian decomposition.

\subsection{Experiments on vision tasks}
\label{section:experiments_activation_decay}
We propose to evaluate our method Activation Decay (AD) on several vision tasks.

\paragraph{Classification with MLP on CIFAR-10}
\label{exp:mlp_classif_cifar10}
This experiment compares different regularization techniques on a 4-layer Multi-Layer Perceptron (MLP) network with GELU activation. The MLP has 3072 input features, and the training was conducted on the CIFAR-10 dataset. The model was trained using Stochastic Gradient Descent (SGD) without momentum, and no weight decay was applied. A learning rate scheduler with annealing was used to adjust the learning rate, which was set to $1\mathrm{e-}1$. We use a batch size of $128$ and standard data augmentation techniques, including random horizontal flips and random crops with padding of 4 pixels. These augmentations are applied to the training data to improve generalization.
Each experiment was repeated 10 times to ensure statistical significance.

We explored various regularization methods:
We apply dropout (DO), parameterized by probability \( p \), on intermediate layers \( \vh^{(l)} \); weight decay (WD) on all layers, parameterized by \( \sigma \); activation decay (AD), parameterized by \( \sigma \), on the last layer \( \vh^{(L)} \); a combination of AD on the last layer \( \vh^{(L)} \), parameterized by \( \sigma \), with weight decay on intermediate layers \( \vh^{(l)} \), where the best parameter is obtained from the previous weight decay experiment, we perform the same process replacing weight decay with spectral norm regularization to showcase its combination with AD;
and SAM parametrized by $\rho$.
\begin{figure}[t]
    \centering
    \includegraphics[width=1\linewidth]{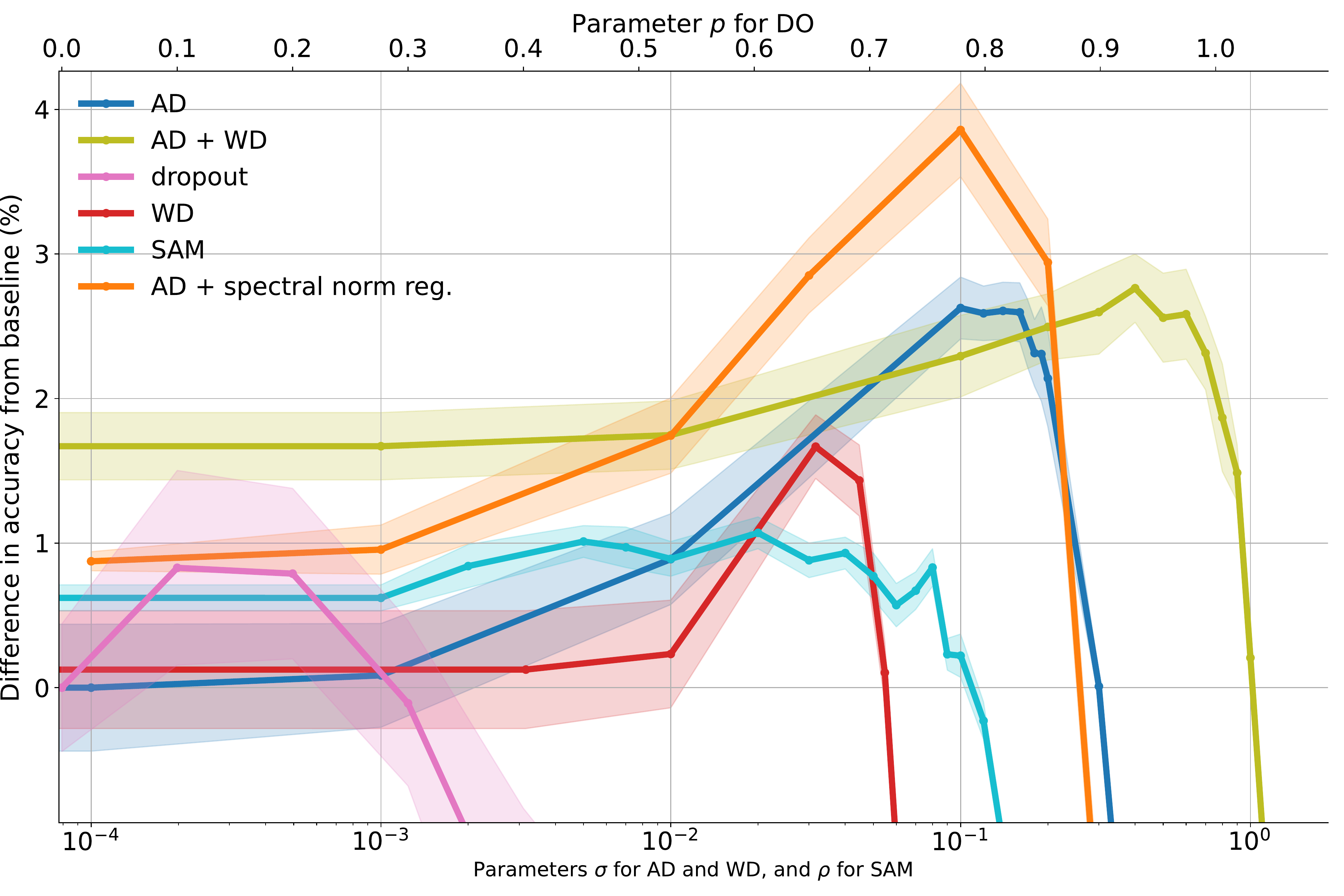}
    \caption{Comparison of accuracies of different regularizations, \textbf{the higher the better}, applied to a 4-layer MLP network trained on the CIFAR-10 dataset. The plots show the evolution of accuracy with varying values of parameter $\sigma$ for activation decay (AD) and weight decay (WD) and drop rate $p$ for dropout (DO).
        The first curve (AD) depicts the effect of $\sigma$ while keeping $p$ at $0.0$.
        The second curve (AD + WD)  combines WD with the best parameter $1\mathrm{e-}3$ and varying $\sigma$ for AD.
        The third curve (DO) illustrates the impact of dropout when $\sigma$ is $0.0$.
        The fourth curve (WD) depicts the effect of $\sigma$ when it parameterizes weight decay.
        The fifth curve (SAM) illustrates the impact of the $\rho$ parameter.
        The sixth curve (AD + spectral norm reg.) illustrates the combined spectral norm regularization with the parameter value $0.1$ for (AD).
        Shell indicates the standard deviation over $10$ runs.
    }
    \label{fig:compare_reg_cifar10_mlp}
\end{figure}
Results are presented in Figure~\ref{fig:compare_reg_cifar10_mlp}. The baseline is at $62.17$\%.

Our results demonstrate that AD increases generalization when \(\sigma = 0.1\), improving accuracy by $2.63$ \%.
Applying dropout alone also slightly enhances generalization, improving by $1.73$\% for best parameter $p=0.1$.
However, combining dropout with AD does not yield additional benefits and performs worse than using AD alone.
The method that combines AD and weight decay performs better than only AD as highlighted by Theorem~\ref{thm:bound_hessian_layers}.
, following this result, we try to combine AD and spectral norm regularization of layers as it is a closer upper bound to the formula from~Theorem~\ref{thm:bound_hessian_layers} and it provides the best-obtained result.
This result confirms the impact of the Lipschitz-constrained network (through spectral norm regularization of intermediate layers) and its link to overall flatness and impact on generalization.
SAM does not provide as good results for the MLP architecture as for CNNs architecture as reported in SAM paper \citep{foret2021sharpnessaware}.
We also provide a comparison with a similar approach proposed by \cite{baek2024why} in Appendix~\ref{appendix:exp:baek_comparison}. Their method applies layer-wise activation regularization across all layers except the last, where weight decay is used. In contrast, our approach requires fewer hyperparameters to tune while achieving comparable results.

We also provide results on MLP-Mixer architecture ~\citep{tolstikhin2021mlpmixer} on ImageNet in the Appendix, showing that our method extends to a bigger dataset and architecture.

\begin{figure}[h]
    \centering
    \includegraphics[width=0.8\linewidth]{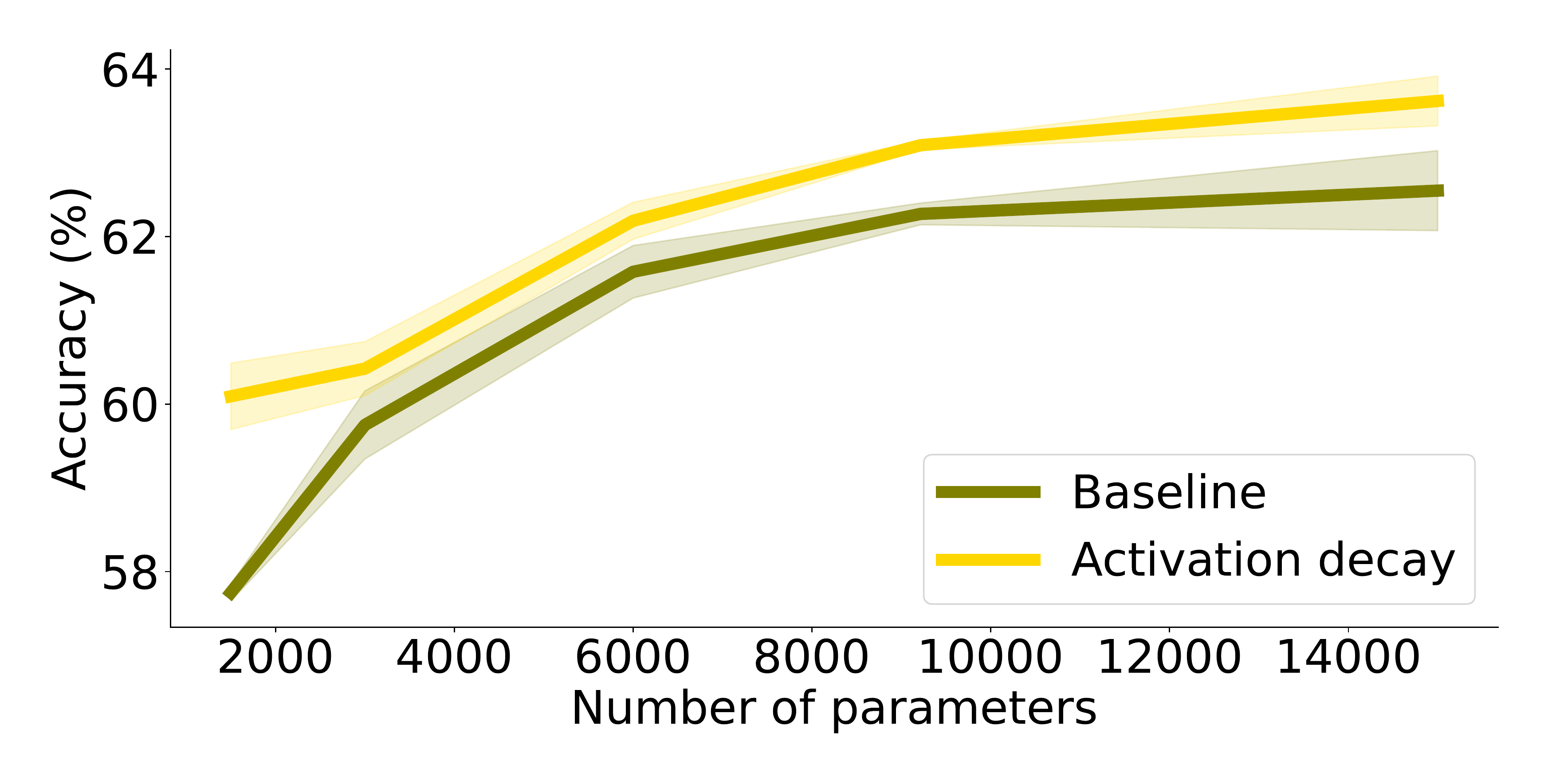}
    \caption{Accuracy on CIFAR-10 for an MLP with depth 3 and varying numbers of hidden features per layer. Our method with regularization is compared to the baseline with no regularization.}
    \label{fig:acc_vs_features}
\end{figure}
\textbf{Effect of overparameterization on regularization performance}
This experiment evaluates how AD regularization behaves when varying the number of parameters in the model. We use a 3-layer MLP on CIFAR-10 and adjust the number of hidden features to transition between underparameterized and overparameterized regimes.
As shown in Figure~\ref{fig:acc_vs_features}, we observe that the gains from our method are consistent across both regimes. Smoothing the loss leads to noticeable improvements in accuracy, demonstrating the effectiveness of our regularization technique regardless of the model's parameter count.
\begin{table}[h!]
    \caption{Accuracies on validation set for baseline, AD ($\sigma = 0.2$), ASAM  ($\rho_{\mathrm{ASAM}} = 2.0$) and AD+SAM, for WideResNet on CIFAR-10. Results were reported after averaging over 3 runs and the standard deviation is $0.03$ for all runs.}
    \centering
    \begin{tabular}{lcc}
        \toprule
        \textbf{Configuration} & \textbf{Validation accuracy (\%)} \\
        \midrule
        Baseline               & 97.09                             \\
        ASAM                   & 97.48                             \\
        AD                     & 97.27                             \\
        AD + ASAM              & \textbf{97.54}                    \\
        \bottomrule
    \end{tabular}
    \label{table:sam_vs_ours_vs_baseline_cifar10}
\end{table}

\textbf{Comparison to ASAM with Wide ResNet on CIFAR-10}
Table~\ref{table:sam_vs_ours_vs_baseline_cifar10} presents a comparative analysis of the average accuracies on the validation set achieved by different training configurations using the WideResNet architecture trained on 300 epochs on the CIFAR-10 dataset. The configurations evaluated are: Baseline, the standard training setup without additional optimization techniques, achieved a validation accuracy of 97.09\%. Ours, a proposed method utilizing AD with $\sigma = 0.2$, improved the validation accuracy to 97.27\%.
For experiments involving SAM we use the upgrade Adaptive SAM (ASAM) \citep{kwon2021asam}, we adopted the official implementation provided by the authors. To ensure a fair evaluation, we used the best parameter $\rho_{\mathrm{ASAM}} = 2.0$ as specified in their code repository for this particular task which achieved a validation accuracy of 97.48\%.
AD + ASAM, a combination of the proposed method and SAM yielded the highest validation accuracy of 97.54\%.
The best values of ASAM and AD parameters were picked by hyperparameter search.
AD and ASAM outperform the baseline individually, while the combination yields the best performance.
This can be explained because ASAM is designed to minimize the worst-case ascent sharpness, specifically targeting regions of the loss landscape where the maximum sharpness is reduced \citep{wen2022how}. In contrast, AD focuses on reducing the average sharpness, promoting a smoother and more stable optimization trajectory. This complementary behavior could explain why the two methods combine effectively, leading to enhanced generalization performance.

\begin{figure}[h]
    \centering
    \includegraphics[width=0.7\linewidth]{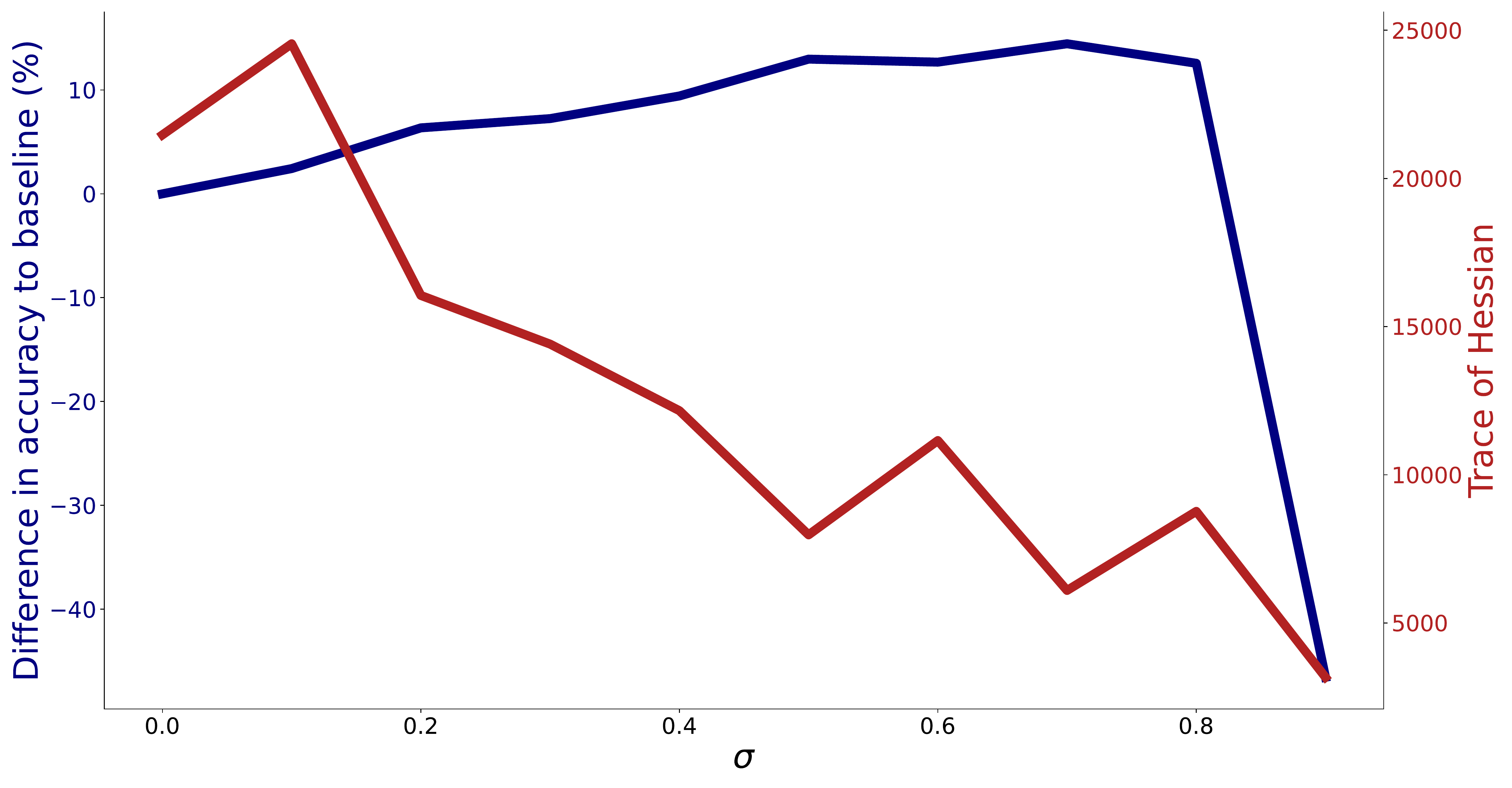}
    \caption{Accuracy difference and Hessian trace of the loss for varying \(\sigma\), with 30\% label noise on CIFAR-10 using ResNet-56.}
    \label{fig:label_noise_acc_and_hessian_trace}
\end{figure}
\textbf{Classification with label noise with ResNet-56 on CIFAR-10}
We evaluate the effect of 30\% label noise effect on CIFAR-10 classification using ResNet-56, varying the parameter \(\sigma\) in AD. Figure~\ref{fig:label_noise_acc_and_hessian_trace} shows the accuracy difference from a baseline model and the trace of the Hessian of the loss w.r.t all parameters, plotted against \(\sigma\). The Hessian trace, a proxy for sharpness, indicates sharper minima and poorer generalization under label noise.
As \(\sigma\) increases, accuracy initially improves but declines as the Hessian trace rises, aligning with findings from~\citet{foret2021sharpnessaware} and~\citet{baek2024why}, where SAM (Sharpness-Aware Minimization) improves regularization under label noise.
Our AD method enhances noise robustness by controlling sharpness as \(\sigma\) increases, improving accuracy without the computational cost of SAM. This makes AD an efficient alternative for handling noisy labels for free. This experiment further reinforced the link between flat minima and label noise robustness.
As Lipschitz continuity is tied to generalization properties~\citep{bartlett2017spectrally} and robustness~{tsuzuku2018lipschitz}, flat minima also provides robustness.

\subsection{Experiment with large language models}
We propose to tackle the problem of multi-task learning with large language models (LLMs) using our Activation Decay (AD) method.

Multi-task learning (MTL)~\cite{zhang2021surveyMTL} is a paradigm in machine learning where a model is trained to
perform multiple tasks simultaneously, leveraging shared representations across tasks.
MTL also offers the benefit of reducing computational
costs and latency, enabling the model to handle multiple tasks in a single inference step, rather than performing separate inferences for
each task.

With the advent of large-scale pretrained models, multi-task learning has become increasingly popular in NLP.
Pretrained on large corpora, these models can capture a wide range of patterns and dependencies in text data. Leveraging this general
knowledge during fine-tuning while specializing in specific tasks leads to new state-of-the-art performances on a variety of downstream tasks.
This task-specific adaptation during fine-tuning, is needed for earlier Large Language Models (LLMs) as BERT~\cite{devlin2019bert} and
RoBERTa~\cite{liu2019roberta} and has recently been reduced to few-shot learning with models like GPT-3~\cite{brown2020language} and
T5~\cite{raffel2020exploring} and even extended to zero-shot settings by~\cite{wei2022finetuned, sanh2022multitask} in the context of very large models.

However, fine-tuning multiple tasks can lead to overfitting on individual tasks, which can degrade the model's performance on other tasks, calling
for specific design choices, adding extra task-specific parameters (adapters), or using specific prompts (see~\cite{stickland2019bert, wang2023multitask}).
Our AD method promotes flat minima to improve generalization across tasks.
The goal is to prevent overfitting to any specific task while retaining the benefits of pretraining on various tasks.
Our flat minima regularization helps preserve the generalization
capabilities of the pretrained model, ensuring robust performance across all tasks during fine-tuning.
As detailed in Section~\ref{section:background_and_related_workt:subsection:flat_minima} flat minima and generalization are often tied together and the experimental results
presented in Tables~\ref{tab:results_bert_vs_sam_vs_do} and ~\ref{tab:results_t5_mmlu} show our regularization helps conserve performance
across diverse tasks by promoting smooth optimization landscapes during fine-tuning.

\textbf{Fine tuning on distinct tasks}
In this experiment, we evaluate the performance of a multi-task NLP model using the RoBERTa \citep{liu2019roberta}, BERT \citep{devlin2019bert},
and T5 \citep{raffel2020exploring} architectures to handle distinct tasks.
The multi-task setup allows the model to process these tasks simultaneously, optimizing latency and resource consumption.
We experimentally show that in the context of fine-tuning and few-shot learning, our Activation Decay (AD) method helps LLMs
to generalize better across tasks, leading to improved performance compared to the baseline models.

Each task is described as follows:
(i) Sentiment Analysis: a binary classification task (positive/negative) on the IMDb dataset, using classification accuracy as the metric;
(ii) NER: named entity recognition on the Snips and CoNLL datasets, evaluated using F1-score, precision, and recall metrics;
(iii) Intent Classification: intent detection on the Snips dataset, evaluated with classification accuracy;
(iv) Entailment Classification (SNLI): a binary classification task predicting whether a sentence entails another, based on the SNLI dataset, with classification accuracy as the metric;
(v) POS Tagging: part-of-speech tagging on the CoNLL dataset, evaluated using F1-score, precision, and recall metrics;
(vi) Query Correctness: a binary classification task assessing the correctness of queries, evaluated with classification accuracy.

The backbone models used for all tasks are BERT (bert-base) and RoBERTa (roberta-base), with dropout probability set to 0. The model's performance is evaluated using a custom smoothed loss function with $\sigma = 0.05$, and the results are compared with SAM regularization at different $\rho$ values.
Weight decay is present by default in the training configuration of the backbone. We use standard training configuration from HuggingFace corresponding models and trainers.

\begin{table}[h!]
    \caption{Evaluation results for BERT baseline with DO ($p=0.1$), SAM ($\rho=0.01$), and AD ($\sigma=0.05$) on 7 tasks.}
    \label{tab:results_bert_vs_sam_vs_do_lean}
    \centering
    \begin{tabular}{lcccc}
        \toprule
        \textbf{Metric}                       & \textbf{DO}    & \textbf{SAM}   & \textbf{AD}    \\
        \midrule
        \textbf{Sentiment Evaluation}         &                &                &                \\
        Classification Accuracy (\%)          & 76.72          & 76.54          & \textbf{77.08} \\
        \midrule
        \textbf{NER Evaluation}               &                &                &                \\
        Snips F1 Score (\%)                   & 78.33          & 69.67          & \textbf{80.90} \\
        Snips Precision (\%)                  & 73.69          & 64.11          & \textbf{76.20} \\
        Snips Recall (\%)                     & 83.59          & 76.28          & \textbf{86.21} \\
        \midrule
        \textbf{Intent Evaluation}            &                &                &                \\
        Classification Accuracy (\%)          & 98.04          & 98.19          & \textbf{98.49} \\
        \midrule
        \textbf{Entailment SNLI Evaluation}   &                &                &                \\
        Classification Accuracy (\%)          & 87.96          & \textbf{89.39} & 88.88          \\
        \midrule
        \textbf{CoNLL NER Evaluation}         &                &                &                \\
        Seqeval F1 Score (\%)                 & 64.43          & 61.01          & \textbf{65.94} \\
        Seqeval Precision (\%)                & 61.87          & 61.48          & \textbf{64.11} \\
        Seqeval Recall (\%)                   & 67.20          & 60.55          & \textbf{67.87} \\
        \midrule
        \textbf{CoNLL POS Evaluation}         &                &                &                \\
        Seqeval F1 Score (\%)                 & 75.95          & 72.48          & \textbf{77.89} \\
        Seqeval Precision (\%)                & 74.89          & 71.59          & \textbf{76.98} \\
        Seqeval Recall (\%)                   & 77.04          & 73.39          & \textbf{78.82} \\
        \midrule
        \textbf{Query Correctness Evaluation} &                &                &                \\
        Classification Accuracy (\%)          & \textbf{69.95} & 69.47          & 69.31          \\
        \bottomrule
    \end{tabular}
\end{table}

Table~\ref{tab:results_bert_vs_sam_vs_do_lean} presents evaluation results for fine-tuning BERT on seven NLP tasks, comparing the baseline model with standard dropout (\(p=0.1\)), SAM regularization with standard values  (\(\rho=0.01\)), and Activation Decay (\(\sigma=0.05\)). The best values for SAM were selected from \(\rho=0.01, 0.05, 0.1\), and Activation Decay from \(\sigma=0.01, 0.05, 0.1\). Metrics include classification accuracy, F1-score, precision, and recall. Activation Decay consistently outperforms both the baseline and SAM across most tasks. The same results for RoBERTa are reported in Table~\ref{tab:res_roberta} in the Appendix, showing the efficiency of AD.
\begin{table}[t]
    \caption{Test accuracy results for T5 configurations on the  MMMLU dataset, for baseline used with DO ($p = 0.1$), and AD  ($\sigma = 0.01$).}
    \centering
    \begin{tabular}{lcc}
        \toprule
        \textbf{Model} & \textbf{DO} & \textbf{AD}    \\
        \midrule
        T5-large       & 52.07       & \textbf{52.95} \\
        T5-base        & 49.89       & \textbf{50.25} \\
        T5-small       & 32.21       & \textbf{33.49} \\
        \bottomrule
    \end{tabular}
    \label{tab:results_t5_mmlu}
\end{table}

\textbf{Few shots learning on MMMLU dataset}
We evaluate our models on the Multilingual Massive Multitask Language Understanding (MMMLU) dataset~\citep{hendrycks2020measuring}, which spans 57 diverse topics ranging from elementary to advanced professional subjects.
We use the newly updated version published by~\citep{openai2023simpleevals}, which expanded the dataset to include 14 languages using professional human translators.
We use the T5 architecture~\citep{raffel2020exploring} for our experiments, fine-tuning the small (60 M), base (220 M), and large (770 M) variants of the model over 3 epochs.
%
%
The results, summarized in Table~\ref{tab:results_t5_mmlu}, show that our AD with $\sigma=0.01$  in a multitask setting, consistently outperforms the standard dropout $p=0.1$ a common baseline for fine-tuning baseline across all model sizes, demonstrating its effectiveness for large models.
This result highlights the importance of our approach in improving accuracy, particularly in large-scale multitasking environments.
All code and implementation details will be made available upon acceptance of the paper to ensure reproducibility.

\section{Conclusion}

We introduced a spectral norm regularization technique that efficiently constrains the Lipschitz constant of individual layers, including convolutions, using the Gram iteration method. This approach stabilizes training, enhances generalization, and offers computational advantages over existing methods, as demonstrated in our CIFAR-10 and ImageNet-1k experiments.

Furthermore, we showed that Lipschitz regularization, which limits network sensitivity to input perturbations, not only improves robustness but also promotes flatter minima (Theorem~\ref{thm:bound_hessian_layers}). To further reduce Hessian curvature, we introduced activation decay as a deterministic alternative to dropout, which relies on noise injection. Activation decay regularizes critical activations via Gaussian smoothing, effectively flattening minima and enhancing robustness without additional computational cost. Unlike Sharpness-Aware Minimization (SAM), which employs a min-max optimization framework with higher overhead, activation decay provides a computationally efficient alternative that integrates seamlessly with weight decay and Lipschitz regularization. Empirical results validate its effectiveness in replacing dropout and confirm the proposed formulation.

A known limitation of norm-based sharpness metrics is their sensitivity to simple reparameterizations such as neuron-wise rescaling in ReLU networks, which preserve the function but alter spectral or Frobenius norms. This lack of invariance may obscure the relationship between sharpness and generalization, motivating the exploration of scale-invariant alternatives~\cite{gonon2025rescaling}.

%% file: content/chapter-conclusion.tex
%
\chapter{Conclusion}\label{sec:conclusion}
\minitoc%

\section{Thesis summary}
In this thesis, we address the problem of stability in deep learning through the lens of sensitivity analysis from various perspectives: sensitivity to inputs with Lipschitz networks and to parameters with loss curvature.
Specifically, we advocate for the use of Lipschitz networks as a means to enhance the stability and robustness of deep learning models. By combining Lipschitz networks with Hessian-bounded loss regularization, this thesis introduces a unified framework for tackling long-standing challenges in stability and robustness.
Through spectral norm computation, randomized smoothing, and curvature regularization, we provide both theoretical insights and practical methodologies.
Our key contributions can be summarized as follows:
\begin{contribframe*}[Summary of the contributions]
    {\quad}
    \begin{my_list_num}
        \item We developed Gram iteration and its derived algorithms for computing the spectral norm of both convolutional and dense layers in deep learning networks.
        \item We established a connection between randomized smoothing—a probabilistic method for certifying robustness—and the Lipschitz constant of neural networks. This interaction improves stability, leading to better certificates and reduced variance.
        \item We proposed a novel method to generate tighter confidence intervals in the context of randomized smoothing, further improving the reliability of robustness estimates.
        \item We introduced a new regularization term based on the Hessian of the loss function to enhance the stability of neural networks.
    \end{my_list_num}
\end{contribframe*}

These contributions provide both theoretical and practical advancements in the study of neural network stability and robustness. By improving spectral norm estimation, establishing connections between Lipschitz continuity and randomized smoothing, refining confidence intervals, and introducing Hessian-based regularization, our work enhances the reliability of robustness guarantees while maintaining computational efficiency. These methods contribute to a better understanding of generalization, adversarial robustness, and certified defenses in deep learning.
However, several challenges remain to be addressed in the field of Lipschitz networks, including scaling, attention mechanisms, and Lipschitz-certified robustness for large language models.

\section{Open problems}
\paragraph{Scaling Lipschitz networks}
Scaling generally improves performance, but Lipschitz networks have yet to match the accuracy and efficiency of standard architectures. Despite this, they offer significant advantages in robustness, stability, and generalization. Unlike traditional networks, Lipschitz models train reliably without requiring tricks such as batch normalization or gradient clipping, making them inherently stable and easier to optimize~\cite{meunier2022dynamical}.

Lipschitz networks require more extensive parameter tuning than standard networks~\citep{bubeck2021law,bubeck2021universal}. Developing foundational models with pre-tuned parameters would simplify their deployment. Indeed, the works of \citet{araujo2023a, hu2023scaling} have shown that adding large feedforward layers after convolutional layers significantly improves performance, as this overparameterizes the network and enhances its capacity. Investigating the scaling laws of robustness in relation to the Lipschitz constant could offer valuable insights for designing scalable Lipschitz networks.
Training large Lipschitz networks on extensive datasets in a highly overparameterized regime is a promising direction. Moreover, similar scaling laws apply to data, where the use of synthetic data has proven highly effective in increasing performance~\citep{hu2023scaling, hu2024recipe}. However, this implies a significant computational cost and energy consumption, which is not always feasible in practice and goes against the goal of frugality in model use and design.

While scaling could be considered the primary explanation for this gap, the inherent limitations of Lipschitz networks suggest that scaling alone may not be sufficient.
Randomized smoothing provides a theoretical framework for defining efficient Lipschitz networks, even if they can only be approximated in practice using Monte Carlo methods.
Recently models combining diffusion-based denoisers with classifiers have demonstrated promising results in terms of robustness and accuracy on large datasets such as ImageNet~\citep{carlini2023certified}, with performance close to standard networks.
This highlights a crucial perspective: the existence of such architectures suggests that the gap can still be bridged by developing innovative designs that balance training stability, performance, and computational efficiency.
Further research is needed to understand the trade-offs in these models and to propose new architectures that push the boundaries of Lipschitz network performance.
Recent works on by-design Lipschitz networks, such as~\citet{hu2023scaling, hu2024recipe} pave the way and encourage the community to explore this direction, demonstrating significant improvements in scalability and robustness.
Another perspective to narrow this gap could be to explore the development of attention mechanisms for Lipschitz networks, as they are a key component in many recent state-of-the-art models.

\paragraph{Lipschitz constraints for attention mechanisms}
Self-attention mechanisms have become a fundamental component of modern deep learning architectures, particularly in natural language processing (NLP)~\citep{vaswani2017attention}.
However, self-attention relies on interactions between keys and queries, leading to a quadratic transformation of the inputs, which inherently makes it non-Lipschitz. This poses a challenge when designing Lipschitz-constrained attention layers.
Modern architectures consistently incorporate residual connections, which suggests a possible workaround: instead of enforcing Lipschitz constraints on self-attention alone, one could consider the Lipschitz behavior of the entire residual transformation, including both the self-attention mechanism and the residual path.

A promising approach in this direction is the framework of convex potential flows introduced by~\citet{meunier2022dynamical}, which enables Lipschitz residual transformations through carefully chosen discretizations of continuous flows. Extending this idea to self-attention would require designing a convex potential that naturally induces an attention-like transformation while preserving Lipschitz's continuity.

\paragraph{Enhancing randomized smoothing certification}
In randomized smoothing, the key quantity of interest is the certified radius that guarantees robustness. This radius is determined by two main factors: the Lipschitz constant of the margin function and the margin of the smoothed classifier at a given point.
Since the true margin of the smoothed classifier is not directly accessible, it is estimated using confidence intervals on each logit coordinate, from which the margin is then computed. Similarly, the Lipschitz constant of the margin function is typically bounded using the Lipschitz constants of the individual logit coordinates, leading to the expression of the radius \(\Rcoord\) in~\eqref{eq:rcoord_bound}.

A potential improvement to this procedure would be to estimate the margin of the smoothed classifier directly, rather than first estimating the margin of the logits and then deriving the margin of the smoothed classifier. Developing a method for direct estimation could reduce variance, leading to tighter confidence intervals and ultimately improving the certified radius. The same principle applies to the Lipschitz constant of the margin function: instead of relying on bounds derived from individual logit coordinates, estimating it directly from the full margin function could provide a more accurate and potentially tighter bound. This would involve leveraging the first expression of the radius in~\eqref{eq:first_radius} and incorporating smooth margins.\\

The main challenge in implementing this approach lies in adapting it to the radius defined by the composition of the Gaussian quantile function with the smoothed classifier. Indeed, the composed function is not bounded, which necessitates a careful analysis to construct a reliable confidence interval.

\paragraph{Certified robustness for large language models}
Certified robustness remains an open challenge for large language models (LLMs), as adversarial vulnerabilities have been observed in these architectures~\citep{jia2019certified, huang2019achieving}.
While certified defenses have been well studied in computer vision, their extension to NLP is less straightforward due to the discrete nature of text inputs and the difficulty of defining meaningful perturbations. Developing robust certification methods for LLMs that account for adversarial attacks while maintaining linguistic coherence remains an important research direction.





%% file: content/appendix/appendix-spectral_norm.tex
\chapter{Spectral norm computation}
\label{app:sec:spectral_norm_computation}

\subsection{Toeplitz matrix}
\label{section:appendix_toeplitz_matrix}

Let $\mT \in \R^{\cout n^2 \times \cin n^2}$ be a $\cout \times  \cin $ block matrix:
\begin{align*}
  \mT = \begin{psmallmatrix}
          \mT_{1,1, ::} & \mT_{1,2, ::} & \cdots & \mT_{1,\cin, ::} \\
          \mT_{2,1, ::} & \mT_{2,2, ::} & \cdots & \mT_{2,\cin, ::} \\
          \vdots & \vdots & \ddots & \vdots \\
          \mT_{\cout, 1, ::} & \mT_{\cout, 2, ::} & \cdots & \mT_{\cout,\cin, ::}
        \end{psmallmatrix} \ ,
\end{align*}

where each $\mT_{j,i, ::}$ is a $n^2 \times n^2$ banded doubly Toeplitz matrix formed with kernel $\tK_{j,i}$. The matrix $\mT_{j,i, ::}$ is represented in Equation~(\ref{eq:mat_tij}), where we define:
\begin{align*}
  \mathrm{toep}(\tK_{j, i, k_1}) =
  \begin{psmallmatrix}
    \tK_{j, i, k_1, \lfloor\frac{k}{2}\rfloor} & \cdots  & \tK_{j, i, k_1, 1} & 0 & \cdots & 0\\
    \vdots & \tK_{j, i, k_1, \lfloor\frac{k}{2}\rfloor} & \ddots & \ddots & \ddots &  \vdots \\
    \tK_{j, i, k_1, k} & \ddots &\ddots &\ddots &\ddots & 0 \\
    0 & \ddots &\ddots &\ddots &\ddots & \tK_{j, i, k_1, 1} \\
    \vdots & \ddots &\ddots &\ddots & \tK_{j, i, k_1, \lfloor\frac{k}{2}\rfloor} & \vdots \\
    0 & \cdots & 0  & \tK_{j, i, k_1, k} & \cdots  & \tK_{j, i, k_1, \lfloor\frac{k}{2}\rfloor}
  \end{psmallmatrix}
\end{align*}

\begin{figure*}[t]
  \begin{align}
    \label{eq:mat_tij}
    \mT_{j,i, ::} =
    \begin{psmallmatrix}
      \mathrm{toep}(\tK_{j, i, \lfloor\frac{k}{2}\rfloor}) & \cdots  & \mathrm{toep}(\tK_{j, i, 1}) & 0 & \cdots & 0\\
      \vdots & \mathrm{toep}(\tK_{j, i, \lfloor\frac{k}{2} \rfloor + 1  }) & \ddots & \ddots & \ddots &  \vdots \\
      \mathrm{toep}(\tK_{j, i, k}) & \ddots &\ddots &\ddots &\ddots & 0 \\
      0 & \ddots &\ddots &\ddots &\ddots & \mathrm{toep}(\tK_{j, i, 1}) \\
      \vdots & \ddots &\ddots &\ddots & \mathrm{toep}(\tK_{j, i, \lfloor\frac{k}{2}\rfloor}) & \vdots \\
      0 & \cdots & 0  & \mathrm{toep}(\tK_{j, i, k}) & \cdots  & \mathrm{toep}(\tK_{j, i, \lfloor\frac{k}{2}\rfloor})
    \end{psmallmatrix}
  \end{align}
\end{figure*}

\subsection{Circulant matrix}
\label{section:appendix_circulant_matrix}

In the same manner, as for zero padding with matrix $\mT$, $\mC$ can also be represented as a $\cout \times  \cin $ block matrix of doubly circulant matrices.
$\mC_{j,i, ::}$ is fully determined by the kernel $\tK$ as:
\begin{align*}
  \mC_{j,i, ::} =
  \begin{psmallmatrix}
    \circulant(\tK_{j, i, 1, :}) & \circulant(\tK_{j, i, 2, :}) & \cdots & \circulant(\tK_{j, i, n, :}) \\
    \circulant(\tK_{j, i, n, :}) & \circulant(\tK_{j, i, 1, :}) & \ddots & \vdots \\
    \vdots & \ddots & \circulant(\tK_{j, i, 1, :}) & \circulant(\tK_{j, i, 2, :}) \\
    \circulant(\tK_{j, i, 2, :}) & \cdots & \circulant(\tK_{j, i, n, :}) & \circulant(\tK_{j, i, 1, :}) \phantom{\vdots}
  \end{psmallmatrix} \ .
\end{align*}
We denote $\mC_{j,i, ::} = \bcirc(\tK_{j, i, 1, :}, \dots, \tK_{j, i, n, :})$.


\newpage
\subsection{Proof of Theorem~\ref{thm:gram_iteration_main_result}}
\label{section:appendix_gram_iteration_main_result}

\begin{proof}
  \emph{Closed form.} By induction, for $t\ge 2$,
  \[
    \mW^{(t)}=(\mW^*\mW)^{\,2^{t-2}}=: \mA^{\,k},\qquad k:=2^{t-2},
  \]
  so
  \[
    s_t=\|\mW^{(t)}\|^{2^{1-t}}=\|\mA^{k}\|^{1/(2k)}.
  \]

  \emph{Convergence and upper bound.} Gelfand’s formula gives
  $\lim_{k\to\infty}\|\mA^{k}\|^{1/k}=\rho(\mA)=\sigma_1(\mW)^2$, hence
  $s_t\to \sigma_1(\mW)$. Since $\rho(\mX)\le\|\mX\|$ for any norm,
  $\sigma_1(\mW)^{2^{t-1}}=\rho(\mA)^{k}\le\|\mA^{k}\|$, and raising to
  $2^{1-t}$ yields $s_t\ge\sigma_1(\mW)$.

  \emph{Rate (general norm).} By norm equivalence in finite dimension,
  $\exists\,0<m\le M<\infty$ such that $m\|\mX\|_2\le\|\mX\|\le M\|\mX\|_2$.
  For $\mA\succeq 0$, $\|\mA^{k}\|_2=\rho(\mA)^{k}$. Thus
  \[
    \sigma_1(\mW)\,e^{\frac{\log m}{2^{t-1}}}\le s_t
    \le \sigma_1(\mW)\,e^{\frac{\log M}{2^{t-1}}}
  \]
  and $s_t-\sigma_1(\mW)=O(2^{-t})$.

  \emph{Frobenius readout.} Suppose the final norm is $\|\cdot\|_F$.
  Let the eigenvalues of $\mA$ be $\lambda_1\ge\lambda_2\ge\cdots\ge 0$ and
  set $q:=\sqrt{\lambda_2/\lambda_1}=\sigma_2(\mW)/\sigma_1(\mW)\in(0,1)$.
  Then
  \[
    \|\mA^{k}\|_F^2=\sum_{i}\lambda_i^{2k}
    =\lambda_1^{2k}\!\left(1+\sum_{i\ge 2}(\lambda_i/\lambda_1)^{2k}\right)
    =\lambda_1^{2k}\,(1+S_k),
  \]
  with $0\le S_k\le (r-1)\,q^{4k}$ ($r=\operatorname{rank}\mA$) and $S_k\ge q^{4k}$ if
  $\lambda_2>0$. Hence
  \[
    s_t=\|\mA^{k}\|_F^{1/(2k)}=\sigma_1(\mW)\,(1+S_k)^{1/(4k)},\qquad
    e_t=\sigma_1(\mW)\big[(1+S_k)^{1/(4k)}-1\big].
  \]
  Using $\log(1+u)=u+O(u^2)$ and $4k=2^{t}$,
  \[
    e_t=\sigma_1(\mW)\,\frac{S_k}{4k}\,(1+o(1)).
  \]
  Consequently, for some constants $0<c\le C<\infty$,
  \[
    c\,\frac{q^{\,2^{t}}}{2^{t}} \;\le\; e_t \;\le\; C\,\frac{q^{\,2^{t}}}{2^{t}}
    \qquad (t\to\infty).
  \]
  Given any $\varepsilon>0$, since
  $\big(\tfrac{q}{q+\varepsilon}\big)^{2^{t}}\,2^{t}\to 0$,
  there exist $C_\varepsilon,T$ with $e_t\le C_\varepsilon\,(q+\varepsilon)^{2^{t}}$
  for all $t\ge T$: this is $K$-supergeometric with $K=q$.

  For the “order”, write $\log e_t=\log c + 2^{t}\log q - t\log 2 + o(1)$
  (with some $c>0$), so
  \[
    p_t:=\frac{\log e_{t+1}}{\log e_t}
    =\frac{\log c + 2^{t+1}\log q -(t+1)\log 2+o(1)}
    {\log c + 2^{t}\log q - t\log 2+o(1)}
    =2-\Theta\!\Big(\frac{t}{2^{t}}\Big),
  \]
  hence $p_t\uparrow 2$.

  \emph{(Not R-quadratic.)} From $e_t = \Theta(q^{2^{t}}/2^{t})$ one gets
  $e_{t+1}/e_t^{2}\sim (2^{t-1}/\sigma_1(\mW))\to\infty$, so no inequality
  $e_{t+1}\le K e_t^2$ can hold for all large $t$.

  \emph{Operator 2-norm.} If the final norm is $\|\cdot\|_2$ then
  $\|\mA^{k}\|_2=\rho(\mA)^k$ and $s_t\equiv\sigma_1(\mW)$ for $t\ge 2$.
\end{proof}

\subsection{Proof of Theorem~\ref{thm:bound_spectral_norm_toeplitz}}

\begin{proof}
  \begin{align*}
    \mT_{j, i, k_2 l_2, k_1 l_1} = \tK_{i, j, k_1 - k_2 + k -1, l_1 - l_2 + k -1}
  \end{align*}

  For $\mT^{(1)} \in \R^{n^2 \cin \times n^2 \cin}$, the first Gram iterate, $1 \leq i_1, i_2 \leq \cin$,
  \begin{align*}
    \mT^{(1)}_{i_1, i_2, k_1 l_1, k_2 l_2} = \left( \sum_{j=1}^\cout   \tK_{j, i_1} \star \tK_{j, i_2} \right)_{k_2 - k_1, l_2 -l_1}
  \end{align*}
  It was first derived in \citet{prach2022almost}.

  We define the Gram iterate for the filter $\tK$, $\tK^{(t+1)}_{i_1, i_2} = \sum_{j=1}^\cout   \tK^{(t)}_{j, i_1} \star \tK^{(t)}_{j, i_2}$, where convolution is defined with maximal non-trivial padding. We pad $\tK^{(t)}$ with zeros corresponding to the current spatial size of the kernel of $\tK^{(t)}$ minus one.

  We can extend the previous result for the $t$-th iterate of Gram:
  \begin{align}
    \label{eq:toeplitz_gram_conv_relation}
    \mT^{(t+1)}_{i_1, i_2, k_1 l_1, k_2 l_2} = \left( \sum_{j=1}^\cout   \tK^{(t)}_{j, i_1} \star \tK^{(t)}_{j, i_2} \right)_{k_2 - k_1, l_2 -l_1} \ .
  \end{align}
  For the norm $\|~.~\|_{\infty}$, we use the tight bound which simplifies to:
  \begin{align*}
    \norm{\mT^{(t+1)}}_\infty & = \max_{i_2, k_2, l_2} \sum_{i_1, k_1, l_1} \left| \sum_{j=1}^\cout   \tK^{(t)}_{j, i_1} \star \tK^{(t)}_{j, i_2} \right|_{k_2 - k_1, l_2 -l_1}        \\
                              & \leq  \max_{i_2} \sum_{i_1, k^\prime, l^\prime} \left| \sum_{j=1}^\cout   \tK^{(t)}_{j, i_1} \star \tK^{(t)}_{j, i_2} \right|_{k^\prime, l^\prime} \ .
  \end{align*}
  For the norm $\|~.~\|_\frob$, it gives:
  \begin{align*}
    \norm{\mT^{(t+1)}}_\frob^2 & =  \sum_{i_1, i_2, k_1, k_2, l_1, l_2} \left| \sum_{j=1}^\cout   \tK^{(t)}_{j, i_1} \star \tK^{(t)}_{j, i_2} \right|_{k_2 - k_1, l_2 -l_1}^2          \\
                               & \leq k^2 \sum_{i_1, i_2, k^\prime, l^\prime} \left| \sum_{j=1}^\cout   \tK^{(t)}_{j, i_1} \star \tK^{(t)}_{j, i_2} \right|_{k^\prime, l^\prime}^2 \ .
  \end{align*}
\end{proof}


\subsection{Proof of Theorem~\ref{thm:bound_approximation_for_lower_input_size}}

\begin{theorem} [Adapted from Theorem~1 of \citet{pfister2019bounding}]
  \label{theorem:approx_trigonometric_polynomial}
  Let $\gamma : \mathbb{R}^2 \rightarrow \mathbb{C}$ be a trigonometric polynomial of degree $d = \left\lfloor k/2 \right \rfloor$ defined by coefficients $\Gamma \in \mathbb{C}^{k \times k}$:
  \begin{equation*}
    \gamma(w_1, w_2) = \sum_{k_1=0}^{2d} \sum_{k_2=0}^{2d}
    \Gamma_{k_1, k_2} \ e^{\ci(w_1 (k_1-d))} e^{\ci(w_2 (k_2-d))} \ .
  \end{equation*}
  Let $\Omega_n$ be the set of $n$ equidistant sampling points on $[0, 2\pi ]$: $\Omega_n = \left \{\omega_l = \frac{2\pi l}{n} ~ |~ l \in \{1, \dots, n \} \right \}$. \\
  Then, for $\alpha = \frac{2d}{n}$, we have:
  \begin{equation*}
    \max_{w_1, w_2 \in [0, 2\pi ]^2} \left| \gamma(w_1, w_2) \right|
    \leq (1 - \alpha)^{-1} \max_{w_1^{\prime}, w_2^{\prime} \in \Omega_n^2} \left| \gamma(w_1^{\prime}, w_2^{\prime}) \right| \ .
  \end{equation*}
\end{theorem}

We define $\mE \in \mathbb{C}^{\cout \times \cin}$ as the spectral density matrix of the filter $\tK$:
\begin{equation}
  \label{eq:spectral_density_matrix}
  \mE (w_1, w_2) = \sum_{k_1=0}^{k-1} \sum_{k_2=0}^{k-1} e^{-\ci k_1 w_1} \ e^{-\ci k_2 w_2} \ \tK_{:,:,k_1, k_2}.
\end{equation}
The spectral density matrix $\mE$ corresponds to the Discrete Time Fourier Transform (DTFT) of the filter $\tK$, where $\tK_{:,:,k_1,k_2}$ represents the coefficients of the convolution kernel.

Although $\mE$ is directly defined in terms of $\tK$, it provides key insights into the spectral properties of the Toeplitz matrix $\mT$, as $\mT$ is constructed based on $\tK$. Iteratively, $\mE^{(t)}$ denotes the $t^\text{th}$ Gram iterate of the matrix $\mE$, which is used to analyze the spectral behavior of $\mT$.

Using Theorem~\ref{theorem:approx_trigonometric_polynomial}, we can bound the Gram iterate maximum norm over $[0, 2\pi]$ of $\mE$ with the maximum taken over uniform sampled points.

\begin{lemma} (Inequality between the maximum of spectral norm density and density uniformly sampled for Gram iterates) \\
  \label{lemma:inequality_btw_max_spectral_norm_density}
  For $n_0 \geq 2^{t}  \lfloor \frac{k}{2}\rfloor + 1$ sampling points, $\alpha = \frac{2^{t} \lfloor \frac{k}{2}\rfloor}{n_0}$:
  \begin{equation*}
    \max_{w_1,w_2 \in [0, 2\pi]} \norm{\mE^{(t)}(w_1, w_2)}_\frob^2  \leq (1 - \alpha)^{-1} \max_{1 \leq u,v \leq n_0} \norm{{\mD^\downarrow}^{(t)}_{u,v}}_\frob^{2}
  \end{equation*}
\end{lemma}

\begin{proof}
  We define the trigonometric polynomial of degree $2^t \lfloor \frac{k}{2}\rfloor$: $\mP = \norm{\mE^{(t)}(~.~, ~.~)}^{2}_\frob$.
  \begin{align*}
     & \max_{w_1,w_2 \in [0, 2\pi]} \norm{\mE^{(t)}(w_1, w_2)}_\frob^2                                                                                  \\
     & = \max_{w_1,w_2 \in [0, 2\pi]^2} \mP(w_1, w_2)                                                                                                   \\
     & \leq  (1 - \alpha)^{-1} \max_{w_1,w_2 \in \Omega_{n_0}^2} \mP(w_1, w_2) ~~ \text{applying Theorem~\ref{theorem:approx_trigonometric_polynomial}} \\
     & = (1 - \alpha)^{-1} \max_{1 \leq u,v \leq n_0} \norm{{\mD^\downarrow}^{(t)}_{u,v}}_\frob^{2} \ .
  \end{align*}
\end{proof}

Now we can prove Theorem~\ref{thm:bound_approximation_for_lower_input_size}.

\begin{proof}
  We can cast $\underset{1 \leq u,v \leq n}{\max} \norm{{\mD_{u,v}}^{(t)}}_\frob^2$ as the maximum over $\Omega_n^2$ of $\norm{\mE^{(t)}(w_1, w_2)}^{2}_\frob$.
  \begin{align*}
    \max_{1 \leq u,v \leq n} \norm{{\mD}^{(t)}_{u,v}}_\frob^2 & = \max_{w_1,w_2 \in \Omega_n^2} \norm{\mE^{(t)}(w_1, w_2)}^{2}_F          \\
                                                              & \leq  \max_{w_1,w_2 \in [0, 2\pi]^2} \norm{\mE^{(t)}(w_1, w_2)}^{2}_\frob
  \end{align*}
  Using Lemma~\ref{lemma:inequality_btw_max_spectral_norm_density}:
  \begin{align*}
    \max_{1 \leq u,v \leq n} \norm{{\mD}^{(t)}_{u,v}}_\frob^2
    \leq  (1 - \alpha)^{-1} \max_{1 \leq u,v \leq n_0} \norm{{\mD^\downarrow}^{(t)}_{u,v}}_\frob^2 \ .
  \end{align*}
  Finally, using that
  \begin{align*}
    \sigma_1(\mC) = \max_{1 \leq u,v \leq n}  \norm{{\mD}_{u,v}}_2
     & \leq \max_{1 \leq u,v \leq n} \norm{{\mD}^{(t)}_{u,v}}_\frob^{2^{1-t}} \ ,
  \end{align*}
  we have
  \begin{equation*}
    \sigma_1(\mC) \leq \left(\frac{1}{1 - \alpha}\right)^{2^{-t}}
    \max_{1 \leq u,v \leq n_0} \norm{{\mD^\downarrow}^{(t)}_{u,v}}_\frob^{2^{1-t}} \ .
  \end{equation*}
\end{proof}


\subsection{Proof of Theorem~\ref{thm:bound_circ_toep}}
\label{appendix:proof_bound_circ_toep}
\begin{proof}
  We use the spectral density matrix $\mE$ defined in Equation~\ref{eq:spectral_density_matrix}.
  Lemma~4 of \citet{yi2020asymptotic} shows that the spectral norm of the Toeplitz matrix is bounded by the spectral norm of its density matrix:
  $\sigma_1(\mT) \leq \sigma_1(\mE)$.
  Then,
  \begin{equation*}
    \sigma_1(\mT)^{2^t} \leq \sigma_1(\mE)^{2^{t}} \leq \norm{\mE^{(t)}}_\frob^{2} \ .
  \end{equation*}
  Using Lemma~\ref{lemma:inequality_btw_max_spectral_norm_density} with $n$ sampling points:
  \begin{align*}
    \max_{w_1,w_2 \in [0, 2\pi]} \norm{\mE^{(t)}(w_1, w_2)}_\frob^2 &
    \leq (1 - \alpha)^{-1} \max_{1 \leq u,v \leq n} \norm{{\mD}^{(t)}_{u,v}}_\frob^{2} \ .
  \end{align*}
  Finally,
  \begin{equation*}
    \sigma_1(\mT) \leq \left(\frac{1}{1 - \alpha}\right)^{2^{-t}}
    \max_{1 \leq u,v \leq n} \norm{{\mD}^{(t)}_{u,v}}_\frob^{2^{1-t}} \ .
  \end{equation*}
\end{proof}

\subsection{Proof of Proposition~\ref{prop:orthogonal_circular_padding}}
\label{appendix:proof_orthogonal_circular_padding}
\begin{proof}
  As in the previous proof, Lemma~4 of \citet{yi2020asymptotic} shows that the spectral norm of the Toeplitz matrix \( \mT \) is bounded by the spectral norm of its density matrix \( \mE \), defined in Eq.~(\ref{eq:spectral_density_matrix}):
  \[
    \sigma_1(\mT) \leq \sigma_1(\mE).
  \]

  Using Lemma~\ref{lemma:inequality_btw_max_spectral_norm_density} with \( n \) sampling points, we have:
  \begin{align*}
    \max_{w_1, w_2 \in [0, 2\pi]} \norm{\mE^{(t)}(w_1, w_2)}_\frob^2
     & \leq (1 - \alpha)^{-1} \max_{1 \leq u, v \leq n} \norm{\mD^{(t)}_{u,v}}_\frob^{2}.
  \end{align*}

  In the case where \( c = \cout = \cin \), for orthogonal circular convolution, we have \( \norm{\mD^{(t)}_{u,v}}_\frob^{2} = c^2 \). Thus:
  \[
    \sigma_1(\mT) \leq \left(\frac{c}{1 - \alpha}\right)^{2^{-t}}.
  \]

  Taking the limits \( n \to \infty \) and \( t \to \infty \), we obtain:
  \[
    \sigma_1(\mT) \leq 1.
  \]

  This completes the proof.
\end{proof}

\subsection{Additional experiments}

\begin{table}[h]
  \caption{Numerical values associated to Figure~\ref{fig:comparaison_methods_compute_spectral_norm_matrix}: error ratios and computational times of methods for spectral norm computation. Error ratio is defined as ${\sigma_1}_{\text{method}} / {\sigma_1}_{\text{ref}} - 1$.}
  \centering
  \footnotesize
  \begin{tabular}{lrrr}\toprule
    \textbf{Method}    & \textbf{Error ratio}          & \textbf{Computational time (s)} \\\midrule
    Reference          & 0                             & 1.47e+0                         \\
    PI (10 iters)      & -3.27e-2 $\pm$ 3.55e\text{-}3 & 1.12e-3                         \\
    PI (100 iters)     & -1.41e-3 $\pm$ 1.75e\text{-}3 & 1.03e-2                         \\
    PI (1000 iters)    & -2.43e-6 $\pm$ 4.01e\text{-}6 & 1.04e-1                         \\
    Ours GI (10 iters) & 6.75e-6                       & 1.83e-3                         \\
    Ours GI (11 iters) & 4.73e-8                       & 1.99e-3                         \\
    Ours GI (12 iters) & 4.33e-12                      & 2.13e-3                         \\
    \bottomrule
  \end{tabular}
  \label{tab:comparaison_methods_compute_spectral_norm_matrix}
\end{table}

\begin{table}[!htp]\centering
  \caption{Times and difference to reference,
    for different $p \times p$ matrices, averaged 10 times. Error is defined as $\sigma_{\text{method}} - \sigma_{\text{ref}}$.  To have comparable time and precision, $100$ iterations were taken for Power iteration and $10$ for Gram iteration. Gram iteration is overall more competitive than Power iteration for all matrix dimensions considered.
  }
  \footnotesize
  \begin{tabular}{lrrrrrr}
    \toprule
    \textbf{Dimension p} & \textbf{PI time (s)} & \textbf{GI time (s)} & \textbf{PI error} & \textbf{GI error} & \textbf{Ref} \\
    \midrule
    10                   & 0.0093               & 0.0012               & 0.0006            & 0                 & 5.28         \\
    100                  & 0.0083               & 0.0010               & -0.0231           & 0                 & 19.31        \\
    1000                 & 0.0093               & 0.0017               & -0.0426           & 0.0005            & 62.95        \\
    10000                & 0.016                & 0.0019               & -0.6004           & 0.0037            & 199.77       \\
    50000                & 0.017                & 0.0040               & -1.70             & 0.1               & 447.21       \\
    \bottomrule
  \end{tabular}
  \label{tab:comparaison_methods_compute_spectral_norm_dimension}
\end{table}

\begin{table*}[h]
  \caption{Comparison of Lipschitz bounds for all convolutional layers ResNet18, reference for real Lipschitz bound is given by \citet{ryu2019plug} method. We observe that our method gives close value to \citet{sedghi2019singular} up to two decimal digits while being significantly faster. We remark that convolution filter $512 \times 512 \times 1 \times 1$ \citet{singla2021fantastic} value $2.03$ underestimates reference value $2.05$.}
  \centering
  \scriptsize
  \resizebox{0.95\textwidth}{!}{
    \begin{tabular}{lllllllllllll}\toprule
      \textbf{Filter Shape}                  & \multicolumn{5}{c}{\textbf{Lipschitz Bound}} &            & \multicolumn{5}{c}{\textbf{Running Time (s)}}                                                                                     \\\cmidrule{2-6}\cmidrule{8-12}
                                             & Ours                                         & Singla2021 & Araujo2021                                    & Sedghi2019 & Ryu2019 &  & Ours   & Singla2021 & Araujo2021 & Sedghi2019 & Ryu2019 \\\cmidrule{2-6}\cmidrule{8-12}
      64 $\times$ 3 $\times$ 7 $\times$ 7    & 15.91                                        & 28.89      & 22.25                                         & 15.91      & 15.91   &  & 0.0024 & 0.0507     & 0.0003     & 18.06      & 0.7579  \\\cmidrule{1-6}\cmidrule{8-12}
      64 $\times$ 64 $\times$ 3 $\times$ 3   & 6.00                                         & 9.32       & 17.59                                         & 6.00       & 6.00    &  & 0.0014 & 0.0264     & 0.0003     & 7.47       & 1.07    \\\cmidrule{1-6}\cmidrule{8-12}
      64 $\times$ 64 $\times$ 3 $\times$ 3   & 5.34                                         & 6.28       & 15.31                                         & 5.34       & 5.33    &  & 0.0015 & 0.025      & 0.0012     & 7.29       & 1.07    \\\cmidrule{1-6}\cmidrule{8-12}
      64 $\times$ 64 $\times$ 3 $\times$ 3   & 7.00                                         & 8.71       & 16.46                                         & 7.00       & 7       &  & 0.0011 & 0.026      & 0.0003     & 7.27       & 1.08    \\\cmidrule{1-6}\cmidrule{8-12}
      64 $\times$ 64 $\times$ 3 $\times$ 3   & 3.82                                         & 5.40       & 13.74                                         & 3.81       & 3.82    &  & 0.0014 & 0.025      & 0.0003     & 7.19       & 1.08    \\\cmidrule{1-6}\cmidrule{8-12}
      128 $\times$ 64 $\times$ 3 $\times$ 3  & 4.71                                         & 5.98       & 16.35                                         & 4.71       & 4.71    &  & 0.0014 & 0.026      & 0.0012     & 9          & 0.473   \\\cmidrule{1-6}\cmidrule{8-12}
      128 $\times$ 128 $\times$ 3 $\times$ 3 & 5.72                                         & 7.21       & 23.58                                         & 5.72       & 5.72    &  & 0.0011 & 0.026      & 0.0013     & 4.85       & 0.78    \\\cmidrule{1-6}\cmidrule{8-12}
      128 $\times$ 64$\times$ 1 $\times$ 1   & 1.91                                         & 1.91       & 6.39                                          & 1.91       & 1.91    &  & 0.0014 & 0.0333     & 0.0001     & 9.25       & 0.042   \\\cmidrule{1-6}\cmidrule{8-12}
      128 $\times$ 128 $\times$ 3 $\times$ 3 & 4.39                                         & 6.78       & 21.93                                         & 4.39       & 4.41    &  & 0.0011 & 0.026      & 0.0003     & 4.844      & 0.78    \\\cmidrule{1-6}\cmidrule{8-12}
      128 $\times$ 128 $\times$ 3 $\times$ 3 & 4.89                                         & 7.57       & 19.79                                         & 4.89       & 4.88    &  & 0.0014 & 0.027      & 0.0004     & 4.78       & 0.78    \\\cmidrule{1-6}\cmidrule{8-12}
      256 $\times$ 128 $\times$ 3 $\times$ 3 & 7.39                                         & 8.45       & 35.86                                         & 7.39       & 7.39    &  & 0.0014 & 0.026      & 0.0017     & 5.36       & 0.46    \\\cmidrule{1-6}\cmidrule{8-12}
      256 $\times$ 256 $\times$ 3 $\times$ 3 & 6.58                                         & 8.03       & 37.98                                         & 6.58       & 6.58    &  & 0.0014 & 0.026      & 0.0025     & 3.46       & 0.85    \\\cmidrule{1-6}\cmidrule{8-12}
      256 $\times$ 128$\times$ 1 $\times$ 1  & 1.21                                         & 1.21       & 5.97                                          & 1.21       & 1.21    &  & 0.0014 & 0.036      & 0.0001     & 5.96       & 0.04    \\\cmidrule{1-6}\cmidrule{8-12}
      256 $\times$ 256 $\times$ 3 $\times$ 3 & 6.36                                         & 7.57       & 30.07                                         & 6.36       & 6.35    &  & 0.001  & 0.026      & 0.0004     & 3.48       & 0.85    \\\cmidrule{1-6}\cmidrule{8-12}
      256 $\times$ 256 $\times$ 3 $\times$ 3 & 7.68                                         & 9.17       & 29.19                                         & 7.68       & 7.67    &  & 0.0011 & 0.027      & 0.0004     & 3.48       & 0.85    \\\cmidrule{1-6}\cmidrule{8-12}
      512 $\times$ 256 $\times$ 3 $\times$ 3 & 9.99                                         & 11         & 46.24                                         & 9.99       & 9.92    &  & 0.0011 & 0.026      & 0.004      & 3.55       & 0.45    \\\cmidrule{1-6}\cmidrule{8-12}
      512 $\times$ 512 $\times$ 3 $\times$ 3 & 9.09                                         & 10.45      & 47.59                                         & 9.09       & 9.04    &  & 0.0014 & 0.026      & 0.0001     & 2.8        & 0.79    \\\cmidrule{1-6}\cmidrule{8-12}
      512 $\times$ 256 $\times$ 1 $\times$ 1 & 2.05                                         & 2.03       & 11.87                                         & 2.05       & 2.05    &  & 0.0012 & 0.024      & 0.0001     & 3.77       & 0.052   \\\cmidrule{1-6}\cmidrule{8-12}
      512 $\times$ 512 $\times$ 3 $\times$ 3 & 17.60                                        & 18.37      & 59.04                                         & 17.60      & 17.5    &  & 0.001  & 0.027      & 0.0005     & 2.79       & 0.79    \\\cmidrule{1-6}\cmidrule{8-12}
      512 $\times$ 512 $\times$ 3 $\times$ 3 & 7.48                                         & 7.6        & 57.33                                         & 7.48       & 7.43    &  & 0.0014 & 0.027      & 0.0006     & 2.83       & 0.79    \\
      \bottomrule
    \end{tabular}
  }
  \label{tab:lipschitz_resnet18}
\end{table*}

%% file: content/appendix/appendix-lipschitz.tex
\chapter{Lipschitz networks}\label{app:chapter:lipschitz_networks}

\section{Computation of Product Upper Bound (PUB) for ResNet architectures}
\label{app:sec:pub_approximation}

The Product Upper Bound ($\PUB$) for the Lipschitz constant of a ResNet model is computed as the product of the Lipschitz constants of its individual layers. Below, we summarize how the Lipschitz constant is estimated for different components in the ResNet architecture.

For convolutional layers (Conv2D), the Lipschitz constant is the spectral norm of the layer, which corresponds to the largest singular value of the weight matrix. This is computed via power iteration applied directly to the convolution operator as in \citet{ryu2019plug}. The iteration alternates between forward and transposed convolutions until convergence.

For fully connected (dense) layers, the Lipschitz constant is also the spectral norm of the weight matrix. Power iteration is similarly applied, where random vectors are iteratively multiplied by the weight matrix and its transpose to approximate the largest singular value.

Batch normalization layers (BatchNorm) are treated as affine transformations during inference. The Lipschitz constant is computed as:
\[
    \Lip_{\text{BN}} = \max_{i} \left| \frac{\gamma_i}{\sqrt{\sigma_i^2 + \epsilon}} \right|,
\]
where \( \gamma_i \) is the scale parameter, \( \sigma_i^2 \) is the running variance, and \( \epsilon \) is a small stability constant. This ensures the Lipschitz constant accurately reflects the effect of batch normalization during evaluation.

Pooling layers, including max pooling and average pooling, are conservatively assumed to have a Lipschitz constant of 1. Max pooling performs a maximum operation over the input, which does not amplify the norm, and average pooling is a linear operation with bounded norm.

Residual connections involve summing the output of a subnetwork with its input. The Lipschitz constant of a residual block is given by:
\[
    \Lip_{\text{residual}} \leq L_{\text{main}} + 1,
\]
where \( \Lip_{\text{main}} \) is the Lipschitz constant of the main path (e.g., convolutional layers). The identity mapping has a Lipschitz constant of 1, and the addition operation preserves the norm.

The overall PUB for a ResNet model is computed by multiplying the Lipschitz constants of all layers, taking into account the components described above. This product provides a conservative upper bound for the Lipschitz constant of the network, which is crucial for robust certification and stability analysis.

\section{Proof of Theorem~\ref{thm:improve_aol}}
\label{app:sec:proof_improve_aol}
\begin{proof}
    Using Theorem~1 of \cite{araujo2023a}, we have to show that $\mW^\top \mW - \mathbb{T} \preceq 0$, for $t \geq 1$.
    \\
    The case $t=1$ comes directly from Theorem~3 of \cite{araujo2023a}.
    \\
    For $t \geq 1$, using Theorem 3 of \cite{araujo2023a} on $\mW^{(t)}$,
    \begin{align*}
        {\mW^{(t)}}^\top \mW^{(t)}
        - \diag{\left( \sum_j \left| {\mW^{(t)}}^\top \mW^{(t)} \right|_{ij}
        \frac{q_j}{q_i}\right)} & \preceq 0     \\
        (\mW^\top \mW)^{2^{t}}  - \diag{\left( \sum_j \left| {\mW^{(t+1)}} \right|_{ij}
        \frac{q_j}{q_i}\right)} & \preceq 0 \ .
    \end{align*}
    Using \cite[Theorem~7.9, p.~210]{zhang2011matrix}, we have that for $A, B \succeq 0$, $A^2 \preceq B^2 \implies A \preceq B$:
    \begin{align*}
        \mW^\top \mW  - \diag{\left( \sum_j \left| {\mW^{(t+1)}} \right|_{ij} \frac{q_j}{q_i} \right)}^{2^{-t}} & \preceq 0 \ .
    \end{align*}
\end{proof}

\section{Proof of Theorem~\ref{thm:thm_bound_lip_sigma_element_wise} Lipschitz bound for the Weierstrass transform with based Lipschitz continuity
 }
\label{app:sec:proof_bound_lip_sigma_element_wise}

Stein's lemma can be easily extended to $h : \mathbb{R}^d \mapsto \mathbb{R}^c$.
We note $\frac{\partial}{\partial x} \tilde{h}(x)$ the Jacobian matrix of $\tilde{h}(x)$.
\begin{lemma}
    \label{lemma:stein_extended}
    Let $\sigma > 0$, let $h : \mathbb{R}^d \mapsto \mathbb{R}^c$ be measurable, and let
    $\tilde{h}(x) = \mathbb{E}_{\delta\sim {\cal N}(0, \sigma^2 I)}[h(x + \delta)]$. Then $\tilde{h}$ is differentiable, and moreover,
    $$
        \frac{\partial \tilde{h}(x)}{\partial x}  = \frac{1}{\sigma^2} \mathbb{E}_{\delta\sim {\cal N}(0, \sigma^2 I)}[ \delta h(x + \delta)^\top] \ .
    $$
\end{lemma}

\begin{proof}
    We are not going to prove differentiability as it is the same argument as in the proof of Lemma~\ref{lemma:stein}, see \citep{stein1981annals}.
    \begin{align*}
        \frac{\partial }{\partial x}  \tilde{h}(x) & = \frac{\partial }{\partial x}  \left( \frac{1}{(2\pi\sigma^2)^{d/2}} \int_{\mathbb{R}^d} h(t) \exp \left( -\frac{1}{2\sigma^2} \|x - t\|^2 \right) dt \right)       \\
                                                   & = \frac{1}{(2\pi\sigma^2)^{d/2}} \int_{\mathbb{R}^d} \frac{\partial }{\partial x}  \left( \exp \left( -\frac{1}{2\sigma^2} \|x - t\|^2 \right) \right) h(t)^\top dt  \\
                                                   & = \frac{1}{(2\pi\sigma^2)^{d/2}} \int_{\mathbb{R}^d}  \frac{\partial }{\partial x}  \left( \exp \left( -\frac{1}{2\sigma^2} \|x - t\|^2 \right) \right) h(t)^\top dt \\
                                                   & = \frac{1}{(2\pi\sigma^2)^{d/2}} \int_{\mathbb{R}^d}  \frac{t - x}{\sigma^2} \exp \left( -\frac{1}{2\sigma^2} \|x - t\|^2 \right) h(t)^\top dt \ ,
    \end{align*}
    with a change of variable and definition of expectation, we get the result.
\end{proof}

\begin{lemma}
    \label{lemma:lip_ho}
    For $L \geq 0, r \geq 0$, let $h : \R^d \mapsto [0, r]$ be defined as follows:
    \[
        h(x) = \frac{1}{2} \ \text{sign}(x_1) \min\{r, 2 L |x_1|\} + \frac{r}{2} \ ,
    \]
    where $\text{sign}$ is the sign function with the convention $\text{sign}(0) = 0$.
    Then, $h$ is $L$-Lipschitz continuous.
\end{lemma}

\begin{proof}
    To show that $h$ is $L$-Lipschitz continuous, we need to demonstrate that for all $x, y \in \R^d$:
    \[
        |h(x) - h(y)| \leq L \norm{x - y}_2 \ .
    \]

    We write $x = (x_1, \dots, x_d)$ and $y = (y_1, \dots, y_d)$.
    In the following cases, only the first coordinate is going to intervene.
    We consider three cases:

    \textbf{Case 1:} $x_1 = 0$ and $y_1 = 0$.

    In this case, $\text{sign}(x_1) = 0$, $\text{sign}(y_1) = 0$, and $h(y) = h(x) = \frac{r}{2}$ for any $x$.

    Thus, $|h(x) - h(y)| = 0 \leq L \norm{x -y}_2$.

    \textbf{Case 2:} $x_1 \neq 0$ and $y_1 = 0$ (without loss of generality same as $x_1 = 0 $ and $y_1 \neq 0$).

    In this case, $h(x)$ is given by:
    \[
        h(x) = \frac{1}{2} \text{sign}(x_1) \min\{r, 2 L |x_1| \} + \frac{r}{2} \ ,
    \]
    and $h(y)$ is given by:
    \[
        h(y) = \frac{r}{2} \ .
    \]

    Now, let's consider the difference $|h(x) - h(y)|$:
    \[
        |h(x) - h(y)| = \left|\frac{1}{2} \text{sign}(x_1) \min\{r, 2 L |x_1| \} + \frac{r}{2} - \frac{r}{2}\right| \ .
    \]

    If $2L |x_1| \leq r$, then $\min\{r, 2L |x_1|\} = 2L|x_1|$ and the expression becomes:
    \[
        \left|\frac{1}{2} (2Lx_1) + \frac{r}{2} - \frac{r}{2}\right| = L|x_1| \leq L \norm{x-y}_2 \ .
    \]

    If $2L |x_1| > r$, then $\min\{r, 2L |x_1|\} = r$ and the expression becomes:
    \[
        \left|\frac{1}{2} \text{sign}(x_1) r + \frac{r}{2} - \frac{r}{2}\right| = \frac{r}{2} \leq L |x_1| \leq L \norm{x-y}_2 \ .
    \]

    In both cases, $|h(x) - h(y)| \leq L \norm{x -y}_2$, therefore, in the case where $x_1 \neq 0$ and $y_1 = 0$, $|h(x) - h(y)|$ is $L$-Lipschitz. Same result goes for  $x_1 = 0$ and $y_1 \neq 0$.

    \textbf{Case 3:} $x_1 \neq 0$ and $y_1 \neq 0$.

    Without loss of generality, assume $|x_1| \geq |y_1|$. Consider
    \[
        |h(x) - h(y)| = \frac{1}{2} \left| \text{sign}(x_1) \min\{r, 2 L |x_{1}|\} - \text{sign}(y_1) \min\{r, 2 L |y_{1}|\}\right| \ .
    \]

    Let's consider two sub-cases:

    \textbf{Sub-case 3a:} If $2 L |x_{1}| \leq r$, then $\min\{r, 2 L |x_{1}| \} = 2 L |x_{1}|$.

    \textbf{Sub-case 3b:} If $2 L |x_{1}| > r$, then $\min\{r, 2 L |x_{1}|\} = r$.

    Similarly, for $|y_1|$, we have $\min\{r, 2 L |y_{1}|\} = 2 L |y_{1}|$ if $2 L |y_{1}| \leq r$ and $\min\{r, 2 L |y_{1}|\} = r$ if $2 L |y_{1}| > r$.

    In both sub-cases, we can write:
    \[
        |h(x) - h(y)| = \frac{1}{2} |2 L x_{1} - 2 L y_{1}| = L |x_{1} - y_{1}| \leq L \norm{x-y}_2 \ .
    \]

    Therefore, $h$ is $L$-Lipschitz continuous.
\end{proof}


Now we are ready to prove the theorem.
\begin{proof}
    For ease of notation we note $L = L(f)$ ,
    we are interested in the following:
    \begin{align*}
        \label{eq: obj}
        J(\sigma, L)
        =
        \sup_{\substack{h} : L(h) = L} \ \
        \sup_{x\in \R^d}
        \|\nabla \tilde{h}(x)\|_2
        =
        \sup_{\substack{h} : L(h) = L} \ \
        \sup_{x\in \R^d} \ \
        \sup_{\substack{v\in \R^d: \|v\|=1}}
        v^\top\nabla \tilde{h}(x) \ .
    \end{align*}

    First, we will derive an upper bound on $J(\sigma, L)$. Consider any $x\in \R^d$, any $h$ Lipschitz continuous s.t $h(x) \in [0,r]$,
    and any $v\in \R^d$ with $\|v\| = 1$.
    Any $\delta\in \R^d$ can be decomposed as $\delta = \delta^\perp + \Tilde{\delta}$, where $\Tilde{\delta} = (v^T\delta)v$ and $\delta^\perp \perp v$.
    Let $\delta' = \delta^\perp - \Tilde{\delta}$.
    That is, $\delta'$ is the reflection of the vector $\delta$ with respect to the hyperplane that is normal to $v$.
    If $\delta\sim {\cal N}(0, \sigma^2 I)$, then $\delta'\sim {\cal N}(0, \sigma^2 I)$ because ${\cal N}(0, \sigma^2 I)$ is radially symmetric. Moreover, $v^T\delta' = -v^T\delta$. Hence,
    \begin{align*}
        \E_{\delta\sim {\cal N}(0, \sigma^2 I)}[v^\top\delta h(x+\delta)]
        =
        \E_{\delta\sim {\cal N}(0, \sigma^2 I)}[v^\top\delta' h(x+\delta')]
        = - \E_{\delta\sim {\cal N}(0, \sigma^2 I)}[v^\top\delta h(x+\delta')] \ .
    \end{align*}
    Using the above, we have the following, using Stein's Lemma~\ref{lemma:stein}:
    \begin{align*}
        v^\top\nabla \tilde{h}(x)
         & = \frac{1}{\sigma^2 }\E_{\delta\sim {\cal N}(0, \sigma^2 I)}[v^T\delta h(x+\delta)]                  \\
         & =
        \frac{1}{2\sigma^2 }\E_{\delta\sim {\cal N}(0, \sigma^2 I)}[v^T\delta (h(x+\delta) - h(x+\delta'))]     \\
         & \leq
        \frac{1}{2\sigma^2 }\E_{\delta\sim {\cal N}(0, \sigma^2 I)}[|v^T\delta|~|(h(x+\delta) - h(x+\delta'))|] \\
         & \stackrel{(i)}{\leq}
        \frac{1}{2\sigma^2 }\E_{\delta\sim {\cal N}(0, \sigma^2 I)}[|v^T\delta| \min\{r, 2L|v^T\delta|\}]       \\
         & \stackrel{(ii)}{=}
        \frac{1}{2\sigma^2 }\E_{\delta\sim {\cal N}(0, \sigma^2 I)}[|\delta_1| \min\{r, 2L|\delta_1|\}]         \\
         & \stackrel{(iii)}{=}
        \frac{1}{2\sigma^2 }\E_{z\sim {\cal N}(0, \sigma^2)}[|z| \min\{r, 2L |z|\}] \ ,
    \end{align*}
    where $(i)$ follows from the Lipschitz assumption on $h$ and the fact that $\|\delta-\delta'\| = \| 2\Tilde{\delta} \| = \|  2(v^T\delta)v\| = 2|v^T\delta|$ and that $|h(x+\delta) - h(x+\delta')| \leq r$,
    $(ii)$ follows by choosing the canonical unit vector $v = (1, 0, \ldots, 0)$ because the previous expression does not depend on the direction of $v$,
    and $(iii)$ follows by simply rewriting the expression in terms of $z$.

    Now, we will derive a lower bound on $J(\sigma, L)$. For this, we choose a specific $\bar{h} \in [0, r]$ as
    $\bar{h}(x) = \frac{1}{2}\sign(x_1) \min\{r, 2 L |x_1|\} + r/2$, with $\sign(0)=0$.
    Using Lemma~\ref{lemma:lip_ho}, we have that $\bar{h}$ is $L$-Lipschitz. We choose a specific $x = 0$ and specific unit vector $v_0 = (1, 0, \ldots, 0)$.
    For this choice, note that $\bar{h}(0) = r/2$ and $v_0^\top\delta = \delta_1$.
    Then,
    \begin{align*}
        J(\sigma, L)
        \geq
        v_0^\top \nabla \tilde{\bar{h}}
         & = \frac{1}{\sigma^2 }\E_{\delta\sim {\cal N}(0, \sigma^2 I)}[v_0^\top\delta \bar{h}(\delta)^\top]       \\
         & = \frac{1}{2\sigma^2 }\E_{\delta\sim {\cal N}(0, \sigma^2 I)}[ |\delta_1| \min\{r, 2L|\delta_1|\} ] + 0 \\
         & =\frac{1}{2\sigma^2 }\E_{z\sim {\cal N}(0, \sigma^2 )}[ |z| \min\{r, 2L|z|\} ] \ .
    \end{align*}

    Combining the upper and lower bounds, we have the following equality:
    \begin{align}
        J(\sigma, L)
        = \frac{1}{2\sigma^2 } \E_{z\sim {\cal N}(0,\sigma^2)}\left[ \min\left\{r|z|, 2Lz^2 \right\}\right] \ .
    \end{align}

    We will now compute the above expression exactly:
    \begin{align*}
        \frac{1}{2\sigma^2 } \E_{z\sim {\cal N}(0,\sigma^2)} & \left[ \min\left\{r|z|, 2Lz^2 \right\}\right]                                                                                                                \\
                                                             & = \frac{1}{2\sigma^2 } \int_{-\frac{r}{2L}}^{\frac{r}{2L}}
        \frac{1}{\sqrt{2 \pi \sigma^2}} \exp{\left(\frac{-z^2}{2\sigma^2}\right)} 2 L z^2 dz                                                                                                                                \\
                                                             & - \frac{1}{2\sigma^2 } \int_{-\infty}^{-\frac{r}{2L}}
        \frac{r}{\sqrt{2 \pi \sigma^2}} \exp{\left(\frac{-z^2}{2\sigma^2}\right)} z dz                                                                                                                                      \\
                                                             & + \frac{1}{2\sigma^2 } \int_{\frac{r}{2L}}^{+\infty}
        \frac{1}{\sqrt{2 \pi \sigma^2}} \exp{\left(\frac{-z^2}{2\sigma^2}\right)} z dz                                                                                                                                      \\
                                                             & = \frac{1}{\sigma^2 } \int_{-\frac{r}{2L}}^{\frac{r}{2L}}
        \frac{1}{\sqrt{2 \pi \sigma^2}} \exp{\left(\frac{-z^2}{2\sigma^2}\right)} L z^2 dz
        + \dfrac{\mathrm{e}^{-\frac{r}{8L^2{\sigma}^2}}}{2^\frac{3}{2}\sqrt{{\pi}}\left|{\sigma}\right|}
        + \dfrac{\mathrm{e}^{-\frac{r}{8L^2{\sigma}^2}}}{2^\frac{3}{2}\sqrt{{\pi}}\left|{\sigma}\right|}                                                                                                                    \\
                                                             & = L\operatorname{erf}\left(\frac{r}{2^\frac{3}{2}L{\sigma}}\right)-\dfrac{\mathrm{e}^{-\frac{r}{8L^2{\sigma}^2}}}{\sqrt{2}\sqrt{{\pi}}\left|{\sigma}\right|}
        + \dfrac{\mathrm{e}^{-\frac{r}{8L^2{\sigma}^2}}}{2^\frac{3}{2}\sqrt{{\pi}}\left|{\sigma}\right|}
        + \dfrac{\mathrm{e}^{-\frac{r}{8L^2{\sigma}^2}}}{2^\frac{3}{2}\sqrt{{\pi}}\left|{\sigma}\right|}                                                                                                                    \\
                                                             & = L\operatorname{erf}\left(\frac{r}{2^\frac{3}{2}L{\sigma}}\right) \ .
    \end{align*}
\end{proof}

\begin{remark}
    Jensen's inequality gives the following simple upper bound on $J(\sigma, L)$:
    \begin{align*}
        J(\sigma, L)
         & = \frac{1}{2\sigma^2 } \E_{z\sim {\cal N}(0,\sigma^2)}\left[ \min\left\{r|z|, 2Lz^2 \right\}\right]                            \\
         & \leq \min\left\{ \E_{z\sim {\cal N}(0,\sigma^2)}[r|z|/{2\sigma^2}], \E_{z\sim {\cal N}(0,\sigma^2)}[Lz^2 /{\sigma^2}] \right\}
        \\
         & \leq \min\left\{ \frac{r}{\sqrt{2\pi\sigma^2}}, L \right\} \ .
    \end{align*}
    Hence, $J(\sigma, L)$ is no worse than the Lipschitz constant of the original classifier $f$, or its Gaussian smoothed counterpart without the Lipschitz assumption, the latter bound is twice smaller than the previous original derivation in \cite[Appendix A]{salman2019provably}.
\end{remark}

\section{Proof of Proposition~\ref{prop:sigma_star_lip} optimal smoothing parameter for Lipschitz continuity}
\label{app:proof_sigma_star_lip}
\begin{proof}
    For \(\gamma \geq 0 \), we seek \(\sigma^*\) that maximizes the gap between the bounds of Eq.~(\ref{eq:bound_lip_sigma_element_wise}) with respect to $\sigma$:
    \begin{align*}
        \sigma^* = \argmax_{\sigma > 0}
        \left\{
        \min\left\{ \gamma, \frac{r}{\sqrt{2 \pi \sigma^2}}\right\}
        - \gamma\operatorname{erf}\left(\frac{r}{2^{3/2} \gamma \sigma}\right)
        \right\} \ .
    \end{align*}
    To find the value of \(\sigma^*\) that maximizes the given function, we'll determine the critical points.
    Let
    \[ g(\sigma) = \min\left\{ \gamma, \frac{r}{\sqrt{2 \pi \sigma^2}}\right\} - \gamma\operatorname{erf}\left(\frac{r}{2^{3/2}\gamma\sigma}\right) \ . \]

    Let's start by setting the two functions inside the \(\min\) function equal to each other and solving for \(\sigma\):
    \[ \gamma = \frac{r}{\sqrt{2 \pi \sigma^2}} \]
    \[ \Rightarrow \sigma^2 = \frac{r}{\gamma^2 2\pi} \]
    \[ \Rightarrow \sigma = \frac{r}{\gamma \sqrt{2\pi}} \ . \]
    This is the point of intersection, hence the value of \(\sigma\) where the two functions inside the \(\min\) change dominance. For that value,
    $g( \frac{r}{\gamma \sqrt{2\pi}}) = \gamma(1 - \operatorname{erf}(\frac{\sqrt{\pi}}{2}))$ .

    Now, for \(\sigma < \frac{r}{\gamma \sqrt{2\pi}}\), \(g(\sigma) = \gamma - \gamma\operatorname{erf}\left(\frac{r}{2^{3/2}\gamma\sigma}\right)\) .

    Let's differentiate \(g(\sigma)\) in the this first region:
    \begin{align*}
        g'(\sigma) & = 0 - \frac{d}{d\sigma} \left[ \gamma\operatorname{erf}\left(\frac{r}{2^{3/2}\gamma\sigma}\right) \right] \\
                   & = \gamma \frac{r}{2^{3/2}\gamma} \frac{2}{\sqrt{\pi}}
        \exp\left( -\left(\frac{r}{2^{3/2}\gamma\sigma} \right)^2 \right)                                                      \\
                   & = \frac{r}{\sqrt{2\pi}} \exp\left( -\left(\frac{r}{2^{3/2}\gamma\sigma} \right)^2 \right) \ .
    \end{align*}
    The supremum is obtained for $\sigma \to 0$, and the associated limit value for $g(\sigma)$ is $0$.

    For the second region, \(\sigma > \frac{r}{\gamma \sqrt{2\pi}}\), \(g(\sigma) = \frac{r}{\sqrt{2 \pi \sigma^2}} - \gamma\operatorname{erf}\left(\frac{r}{2^{3/2}\gamma\sigma}\right)\).
    Supremum is obtained for $\sigma  \to \frac{r}{\gamma \sqrt{2\pi}}$ as $g$ is a decreasing function of $\sigma$.
    The associated limit value for $g(\sigma)$ is $\gamma(1 - \operatorname{erf}(\frac{\sqrt{\pi}}{2}))$.

    Finally, taking $\gamma = L(s_k^r \circ f)$,  $\sigma^* = \frac{r}{L(s_k^r \circ f) \sqrt{2\pi}}$ gives maximum value for $g$ on all domain.

    We get a similar result $\sigma^* = \frac{r}{L(s^r \circ f) \sqrt{\pi}}$ for $L(s^r \circ f)$.
\end{proof}

\section{Proof of Theorem~\ref{thm:lip_quantile_smoothed_classifier_ours} Weierstrass transform with local Lipschitz continuity}
\label{app:sec:proof_lip_quantile_smoothed_classifier_ours}

We recall the Pontryagin's maximum principle (PMP).

\begin{theorem}[Pontryagin's Maximum Principle] \citep{pontryagin1962mathematical}
    Consider the optimal control problem:

    Minimize (or maximize) the cost functional
    \[
        J(u) = \int_{t_0}^{t_f} L(t, x(t), u(t)) \, dt + \Phi(x(t_f)),
    \]
    subject to the state (dynamical) equations
    \[
        \dot{x}(t) = f(t, x(t), u(t)), \quad x(t_0) = x_0,
    \]
    and the control constraints
    \[
        u(t) \in U \subset \mathbb{R}^m,
    \]
    where \( x(t) \in \mathbb{R}^n \), \( u(t) \in \mathbb{R}^m \), and \( t \in [t_0, t_f] \).

    Assume that \( L \), \( f \), and \( \Phi \) are continuously differentiable with respect to their arguments and that an optimal control \( u^*(t) \) and corresponding state trajectory \( x^*(t) \) exist.

    Then there exists an absolutely continuous costate (adjoint) function \( p(t) \in \mathbb{R}^n \) and a constant \( p_0 \leq 0 \) (for minimization) or \( p_0 \geq 0 \) (for maximization) such that the following conditions are satisfied almost everywhere on \( [t_0, t_f] \):

    1. State equation:
    \[
        \dot{x}^*(t) = f\big(t, x^*(t), u^*(t)\big).
    \]

    2. Costate equation:
    \[
        \dot{p}(t) = -\frac{\partial H}{\partial x}\big(t, x^*(t), u^*(t), p(t)\big),
    \]
    with terminal condition
    \[
        p(t_f) = \frac{\partial \Phi}{\partial x}\big(x^*(t_f)\big).
    \]
    The Hamiltonian \( H \) is defined by
    \[
        H(t, x, u, p) = L(t, x, u) + p^\top f(t, x, u).
    \]

    Maximum principle (optimality condition):
    \[
        H\big(t, x^*(t), u^*(t), p(t)\big) = \max_{v \in U} H\big(t, x^*(t), v, p(t)\big).
    \]

    Transversality conditions:
    If the final time \( t_f \) is free, then
    \[
        H\big(t_f, x^*(t_f), u^*(t_f), p(t_f)\big) = 0.
    \]

    Non-triviality condition:
    \[
        (p_0, p(t)) \neq 0 \quad \text{for all} \quad t \in [t_0, t_f].
    \]
\end{theorem}

We are going to prove the following lemma

\begin{lemma}
    \label{lemma:optimal_control_problem}
    Given the optimal control problem:

    Maximize the functional
    \[
        I(y) = \int_{-\infty}^{+\infty} s y(s) \phi(s) \, ds
    \]
    subject to the state equation
    \[
        \frac{d y(s)}{d s} = u(s),
    \]
    the control constraint
    \[
        |u(s)| \leq L,
    \]
    the state constraints
    \[
        0 \leq y(s) \leq 1,
    \]
    and the integral constraint
    \[
        \int_{-\infty}^{+\infty} y(s) \phi(s) \, ds = p,
    \]
    where \(\phi(s)\) is the standard normal probability density function (PDF),  \(\Phi(s)\) is the cumulative distribution function (CDF) and $p \in [0, 1]$.

    The optimal control \(u^*(s)\) and the optimal state \(y^*(s)\) are given by:

    When \( s \leq s_0 \):
    \[
        y^*(s) = 0, \quad u^*(s) = 0.
    \]

    When \( s_0 < s < s_1 \):
    \[
        y^*(s) = L (s - s_0), \quad u^*(s) = L.
    \]

    When \( s \geq s_1 \):
    \[
        y^*(s) = 1, \quad u^*(s) = 0.
    \]

    Here, \( s_0 \) is determined by the equation

    %
    \[
        p  - \left(1 - L \int_{s_0}^{s_0 + \frac{1}{L}} \Phi(s) \, ds \right) = 0.
    \]
    The optimal value of the objective functional \( I(y^*) \) is:
    \[
        I(y^*) = L [ \Phi(s_0 + \frac{1}{L}) - \Phi(s_0) ].
    \]
\end{lemma}

\begin{proof}
    To solve the optimal control problem, we apply the PMP with state constraints incorporated via Lagrange multipliers.
    To formulate the Hamiltonian with state constraints,
    we introduce Lagrange multipliers \(\mu_0(s) \geq 0\) and \(\mu_1(s) \geq 0\) for the state constraints \( y(s) \geq 0 \) and \( y(s) \leq 1 \), respectively.

    The augmented Hamiltonian is:
    \[
        H(s, y, p_y, u, \mu_0, \mu_1) = [s y - \lambda y] \phi(s) + p_y u + \mu_0 y + \mu_1 ( - y + 1 ),
    \]
    where \(\lambda\) is the Lagrange multiplier associated with the integral constraint.

    Necessary Conditions from the PMP are:

    State equation:
    \[
        \frac{d y(s)}{d s} = u(s).
    \]

    Costate equation:
    \[
        \frac{d p_y(s)}{d s} = -\frac{\partial H}{\partial y} = -[s - \lambda] \phi(s) - \mu_0(s) + \mu_1(s).
    \]

    Hamiltonian maximization condition:
    \[
        u^*(s) = \operatorname*{arg\,max}_{|u(s)| \leq L} [p_y(s) u(s)].
    \]

    Complementary slackness conditions:
    \begin{align*}
        \mu_0(s) \geq 0, \quad y(s) \geq 0, \quad \mu_0(s) y(s) = 0, \\
        \mu_1(s) \geq 0, \quad y(s) \leq 1, \quad \mu_1(s) [ y(s) - 1 ] = 0.
    \end{align*}

    We consider three regions based on the value of \( y(s) \):

    \paragraph{Region where \( y(s) = 0 \) (lower state constraint active):}

    Complementary slackness: \( \mu_0(s) \geq 0 \), \( \mu_1(s) = 0 \).

    Costate equation:
    \[
        \frac{d p_y(s)}{d s} = -[s - \lambda] \phi(s) - \mu_0(s).
    \]
    Admissible controls: \( u(s) \geq 0 \) (to prevent \( y(s) \) from decreasing below zero).

    Optimal control: Since \( p_y(s) \leq 0 \) (due to the effect of \( \mu_0(s) \)), the Hamiltonian is maximized by \( u^*(s) = 0 \).

    Resulting State: \( y(s) = 0 \).

    \paragraph{Region where \( 0 < y(s) < 1 \) (state constraints inactive):}

    Complementary slackness: \( \mu_0(s) = 0 \), \( \mu_1(s) = 0 \).

    Costate equation:
    \[
        \frac{d p_y(s)}{d s} = -[s - \lambda] \phi(s).
    \]

    Solution of costate equation:
    \[
        p_y(s) = -\int_s^{+\infty} [\sigma - \lambda] \phi(\sigma) \, d\sigma = \lambda Q(s) - \phi(s),
    \]
    where \( Q(s) = \int_s^{+\infty} \phi(\sigma) \, d\sigma \).

    The optimal control is:
    \[
        u^*(s) =
        \begin{cases}
            L,  & \text{if } p_y(s) > 0, \\
            -L, & \text{if } p_y(s) < 0.
        \end{cases}
    \]

    We determining the switching point \( s_0 \):
    \[
        p_y(s_0) = 0 \implies \lambda Q(s_0) = \phi(s_0).
    \]

    Since \( p_y(s) > 0 \) for \( s > s_0 \), the optimal control in this region is \( u^*(s) = L \).

    The resulting state is:
    \[
        y(s) = y(s_0) + L (s - s_0).
    \]

    \paragraph{Region where \( y(s) = 1 \) (upper state constraint active):}

    Complementary slackness: \( \mu_0(s) = 0 \), \( \mu_1(s) \geq 0 \).

    Costate equation:
    \[
        \frac{d p_y(s)}{d s} = -[s - \lambda] \phi(s) + \mu_1(s).
    \]

    Admissible controls: \( u(s) \leq 0 \) (to prevent \( y(s) \) from increasing above one).

    Optimal control: since \( p_y(s) \geq 0 \) (due to the effect of \( \mu_1(s) \)), the Hamiltonian is maximized by \( u^*(s) = 0 \).

    Resulting state: \( y(s) = 1 \).

    Determining \( s_0 \) and \( s_1 \):

    At \( s = s_0 \), \( y(s_0) = 0 \) (transition from \( y(s) = 0 \) to \( y(s) > 0 \)).
    At \( y(s_1) = 1 \), \( s_1 = s_0 + \frac{1}{L} \).

    Integral constraint:
    \[
        p = \int_{s_0}^{s_1} y(s) \phi(s) \, ds + \int_{s_1}^{+\infty} \phi(s) \, ds.
    \]

    Solving for \( \lambda \), \( s_0 \), and \( s_1 \) requires numerical methods due to the integrals involved.
    By incorporating the state constraints into the Hamiltonian via Lagrange multipliers and applying the necessary conditions from the PMP, we derive the optimal control and state trajectories that satisfy all constraints:

    Optimal Control \( u^*(s) \):
    \[
        u^*(s) =
        \begin{cases}
            0, & s \leq s_0,    \\
            L, & s_0 < s < s_1, \\
            0, & s \geq s_1.
        \end{cases}
    \]

    Optimal state \( y^*(s) \):
    \[
        y^*(s) =
        \begin{cases}
            0,           & s \leq s_0,    \\
            L (s - s_0), & s_0 < s < s_1, \\
            1,           & s \geq s_1.
        \end{cases}
    \]

    This solution ensures that the Hamiltonian is maximized over the admissible controls while satisfying the state and control constraints.

    We will derive the expressions for \( p \) and \( I(y^*) \) in terms of \( s_0 \), \( L \), and the standard normal CDF \( \Phi(s) \).

    \paragraph{Expression for \( p \):}

    The integral constraint is:
    \[
        p = \int_{-\infty}^{+\infty} y^*(s) \phi(s) \, ds.
    \]

    Using the structure of \( y^*(s) \):
    \[
        y^*(s) =
        \begin{cases}
            0,           & s \leq s_0,    \\
            L (s - s_0), & s_0 < s < s_1, \\
            1,           & s \geq s_1.
        \end{cases}
    \]

    We can split the integral into regions:
    \[
        p = \int_{s_0}^{s_1} y^*(s) \phi(s) \, ds + \int_{s_1}^{+\infty} y^*(s) \phi(s) \, ds.
    \]

    Compute each part:

    first integral (\( p_1 \)):
    \[
        p_1 = \int_{s_0}^{s_1} L (s - s_0) \phi(s) \, ds = L \int_{s_0}^{s_1} (s - s_0) \phi(s) \, ds.
    \]

    We perform integration by parts:

    Let \( u(s) = (s - s_0) \), \( dv(s) = \phi(s) \, ds \).

    Then \( du(s) = ds \), \( v(s) = \Phi(s) \).

    Integration by parts gives:
    \[
        \int_{s_0}^{s_1} (s - s_0) \phi(s) \, ds = \left[ (s - s_0) \Phi(s) \right]_{s_0}^{s_1} - \int_{s_0}^{s_1} \Phi(s) \, ds.
    \]

    Evaluating the boundaries:
    \[
        (s_1 - s_0) \Phi(s_1) - (s_0 - s_0) \Phi(s_0) = \frac{1}{L} \Phi(s_1).
    \]

    Therefore,
    \[
        \int_{s_0}^{s_1} (s - s_0) \phi(s) \, ds = \frac{1}{L} \Phi(s_1) - \int_{s_0}^{s_1} \Phi(s) \, ds.
    \]

    Multiply both sides by \( L \):
    \[
        p_1 = \Phi(s_1) - L \int_{s_0}^{s_1} \Phi(s) \, ds.
    \]

    Second integral (\( p_2 \)):
    \[
        p_2 = \int_{s_1}^{+\infty} 1 \cdot \phi(s) \, ds = 1 - \Phi(s_1).
    \]

    Total \( p \):
    \[
        p = p_1 + p_2 = \Phi(s_1) - L \int_{s_0}^{s_1} \Phi(s) \, ds + 1 - \Phi(s_1) = 1 - L \int_{s_0}^{s_1} \Phi(s) \, ds.
    \]

    \paragraph{Expression for \( I(y^*) \):}

    The objective functional evaluated at \( y^*(s) \) is:
    \[
        I(y^*) = \int_{-\infty}^{+\infty} s y^*(s) \phi(s) \, ds.
    \]

    Again, split the integral:
    \[
        I(y^*) = \int_{s_0}^{s_1} s y^*(s) \phi(s) \, ds + \int_{s_1}^{+\infty} s y^*(s) \phi(s) \, ds.
    \]

    Compute each part:

    First integral (\( I_1 \)):
    \[
        I_1 = \int_{s_0}^{s_1} s [ L (s - s_0) ] \phi(s) \, ds = L \int_{s_0}^{s_1} s (s - s_0) \phi(s) \, ds.
    \]

    Expand \( s (s - s_0) = s^2 - s_0 s \):
    \[
        I_1 = L \left( \int_{s_0}^{s_1} s^2 \phi(s) \, ds - s_0 \int_{s_0}^{s_1} s \phi(s) \, ds \right).
    \]

    Recall standard integrals:
    \begin{align*}
        \int s^2 \phi(s) \, ds & = -s \phi(s) + \Phi(s), \\
        \int s \phi(s) \, ds   & = -\phi(s).
    \end{align*}

    Evaluate the integrals:
    \begin{align*}
        \int_{s_0}^{s_1} s^2 \phi(s) \, ds & = [ -s \phi(s) + \Phi(s) ]_{s_0}^{s_1} = [ -s_1 \phi(s_1) + \Phi(s_1) ] - [ -s_0 \phi(s_0) + \Phi(s_0) ], \\
        \int_{s_0}^{s_1} s \phi(s) \, ds   & = [ -\phi(s) ]_{s_0}^{s_1} = -\phi(s_1) + \phi(s_0).
    \end{align*}

    Substitute back into \( I_1 \):
    \begin{align*}
        I_1 & = L \left( [ -s_1 \phi(s_1) + \Phi(s_1) ] - [ -s_0 \phi(s_0) + \Phi(s_0) ] - s_0 ( -\phi(s_1) + \phi(s_0) ) \right) \\
            & = L \left( \Phi(s_1) - \Phi(s_0) - s_1 \phi(s_1) + s_0 \phi(s_0) + s_0 \phi(s_1) - s_0 \phi(s_0) \right)            \\
            & = L \left( \Phi(s_1) - \Phi(s_0) - (s_1 - s_0) \phi(s_1) \right).
    \end{align*}

    Since \( s_1 - s_0 = \frac{1}{L} \):
    \[
        I_1 = L \left( \Phi(s_1) - \Phi(s_0) - \frac{1}{L} \phi(s_1) \right) = L [ \Phi(s_1) - \Phi(s_0) ] - \phi(s_1).
    \]

    Second integral (\( I_2 \)):
    \[
        I_2 = \int_{s_1}^{+\infty} s \cdot 1 \cdot \phi(s) \, ds.
    \]

    Recall that:
    \[
        \int_{s}^{+\infty} s \phi(s) \, ds = \phi(s).
    \]

    Therefore:
    \[
        I_2 = \phi(s_1).
    \]

    Total \( I(y^*) \):
    \[
        I(y^*) = I_1 + I_2 = \left( L [ \Phi(s_1) - \Phi(s_0) ] - \phi(s_1) \right) + \phi(s_1) = L [ \Phi(s_1) - \Phi(s_0) ].
    \]

    Thus, the optimal value of the objective functional is:
    \[
        I(y^*) = L [ \Phi(s_1) - \Phi(s_0) ].
    \]
\end{proof}

Now we can prove Theorem 2.

\begin{proof}

    Let us assume \( \sigma = 1 \). We start by expressing the gradient of \( \phi^{-1}(\tilde{F}(\vx)) \):
    \[
        \nabla \phi^{-1}(\tilde{F}(\vx)) = \frac{\nabla \tilde{F}(\vx)}{\phi'(\phi^{-1}(\tilde{F}(\vx)))} \ .
    \]

    We aim to show that for any unit vector \( \vu \), the following inequality holds:
    \[
        \vu^\top \nabla \tilde{F}(\vx) \leq L(F) \left[ \Phi\left(s_0 + \frac{1}{L(F)}\right) - \Phi(s_0) \right],
    \]
    where \( s_0 \) is determined by:
    \[
        \tilde{F}(\vx) = 1 - L(F) \int_{s_0}^{s_0 + \frac{1}{L(F)}} \Phi(s) \, ds \ .
    \]

    \textbf{Applying Stein's Lemma}

    Using Stein's lemma, we obtain the expression for \( \vu^\top \nabla \tilde{F}(\vx) \):
    \[
        \vu^\top \nabla \tilde{F}(\vx) = \mathbb{E}_{\delta \sim \mathcal{N}(0, \mI)} \left[ \vu^\top \delta F(\vx + \delta) \right],
    \]
    where \( \delta \) is a Gaussian vector with mean zero and identity covariance matrix. Our goal is to bound the maximum of this expression under the following constraints:

    \begin{itemize}
        \item \( 0 \leq F(\vx) \leq 1 \)
        \item \( \mathbb{E}_{\delta \sim \mathcal{N}(0, \mI)} \left[ F(\vx + \delta) \right] = p \)
        \item \( F \) is Lipschitz continuous with a Lipschitz constant \( L(F) \).
    \end{itemize}

    Let \( y(z) = F(z + \vx) \). Then the problem can be recast as:
    \[
        (P) \quad \max_{||\vu|| = 1} \max_{y} \quad I(y) = \int_{\mathbb{R}^d} \vu^\top s \, y(s) \, \phi(s) \, ds
    \]
    subject to:
    \[
        L(y) \leq L(F), \quad 0 \leq y(s) \leq 1, \quad \int_{\mathbb{R}^d} y(s) \phi(s) \, ds = p.
    \]

    Without loss of generality, we can take \( \vu = (1, 0, \ldots, 0)^\top \), as Gaussian vectors are rotationally invariant and rotation preserves the \( \ell^2 \)-norm.

    \textbf{Reducing to a One-Dimensional Problem}

    Decompose \( \vs \) as \( \vs = (s_1, \vt) \), where \( s_1 = \vu^\top \vs \) and \( \vt = (\vs_2, \ldots, \vs_d) \). The density function \( \phi(\vs) \) can be factorized as:
    \[
        \phi(\vs) = \phi_1(s_1) \phi_{d-1}(\vt),
    \]
    where \( \phi_1 \) and \( \phi_{d-1} \) are the density functions of a standard Gaussian in one dimension and \( d-1 \) dimensions, respectively.

    Rewrite the objective function as:
    \[
        I(y) = \int_{\mathbb{R}^{d-1}} \int_{\mathbb{R}} s_1 y(s_1, \vt) \phi_1(s_1) \phi_{d-1}(\vt) \, ds_1 \, d\vt.
    \]

    Define \( g(s_1) = \int_{\mathbb{R}^{d-1}} y(s_1, \vt) \phi_{d-1}(\vt) \, d\vt \). Then, the integral becomes:
    \[
        I(y) = I(g) = \int_{\mathbb{R}} s_1 g(s_1) \phi_1(s_1) \, ds_1.
    \]

    \textbf{Reformulating the Constraints for \( g(s_1) \)}

    The integral constraint on \( p \) directly translates to:
    \[
        \int_{\mathbb{R}} g(s_1) \phi_1(s_1) \, ds_1 = p.
    \]

    Since \( 0 \leq y(\vs) \leq 1 \) for all \( \vs \in \mathbb{R}^d \), it follows that \( 0 \leq g(s_1) \leq 1 \) for all \( s_1 \in \mathbb{R} \).

    For the Lipschitz condition, we have:
    \[
        |\nabla g(s_1)| \leq \int_{\mathbb{R}^{d-1}} |\nabla y(s_1, \vt)| \phi_{d-1}(\vt) \, d\vt \leq L(F).
    \]

    \textbf{Solving the One-Dimensional Problem}

    The reduced problem is now:
    \[
        (Q) \quad \max_{g} \quad I(g) = \int_{\mathbb{R}} s_1 g(s_1) \phi_1(s_1) \, ds_1
    \]
    subject to:
    \[
        L(g) \leq L(F), \quad 0 \leq g(s_1) \leq 1, \quad \int_{\mathbb{R}} g(s_1) \phi_1(s_1) \, ds_1 = p.
    \]

    Applying Lemma~\ref{lemma:optimal_control_problem}, the optimal solution \( g^*(s_1) \) is given by:
    \[
        g^*(s_1) =
        \begin{cases}
            0,                & s_1 \leq s_0,                     \\
            L(F) (s_1 - s_0), & s_0 < s_1 < s_0 + \frac{1}{L(F)}, \\
            1,                & s_1 \geq s_0 + \frac{1}{L(F)}.
        \end{cases}
    \]

    Here, \( s_0 \) is determined by the equation:
    \[
        \tilde{F}(\vx) = 1 - L(F) \int_{s_0}^{s_0 + \frac{1}{L(F)}} \Phi(s) \, ds.
    \]

    Evaluating \( I(g^*) \), we find:
    \[
        \vu^\top \nabla \tilde{F}(\vx) \leq L(F) \left[ \Phi\left(s_0 + \frac{1}{L(F)}\right) - \Phi(s_0) \right].
    \]

    For arbitrary \( \sigma \), the result follows by scaling the variables appropriately and performing a change of integration variable.

    Finally,
    \begin{align*}
         & L\left( \Phi^{-1} \circ \tilde{F}, B(\vx, \rho) \right)
        =
         & \sup_{\vx^\prime \in B(\vx, \rho)} \left\{ \frac{L(F) \left[ \Phi_\sigma\left(s_0(\vx^\prime) + \dfrac{1}{L(F)}\right) - \Phi_\sigma(s_0(\vx^\prime)) \right]}{
            \dfrac{1}{\sqrt{2\pi}} \exp\left( -\dfrac{1}{2} \left( \Phi^{-1}(\tilde{F}(\vx^\prime)) \right)^2 \right)
        } \right\} \
    \end{align*}

\end{proof}

%% file: content/appendix/appendix-robustness.tex

%% file: content/appendix/appendix-risk_management.tex
\chapter{Risk management}
\label{app:chapter:risk_management}

\section{Counterexample for Bonferroni correction}
\label{app:sec:counterexample_bonferroni_correction}

The Bonferroni correction is a method used to adjust the significance level of statistical tests when multiple comparisons are made simultaneously.
It helps control the family-wise error rate by reducing the risk of false positives due to multiple hypothesis testing.
This section provides a counterexample to illustrate the importance of properly applying the Bonferroni correction in statistical inference.
The corresponding code can be found in the following repository: \href{https://github.com/blaisedelattre/bridging_the_gap_rs/blob/main/counter_example.py}{\textcolor{blue}{counter example}}.
The code defines a function to compute the probability of an event under a multinomial distribution and demonstrates the effect of applying and not applying the Bonferroni correction.

\textbf{Proper Bonferroni Correction}: Adjusts the significance level \( \alpha \) by dividing it by the number of comparisons (e.g., \( \alpha' = \alpha / k \)) to control the family-wise error rate.

\textbf{Improper Application}: Using the unadjusted \( \alpha \) fails to account for multiple comparisons, leading to narrower confidence intervals and an increased risk of Type I errors.

When the Bonferroni correction is properly applied, the calculated probability of the event meets or exceeds the theoretical minimum probability (\( 1 - \alpha \)), maintaining the desired confidence level.
Without the correction, the probability of the event may fall below the theoretical minimum, indicating that the confidence intervals are too narrow and do not provide the intended level of confidence.
Properly adjusting for multiple comparisons is crucial for valid statistical inference, especially when making simultaneous inferences about multiple parameters.
Ensuring accurate confidence intervals helps maintain the integrity of research findings and prevents the reporting of false positives.

\section{Additional experiments on CPM}
\label{app:sec:additional_experiments_cpm}

We conduct additional experiments on certified prediction margin (CPM) using ResNet-110 for CIFAR-10 and ResNet-50 for ImageNet, both trained with noise injection. The training procedure follows the approach of Cohen et al. (2019), and all settings remain consistent with those described in Section 4.1 of the paper.

\begin{figure*}
    \centering
    \includegraphics[width=0.7\linewidth]{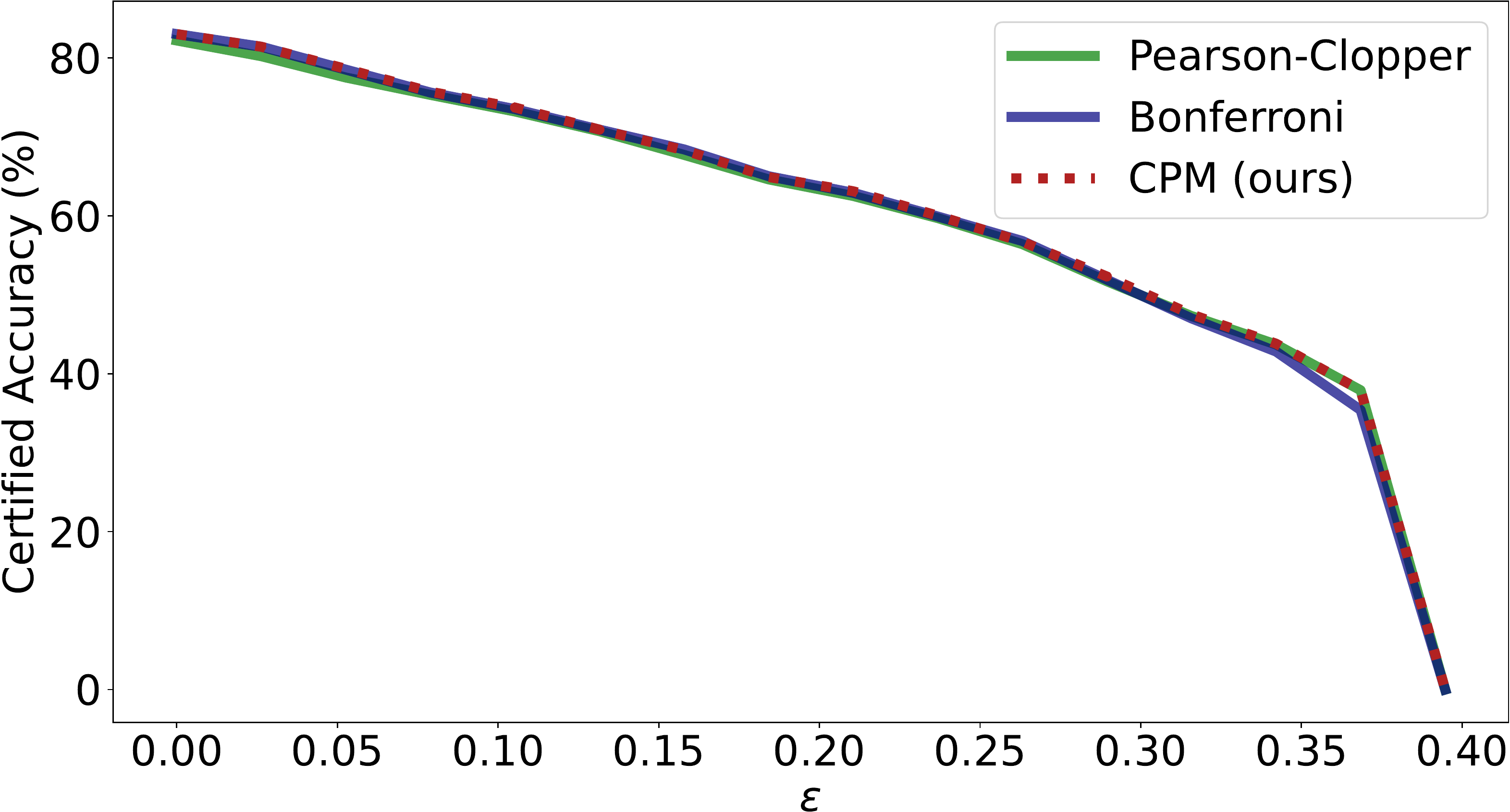}
    \caption{Comparison of various confidence interval methods for certified accuracy estimation with smoothing standard deviation $\sigma=0.12$ on the CIFAR-10 dataset.}
    \label{fig:cifar10_sigma_0.12}
\end{figure*}

\begin{figure*}
    \centering
    \includegraphics[width=0.7\linewidth]{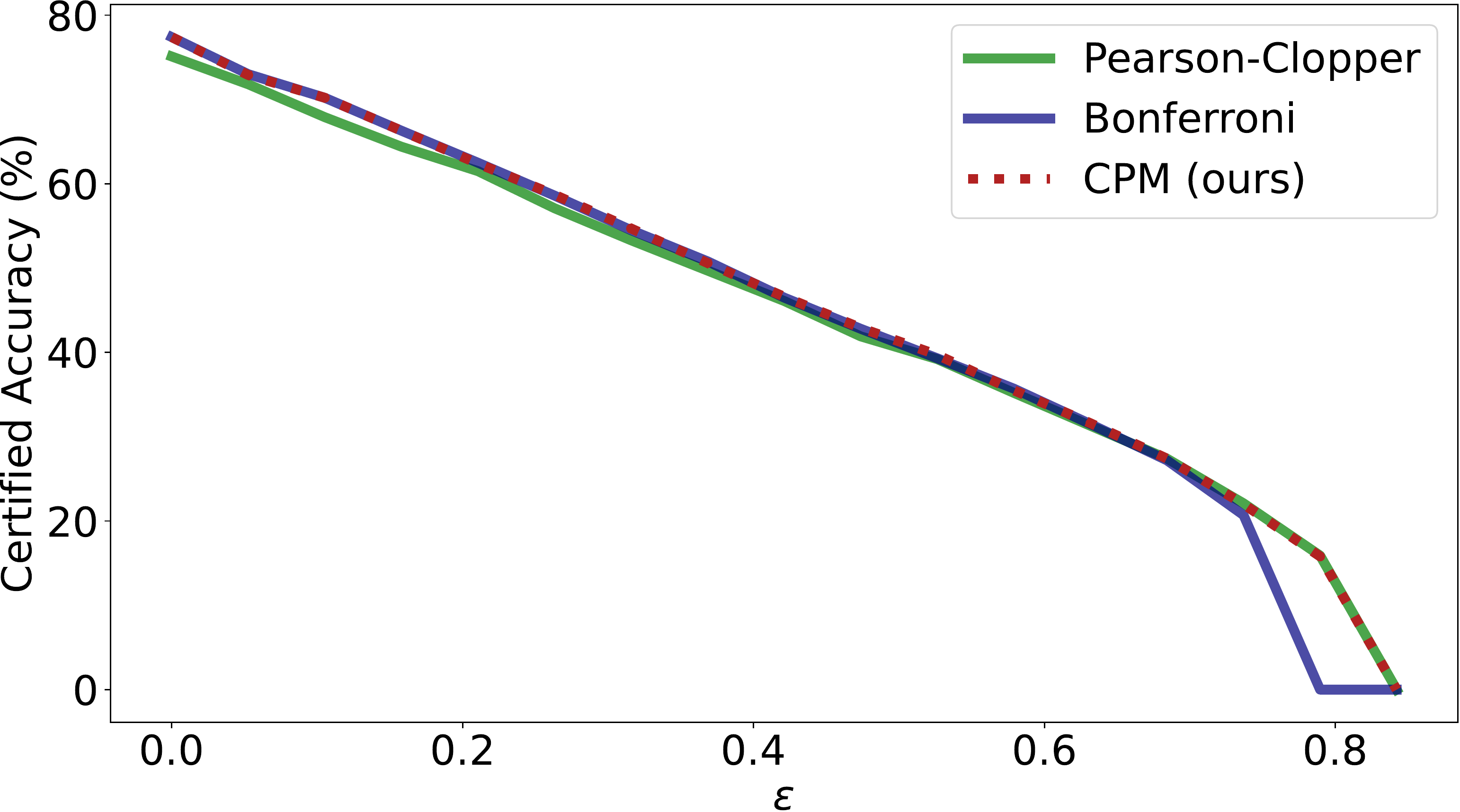}
    \caption{Comparison of various confidence interval methods for certified accuracy estimation with smoothing standard deviation $\sigma=0.25$ on the CIFAR-10 dataset.}
    \label{fig:cifar10_sigma_0.25}
\end{figure*}

\begin{figure*}
    \centering
    \includegraphics[width=0.7\linewidth]{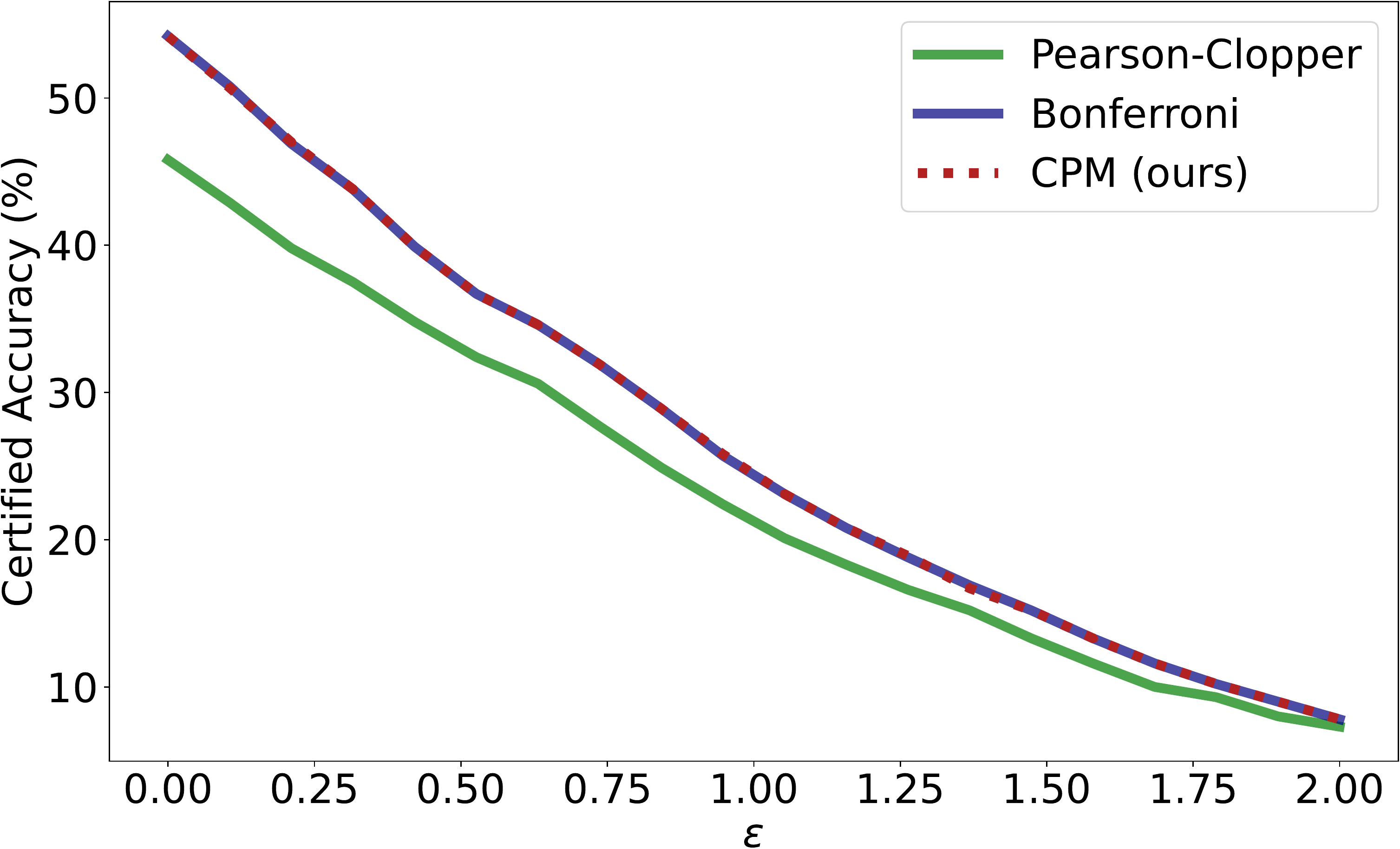}
    \caption{Comparison of various confidence interval methods for certified accuracy estimation with smoothing standard deviation $\sigma=1.0$ on the CIFAR-10 dataset.}
    \label{fig:cifar10_sigma_1.0}
\end{figure*}

For CIFAR-10, we report certified accuracy for ResNet-110, trained with different noise standard deviations (0.12, 0.25, 0.5, 1), comparing confidence intervals from Pearson-Clopper, Bonferroni, and CPM. The results are illustrated in Figure~\ref{fig:cifar10_sigma_0.12}, Figure~\ref{fig:cifar10_sigma_0.25}, Figure~\ref{fig:cifar10_sigma_0.5}, and Figure~\ref{fig:cifar10_sigma_1.0}.

\begin{figure*}
    \centering
    \includegraphics[width=0.7\linewidth]{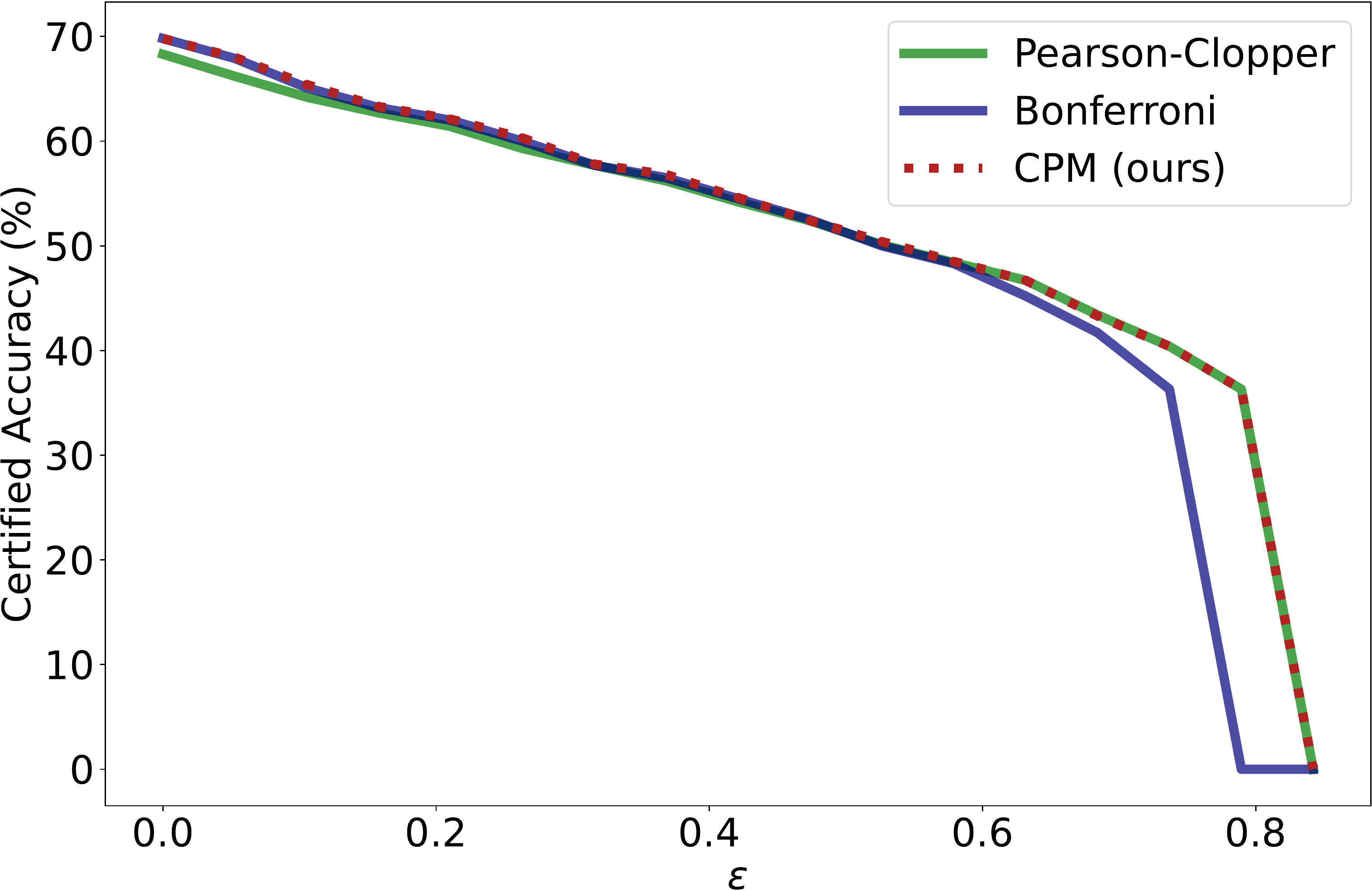}
    \caption{Comparison of various confidence interval methods for certified accuracy estimation with smoothing standard deviation $\sigma=0.25$ on the ImageNet dataset.}
    \label{fig:imagenet_sigma_0.25}
\end{figure*}

\begin{figure*}
    \centering
    \includegraphics[width=0.7\linewidth]{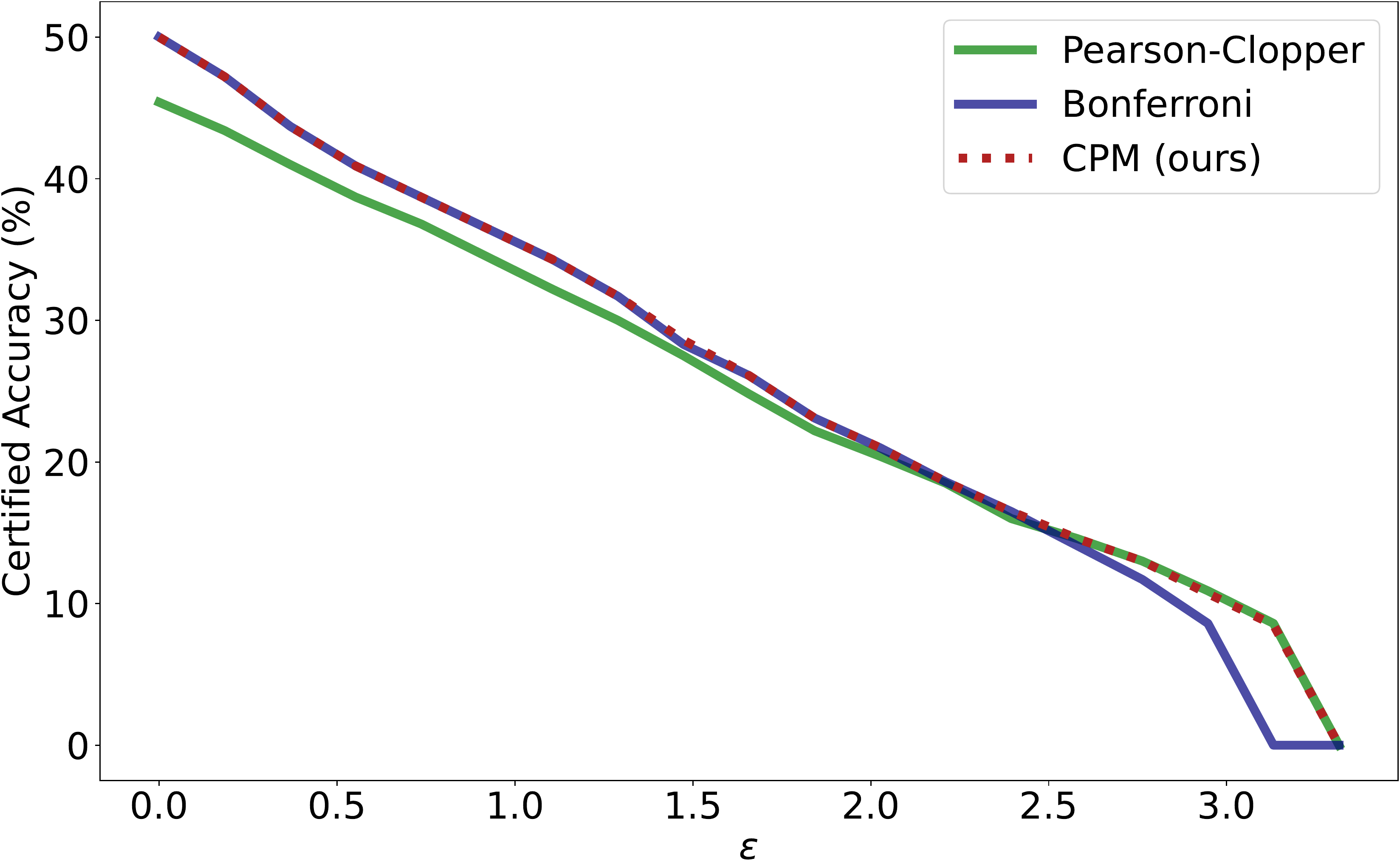}
    \caption{Comparison of various confidence interval methods for certified accuracy estimation with smoothing standard deviation $\sigma=1.0$ on the ImageNet dataset.}
    \label{fig:imagenet_sigma_1.0}
\end{figure*}

For ImageNet, we evaluate certified accuracy for ResNet-50, trained with different noise standard deviations(0.25, 1), using the same confidence intervals. The results are shown in Figure~\ref{fig:imagenet_sigma_0.25},and Figure~\ref{fig:imagenet_sigma_1.0}.
The figure for $\sigma=0.5$ is already presented in the main body.

The results indicate that for high variance in probability outputs, often associated with larger noise and higher entropy, Bonferroni performs better than Pearson-Clopper, while CPM achieves results comparable to Bonferroni. Conversely, when the probability outputs exhibit lower entropy, typically with smaller noise levels, Pearson-Clopper performs better, and CPM closely mimics Pearson-Clopper while also providing superior results in intermediate scenarios.

We observed slightly better gains on ImageNet, likely due to the larger number of classes, where CPM has a more significant impact, as Bonferroni tends to be overly conservative in such settings.



\section{Example on hardmax}
\label{ssec:example_var_argmax}
\begin{example}[Variance inflation by hard threshold]
    \label{example:variance_inflation}
    Let \(X_\epsilon\) be a random variable uniformly distributed on the interval
    \[
        \Bigl(\tfrac{1}{2}-\epsilon,\; \tfrac{1}{2}+\epsilon\Bigr),
        \quad\text{where }0 < \epsilon \le \tfrac{1}{2}.
    \]
    Define a new random variable
    \[
        Y_\epsilon
        \;=\; s\bigl(X_\epsilon\bigr)
        \;=\; \mathbbm{1}_{\bigl\{X_\epsilon > \tfrac{1}{2}\bigr\}}
        \quad
        \bigl(\text{hard threshold at } \tfrac{1}{2}\bigr).
    \]
    Then
    \[
        \V(X_\epsilon)
        \;=\; \frac{\epsilon^2}{3}
        \quad\text{while}\quad
        \V(Y_\epsilon)
        \;=\; \tfrac{1}{4}.
    \]
    Consequently, as \(\epsilon\to 0\),
    \(\V(X_\epsilon)\to 0\),
    but \(\V(Y_\epsilon)\) remains at \(\tfrac14\),
    showing how the hard threshold can inflate a small variance into its maximal Bernoulli value.
\end{example}

\begin{proof}
    \textbf{Step 1 (Mean and Variance of \(X_\epsilon\)):}
    By construction, \(X_\epsilon \sim \mathrm{Uniform}\bigl(\tfrac12-\epsilon,\;\tfrac12+\epsilon\bigr)\).
    The mean of a uniform random variable on \((a,b)\) is \(\tfrac{a+b}{2}\), so
    \[
        \mathbb{E}[X_\epsilon]
        = \frac{\bigl(\tfrac{1}{2}-\epsilon\bigr) + \bigl(\tfrac{1}{2}+\epsilon\bigr)}{2}
        = \frac{1}{2}.
    \]
    Its variance is \(\tfrac{(b-a)^2}{12}\); hence
    \[
        \V(X_\epsilon)
        = \frac{(2\,\epsilon)^2}{12}
        = \frac{\epsilon^2}{3}.
    \]

    \textbf{Step 2 (Distribution of \(Y_\epsilon\)):}
    The hard-threshold map
    \[
        s(x) =
        \begin{cases}
            1, & x > 1/2,  \\
            0, & x \le 1/2
        \end{cases}
    \]
    transforms \(X_\epsilon\) into a Bernoulli-type variable \(Y_\epsilon\).
    Because \(X_\epsilon\) is \emph{uniform} on \(\bigl(\tfrac12-\epsilon,\tfrac12+\epsilon\bigr)\), exactly half of that interval lies above \(1/2\). Consequently,
    \[
        \mathbb{P}\bigl(X_\epsilon > 1/2\bigr)
        = \frac{1}{2},
        \quad
        \mathbb{P}\bigl(X_\epsilon \le 1/2\bigr)
        = \frac{1}{2},
    \]
    so \(Y_\epsilon \sim \mathrm{Bernoulli}\bigl(1/2\bigr)\).

    \textbf{Step 3 (Mean and Variance of \(Y_\epsilon\)):}
    A \(\mathrm{Bernoulli}\bigl(1/2\bigr)\) random variable \(Y_\epsilon\) has:
    \[
        \mathbb{E}[Y_\epsilon]
        = \frac{1}{2},
        \quad
        \V(Y_\epsilon)
        = \frac{1}{2}\,\bigl(1 - \tfrac{1}{2}\bigr)
        = \frac{1}{4}.
    \]
    Therefore we obtain
    \[
        \V(X_\epsilon)
        = \frac{\epsilon^2}{3}
        \quad\text{and}\quad
        \V(Y_\epsilon)
        = \frac{1}{4}.
    \]
    As \(\epsilon\) shrinks, \(\V(X_\epsilon)\to 0\), yet \(\V(Y_\epsilon)\) remains at \(\tfrac{1}{4}\), which is the maximal value for a \(\mathrm{Bernoulli}(1/2)\).
\end{proof}

This example illustrates the key point: \emph{discontinuous transformations} (like a hard threshold) can magnify small input variations into large output variations, since the output ``jumps'' from 0 to 1. Such a function can thus inflate the variance of the resulting variable significantly---even to its maximal Bernoulli value \(1/4\) when the mean is \(1/2\).



\newpage 

\section{Ablation study in Lipschitz variance margin trade-off}
\label{appendix:ablation_study}
\begin{figure}[h]
    \centering
    \includegraphics[width=0.8\linewidth]{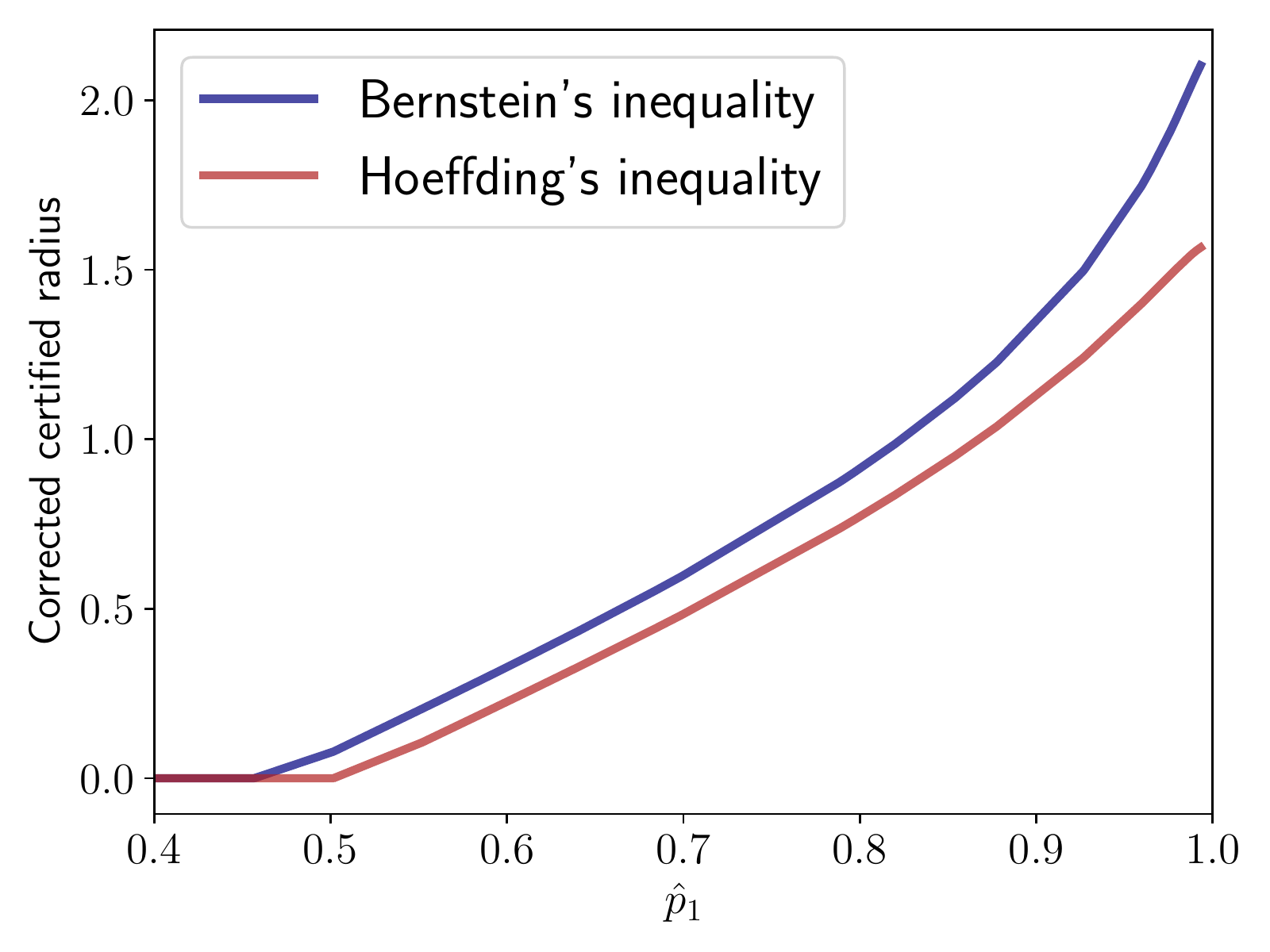}
    \caption{Comparison between corrected certified radii $R_2(\bar{\vp})$ produced by Bernstein's and Hoeffding's inequalities, for a random subset of $1000$ images of ImageNet dataset using RS with a smoothing noise $\sigma=1.0$. We use the ViT-denoiser baseline from~\cite{carlini2023certified}.
    }
    \label{fig:comparison_concentration}
\end{figure}

\begin{figure}[h]
    \centering
    \includegraphics[width=0.8\textwidth]{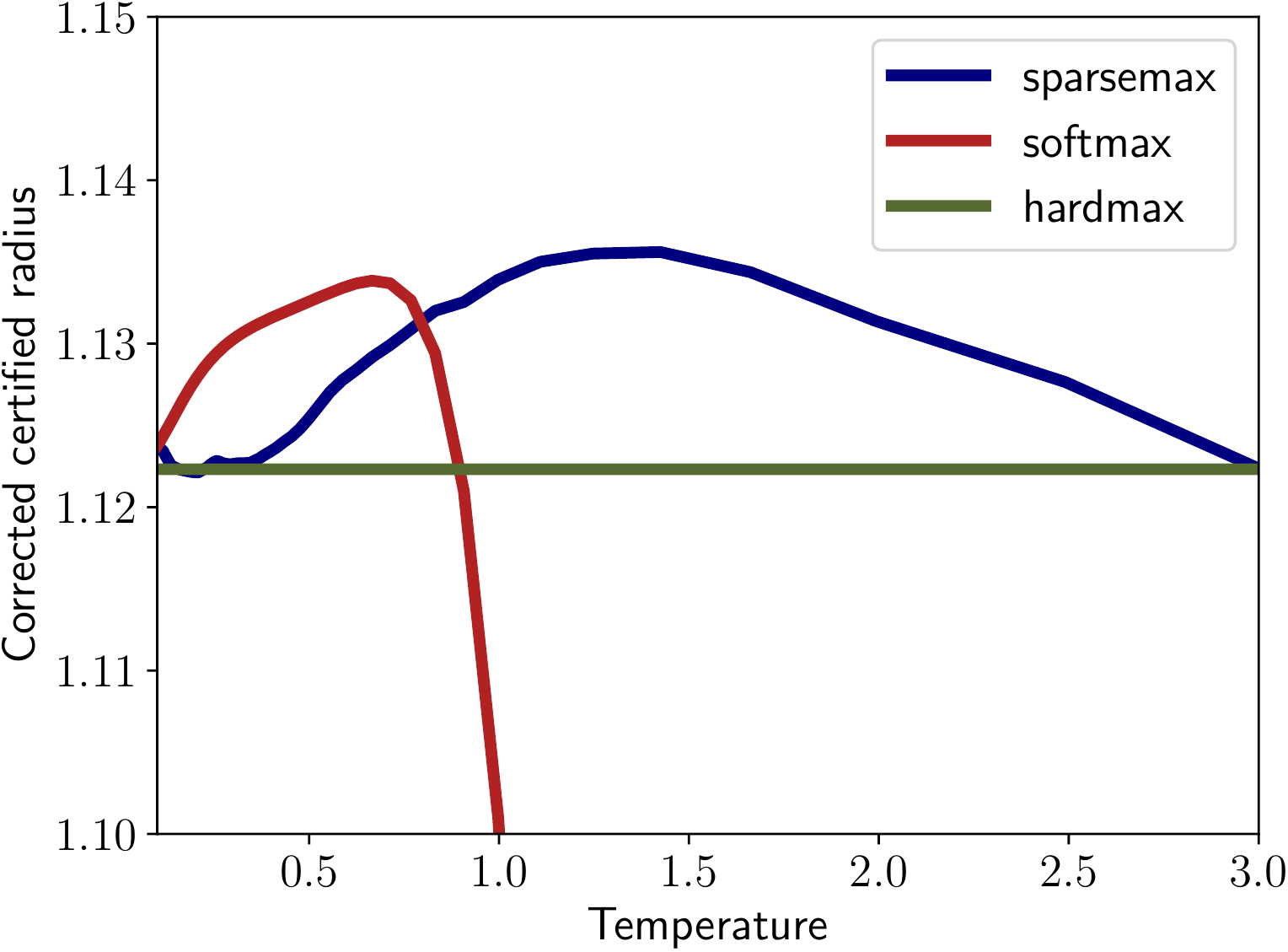}
    \caption{Comparison of the effect on corrected certified radii $\Rmult(\underline{\mathbf{p}_{I_1}}, \overline{\mathbf{p}_{I_2}})$ of the choice of the simplex map $\tau$ and associated temperature $T$. Simplex maps considered are $s \in \{\mathrm{sparsemax}, \mathrm{softmax}, \mathrm{hardmax}\}$.
        The base network $f$ is the one from~\cite{carlini2023certified} and the corrected certified radii were generated with one image from ImageNet with smoothing variance $\sigma=1.0$.
        Radii are risk corrected with Empirical Bernstein inequality for a risk $\alpha=1\mathrm{e-}3$ and $n=10^4$.
        We see that by varying the temperature $T$, $\mathrm{softmax}$ and $\mathrm{sparsemax}$ can find a better solution than $\mathrm{hardmax}$ to the variance-margin trade-off.
    }
    \label{fig:comparison_projection_simplex}
\end{figure}

This ablation study provides two comparisons:
\begin{itemize}
    \item A comparison between corrected certified radii produced by Hoeffding's and Bernstein's inequalities in Fig.~\ref{fig:comparison_concentration}. The Clopper-Pearson is not included as it is only applicable to binomial values.
    \item A comparison between corrected certified radii produced by different simplex maps and temperatures in Fig.\ref{fig:comparison_projection_simplex}.
\end{itemize}

\section{LVM-RS additional experiments~\ref{ssec:expe_lvm_rs}}
\label{appendix:expe_lvm_rs}
\begin{figure}[h]
    \centering
    \includegraphics[width=1.0\textwidth]{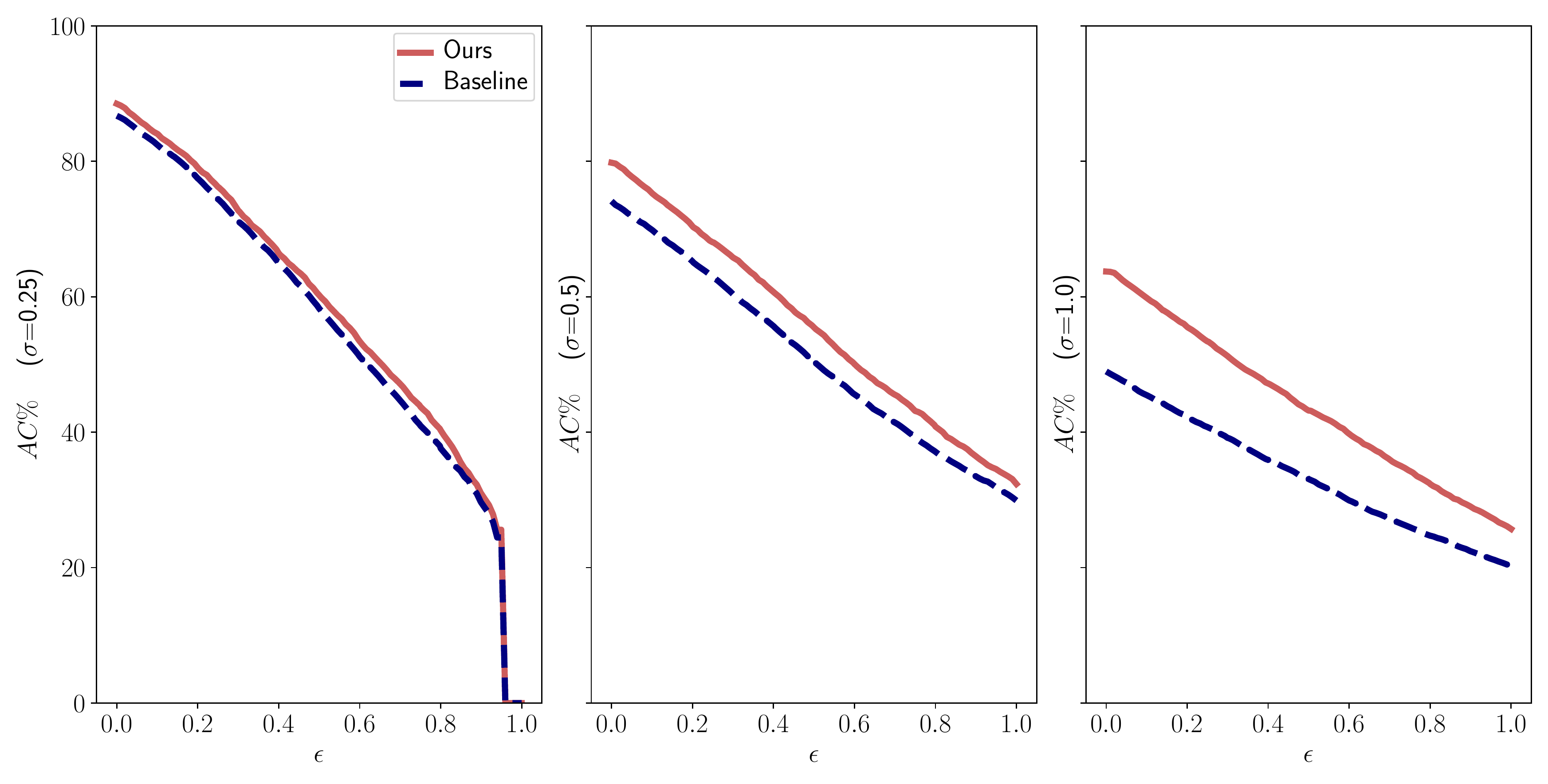}
    \caption{Certified accuracies ($CA$ in $\%$) in function of level of perturbations $\epsilon$ on CIFAR-10, for different noise levels $\sigma \in \{0.25, 0.5, 1\}$. Number of samples is $n=10^5$ and risk $\alpha = 1e\text{-}3$.
        Our method is compared to the baseline chosen as in~\cite{carlini2023certified}.}
    \label{fig:certified_accuracy_for_different_eps_cifar10}
\end{figure}

\begin{figure}[h]
    \centering
    \includegraphics[width=1.0\textwidth]{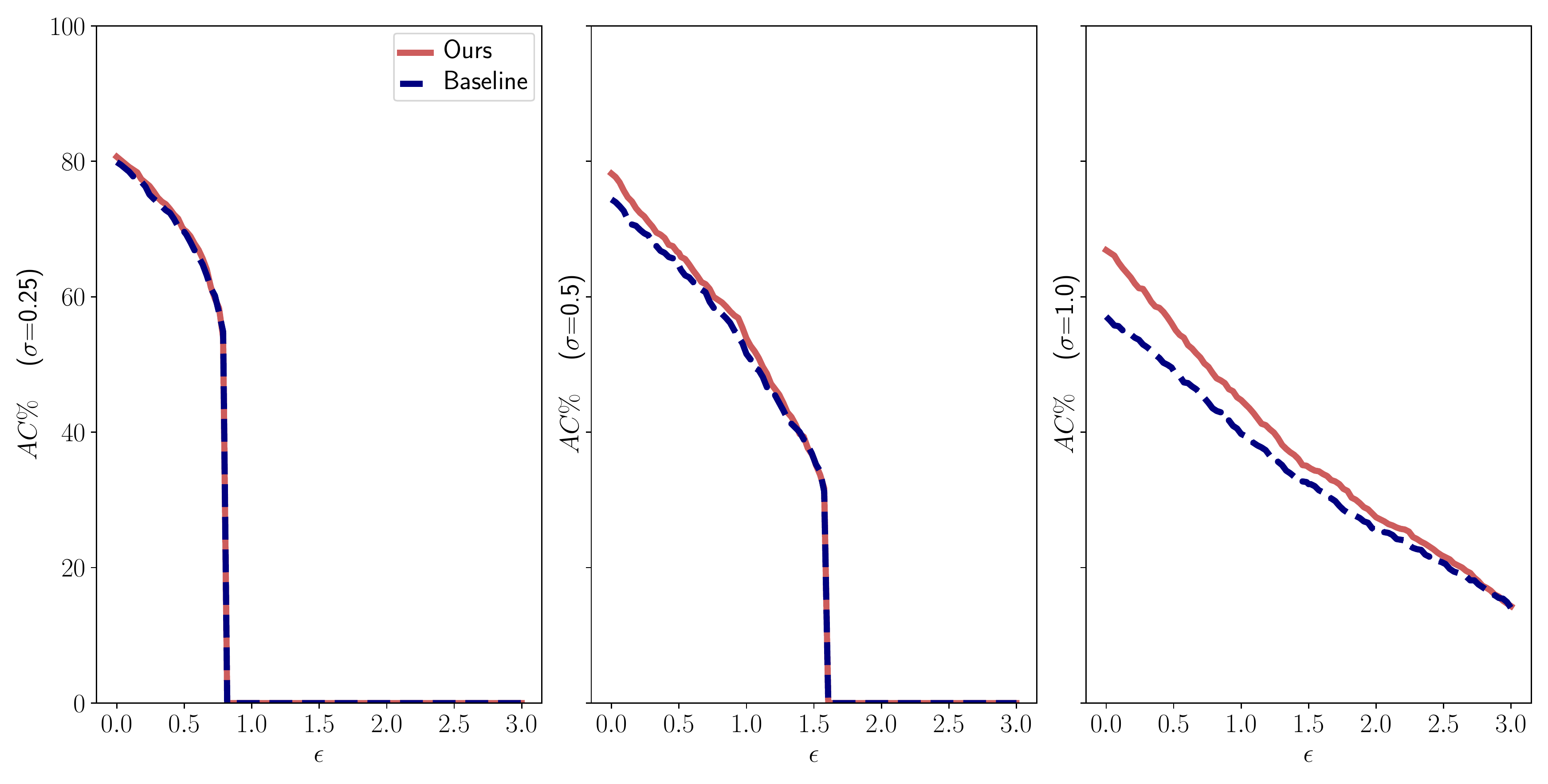}
    \caption{Certified accuracies ($CA$ in $\%$) in function of level of perturbations $\epsilon$ on ImageNet, for different noise levels $\sigma \in \{0.25, 0.5, 1\}$. Number of samples is $n=10^4$ and risk $\alpha = 1e\text{-}3$.
        Our method is compared to the baseline chosen as in~\cite{carlini2023certified}.}
    \label{fig:certified_accuracy_for_different_eps_imagenet}
\end{figure}

\begin{table}[h]
    \caption{Certified accuracies comparison for different perturbation $\epsilon$ values, for $n=10^4$ samples and $\alpha=1\mathrm{e-}3$. On ImageNet dataset. Here the baseline is a ResNet-150 from~\cite{salman2019provably}.}
    \centering
    \begin{tabular}{lcccccccc}
        \toprule
        \multirow{2}[2]{*}{\textbf{Methods}} & \multicolumn{8}{c}{{\boldmath{}\textbf{Certified accuracy ($\varepsilon$)}\unboldmath{}}}                                                                                                             \\
        \cmidrule{2-9}
                                             & 0.14                                                                                      & 0.2            & 0.3            & 0.4            & 0.5            & 0.6            & 0.7            & 0.8 \\
        \midrule
        {\cite{salman2019provably}}          & 74.49                                                                                     & 73.08          & 69.84          & 66.41          & 62.42          & 57.75          & 51.24          & 0.0 \\
        \textbf{LVM-RS (ours)}               & \textbf{76.77}                                                                            & \textbf{74.99} & \textbf{71.26} & \textbf{67.55} & \textbf{63.43} & \textbf{58.59} & \textbf{51.39} & 0.0 \\
        \bottomrule
    \end{tabular}
    \label{tab:accuracy_comparison_transposed}
\end{table}

\begin{table}[h]
    \label{tab:ca_sigma_025_cifar10}
    \caption{Certified accuracy for \(\sigma = 0.25\) on CIFAR-10, for risk $\alpha=1\mathrm{e-}3$ and $n=10^5$ samples.}
    \centering
    \begin{tabular}{lcccccccccccc}
        \toprule
        \multirow{2}[2]{*}{\textbf{Methods}} & \multicolumn{11}{c}{{\boldmath{}\textbf{Certified accuracy ($\varepsilon$)}\unboldmath{}}}                                                                                                                                                                \\
        \cmidrule{2-12}
                                             & 0.0                                                                                        & 0.14           & 0.2            & 0.25           & 0.3            & 0.4            & 0.5            & 0.6            & 0.75           & 0.8            & 1.0 \\
        \midrule
        \cite{carlini2023certified}          & 86.72                                                                                      & 80.73          & 77.47          & 74.41          & 71.15          & 65.01          & 58.25          & 51.15          & 40.96          & 37.6           & 0.0 \\
        \textbf{LVM-RS (ours)}               & \textbf{88.49}                                                                             & \textbf{82.15} & \textbf{79.06} & \textbf{76.21} & \textbf{72.73} & \textbf{66.41} & \textbf{60.22} & \textbf{53.41} & \textbf{43.76} & \textbf{40.27}
                                             & 0.0                                                                                                                                                                                                                                                       \\
        \bottomrule
    \end{tabular}
\end{table}

\begin{table}[h]
    \label{tab:ca_sigma_05_cifar10}
    \caption{Certified accuracy for \(\sigma = 0.5\) on CIFAR-10, for risk $\alpha=1\mathrm{e-}3$ and $n=10^5$ samples.}
    \centering
    \begin{tabular}{lcccccccccccc}
        \toprule
        \multirow{2}[2]{*}{\textbf{Methods}} & \multicolumn{11}{c}{{\boldmath{}\textbf{Certified accuracy ($\varepsilon$)}\unboldmath{}}}                                                                                                                                                                          \\
        \cmidrule{2-12}
                                             & 0.0                                                                                        & 0.14           & 0.2            & 0.25           & 0.3           & 0.4            & 0.5            & 0.6            & 0.75           & 0.8            & 1.0            \\
        \midrule
        \cite{carlini2023certified}          & 74.11                                                                                      & 67.99          & 65.22          & 62.89          & 60.38         & 55.67          & 50.43          & 45.59          & 39.26          & 37.11          & 29.91          \\
        \textbf{LVM-RS (ours)}               & \textbf{79.79}                                                                             & \textbf{73.45} & \textbf{70.41} & \textbf{68.04} & \textbf{65.8} & \textbf{60.71} & \textbf{55.48} & \textbf{50.07} & \textbf{43.13} & \textbf{40.83} & \textbf{32.35} \\
        \bottomrule
    \end{tabular}
\end{table}

\begin{table}[h]
    \label{tab:ca_sigma_1_cifar10}
    \caption{Certified accuracy for \(\sigma = 1\) on CIFAR-10, for risk $\alpha=1\mathrm{e-}3$ and $n=10^5$ samples.}
    \centering
    \begin{tabular}{lcccccccccccc}
        \toprule
        \multirow{2}[2]{*}{\textbf{Methods}} & \multicolumn{11}{c}{{\boldmath{}\textbf{Certified accuracy ($\varepsilon$)}\unboldmath{}}}                                                                                                                                                                          \\
        \cmidrule{2-12}
                                             & 0.0                                                                                        & 0.14           & 0.2            & 0.25          & 0.3            & 0.4            & 0.5            & 0.6            & 0.75           & 0.8            & 1.0            \\
        \midrule
        \cite{carlini2023certified}          & 48.97                                                                                      & 44.24          & 42.26          & 40.76         & 39.15          & 35.91          & 33.08          & 29.92          & 25.97          & 24.72          & 20.09          \\
        \textbf{LVM-RS (ours)}               & \textbf{63.72}                                                                             & \textbf{57.99} & \textbf{55.54} & \textbf{53.4} & \textbf{51.23} & \textbf{47.19} & \textbf{43.19} & \textbf{39.76} & \textbf{34.27} & \textbf{32.35} & \textbf{25.71} \\
        \bottomrule
    \end{tabular}
\end{table}

\begin{table}[h]
    \label{tab:ca_sigma_025_imagnet}
    \caption{Certified accuracy for \(\sigma = 0.25\) on ImageNet, for risk $\alpha=1\mathrm{e-}3$ and $n=10^4$ samples.}
    \centering
    \begin{tabular}{lccccccc}
        \toprule
        \multirow{2}[2]{*}{\textbf{Methods}} & \multicolumn{6}{c}{{\boldmath{}\textbf{Certified accuracy ($\varepsilon$)}\unboldmath{}}}                                          \\
        \cmidrule{2-7}
                                             & 0.0                                                                                       & 0.5            & 1.0 & 1.5 & 2   & 3   \\
        \midrule
        \cite{carlini2023certified}          & 79.88                                                                                     & 69.57          & 0.0 & 0.0 & 0.0 & 0.0 \\
        \textbf{LVM-RS (ours)}               & \textbf{80.66}                                                                            & \textbf{69.84} & 0.0 & 0.0 & 0.0 & 0.0 \\
        \bottomrule
    \end{tabular}
\end{table}

\begin{table}[h]
    \label{tab:ca_sigma_05_imagnet}
    \caption{Certified accuracy for \(\sigma = 0.5\) on ImageNet, for risk $\alpha=1\mathrm{e-}3$ and $n=10^4$ samples.}
    \centering
    \begin{tabular}{lccccccc}
        \toprule
        \multirow{2}[2]{*}{\textbf{Methods}} & \multicolumn{6}{c}{{\boldmath{}\textbf{Certified accuracy ($\varepsilon$)}\unboldmath{}}}                                                                \\
        \cmidrule{2-7}
                                             & 0.0                                                                                       & 0.5            & 1.0            & 1.5            & 2   & 3   \\
        \midrule
        \cite{carlini2023certified}          & 74.37                                                                                     & 64.56          & 51.55          & \textbf{36.04} & 0.0 & 0.0 \\
        \textbf{LVM-RS (ours)}               & \textbf{78.18}                                                                            & \textbf{66.47} & \textbf{53.85} & \textbf{36.04} & 0.0 & 0.0 \\
        \bottomrule
    \end{tabular}
\end{table}

\begin{table}[h]
    \label{tab:ca_sigma_1_imagnet}
    \caption{Certified accuracy for \(\sigma = 1\) on ImageNet, for risk $\alpha=1\mathrm{e-}3$ and $n=10^4$ samples.}
    \centering
    \begin{tabular}{lccccccc}
        \toprule
        \multirow{2}[2]{*}{\textbf{Methods}} & \multicolumn{6}{c}{{\boldmath{}\textbf{Certified accuracy ($\varepsilon$)}\unboldmath{}}}                                                                                      \\
        \cmidrule{2-7}
                                             & 0.0                                                                                       & 0.5            & 1.0            & 1.5            & 2              & 3              \\
        \midrule
        \cite{carlini2023certified}          & 57.06                                                                                     & 49.05          & 39.74          & 32.33          & 25.53          & 14.01          \\
        \textbf{LVM-RS (ours)}               & \textbf{66.87}                                                                            & \textbf{55.56} & \textbf{44.74} & \textbf{34.83} & \textbf{27.43} & \textbf{14.31} \\
        \bottomrule
    \end{tabular}
\end{table}

%% file: content/appendix/appendix-regularization.tex
\chapter{Regularization}\label{app:chapter:regularization}

\section{Additional experiments}

\subsection{Activation Decay Faithfully Approximates Monte Carlo Smoothing of the Last Layer}
\label{sec:mc_vs_ad}

Monte Carlo smoothing offers a way to flatten sharp minima by averaging the loss over random perturbations.
However, it requires multiple stochastic forward-backward passes, incurring high computational cost.
Activation Decay provides a deterministic and efficient surrogate based on a bound (obtained with Jensen's inequality) of the smoothed loss restricted to the final layer.

We train a 3-layer MLP with 512 hidden units and GELU activations on CIFAR-10 for 10 epochs using SGD (learning rate 0.1, cosine annealing, no momentum, no weight decay) and batch size 128.
At each iteration, for the same mini-batch, we compute:

\begin{itemize}
    \item the \emph{Monte Carlo smoothed loss}:
          \[
              \mathcal{L}^\sigma \left(\mW^{(L)} \vh^{(L-1)}, \vy\right)
              = \mathbb{E}_{\mathbf{\Delta} \sim \mathcal{N}(0, \sigma^2 \mI)}
              \left[
                  \mathcal{L} \left((\mW^{(L)} + \mathbf{\Delta}) \vh^{(L-1)}, \vy \right)
                  \right]
          \]
          estimated via $n_{\text{MC}} = 100$ samples, along with its gradient
          \( \nabla_{\mW^{(L)}} \mathcal{L}^\sigma \);

    \item the \emph{Activation Decay surrogate loss}:
          \[
              \tilde{\mathcal{L}}\left(\mW^{(L)} \vh^{(L-1)}, \vy\right)
              = \mathcal{L}\left(\mW^{(L)} \vh^{(L-1)}, \vy\right)
              + \frac{\sigma^2}{2} \left\| \vh^{(L-1)} \right\|^2
          \]
          with corresponding gradient \( \nabla_{\mW^{(L)}} \tilde{\mathcal{L}} \).
\end{itemize}

For each value of $\sigma \in \{0.05, 0.10, \dots, 1.00\}$, we log the following diagnostics, averaged over each epoch: cosine similarity between the two gradients, norm ratio \( \|\nabla_{\text{MC}}\|_2 / \|\nabla_{\text{AD}}\|_2 \), and the loss gap \( \mathcal{L}^\sigma - \tilde{\mathcal{L}} \).

\begin{table}[h]
    \centering
    \small
    \begin{tabular}{cccc}
        \toprule
        $\sigma$ & Cosine $\uparrow$       & Norm ratio                                      & Loss gap                                   \\[-0.2ex]
                 & $(\text{MC},\text{AD})$ & $\|\nabla_{\text{MC}}\|/\|\nabla_{\text{AD}}\|$ & $\mathcal{L}^\sigma - \tilde{\mathcal{L}}$ \\
        \midrule
        0.05     & 0.916                   & 1.085                                           & \phantom{$-$}0.021                         \\
        0.10     & 0.938                   & 1.065                                           & \phantom{$-$}0.009                         \\
        0.15     & 0.939                   & 1.064                                           & \phantom{$-$}0.004                         \\
        0.20     & 0.941                   & 1.062                                           & $-0.001$                                   \\
        0.25     & 0.938                   & 1.068                                           & $-0.003$                                   \\
        0.30     & 0.939                   & 1.068                                           & $-0.006$                                   \\
        0.35     & 0.935                   & 1.071                                           & $-0.007$                                   \\
        0.40     & 0.931                   & 1.072                                           & $-0.007$                                   \\
        0.50     & 0.933                   & 1.078                                           & $-0.009$                                   \\
        1.00     & 0.936                   & 1.075                                           & $-0.009$                                   \\
        \bottomrule
    \end{tabular}
    \caption{Comparison between Monte Carlo smoothing and Activation Decay surrogate. Values are averaged over ten training epochs. Cosine similarity close to 1 and near-unit norm ratios confirm gradient alignment; loss gaps remain below $1\%$ of total loss.}
    \label{tab:mc_vs_ad}
\end{table}

The empirical results, see Table~\ref{tab:mc_vs_ad}, confirm the tight match between the gradients and loss values of the smoothed objective and its Activation Decay surrogate across a wide range of $\sigma$.
Even at higher smoothing levels, gradients remain highly aligned (cosine $\geq 0.93$), and loss discrepancies are negligible.
This validates Activation Decay as an efficient and reliable approximation to Monte Carlo smoothing for last-layer perturbations.

\subsection{Classification with MLP on CIFAR-10}
\label{appendix:exp:baek_comparison}
In this experiment we use the same setting as in experiment~\ref{exp:mlp_classif_cifar10}.
We conducted a comparison of our method AD, with the approach outlined in~\citep{baek2024why} Section~4.3, whose  framework, designed specifically for label noise robustness, requires tuning decay parameters for each layer to achieve effective regularization. This introduces additional hyperparameter complexity but shares conceptual similarities with our approach.

To ensure a fair comparison, both methods were evaluated under identical experimental settings, including hyperparameter tuning for ~\cite{baek2024why}'s regularization coefficients. The results are summarized in Table~\ref{tab:comparison}.

\begin{table}[h!]
    \centering
    \caption{Comparison with~\citep{baek2024why} and Activation Decay (AD) under identical experimental settings.}
    \label{tab:comparison}
    \begin{tabular}{lcc}
        \toprule
        \textbf{Metric}          & \textbf{\citep{baek2024why}} & \textbf{AD (ours)} \\
        \midrule
        Mean test accuracy (\%)  & 65.05                        & 65.11              \\
        95\% confidence interval & [64.86, 65.23]               & [64.90, 65.33]     \\
        \bottomrule
    \end{tabular}
\end{table}

Our findings demonstrate that AD, which applies $\ell_2$-regularization to the penultimate activations, achieves comparable results to~\citep{baek2024why}'s method without the need for layer-wise tuning. Specifically, ~\citep{baek2024why}'s method requires tuning separate coefficients for intermediate layers ($\sigma = 1\mathrm{e}{-3}$) and for the last layer parameters ($\sigma = 1\mathrm{e}{-2}$). In contrast, our approach eliminates this complexity and achieves similar performance with a single hyperparameter, $\sigma$, set to $0.1$.

The simplicity of our method reduces the number of hyperparameters to tune, making it both more practical and easier to analyze. Furthermore, regularizing the penultimate activation indirectly regularizes preceding layers, as the penultimate activation encapsulates their contributions.

\subsection{Classification with MLP-Mixer on ImageNet-1k}
\begin{figure}[h]
    \centering
    \includegraphics[width=0.7\linewidth]{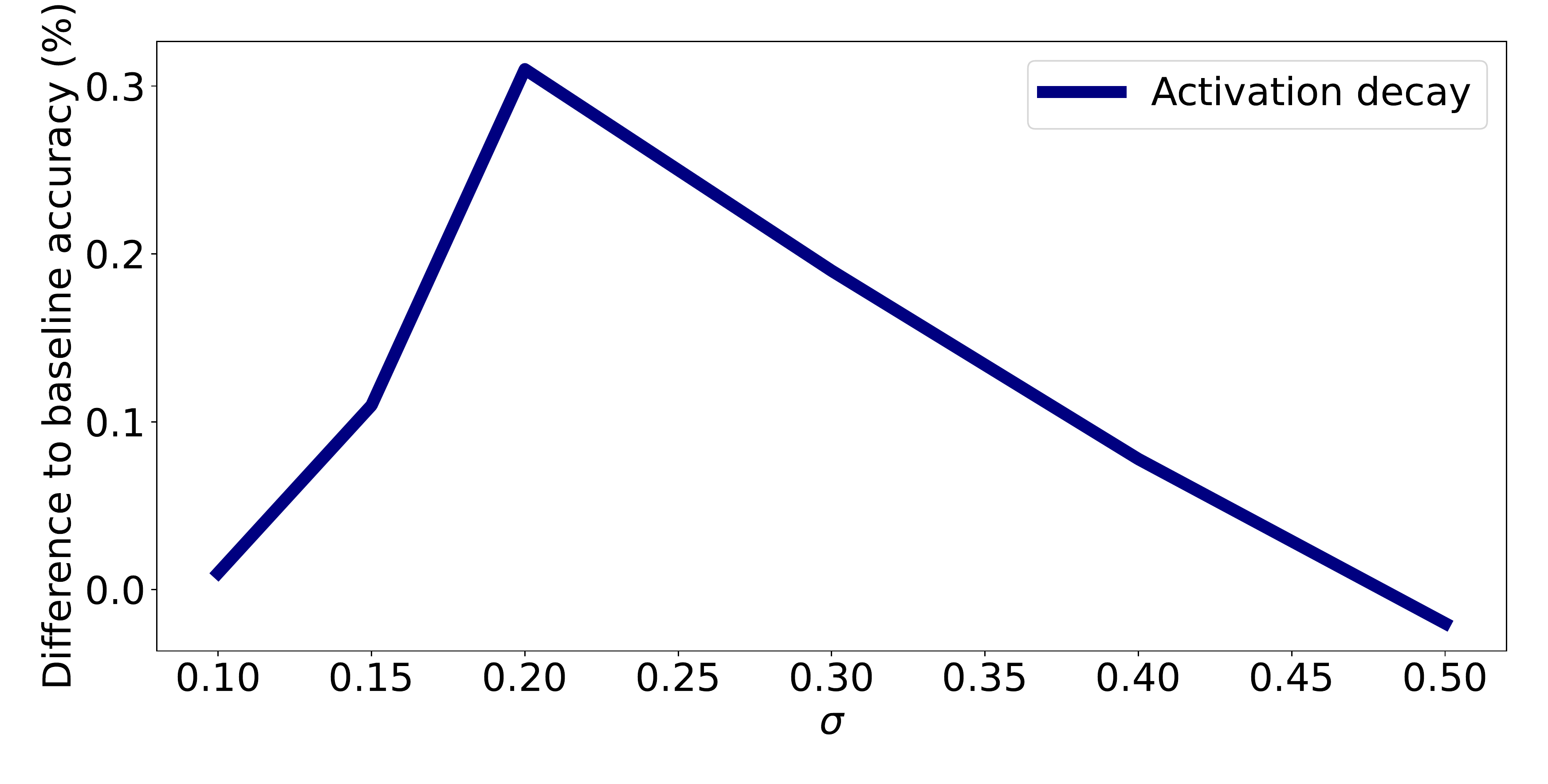}
    \caption{Plot of different pieces of training on ImageNet of MLP-Mixer with AD, with varying $\sigma$.}
    \label{fig:mlp_mixer_ad}
\end{figure}

We test our method on the MLP-Mixer architecture ~\citep{tolstikhin2021mlpmixer}, using the same settings as \cite{liu2023dropout} and \cite{liu2022convnet}. We applied AD to the final layer and the cross-entropy loss of the Mixer-S/32 architecture on ImageNet-1k and reported the accuracy. Figure~\ref{fig:mlp_mixer_ad} results indicate that AD improves model performance, leading to higher accuracy than the baseline.

\subsection{Experiments on LLM}

\begin{table}[h!]
    \centering
    \caption{Evaluation results for RoBERTa baseline and AD ($\sigma=0.05$), on 7 tasks.}
    \begin{tabular}{lcc}
        \toprule
        \textbf{Metric}                       & DO             & AD             \\
        \midrule
        \textbf{Sentiment Evaluation}         &                &                \\
        Classification Accuracy (\%)          & 77.66          & \textbf{77.68} \\
        \midrule
        \textbf{NER Evaluation}               &                &                \\
        Snips F1 Score (\%)                   & 72.70          & \textbf{73.99} \\
        Snips Precision (\%)                  & 67.78          & \textbf{69.25} \\
        Snips Recall (\%)                     & 78.39          & \textbf{79.44} \\
        \midrule
        \textbf{Intent Evaluation}            &                &                \\
        Classification Accuracy (\%)          & \textbf{97.59} & 96.83          \\
        \midrule
        \textbf{Entailment SNLI Evaluation}   &                &                \\
        Classification Accuracy (\%)          & 88.60          & \textbf{89.33} \\
        \midrule
        \textbf{CoNLL NER Evaluation}         &                &                \\
        Seqeval F1 Score (\%)                 & 67.62          & \textbf{67.90} \\
        Seqeval Precision (\%)                & 65.08          & \textbf{65.38} \\
        Seqeval Recall (\%)                   & 70.36          & \textbf{70.61} \\
        \midrule
        \textbf{CoNLL POS Evaluation}         &                &                \\
        Seqeval F1 Score (\%)                 & 71.86          & \textbf{72.14} \\
        Seqeval Precision (\%)                & 70.68          & \textbf{70.89} \\
        Seqeval Recall (\%)                   & 73.07          & \textbf{73.45} \\
        \midrule
        \textbf{Query Correctness Evaluation} &                &                \\
        Classification Accuracy (\%)          & \textbf{68.08} & 68.06          \\
        \bottomrule
    \end{tabular}
    \label{tab:res_roberta}
\end{table}

\begin{table}[h!]
    \caption{Evaluation results for BERT baseline with DO ($p=0.1$), SAM for different $\rho$ values, and AD with ($\sigma=0.05$)  on 7 tasks.}
    \label{tab:results_bert_vs_sam_vs_do}
    \centering
    \begin{tabular}{lccccccc}
        \toprule
        \textbf{Metric}                       & \textbf{DO}    & \multicolumn{3}{c}{\textbf{SAM}} & \textbf{AD}                                                 \\
        \cmidrule(lr){3-5}
                                              &                & \textbf{$\rho=0.01$}             & \textbf{$\rho=0.05$} & \textbf{$\rho=0.1$} &                \\
        \midrule
        \textbf{Sentiment Evaluation}         &                &                                  &                      &                     &                \\
        Classification Accuracy (\%)          & 76.72          & 76.54                            & 75.38                & 62.28               & \textbf{77.08} \\
        \midrule
        \textbf{NER Evaluation}               &                &                                  &                      &                     &                \\
        Snips F1 Score (\%)                   & 78.33          & 69.67                            & 62.29                & 63.95               & \textbf{80.90} \\
        Snips Precision (\%)                  & 73.69          & 64.11                            & 56.58                & 58.08               & \textbf{76.20} \\
        Snips Recall (\%)                     & 83.59          & 76.28                            & 69.27                & 71.14               & \textbf{86.21} \\
        \midrule
        \textbf{Intent Evaluation}            &                &                                  &                      &                     &                \\
        Classification Accuracy (\%)          & 98.04          & 98.19                            & 97.43                & 97.28               & \textbf{98.49} \\
        \midrule
        \textbf{Entailment SNLI Evaluation}   &                &                                  &                      &                     &                \\
        Classification Accuracy (\%)          & 87.96          & \textbf{89.39}                   & 86.77                & 83.85               & 88.88          \\
        \midrule
        \textbf{CoNLL NER Evaluation}         &                &                                  &                      &                     &                \\
        Seqeval F1 Score (\%)                 & 64.43          & 61.01                            & 51.09                & 52.05               & \textbf{65.94} \\
        Seqeval Precision (\%)                & 61.87          & 61.48                            & 51.64                & 51.72               & \textbf{64.11} \\
        Seqeval Recall (\%)                   & 67.20          & 60.55                            & 50.56                & 52.39               & \textbf{67.87} \\
        \midrule
        \textbf{CoNLL POS Evaluation}         &                &                                  &                      &                     &                \\
        Seqeval F1 Score (\%)                 & 75.95          & 72.48                            & 69.31                & 71.20               & \textbf{77.89} \\
        Seqeval Precision (\%)                & 74.89          & 71.59                            & 68.79                & 70.46               & \textbf{76.98} \\
        Seqeval Recall (\%)                   & 77.04          & 73.39                            & 69.84                & 71.97               & \textbf{78.82} \\
        \midrule
        \textbf{Query Correctness Evaluation} &                &                                  &                      &                     &                \\
        Classification Accuracy (\%)          & \textbf{69.95} & 69.47                            & 66.24                & 64.74               & 69.31          \\
        \bottomrule
    \end{tabular}
\end{table}

\newpage

\section{Proofs}

\subsection{Proof of Corollary~\ref{corol:bound_hessian_smoothed} }
\label{proof:corol:bound_hessian_smoothed}
\begin{proof}
    Let \( F(\tens{\theta}) := \nabla_{\tens{\theta}} \mathcal{L}(\tens{\theta}) \in \mathbb{R}^d \).
    By assumption, we have
    \[
        \|F(\tens{\theta})\|_2 \leq \epsilon,
        \qquad
        \|F(\tens{\theta}) - F(\tens{\theta}')\|_2 \leq H \|\tens{\theta} - \tens{\theta}'\|_2
        \quad \text{for all } \tens{\theta}, \tens{\theta}' \in \mathbb{R}^d,
    \]
    which is equivalent to \( \| \nabla^2_{\tens{\theta}} \mathcal{L}(\tens{\theta}) \|_2 \leq H \).

    For any unit vector \( u \in \mathbb{R}^d \), define
    \[
        f_u(\tens{\theta}) := u^\top F(\tens{\theta}) + \epsilon.
    \]
    Then \( f_u : \mathbb{R}^d \to [0, 2\epsilon] \) is \( H \)-Lipschitz.
    Recall its Gaussian smoothed version as
    \(
    f_u^\sigma(\tens{\theta}) = \mathbb{E}_{\Delta \sim \mathcal{N}(0, \sigma^2 I)} [f_u(\tens{\theta} + {\Delta})].
    \)

    Applying Theorem~\ref{thm:lipschitz_gaussian_smoothing} to \( f_u \) with \( r = 2\epsilon \) and \( L(f_u) = H \), we obtain
    \[
        \Lip(f_u^\sigma)
        \leq
        H  \operatorname{erf}\left( \frac{2\epsilon}{2^{3/2} H \sigma} \right)
        =
        H  \operatorname{erf}\left( \frac{\epsilon}{\sqrt{2} H \sigma} \right).
        \tag{1}
    \]

    Since \( \mathcal{L} \) is twice differentiable, we can exchange the gradient and expectation:
    \[
        \nabla_{\tens{\theta}} \mathcal{L}^\sigma(\tens{\theta}) = \mathbb{E}_{{\Delta}}[F(\tens{\theta} + {\Delta})].
    \]
    Therefore,
    \[
        \nabla_{\tens{\theta}} f_u^\sigma(\tens{\theta})
        =
        \left( \nabla_{\tens{\theta}} F^\sigma(\tens{\theta}) \right)^\top u
        =
        \nabla^2_{\tens{\theta}} \mathcal{L}^\sigma(\tens{\theta}) \, u.
    \]
    Hence,
    \[
        \Lip(f_u^\sigma)
        =
        \sup_{\tens{\theta}}
        \left\| \nabla^2_{\tens{\theta}} \mathcal{L}^\sigma(\tens{\theta}) \, u \right\|_2.
        \tag{2}
    \]

    Combining (1) and (2), we obtain for all unit vectors \( u \),
    \[
        \sup_{\tens{\theta}} \left\| \nabla^2_{\tens{\theta}} \mathcal{L}^\sigma(\tens{\theta}) \, u \right\|_2
        \leq
        H  \operatorname{erf}\left( \frac{\epsilon}{\sqrt{2} H \sigma} \right).
    \]
    Taking the supremum over all unit vectors \( u \) yields
    \[
        \left\| \nabla^2_{\tens{\theta}} \mathcal{L}^\sigma(\tens{\theta}) \right\|_2
        \leq
        H  \operatorname{erf}\left( \frac{\epsilon}{\sqrt{2} H \sigma} \right),
    \]
    which concludes the proof.
\end{proof}

\subsection{Proof of Theorem~\ref{thm:bound_hessian_layers}}
\label{proof:thm:bound_hessian_layers}

\begin{proof}
    Write $\vz=\vh^{(L-1)}$ and $J=\frac{\partial \vz}{\partial \theta}$.
    By the second-order chain rule,
    \[
        \nabla^2_{\theta}\mathcal L
        = J^\top \,\nabla^2_{\vz}\mathcal L\, J \;+\;
        \Big(\tfrac{\partial^2 \vz}{\partial \theta^2}\Big)^\top \nabla_{\vz}\mathcal L.
    \]
    Stationarity in logits implies
    $\nabla_{\va^{(L)}}\ell(\va^{(L)}(\theta^\star),y)=0$ sample-wise, hence
    $\nabla_{\vz}\mathcal L(\theta^\star)
        = \big(\tfrac{\partial \va^{(L)}}{\partial \vz}\big)^\top \nabla_{\va^{(L)}}\ell(\va^{(L)}(\theta^\star),y)=0$,
    So the residual term vanishes at $\theta^\star$ and
    \[
        \nabla^2_{\theta}\mathcal L(\theta^\star)=
        J(\theta^\star)^\top \,\nabla^2_{\vz}\mathcal L(\theta^\star)\, J(\theta^\star).
    \]
    Taking operator norms and using submultiplicativity,
    \[
        \|\nabla^2_{\theta}\mathcal L(\theta^\star)\|_2
        \;\le\; \|J(\theta^\star)\|_2^2 \,\|\nabla^2_{\vz}\mathcal L(\theta^\star)\|_2.
    \]
    Finally, expand $J$ layerwise:
    \[
        \frac{\partial \vz}{\partial \theta}
        =\sum_{j=1}^{L-1}\!\left(\prod_{l=j+1}^{L-1}\frac{\partial f^{(l)}}{\partial \vh^{(l-1)}}\right)\!
        \frac{\partial \vh^{(j)}}{\partial \theta},
    \]
    whence, by the triangle inequality and the $1$-Lipschitz property,
    \[
        \|J(\theta^\star)\|_2
        \;\le\;\sum_{j=1}^{L-1}
        \left\|\frac{\partial \vh^{(j)}}{\partial \theta}(\theta^\star)\right\|_2
        \prod_{l=j+1}^{L-1}\|\mW^{(l)}\|_2.
    \]
    Combine the two displays to obtain the claimed bound.
\end{proof}

\subsection{Proof of Theorem~\ref{thm:smoothed_loss}}

\begin{proof}
    The original cross-entropy loss for the correct class \( c \) is given by:
    \[
        \mathcal{L}(h_L(\vh_{L-1}), \vy) = - \mW_c^\top \vh_{L-1} + \log \left(\sum_{j=1}^{d} \exp(\mW_j^\top \vh_{L-1})\right) \ .
    \]

    Consider the smoothed loss by introducing Gaussian noise \(\mathbf{\Delta} \sim \mathcal{N}(0, \mI \sigma^2)\):
    \[
        \mathcal{L}^\sigma(h_L(\vh_{L-1}), \vy) = \mathbb{E}_{\mathbf{\Delta} \sim \mathcal{N}(0, \mI \sigma^2)}\left[ - (\mW_c + \mathbf{\Delta}_c)^\top \vh_{L-1} + \log\left( \sum_{j=1}^{d} \exp((\mW_j + \mathbf{\Delta}_j)^\top \vh_{L-1}) \right) \right]
    \]
    Separating the terms:
    \[
        \mathcal{L}^\sigma(h_L(\vh_{L-1}), \vy) = -\mW_c^\top \vh_{L-1} + \mathbb{E}_{\mathbf{\Delta} \sim \mathcal{N}(0, \mI \sigma^2)} \left[ \log\left( \sum_{j=1}^{d} \exp((\mW_j + \mathbf{\Delta}_j)^\top \vh_{L-1}) \right) \right] \ .
    \]

    By applying Jensen's inequality on the expectation inside the logarithm, we get:
    \[
        \mathbb{E}_{\mathbf{\Delta}} \left[ \log\left( \sum_{j=1}^{d} \exp((\mW_j + \mathbf{\Delta}_j)^\top \vh_{L-1}) \right) \right] \leq \log \left( \mathbb{E}_{\mathbf{\Delta}} \left[ \sum_{j=1}^{d} \exp((\mW_j + \mathbf{\Delta}_j)^\top \vh_{L-1}) \right] \right) \ .
    \]

    Since \(\mathbf{\Delta}_j \sim \mathcal{N}(0, \sigma^2 \mI)\), we use the moment generating function of the Gaussian distribution:
    \[
        \mathbb{E}[e^Z] = e^{\mu + \frac{1}{2} \sigma^2}
    \]
    Applying this to our case for each \(\mW_j^\top \vh_{L-1} + \mathbf{\Delta}_j^\top \vh_{L-1}\):
    \[
        \mathbb{E}\left[ \exp((\mW_j + \mathbf{\Delta}_j)^\top \vh_{L-1}) \right] = \exp(\mW_j^\top \vh_{L-1})  \mathbb{E}\left[ \exp(\mathbf{\Delta}_j^\top \vh_{L-1}) \right]
    \]
    Given \(\mathbf{\Delta}_j \sim \mathcal{N}(0, \sigma^2 \mI)\) and \(\mathbf{\Delta}_j^\top \vh_{L-1}\) is a Gaussian with mean 0 and variance \(\sigma^2 \|\vh_{L-1}\|^2\), we get:
    \[
        \mathbb{E}\left[ \exp(\mathbf{\Delta}_j^\top \vh_{L-1}) \right] = \exp\left(0 + \frac{1}{2} \sigma^2 \|\vh_{L-1}\|^2 \right)
    \]
    Therefore,
    \[
        \mathbb{E}\left[ \exp((\mW_j + \mathbf{\Delta}_j)^\top \vh_{L-1}) \right] = \exp\left(\mW_j^\top \vh_{L-1} + \frac{1}{2} \sigma^2 \|\vh_{L-1}\|^2 \right)
    \]

    Summing over \( j \):
    \[
        \mathbb{E}_{\mathbf{\Delta}} \left[ \sum_{j=1}^{d} \exp((\mW_j + \mathbf{\Delta}_j)^\top \vh_{L-1}) \right] = \sum_{j=1}^{d} \exp\left(\mW_j^\top \vh_{L-1} + \frac{1}{2} \sigma^2 \|\vh_{L-1}\|^2 \right)
    \]

    Substituting this back into the expression for the smoothed loss, we obtain:
    \[
        \mathcal{L}^\sigma(h_L(\vh_{L-1}), \vy) \leq -\mW_c^\top \vh_{L-1} + \log \left( \sum_{j=1}^{d} \exp\left(\mW_j^\top \vh_{L-1} + \frac{1}{2} \sigma^2 \|\vh_{L-1}\|^2 \right) \right)
    \]

    This result shows that the smoothed loss \(\mathcal{L}^\sigma(h_L(\vh_{L-1}), \vy)\) is bounded above by the original loss with an additional offset term \(\frac{1}{2} \sigma^2 \|\vh_{L-1}\|^2\). This offset is akin to the regularization term observed in the loss from the work of~\citet{tsuzuku2018lipschitz} in term of $L \epsilon$ where $L$ is the Lipschitz constant and $\epsilon$ the size of the perturbation.
\end{proof}

\subsection{Theorem and proof on tighter approximation using Taylor expansion}

\begin{theorem}[Tighter approximation via Taylor expansion]
    Let \(\mW^{(L)} \in \mathbb{R}^{c \times d}\), \(\vh^{(L-1)} \in \mathbb{R}^d\), and \(\mathbf{\Delta} \in \mathbb{R}^{c \times d}\) with elements drawn independently from \(\mathcal{N}(0, \sigma^2)\). Denote \(\hat{\vy} = \mathrm{softmax}(\mW^{(L)} \vh^{(L-1)})\). For small \(\sigma\), the expected cross-entropy loss under perturbations \(\mathbf{\Delta}\) is approximately:
    \[
        \mathbb{E}\left[\mathcal{L}_{\mathrm{CE}}\big((\mW^{(L)} + \mathbf{\Delta}) \vh^{(L-1)}, \vy\big)\right] \approx \mathcal{L}_{\mathrm{CE}}(\mW^{(L)} \vh^{(L-1)}, \vy) + \tfrac{1}{2} \sigma^2 \|\vh^{(L-1)}\|_2^2 \sum_{i=1}^c \hat{y}_i (1 - \hat{y}_i) \ .
    \]
\end{theorem}

Note that the obtained approximation of the smoothed loss is not an upper bound on the exact smoothed loss.

\begin{proof}
    We start by expanding the cross-entropy loss around \(\mW^{(L)} \vh^{(L-1)}\) using a first-order Taylor expansion:
    \[
        \mathcal{L}_{\mathrm{CE}}\big((\mW^{(L)} + \mathbf{\Delta}) \vh^{(L-1)}, \vy\big) \approx \mathcal{L}_{\mathrm{CE}}(\mW^{(L)} \vh^{(L-1)}, \vy) + \nabla \mathcal{L}_{\mathrm{CE}}(\mW^{(L)} \vh^{(L-1)}, \vy)^\top (\mathbf{\Delta} \vh^{(L-1)}) \ ,
    \]
    where \(\nabla \mathcal{L}_{\mathrm{CE}}(\mW^{(L)} \vh^{(L-1)}, \vy) = \hat{\vy} - \vy\). The first-order term is then:
    \[
        (\hat{\vy} - \vy)^\top \mathbf{\Delta} \vh^{(L-1)} \ .
    \]
    Taking the expectation of this term with respect to \(\mathbf{\Delta}\), we use the fact that \(\mathbb{E}[\mathbf{\Delta}] = 0\), so the expectation of the first-order term is zero:
    \[
        \mathbb{E}\left[(\hat{\vy} - \vy)^\top \mathbf{\Delta} \vh^{(L-1)}\right] = 0 \ .
    \]

    We then proceed with the second-order Taylor expansion:
    \[
        \frac{1}{2} (\mathbf{\Delta} \vh^{(L-1)})^\top \nabla^2 \mathcal{L}_{\mathrm{CE}}(\mW^{(L)} \vh^{(L-1)}, \vy) (\mathbf{\Delta} \vh^{(L-1)}) \ ,
    \]
    where the Hessian \(\nabla^2 \mathcal{L}_{\mathrm{CE}}(\mW^{(L)} \vh^{(L-1)}, \vy)\) is given by:
    \[
        \nabla^2 \mathcal{L}_{\mathrm{CE}}(\mW^{(L)} \vh^{(L-1)}, \vy) = \mathrm{diag}(\hat{\vy}) - \hat{\vy} \hat{\vy}^\top \ .
    \]
    Now, we compute the expectation of the second-order term. Using the property of quadratic forms for Gaussian random variables, we have:
    \[
        \mathbb{E}[\mathbf{\Delta} \vh^{(L-1)} (\mathbf{\Delta} \vh^{(L-1)})^\top] = \sigma^2 \|\vh^{(L-1)}\|_2^2 \mI_c \ ,
    \]
    where \(\mI_c\) is the identity matrix in \(\mathbb{R}^{c \times c}\). Thus, the second-order term simplifies to:
    \[
        \frac{\sigma^2}{2} \|\vh^{(L-1)}\|_2^2 \mathrm{tr}\left(\nabla^2 \mathcal{L}_{\mathrm{CE}}(\mW^{(L)} \vh^{(L-1)}, \vy)\right) \ .
    \]

    Finally, we compute the trace of the Hessian:
    \[
        \mathrm{tr}\left(\mathrm{diag}(\hat{\vy}) - \hat{\vy} \hat{\vy}^\top\right) = \sum_{i=1}^c \hat{y}_i (1 - \hat{y}_i) \ ,
    \]
    as the trace of \(\hat{\vy} \hat{\vy}^\top\) is 1. Therefore, the second-order term becomes:
    \[
        \frac{\sigma^2}{2} \|\vh^{(L-1)}\|_2^2 \sum_{i=1}^c \hat{y}_i (1 - \hat{y}_i) \ .
    \]

    Thus, the total approximation including both the first- and second-order terms is:
    \[
        \mathbb{E}\left[\mathcal{L}_{\mathrm{CE}}\big((\mW^{(L)} + \mathbf{\Delta}) \vh^{(L-1)}, \vy\big)\right] \approx \mathcal{L}_{\mathrm{CE}}(\mW^{(L)} \vh^{(L-1)}, \vy) + \frac{\sigma^2}{2} \|\vh^{(L-1)}\|_2^2 \sum_{i=1}^c \hat{y}_i (1 - \hat{y}_i) \ .
    \]
\end{proof}

%% file: bibliography/aaai.bib
@article{zhang2019recurjac,
	title = {{RecurJac}: {An} {Efficient} {Recursive} {Algorithm} for {Bounding} {Jacobian} {Matrix} of {Neural} {Networks} and {Its} {Applications}},
	journal = {AAAI Conference on Artificial Intelligence},
	author = {Zhang, Huan and Zhang, Pengchuan and Hsieh, Cho-Jui},
	year = {2019},
}

@article{araujo2021lipschitz,
  title={On Lipschitz Regularization of Convolutional Layers using Toeplitz Matrix Theory},
  author={Araujo, Alexandre and Negrevergne, Benjamin and Chevaleyre, Yann and Atif, Jamal},
  journal={Proceedings of the 35th AAAI Conference on Artificial Intelligence},
  url={https://arxiv.org/abs/2006.08391},
  year={2021}
}

@inproceedings{ross2018improving,
  title={Improving the Adversarial Robustness and Interpretability of Deep Neural Networks by Regularizing their Input Gradients},
  author={Ross, Andrew Slavin and Doshi-Velez, Finale},
  booktitle={Proceedings of the 32nd AAAI Conference on Artificial Intelligence},
  year={2018}
}


%% file: bibliography/arxiv.bib
@misc{baevski20wav2vec,
  title   = {wav2vec 2.0: A Framework for Self-Supervised Learning of Speech Representations},
  author  = {Alexei Baevski and Henry Zhou and Abdelrahman Mohamed and Michael Auli},
  year    = {2020},
  journal = {arXiv}
}

@article{gao2018propertiessoftmaxfunctionapplication,
  title         = {On the Properties of the Softmax Function with Application in Game Theory and Reinforcement Learning},
  author        = {Bolin Gao and Lacra Pavel},
  year          = {2018},
  archiveprefix = {arXiv}
}

@article{li2018secondorder,
  title   = {Second-order adversarial attack and certifiable robustness},
  author  = {Li, Baoyuan and Chen, Changyou and Wang, Wenlin and Carin, Lawrence},
  journal = {arXiv},
  year    = {2018}
}

@article{cho2014learning,
  title   = {Learning Phrase Representations using RNN Encoder-Decoder for Statistical Machine Translation},
  author  = {Cho, Kyunghyun and van Merrienboer, Bart and Gulcehre, Caglar and Bahdanau, Dzmitry and Bougares, Fethi and Schwenk, Holger and Bengio, Yoshua},
  journal = {arXiv},
  year    = {2014}
}

@article{wei2022emergent,
  title   = {Emergent Abilities of Large Language Models},
  author  = {Wei, Jason and Tay, Yi and Bommasani, Rishi and Raffel, Colin and Zoph, Barret and Borgeaud, Sebastian and Yogatama, Dani and Bosma, Maarten and Zhou, Denny and Metzler, Donald and Chi, Ed H.},
  journal = {arXiv},
  year    = {2022}
}

@article{you2020large,
  title   = {Large batch optimization for deep learning: Training bert in 76 minutes},
  author  = {You, Yang and Li, Jingyu and Reddi, Sashank J and Hseu, Jonathan and Kumar, Sanjiv and Bhojanapalli, Srinadh and Song, Xiaodan and Demmel, James and Hsieh, Cho-Jui and Keutzer, Kurt},
  journal = {arXiv},
  year    = {2020}
}

@misc{vuckovic2021regularityattention,
  title   = {On the Regularity of Attention},
  author  = {James Vuckovic and Aristide Baratin and Remi Tachet des Combes},
  year    = {2021},
  journal = {arXiv}
}

@inproceedings{ba2016layer,
  title     = {Layer normalization},
  author    = {Ba, Jimmy Lei and Kiros, Jamie Ryan and Hinton, Geoffrey E},
  booktitle = {arXiv},
  year      = {2016}
}

@article{hendrycks2016gelu,
  title   = {Gaussian error linear units (GELUs)},
  author  = {Hendrycks, Dan and Gimpel, Kevin},
  journal = {arXiv preprint arXiv:1606.08415},
  year    = {2016}
}

@misc{kingma2017adam,
  title     = {Adam: {A} {Method} for {Stochastic} {Optimization}},
  publisher = {arXiv},
  author    = {Kingma, Diederik P. and Ba, Jimmy},
  year      = {2017}
}

@article{voracek2024treatment,
  title   = {Treatment of Statistical Estimation Problems in Randomized Smoothing for Adversarial Robustness},
  author  = {Vor\'a{\v{c}}ek, V\'aclav},
  journal = {arXiv preprint arXiv:2406.17830},
  year    = {2024}
}

@inproceedings{franco2023diffusion,
  title     = {Diffusion {Denoised} {Smoothing} for {Certified} and {Adversarial} {Robust} {Out}-{Of}-{Distribution} {Detection}},
  booktitle = {arXiv},
  author    = {Franco, Nicola and Korth, Daniel and Lorenz, Jeanette Miriam and Roscher, Karsten and Guennemann, Stephan},
  year      = {2023}
}

@article{yoshida2017spectral,
  title   = {Spectral Norm Regularization for Improving the Generalizability of Deep Learning},
  author  = {Yoshida, Yuichi and Miyato, Takeru},
  journal = {arXiv preprint arXiv:1705.10941},
  year    = {2017}
}

@article{xing2018walk,
  author  = {{Xing}, C. and {Arpit}, D. and {Tsirigotis}, C. and {Bengio}, Y.},
  title   = {A Walk with {SGD}},
  journal = {arXiv},
  year    = 2018
}

@article{hendrycks2020measuring,
  title   = {Measuring Massive Multitask Language Understanding},
  author  = {Hendrycks, Dan and Burns, Collin and Basart, Steven and Zou, Andy and Mazeika, Mantas and Song, Dawn and Steinhardt, Jacob},
  journal = {arXiv},
  year    = {2020}
}

@article{zhang2021surveyMTL,
  title   = {A Survey of Multi-task Learning in Natural Language Processing: Regarding Task Relatedness and Training Methods},
  author  = {Zhang, Zhihan and Yu, Wenhao and Yu, Mengxia and Guo, Zhichun and Jiang, Meng},
  journal = {arXiv},
  year    = {2021}
}

@article{liu2019roberta,
  title   = {{RoBERTa}: A robustly optimized {BERT} pretraining approach},
  author  = {Liu, Yinhan and Ott, Myle and Goyal, Naman and Du, Jingfei and Joshi, Mandar and Chen, Danqi and Levy, Omer and Lewis, Mike and Zettlemoyer, Luke and Stoyanov, Veselin},
  journal = {arXiv},
  year    = {2019}
}

@article{lu2021meanfield,
  title   = {Mean-Field Approximation to Gaussian-Softmax Integral with Application to Uncertainty Estimation},
  author  = {Zhiyun Lu and Eugene Ie and Fei Sha},
  year    = {2021},
  journal = {arXiv}
}

@article{kurakin2016adversarial,
  title   = {Adversarial Examples in the Physical World},
  author  = {Kurakin, Alexey and Goodfellow, Ian and Bengio, Samy},
  journal = {arXiv preprint arXiv:1607.02533},
  year    = {2016}
}

@misc{salimans2016weight,
  title   = {Weight Normalization: A Simple Reparameterization to Accelerate Training of Deep Neural Networks},
  author  = {Tim Salimans and Diederik P. Kingma},
  year    = {2016},
  journal = {arXiv}
}

@article{gnassounou2024multi,
  title   = {Multi-Source and Test-Time Domain Adaptation on Multivariate Signals using Spatio-Temporal Monge Alignment},
  author  = {Gnassounou, Th{\'e}o and Collas, Antoine and Flamary, R{\'e}mi and Lounici, Karim and Gramfort, Alexandre},
  journal = {arXiv},
  year    = {2024}
}

@article{boissin2025adaptive,
  author = {Thibault Boissin and Fran{\c{c}}ois Mamalet and Thibaut Fel and Anne-Marie Picard and Thibault Massena and others},
  title  = {An Adaptive Orthogonal Convolution Scheme for Efficient and Flexible CNN Architectures},
  year   = {2025}
}

@article{simonyan2014vgg,
  title   = {Very Deep Convolutional Networks for Large-Scale Image Recognition},
  author  = {Simonyan, Karen and Zisserman, Andrew},
  journal = {arXiv preprint arXiv:1409.1556},
  year    = {2014}
}

@article{li2018second,
  author  = {Bai Li and Changyou Chen and Wenlin Wang and Lawrence Carin},
  title   = {Second-Order Adversarial Attack and Certifiable Robustness},
  journal = {arXiv preprint arXiv:1809.03113},
  year    = {2018}
}

@article{finlay2018lipschitz,
  title   = {Lipschitz Regularized Deep Neural Networks Generalize and are Adversarially Robust},
  author  = {Finlay, Chris and Calder, Jeff and Abbasi, Bilal and Oberman, Adam},
  journal = {arXiv preprint arXiv:1808.09540},
  year    = {2018}
}

@article{kaplan2020scaling,
  title   = {Scaling Laws for Neural Language Models},
  author  = {Kaplan, Jared and McCandlish, Sam and Henighan, Tom and Brown, Tom B and Chess, Benjamin and Child, Rewon and Gray, Scott and Radford, Alec and Wu, Jeffrey and Amodei, Dario},
  journal = {arXiv preprint arXiv:2001.08361},
  year    = {2020}
}

@article{yi2020asymptotic,
  title   = {Asymptotic Singular Value Distribution of Linear Convolutional Layers},
  author  = {Yi, Xinping},
  journal = {arXiv preprint arXiv:2006.07117},
  year    = {2020}
}


%% file: bibliography/cvpr.bib
@article{liu2022convnet,
  author  = {Zhuang Liu and Hanzi Mao and Chao-Yuan Wu and Christoph Feichtenhofer and Trevor Darrell and Saining Xie},
  title   = {A ConvNet for the 2020s},
  journal = {Conference on Computer Vision and Pattern Recognition},
  year    = {2022}
}

@inproceedings{yashwanth2024minimizing,
  author    = {Yashwanth, M. and Nayak, Gaurav Kumar and Rangwani, Harsh and Singh, Arya and Babu, R. Venkatesh and Chakraborty, Anirban},
  title     = {Minimizing Layerwise Activation Norm Improves Generalization in Federated Learning},
  booktitle = {Winter Conference on Applications of Computer Vision (WACV)},
  year      = {2024}
}

@inproceedings{prach2022almost,
  title     = {Almost-orthogonal layers for efficient general-purpose Lipschitz networks},
  author    = {Prach, Bernd and Lampert, Christoph H},
  booktitle = {Computer Vision--ECCV 2022: 17th European Conference},
  year      = {2022}
}

@inproceedings{kayhan2020translation,
  title     = {On {Translation} {Invariance} in {CNNs}: {Convolutional} {Layers} {Can} {Exploit} {Absolute} {Spatial} {Location}},
  booktitle = {{Computer} {Vision} and {Pattern} {Recognition}},
  author    = {Semih Kayhan, Osman and Van Gemert, Jan C.},
  year      = {2020}
}

@inproceedings{redmon2016you,
  title     = {You Only Look Once: Unified, Real-Time Object Detection},
  author    = {Redmon~, Joseph and Divvala, Santosh and Girshick, Ross and Farhadi, Ali},
  booktitle = {Proceedings of the IEEE Conference on Computer Vision and Pattern Recognition \emph{(CVPR)}},
  year      = {2016}
}

@inproceedings{he2016deep,
  title     = {Deep Residual Learning for Image Recognition},
  author    = {He, Kaiming and Zhang, Xiangyu and Ren, Shaoqing and Sun, Jian},
  booktitle = {Proceedings of the IEEE Conference on Computer Vision and Pattern Recognition \emph{(CVPR)}},
  year      = {2016}
}

@inproceedings{wang2020orthogonal,
  title     = {Orthogonal Convolutional Neural Networks},
  author    = {Wang, Jiayun and Chen, Yubei and Chakraborty, Rudrasis and Yu, Stella X.},
  booktitle = {Proceedings of the IEEE Conference on Computer Vision and Pattern Recognition \emph{(CVPR)}},
  year      = {2020}
}

@inproceedings{huang2020controllable,
  title     = {Controllable Orthogonalization in Training DNNs},
  author    = {Huang, Lei and Liu, Li and Zhu, Fan and Wan, Diwen and Yuan, Zehuan and Li, Bo and Shao, Ling},
  booktitle = {Proceedings of the IEEE Conference on Computer Vision and Pattern Recognition \emph{(CVPR)}},
  year      = {2020}
}


%% file: bibliography/eccv.bib
@inproceedings{wu2018group,
  title     = {Group normalization},
  author    = {Wu, Yuxin and He, Kaiming},
  booktitle = {Proceedings of the European Conference on Computer Vision (ECCV)},
  year      = {2018}
}


%% file: bibliography/iccv.bib
@inproceedings{gavrikov2023interplay,
  author    = {Gavrikov, Paul and Keuper, Janis},
  title     = {On the Interplay of Convolutional Padding and Adversarial Robustness},
  booktitle = {Proceedings of the IEEE/CVF International Conference on Computer Vision (ICCV) Workshops},
  year      = {2023}
}

@inproceedings{glorot2010understanding,
  title     = {Understanding the difficulty of training deep feedforward neural networks},
  author    = {Glorot, Xavier and Bengio, Yoshua},
  booktitle = {Proceedings of the Thirteenth International Conference on Artificial Intelligence and Statistics},
  year      = {2010}
}

@inproceedings{he2015delving,
  title     = {Delving Deep into Rectifiers: Surpassing Human-Level Performance on ImageNet Classification},
  author    = {K. He and X. Zhang and S. Ren and J. Sun},
  booktitle = {IEEE International Conference on Computer Vision \emph{(ICCV)}},
  year      = {2015}
}


%% file: bibliography/iclr.bib
@inproceedings{kodali2018on,
  title     = {On Convergence and Stability of GANs},
  author    = {Kodali, Naveen and Hays, James and Abernethy, Jacob and Kira, Zsolt},
  booktitle = {ICLR (Workshop or Blind Sub. if applicable)},
  year      = {2018}
}

@inproceedings{raghunathan2018certified,
  title     = {Certified defenses against adversarial examples},
  author    = {Raghunathan, Aditi and Steinhardt, Jacob and Liang, Percy},
  booktitle = {International Conference on Learning Representations (ICLR)},
  year      = {2018}
}

@inproceedings{singla2021improved,
  title     = {Improved Lipschitz bounds for deep networks: Scaling law and effective approximation},
  author    = {Singla, Sahil and Feizi, Soheil},
  booktitle = {International Conference on Learning Representations (ICLR)},
  year      = {2021}
}

@inproceedings{kingma2014auto,
  title     = {Auto-encoding variational Bayes},
  author    = {Kingma, Diederik P and Welling, Max},
  booktitle = {International Conference on Learning Representations},
  year      = {2014}
}

@misc{kim2021the,
  title     = {The Lipschitz Constant of Self-Attention},
  author    = {Hyunjik Kim and George Papamakarios and Andriy Mnih},
  year      = {2021},
  booktitle = {International Conference on Learning Representations (ICLR)}
}

@inproceedings{kingma2014adam,
  title     = {Adam: A method for stochastic optimization},
  author    = {Kingma, Diederik P and Ba, Jimmy},
  booktitle = {International Conference on Learning Representations (ICLR)},
  year      = {2015}
}

@article{loshchilov2019decoupled,
  title   = {Decoupled weight decay regularization},
  author  = {Loshchilov, Ilya and Hutter, Frank},
  journal = {International Conference on Learning Representations (ICLR)},
  year    = {2019}
}

@inproceedings{dosovitskiy2020image,
  title     = {An Image is Worth 16x16 Words: Transformers for Image Recognition at Scale},
  author    = {Dosovitskiy, Alexey and Beyer, Lucas and Kolesnikov, Alexander and Weissenborn, Dirk and Zhai, Xiaohua and Unterthiner, Thomas and Dehghani, Mostafa and Minderer, Matthias and Heigold, Georg and Gelly, Sylvain and Uszkoreit, Jakob and Houlsby, Neil},
  booktitle = {International Conference on Learning Representations (ICLR)},
  year      = {2021}
}

@misc{chen2024robust,
  title  = {Robust Classification via a Single Diffusion Model},
  author = {Huanran Chen and Yinpeng Dong and Zhengyi Wang and Xiao Yang and Chengqi Duan and Hang Su and Jun Zhu},
  year   = {2024}
}

@inproceedings{yu2022constructing,
  title     = {Constructing Orthogonal Convolutions in an Explicit Manner},
  author    = {Tan Yu and Jun Li and YUNFENG CAI and Ping Li},
  booktitle = {International Conference on Learning Representations},
  year      = {2022}
}

@inproceedings{hu2024recipe,
  title     = {A Recipe for Improved Certifiable Robustness},
  author    = {Kai Hu and Klas Leino and Zifan Wang and Matt Fredrikson},
  booktitle = {The Twelfth International Conference on Learning Representations},
  year      = {2024}
}

@inproceedings{zhang2022boosting,
  title     = {Boosting the certified robustness of $\ell_\infty$ distance nets},
  author    = {Zhang, Bo and Jiang, Difan and He, Di and Wang, Liwei},
  booktitle = {International Conference on Learning Representations},
  year      = {2022}
}

@inproceedings{ghazanfari2024lipsim,
  title     = {LipSim: A Provably Robust Perceptual Similarity Metric},
  author    = {Sara Ghazanfari and Alexandre Araujo and Prashanth Krishnamurthy and Farshad Khorrami and Siddharth Garg},
  booktitle = {The Twelfth International Conference on Learning Representations},
  year      = {2024}
}

@inproceedings{bethune2024dpsgd,
  title     = {{DP}-{SGD} Without Clipping: The Lipschitz Neural Network Way},
  author    = {Louis B{\'e}thune and Thomas Massena and Thibaut Boissin and Aur{\'e}lien Bellet and Franck Mamalet and Yannick Prudent and Corentin Friedrich and Mathieu Serrurier and David Vigouroux},
  booktitle = {The Twelfth International Conference on Learning Representations},
  year      = {2024}
}

@inproceedings{li2018second-order,
  title     = {Second-{Order} {Adversarial} {Attack} and {Certifiable} {Robustness}},
  author    = {Li, Bai and Chen, Changyou and Wang, Wenlin and Carin, Lawrence},
  booktitle = {International Conference on Learning Representations},
  year      = {2018}
}

@inproceedings{zhang2017rethinking,
  title     = {Understanding deep learning requires rethinking generalization},
  author    = {Zhang, Chiyuan and Bengio, Samy and Hardt, Moritz and Recht, Benjamin and Vinyals, Oriol},
  booktitle = {International Conference on Learning Representations (ICLR)},
  year      = {2017}
}

@inproceedings{singla2022improved,
  title     = {Improved deterministic l2 robustness on {CIFAR}-10 and {CIFAR}-100},
  author    = {Sahil Singla and Surbhi Singla and Soheil Feizi},
  booktitle = {International Conference on Learning Representations},
  year      = {2022}
}

@inproceedings{sedghi2019singular,
  title     = {The Singular Values of Convolutional Layers},
  author    = {Sedghi, Hanie and Gupta, Vineet and Long, Philip},
  booktitle = {International Conference on Learning Representations},
  year      = {2019}
}

@inproceedings{sanh2022multitask,
  title     = {Multitask Prompted Training Enables Zero-Shot Task Generalization},
  author    = {Victor Sanh and Albert Webson and Colin Raffel and Stephen Bach and Lintang Sutawika and Zaid Alyafeai and Antoine Chaffin and Arnaud Stiegler and Arun Raja and Manan Dey and M Saiful Bari and Canwen Xu and Urmish Thakker and Shanya Sharma Sharma and Eliza Szczechla and Taewoon Kim and Gunjan Chhablani and Nihal Nayak and Debajyoti Datta and Jonathan Chang and Mike Tian-Jian Jiang and Han Wang and Matteo Manica and Sheng Shen and Zheng Xin Yong and Harshit Pandey and Rachel Bawden and Thomas Wang and Trishala Neeraj and Jos Rozen and Abheesht Sharma and Andrea Santilli and Thibault Fevry and Jason Alan Fries and Ryan Teehan and Teven Le Scao and Stella Biderman and Leo Gao and Thomas Wolf and Alexander M Rush},
  booktitle = {International Conference on Learning Representations},
  year      = {2022}
}

@inproceedings{wei2022finetuned,
  title     = {Finetuned Language Models are Zero-Shot Learners},
  author    = {Jason Wei and Maarten Bosma and Vincent Zhao and Kelvin Guu and Adams Wei Yu and Brian Lester and Nan Du and Andrew M. Dai and Quoc V Le},
  booktitle = {International Conference on Learning Representations},
  year      = {2022}
}

@inproceedings{wang2023multitask,
  title     = {Multitask Prompt Tuning Enables Parameter-Efficient Transfer Learning},
  author    = {Zhen Wang and Rameswar Panda and Leonid Karlinsky and Rogerio Feris and Huan Sun and Yoon Kim},
  booktitle = {International Conference on Learning Representations},
  year      = {2023}
}

@inproceedings{baek2024why,
  title     = {Why is {SAM} Robust to Label Noise?},
  author    = {Christina Baek and J Zico Kolter and Aditi Raghunathan},
  booktitle = {International Conference on Learning Representations},
  year      = {2024}
}

@inproceedings{foret2021sharpnessaware,
  title     = {Sharpness-aware Minimization for Efficiently Improving Generalization},
  author    = {Pierre Foret and Ariel Kleiner and Hossein Mobahi and Behnam Neyshabur},
  booktitle = {International Conference on Learning Representations},
  year      = {2021}
}

@inproceedings{jiang2020fantastic,
  title     = {Fantastic Generalization Measures and Where to Find Them},
  author    = {Yiding Jiang* and Behnam Neyshabur* and Hossein Mobahi and Dilip Krishnan and Samy Bengio},
  booktitle = {International Conference on Learning Representations},
  year      = {2020}
}

@article{keskar2017large-batch,
  title     = {{ON} {LARGE}-{BATCH} {TRAINING} {FOR} {DEEP} {LEARNING}: {GENERALIZATION} {GAP} {AND} {SHARP} {MINIMA}},
  author    = {Keskar, Nitish Shirish and Mudigere, Dheevatsa and Nocedal, Jorge and Smelyanskiy, Mikhail and Tang, Ping Tak Peter},
  booktitle = {International Conference on Learning Representations},
  year      = {2017}
}

@inproceedings{carlini2023certified,
  title     = {{(Certified!!)} Adversarial Robustness for Free!},
  author    = {Nicholas Carlini and Florian Tramer and Krishnamurthy Dj Dvijotham and Leslie Rice and Mingjie Sun and J Zico Kolter},
  booktitle = {International Conference on Learning Representations},
  year      = {2023}
}

@article{farnia2019generalizable,
  title     = {Generalizable Adversarial Training via Spectral Normalization},
  author    = {Farnia, Farzan and Zhang, Jesse M and Tse, David N},
  year      = {2019},
  booktitle = {International Conference on Learning Representations}
}

@inproceedings{gu2022efficiently,
  title     = {Efficiently Modeling Long Sequences with Structured State Spaces},
  author    = {Albert Gu and Karan Goel and Christopher Re},
  booktitle = {International Conference on Learning Representations},
  year      = {2022}
}

@inproceedings{islam2019position,
  title     = {How much {Position} {Information} {Do} {Convolutional} {Neural} {Networks} {Encode}?},
  booktitle = {International Conference on Learning Representations \emph{(ICLR)}},
  author    = {Islam, Md Amirul and Jia, Sen and Bruce, Neil D. B.},
  year      = {2019}
}

@inproceedings{trockman2021orthogonalizing,
  title     = {Orthogonalizing Convolutional Layers with the Cayley Transform},
  author    = {Asher Trockman and J Zico Kolter},
  booktitle = {International Conference on Learning Representations},
  year      = {2021}
}

@inproceedings{singla2021fantastic,
  title     = {Fantastic Four: Differentiable and Efficient Bounds on Singular Values of Convolution Layers},
  author    = {Sahil Singla and Soheil Feizi},
  booktitle = {International Conference on Learning Representations},
  year      = {2021}
}

@misc{bahdanau2015neural,
  title     = {Neural Machine Translation by Jointly Learning to Align and Translate},
  author    = {Dzmitry Bahdanau and Kyunghyun Cho and Yoshua Bengio},
  booktitle = {International Conference on Learning Representations},
  year      = {2015}
}

@inproceedings{araujo2023a,
  title     = {A Unified Algebraic Perspective on Lipschitz Neural Networks},
  author    = {Alexandre Araujo and Aaron J Havens and Blaise Delattre and Alexandre Allauzen and Bin Hu},
  booktitle = {International Conference on Learning Representations \emph{(ICLR)}},
  year      = {2023}
}

@inproceedings{szegedy2013intriguing,
  title     = {Intriguing properties of neural networks},
  author    = {Szegedy, Christian and Zaremba, Wojciech and Sutskever, Ilya and Bruna, Joan and Erhan, Dumitru and Goodfellow, Ian and Fergus, Rob},
  booktitle = {International Conference on Learning Representations \emph{(ICLR)}},
  year      = {2013}
}

@inproceedings{goodfellow2014explaining,
  title     = {Explaining and Harnessing Adversarial Examples},
  author    = {Ian Goodfellow and Jonathon Shlens and Christian Szegedy},
  booktitle = {International Conference on Learning Representations \emph{(ICLR)}},
  year      = {2015}
}

@inproceedings{arjovsky2017wasserstein,
  title     = {{W}asserstein Generative Adversarial Networks},
  author    = {Martin Arjovsky and Soumith Chintala and L{\'e}on Bottou},
  booktitle = {International Conference on Learning Representations \emph{(ICLR)}},
  year      = {2017}
}

@inproceedings{miyato2018spectral,
  title     = {Spectral Normalization for Generative Adversarial Networks},
  author    = {Takeru Miyato and Toshiki Kataoka and Masanori Koyama and Yuichi Yoshida},
  booktitle = {International Conference on Learning Representations \emph{(ICLR)}},
  year      = {2018}
}

@inproceedings{madry2018towards,
  title     = {Towards Deep Learning Models Resistant to Adversarial Attacks},
  author    = {Aleksander Madry and Aleksandar Makelov and Ludwig Schmidt and Dimitris Tsipras and Adrian Vladu},
  booktitle = {International Conference on Learning Representations \emph{(ICLR)}},
  year      = {2018}
}

@inproceedings{neyshabur2018pacbayesian,
  title     = {A {PAC}-Bayesian Approach to Spectrally-Normalized Margin Bounds for Neural Networks},
  author    = {Behnam Neyshabur and Srinadh Bhojanapalli and Nathan Srebro},
  booktitle = {International Conference on Learning Representations \emph{(ICLR)}},
  year      = {2018}
}

@inproceedings{farnia2018generalizable,
  title     = {Generalizable Adversarial Training via Spectral Normalization},
  author    = {Farzan Farnia and Jesse Zhang and David Tse},
  booktitle = {International Conference on Learning Representations \emph{(ICLR)}},
  year      = {2019}
}

@inproceedings{bibi2019deep,
  title     = {Deep Layers as Stochastic Solvers},
  author    = {Adel Bibi and Bernard Ghanem and Vladlen Koltun and Rene Ranftl},
  booktitle = {International Conference on Learning Representations \emph{(ICLR)}},
  year      = {2019}
}

@inproceedings{latorre2020lipschitz,
  title     = {Lipschitz Constant Estimation for Neural Networks via Sparse Polynomial Optimization},
  author    = {Fabian Latorre and Paul Rolland and Volkan Cevher},
  booktitle = {International Conference on Learning Representations \emph{(ICLR)}},
  year      = {2020}
}


%% file: bibliography/icml.bib
@inproceedings{bethune2023robust,
  title     = {Robust One-Class Classification with Signed Distance Function using 1-Lipschitz Neural Networks},
  author    = {B{\'e}thune, Louis and Novello, Paul and Coiffier, Guillaume and Boissin, Thibaut and Serrurier, Mathieu and Vincenot, Quentin and Troya-Galvis, Andr{\'e}s},
  booktitle = {Proceedings of the 40th International Conference on Machine Learning},
  location  = {Honolulu, Hawaii, USA},
  year      = {2023}
}

@inproceedings{gonon2025rescaling,
  title     = {A Rescaling-Invariant Lipschitz Bound Based on Path-Metrics for Modern ReLU Network Parameterizations},
  author    = {Gonon, Antoine and Brisebarre, Nicolas and Riccietti, Elisa and Gribonval, Rémi},
  booktitle = {Proceedings of the 42nd International Conference on Machine Learning (ICML)},
  year      = {2025}
}

@inproceedings{balduzzi2017shattered,
  title     = {The Shattered Gradients Problem: If ResNets are the Answer, Then What is the Question?},
  author    = {Balduzzi, David and Frean, Marcus and Leary, Lennox and Lewis, JP and Ma, Kurt Wan-Duo and McWilliams, Brian},
  booktitle = {Proceedings of the 34th International Conference on Machine Learning (ICML)},
  year      = {2017}
}

@inproceedings{castin2024how,
  title     = {How Smooth Is Attention?},
  author    = {Castin, Val\'{e}rie and Ablin, Pierre and Peyr\'{e}, Gabriel},
  booktitle = {Proceedings of the 41st International Conference on Machine Learning},
  year      = {2024}
}

@inproceedings{pascanu2013difficulty,
  title     = {On the difficulty of training recurrent neural networks},
  author    = {Pascanu, Razvan and Mikolov, Tomas and Bengio, Yoshua},
  booktitle = {International Conference on Machine Learning},
  year      = {2013}
}

@inproceedings{havens2024fine,
  title     = {Fine-grained Local Sensitivity Analysis of Standard Dot-Product Self-Attention},
  author    = {Havens, Aaron J and Araujo, Alexandre and Zhang, Huan and Hu, Bin},
  booktitle = {Proceedings of the 41st International Conference on Machine Learning},
  year      = {2024}
}

@inproceedings{li2020implicit,
  title     = {Implicit Euler skip connections: Enhancing adversarial robustness via numerical stability},
  author    = {Li, Mingjie and He, Lingshen and Lin, Zhouchen},
  booktitle = {International Conference on Machine Learning (ICML)},
  year      = {2020}
}

@inproceedings{leino21gloro,
  title     = {Globally-Robust Neural Networks},
  author    = {Klas Leino and Zifan Wang and Matt Fredrikson},
  booktitle = {International Conference on Machine Learning (ICML)},
  year      = {2021}
}

@inproceedings{andriushchenko2022understanding,
  title     = {Towards Understanding Sharpness-Aware Minimization},
  author    = {Andriushchenko, Maksym and Flammarion, Nicolas},
  booktitle = {Proceedings of the 39th International Conference on Machine Learning},
  year      = {2022}
}

@inproceedings{wang2023direct,
  title     = {Direct parameterization of lipschitz-bounded deep networks},
  author    = {Wang, Ruigang and Manchester, Ian},
  booktitle = {International Conference on Machine Learning},
  year      = {2023}
}

@inproceedings{zhang2021towards,
  title     = {Towards certifying $\ell_\infty$ robustness using neural networks with $\ell_\infty$-dist neurons},
  author    = {Zhang, Bo and Cai, Tianle and Lu, Ziyu and He, Di and Wang, Liwei},
  booktitle = {International Conference on Machine Learning},
  year      = {2021}
}

@inproceedings{stickland2019bert,
  title     = {Bert and pals: Projected attention layers for efficient adaptation in multi-task learning},
  author    = {Stickland, Asa Cooper and Murray, Iain},
  booktitle = {International Conference on Machine Learning},
  year      = {2019}
}

@inproceedings{liu2023dropout,
  title     = {Dropout Reduces Underfitting},
  author    = {Liu, Zhuang and Xu, Zhiqiu and Jin, Joseph and Shen, Zhiqiang and Darrell, Trevor},
  booktitle = {International Conference on Machine Learning},
  year      = {2023}
}

@article{jin2017escape,
  title   = {How to Escape Saddle Points Efficiently},
  author  = {Jin, Chi and Ge, Rong and Netrapalli, Praneeth and Kakade, Sham and Jordan, Michael I},
  journal = {International Conference on Machine Learning},
  year    = {2017}
}

@inproceedings{dinh2017sharp,
  title     = {Sharp minima can generalize for deep nets},
  author    = {Dinh, L. and Pascanu, R. and Bengio, S. and Bengio, Y.},
  booktitle = {International Conference on Machine Learning},
  year      = {2017}
}

@inproceedings{orvieto2022anticorrelated,
  title     = {Anticorrelated {Noise} {Injection} for {Improved} {Generalization}},
  booktitle = {Proceedings of the 39th {International} {Conference} on {Machine} {Learning}},
  author    = {Orvieto, Antonio and Kersting, Hans and Proske, Frank and Bach, Francis and Lucchi, Aurelien},
  year      = {2022}
}

@inproceedings{wei2020implicit,
  title     = {The Implicit and Explicit Regularization Effects of Dropout},
  author    = {Wei, Colin and Schoenholz, Samuel S. and Dai, Zhi and Ma, Tengyu},
  booktitle = {International Conference on Machine Learning},
  year      = {2020}
}

@inproceedings{voracek2023improving,
  title     = {Improving l1-{Certified} {Robustness} via {Randomized} {Smoothing} by {Leveraging} {Box} {Constraints}},
  booktitle = {{International} {Conference} on {Machine} {Learning}},
  author    = {Vor\'a{\v{c}}ek, V\'aclav and Hein, Matthias},
  year      = {2023}
}

@inproceedings{martins2016softmax,
  title     = {From softmax to sparsemax: A sparse model of attention and multi-label classification},
  author    = {Martins, Andre and Astudillo, Ramon},
  booktitle = {International Conference on Machine Learning},
  year      = {2016}
}

@inproceedings{pinson2023linear,
  title     = {Linear {CNNs} {Discover} the {Statistical} {Structure} of the {Dataset} {Using} {Only} the {Most} {Dominant} {Frequencies}},
  booktitle = {{International} {Conference} on {Machine} {Learning}},
  author    = {Pinson, Hannah and Lenaerts, Joeri and Ginis, Vincent},
  year      = {2023}
}

@inproceedings{tang2023defects,
  title     = {Defects of {Convolutional} {Decoder} {Networks} in {Frequency} {Representation}},
  booktitle = {{International} {Conference} on {Machine} {Learning}},
  author    = {Tang, Ling and Shen, Wen and Zhou, Zhanpeng and Chen, Yuefeng and Zhang, Quanshi},
  year      = {2023}
}

@inproceedings{singh2023hessian,
  title     = {The {Hessian} perspective into the {Nature} of {Convolutional} {Neural} {Networks}},
  booktitle = {{International} {Conference} on {Machine} {Learning}},
  author    = {Singh, Sidak Pal and Hofmann, Thomas and Schölkopf, Bernhard},
  year      = {2023}
}

@inproceedings{tan2021efficientnetv2,
  title     = {{EfficientNetV2}: {Smaller} {Models} and {Faster} {Training}},
  booktitle = {{International} {Conference} on {Machine} {Learning}},
  author    = {Tan, Mingxing and Le, Quoc},
  year      = {2021}
}

@inproceedings{zhang2019making,
  title     = {Making {Convolutional} {Networks} {Shift}-{Invariant} {Again}},
  booktitle = {{International} {Conference} on {Machine} {Learning}},
  author    = {Zhang, Richard},
  year      = {2019}
}

@inproceedings{singla2021skew,
  title     = {Skew Orthogonal Convolutions},
  author    = {Singla, Sahil and Feizi, Soheil},
  booktitle = {Proceedings of the 38th International Conference on Machine Learning},
  year      = {2021}
}

@inproceedings{behrmann2019invertible,
  author    = {Jens Behrmann and Will Grathwohl and Ricky T. Q. Chen and David Duvenaud and Jörn-Henrik Jacobsen},
  title     = {Invertible Residual Networks},
  booktitle = {Proceedings of the 37th International Conference on Machine Learning (ICML)},
  year      = {2019}
}

@inproceedings{meunier2022dynamical,
  title     = {A {Dynamical} {System} {Perspective} for {Lipschitz} {Neural} {Networks}},
  booktitle = {Proceedings of the 39th {International} {Conference} on {Machine} {Learning}},
  author    = {Meunier, Laurent and Delattre, Blaise J. and Araujo, Alexandre and Allauzen, Alexandre},
  year      = {2022}
}

@inproceedings{nair2010rectified,
  title     = {Rectified Linear Units Improve Restricted Boltzmann Machines},
  author    = {Nair, Vinod and Hinton, Geoffrey E},
  booktitle = {Proceedings of the 27th International Conference on Machine Learning \emph{(ICML)}},
  year      = {2010}
}

@inproceedings{maas2013rectifier,
  title     = {Rectifier Non-Linearities Improve Neural Network Acoustic Models},
  author    = {Andrew L. Maas and Awni Y. Hannun and Andrew Y. Ng},
  booktitle = {In ICML Workshop on Deep Learning for Audio, Speech and Language Processing},
  year      = {2013}
}

@inproceedings{ioffe2015batch,
  title     = {Batch Normalization: Accelerating Deep Network Training by Reducing Internal Covariate Shift},
  author    = {Ioffe, Sergey and Szegedy~, Christian},
  booktitle = {Proceedings of the 32nd International Conference on Machine Learning \emph{(ICML)}},
  year      = {2015}
}

@inproceedings{cisse2017parseval,
  title     = {Parseval Networks: Improving Robustness to Adversarial Examples},
  author    = {Cisse, Moustapha and Bojanowski, Piotr and Grave, Edouard and Dauphin, Yann and Usunier, Nicolas},
  booktitle = {Proceedings of the 34th International Conference on Machine Learning \emph{(ICML)}},
  year      = {2017}
}

@inproceedings{athalye2018obfuscated,
  title     = {Obfuscated Gradients Give a False Sense of Security: Circumventing Defenses to Adversarial Examples},
  author    = {Athalye, Anish and Carlini, Nicholas and Wagner, David},
  booktitle = {Proceedings of the 35th International Conference on Machine Learning \emph{(ICML)}},
  year      = {2018}
}

@inproceedings{wong2018provable,
  title     = {Provable defenses against adversarial examples via the convex outer adversarial polytope},
  author    = {Wong, Eric and Kolter, Zico},
  booktitle = {Proceedings of the 35th International Conference on Machine Learning \emph{(ICML)}},
  year      = {2018}
}

@inproceedings{cohen2019certified,
  title     = {Certified Adversarial Robustness via Randomized Smoothing},
  author    = {Jeremy Cohen and Elan Rosenfeld and Zico Kolter},
  booktitle = {Proceedings of the 36th International Conference on Machine Learning \emph{(ICML)}},
  year      = {2019}
}

@inproceedings{anil2019sorting,
  title     = {Sorting out Lipschitz Function Approximation},
  author    = {Anil, Cem and Lucas, James and Grosse, Roger},
  booktitle = {Proceedings of the 36th International Conference on Machine Learning \emph{(ICML)}},
  year      = {2019}
}

@inproceedings{ryu2019plug,
  title     = {Plug-and-Play Methods Provably Converge with Properly Trained Denoisers},
  author    = {Ryu, Ernest and Liu, Jialin and Wang, Sicheng and Chen, Xiaohan and Wang, Zhangyang and Yin, Wotao},
  booktitle = {Proceedings of the 36th International Conference on Machine Learning \emph{(ICML)}},
  year      = {2019}
}


%% file: bibliography/linear_algebra.bib
@article{friedland1991revisiting,
  title={Revisiting matrix squaring},
  author={Friedland, Shmuel},
  journal={Linear algebra and its applications},
  year={1991},
}

@book{horn2012matrix,
  title={Matrix Analysis},
  author={Horn, R.A. and Johnson, C.R.},
  year={2012},
  publisher={Cambridge University Press}
}


%% file: bibliography/neurips.bib
@article{shi2022efficiently,
  title   = {Efficiently {Computing} {Local} {Lipschitz} {Constants} of {Neural} {Networks} via {Bound} {Propagation}},
  journal = {Advances in Neural Information Processing Systems},
  author  = {Shi, Zhouxing and Wang, Yihan and Zhang, Huan and Kolter, J. Zico and Hsieh, Cho-Jui},
  year    = {2022}
}

@inproceedings{senderovich2022practical,
  title     = {Towards Practical Control of Singular Values of Convolutional Layers},
  author    = {Alexandra Senderovich and Ekaterina Bulatova and Anton Obukhov and Maxim Rakhuba},
  booktitle = {Advances in Neural Information Processing Systems (NeurIPS)},
  year      = {2022},
  note      = {36th Conference on Neural Information Processing Systems}
}

@inproceedings{goodfellow2014generative,
  title     = {Generative adversarial networks},
  author    = {Goodfellow, Ian and Pouget-Abadie, Jean and Mirza, Mehdi and Xu, Bing and Warde-Farley, David and Ozair, Sherjil and Courville, Aaron and Bengio, Yoshua},
  booktitle = {Advances in Neural Information Processing Systems},
  year      = {2014}
}

@article{ho2020denoising,
  title   = {Denoising diffusion probabilistic models},
  author  = {Ho, Jonathan and Jain, Ajay and Abbeel, Pieter},
  journal = {Advances in Neural Information Processing Systems},
  year    = {2020}
}

@inproceedings{chen2024diffusion,
  title     = {Diffusion Models are Certifiably Robust Classifiers},
  author    = {Huanran Chen and Yinpeng Dong and Shitong Shao and Zhongkai Hao and Xiao Yang and Hang Su and Jun Zhu},
  booktitle = {The Thirty-eighth Annual Conference on Neural Information Processing Systems},
  year      = {2024}
}

@inproceedings{yang2020closer,
  author    = {Yang, Yao-Yuan and Rashtchian, Cyrus and Zhang, Hongyang and Salakhutdinov, Russ R and Chaudhuri, Kamalika},
  booktitle = {Advances in Neural Information Processing Systems},
  editor    = {H. Larochelle and M. Ranzato and R. Hadsell and M.F. Balcan and H. Lin},
  title     = {A Closer Look at Accuracy vs. Robustness},
  year      = {2020}
}

@inproceedings{fu2023dreamsim,
  title     = {DreamSim: Learning New Dimensions of Human Visual Similarity using Synthetic Data},
  author    = {Stephanie Fu and Netanel Yakir Tamir and Shobhita Sundaram and Lucy Chai and Richard Zhang and Tali Dekel and Phillip Isola},
  booktitle = {Thirty-seventh Conference on Neural Information Processing Systems},
  year      = {2023}
}

@inproceedings{bubeck2021universal,
  title     = {A Universal Law of Robustness via Isoperimetry},
  author    = {Sebastien Bubeck and Mark Sellke},
  booktitle = {Advances in Neural Information Processing Systems},
  editor    = {A. Beygelzimer and Y. Dauphin and P. Liang and J. Wortman Vaughan},
  year      = {2021}
}

@inproceedings{santurkar2018does,
  title     = {How does batch normalization help optimization?},
  author    = {Santurkar, Shibani and Tsipras, Dimitris and Ilyas, Andrew and Madry, Aleksander},
  booktitle = {Advances in Neural Information Processing Systems (NeurIPS)},
  year      = {2018}
}

@inproceedings{wen2022how,
  title     = {How Does Sharpness-Aware Minimization Minimizes Sharpness?},
  author    = {Kaiyue Wen and Tengyu Ma and Zhiyuan Li},
  booktitle = {OPT 2022: Optimization for Machine Learning (NeurIPS 2022 Workshop)},
  year      = {2022}
}

@inproceedings{kwon2021asam,
  title     = {ASAM: Adaptive Sharpness-Aware Minimization for Scale-Invariant Learning of Deep Neural Networks},
  author    = {Kwon, Jungmin and Park, Hae Beom and Kim, Junho and Choi, Il Dong},
  booktitle = {Advances in Neural Information Processing Systems},
  year      = {2021}
}

@inproceedings{scholten2023hierarchical,
  title     = {Hierarchical Randomized Smoothing},
  author    = {Scholten, Yan and Schuchardt, Jan and Bojchevski, Aleksandar and G\"{u}nnemann, Stephan},
  booktitle = {Conference on Neural Information Processing Systems},
  year      = {2023}
}

@article{elsayed2018large,
  title   = {Large margin deep networks for classification},
  author  = {Elsayed, Gamaleldin and Krishnan, Dilip and Mobahi, Hossein and Regan, Kevin and Bengio, Samy},
  journal = {Advances in neural information processing systems},
  volume  = {31},
  year    = {2018}
}

@inproceedings{pautov2022smoothed,
  title     = {Smoothed Embeddings for Certified Few-Shot Learning},
  author    = {Mikhail Pautov and Olesya Kuznetsova and Nurislam Tursynbek and Aleksandr Petiushko and Ivan Oseledets},
  booktitle = {Advances in Neural Information Processing Systems},
  year      = {2022}
}

@inproceedings{wong2018scaling,
  title     = {Scaling provable adversarial defenses},
  author    = {Wong, Eric and Schmidt, Ludwig and Kolter, J Zico},
  booktitle = {Advances in Neural Information Processing Systems (NeurIPS)},
  year      = {2018}
}

@inproceedings{leino2021relaxing,
  title     = {Relaxing Local Robustness},
  author    = {Klas Leino and Matt Fredrikson},
  booktitle = {Neural Information Processing Systems (NIPS)},
  year      = {2021}
}

@inproceedings{hu2023scaling,
  title     = {Unlocking Deterministic Robustness Certification on ImageNet},
  author    = {Kai Hu and Andy Zou and Zifan Wang and Klas Leino and Matt Fredrikson},
  booktitle = {Conference on Neural Information Processing Systems},
  year      = {2023}
}

@article{huang2021local,
  title     = {Training Certifiably Robust Neural Networks with Efficient Local Lipschitz Bounds},
  author    = {Huang, Yujia and Zhang, Huan and Shi, Yuanyuan and Kolter, J Zico and Anandkumar, Anima},
  booktitle = {Advances in Neural Information Processing Systems},
  year      = {2021}
}

@inproceedings{zhang2022rethinking,
  title     = {Rethinking {Lipschitz} {Neural} {Networks} and {Certified} {Robustness}: {A} {Boolean} {Function} {Perspective}},
  author    = {Zhang, Bohang and Jiang, Du and He, Di and Wang, Liwei},
  booktitle = {Advances in Neural Information Processing Systems},
  year      = {2022}
}

@inproceedings{jordan2020exactly,
  title     = {Exactly {Computing} the {Local} {Lipschitz} {Constant} of {ReLU} {Networks}},
  booktitle = {Advances in {Neural} {Information} {Processing} {Systems}},
  author    = {Jordan, Matt and Dimakis, Alexandros G},
  year      = {2020}
}

@inproceedings{tolstikhin2021mlpmixer,
  title     = {{MLP}-Mixer: An all-{MLP} Architecture for Vision},
  author    = {Ilya Tolstikhin and Neil Houlsby and Alexander Kolesnikov and Lucas Beyer and Xiaohua Zhai and Thomas Unterthiner and Jessica Yung and Andreas Peter Steiner and Daniel Keysers and Jakob Uszkoreit and Mario Lucic and Alexey Dosovitskiy},
  booktitle = {Advances in Neural Information Processing Systems},
  year      = {2021}
}

@inproceedings{zhang2018local,
  title     = {On the Local Hessian in Back-propagation},
  author    = {Zhang, Huishuai and Chen, Wei and Liu, Tie-Yan},
  booktitle = {Conference on Neural Information Processing Systems},
  year      = {2018}
}

@inproceedings{laha2018controllable,
  title     = {On {Controllable} {Sparse} {Alternatives} to {Softmax}},
  booktitle = {Advances in {Neural} {Information} {Processing} {Systems}},
  author    = {Laha, Anirban and Chemmengath, Saneem Ahmed and Agrawal, Priyanka and Khapra, Mitesh and Sankaranarayanan, Karthik and Ramaswamy, Harish G},
  year      = {2018}
}

@inproceedings{bethune2022pay,
  author    = {B\'{e}thune, Louis and Boissin, Thibaut and Serrurier,   Mathieu  and Mamalet, Franck and Friedrich, Corentin and Gonzalez Sanz, Alberto},
  booktitle = {Advances in Neural Information Processing Systems},
  title     = {Pay attention to your loss : understanding misconceptions about Lipschitz neural networks},
  year      = {2022}
}

@inproceedings{krogh1992simple,
  title     = {A Simple Weight Decay can Improve Generalization},
  author    = {Krogh, Anders and Hertz, John A},
  booktitle = {Advances in Neural Information Processing Systems \emph{(NeurIPS)}},
  year      = {1992}
}

@incollection{krizhevsky2012imagenet,
  title     = {ImageNet Classification with Deep Convolutional Neural Networks},
  author    = {Alex Krizhevsky and Sutskever, Ilya and Hinton, Geoffrey E},
  booktitle = {Advances in Neural Information Processing Systems \emph{(NeurIPS)}},
  year      = {2012}
}

@inproceedings{bartlett2017spectrally,
  title     = {Spectrally-normalized margin bounds for neural networks},
  author    = {Bartlett, Peter L and Foster, Dylan J and Telgarsky, Matus J},
  booktitle = {Advances in Neural Information Processing Systems \emph{(NeurIPS)}},
  year      = {2017}
}

@inproceedings{gulrajani2017improved,
  title     = {Improved training of wasserstein gans},
  author    = {Gulrajani, Ishaan and Ahmed, Faruk and Arjovsky, Martin and Dumoulin, Vincent and Courville, Aaron C},
  booktitle = {Advances in Neural Information Processing Systems \emph{(NeurIPS)}},
  year      = {2017}
}

@inproceedings{vaswani2017attention,
  title     = {Attention is all you need},
  author    = {Vaswani, Ashish and Shazeer, Noam and Parmar, Niki and Uszkoreit, Jakob and Jones, Llion and Gomez, Aidan N and Kaiser, {\L}ukasz and Polosukhin, Illia},
  booktitle = {Advances in Neural Information Processing Systems \emph{(NeurIPS)}},
  year      = {2017}
}

@inproceedings{tsuzuku2018lipschitz,
  title     = {Lipschitz-margin training: Scalable certification of perturbation invariance for deep neural networks},
  author    = {Tsuzuku, Yusuke and Sato, Issei and Sugiyama, Masashi},
  booktitle = {Advances in Neural Information Processing Systems \emph{(NeurIPS)}},
  year      = {2018}
}

@incollection{scaman2018lipschitz,
  title     = {Lipschitz regularity of deep neural networks: analysis and efficient estimation},
  author    = {Virmaux, Aladin and Scaman, Kevin},
  booktitle = {Advances in Neural Information Processing Systems \emph{(NeurIPS)}},
  year      = {2018}
}

@inproceedings{pinot2019theoretical,
  title     = {Theoretical Evidence for Adversarial Robustness through Randomization},
  author    = {Pinot, Rafael and Meunier, Laurent and Araujo, Alexandre and Kashima, Hisashi and Yger, Florian and Gouy-Pailler, Cedric and Atif, Jamal},
  booktitle = {Advances in Neural Information Processing Systems \emph{(NeurIPS)}},
  year      = {2019}
}

@incollection{li2019preventing,
  title     = {Preventing Gradient Attenuation in Lipschitz Constrained Convolutional Networks},
  author    = {Li, Qiyang and Haque, Saminul and Anil, Cem and Lucas, James and Grosse, Roger B and Jacobsen, Joern-Henrik},
  booktitle = {Advances in Neural Information Processing Systems \emph{(NeurIPS)}},
  year      = {2019}
}

@incollection{fazlyab2019efficient,
  title     = {Efficient and Accurate Estimation of Lipschitz Constants for Deep Neural Networks},
  author    = {Fazlyab, Mahyar and Robey, Alexander and Hassani, Hamed and Morari, Manfred and Pappas, George},
  booktitle = {Advances in Neural Information Processing Systems \emph{(NeurIPS)}},
  year      = {2019}
}

@inproceedings{paszke2019pytorch,
  title     = {PyTorch: An Imperative Style, High-Performance Deep Learning Library},
  author    = {Paszke, Adam and Gross, Sam and Massa, Francisco and Lerer, Adam and Bradbury, James and Chanan, Gregory and Killeen, Trevor and Lin, Zeming and Gimelshein, Natalia and Antiga, Luca and Desmaison, Alban and Kopf, Andreas and Yang, Edward and DeVito, Zachary and Raison, Martin and Tejani, Alykhan and Chilamkurthy, Sasank and Steiner, Benoit and Fang, Lu and Bai, Junjie and Chintala, Soumith},
  booktitle = {Advances in Neural Information Processing Systems \emph{(NeurIPS)}},
  year      = {2019}
}

@article{salman2020denoised,
  title   = {Denoised smoothing: A provable defense for pretrained classifiers},
  author  = {Salman, Hadi and Sun, Mingjie and Yang, Greg and Kapoor, Ashish and Kolter, J Zico},
  journal = {Advances in Neural Information Processing Systems},
  year    = {2020}
}

@inproceedings{salman2019provably,
  title     = {Provably robust deep learning via adversarially trained smoothed classifiers},
  author    = {Salman, Hadi and Li, Jerry and Razenshteyn, Ilya and Zhang, Pengchuan and Zhang, Huan and Bubeck, Sebastien and Yang, Greg},
  booktitle = {Advances in Neural Information Processing Systems \emph{(NeurIPS)}},
  year      = {2019}
}

@inproceedings{brown2020language,
  title     = {Language Models are Few-Shot Learners},
  author    = {
               Brown, Tom B. and
               Mann, Benjamin and
               Ryder, Nick and
               Subbiah, Melanie and
               Kaplan, Jared and
               Dhariwal, Prafulla and
               Neelakantan, Arvind and
               Shyam, Pranav and
               Sastry, Girish and
               Askell, Amanda and
               Agarwal, Sandhini and
               Herbert-Voss, Arieland
               Krueger, Gretchen and
               Henighan, Tom and
               Child, Rewon and
               Ramesh, Aditya and
               Ziegler, Daniel M. and
               Wu, Jeffrey and
               Winter, Clemens and
               Hesse, Christopher and
               Chen, Mark and
               Sigler, Eric and
               Litwin, Mateusz and
               Gray, Scott and
               Chess, Benjamin and
               Clark, Jack and
               Berner, Christopher and
               McCandlish, Sam and
               Radford, Alec and
               Sutskever, Ilya and
               Amodei, Dario},
  booktitle = {Advances in Neural Information Processing Systems \emph{(NeurIPS)}},
  year      = {2020}
}


%% file: bibliography/other.bib
@article{hochreiter1997long,
  title   = {Long short-term memory},
  author  = {Hochreiter, Sepp and Schmidhuber, J{\"u}rgen},
  journal = {Neural computation},
  year    = {1997}
}

@inproceedings{jia2019certified,
  title     = {Certified robustness to adversarial word substitutions},
  author    = {Jia, Robin and Liang, Percy},
  booktitle = {Empirical Methods in Natural Language Processing (EMNLP)},
  year      = {2019}
}

@article{CombettesPesquet2020_VI,
  author  = {Patrick L. Combettes and Jean{-}Christophe Pesquet},
  title   = {Deep neural network structures solving variational inequalities},
  journal = {Set-Valued and Variational Analysis},
  volume  = {28},
  number  = {3},
  pages   = {491--518},
  year    = {2020},
  month   = sep
}

@article{CombettesPesquet2020_LipCert,
  author  = {Patrick L. Combettes and Jean{-}Christophe Pesquet},
  title   = {Lipschitz certificates for layered network structures driven by averaged activation operators},
  journal = {SIAM Journal on Mathematics of Data Science},
  volume  = {2},
  number  = {2},
  pages   = {529--557},
  year    = {2020},
  month   = jun
}

@inproceedings{prach2024lipschitz,
  author    = {Benedikt Prach and Florian Brau and Giuseppe Buttazzo and Christoph H. Lampert},
  title     = {1-Lipschitz Layers Compared: Memory, Speed, and Certifiable Robustness},
  booktitle = {Proceedings of the IEEE/CVF Conference on Computer Vision and Pattern Recognition (CVPR)},
  year      = {2024}
}

@inproceedings{ablin2022fast,
  title     = {Fast and accurate optimization on the orthogonal manifold without retraction},
  author    = {Ablin, Pierre and Peyr{\'e}, Gabriel},
  booktitle = {International Conference on Artificial Intelligence and Statistics},
  year      = {2022}
}

@article{drucker1992double,
  author  = {Harris Drucker and Yann LeCun},
  title   = {Improving generalization performance using double backpropagation},
  journal = {IEEE Transactions on Neural Networks},
  year    = {1992}
}

@inproceedings{huang2019achieving,
  title     = {Achieving certified robustness to word substitutions with differential privacy},
  author    = {Huang, Po-Sen and He, He and Chen, Anshul and Li, Jinfeng and Zettlemoyer, Luke and Chang, Kai-Wei},
  booktitle = {Annual Meeting of the Association for Computational Linguistics (ACL)},
  year      = {2019}
}

@article{halko2011finding,
  title   = {Finding structure with randomness: Probabilistic algorithms for constructing approximate matrix decompositions},
  author  = {Halko, Nathan and Martinsson, Per-Gunnar and Tropp, Joel A.},
  journal = {SIAM Review},
  year    = {2011}
}

@article{chellapilla2006high,
  author  = {K. Chellapilla and Siddharth Puri and Patrice Simard},
  title   = {High Performance Convolutional Neural Networks for Document Processing},
  journal = {Tenth International Workshop on Frontiers in Handwriting Recognition},
  year    = {2006}
}

@article{jerrum1986median,
  author  = {Mark Jerrum, Leslie G. Valiant, Vijay V. Vazirani},
  title   = {Random generation of combinatorial structures from a uniform distribution},
  journal = {Theoretical Computer Science},
  year    = {1986}
}

@book{nemirovsky1983problem,
  author    = {Arkadi S. Nemirovsky, David B. Yudin},
  title     = {Problem Complexity and Method Efficiency in Optimization},
  publisher = {John Wiley \& Sons},
  year      = {1983}
}

@article{lugosi2019mean,
  author  = {Gábor Lugosi and Shahar Mendelson},
  title   = {Mean Estimation and Regression Under Heavy-Tailed Distributions: A Survey},
  journal = {Foundations of Computational Mathematics},
  year    = {2019}
}

@article{von2007tutorial,
  title   = {A tutorial on spectral clustering},
  author  = {Von Luxburg, Ulrike},
  journal = {Statistics and Computing},
  year    = {2007}
}

@article{devroye1979distribution_free,
  title   = {Distribution-free inequalities for the deleted and holdout error estimates},
  author  = {Devroye, Luc and Wagner, T. J.},
  journal = {IEEE Transactions on Information Theory},
  year    = {1979}
}

@inproceedings{bubeck2021law,
  title     = {A Law of Robustness for Two-Layers Neural Networks},
  author    = {Bubeck, S{\'e}bastien and Li, Yuanzhi and Nagaraj, Dheeraj},
  booktitle = {Proceedings of the 34th Annual Conference on Learning Theory},
  year      = {2021},
  publisher = {PMLR}
}

@book{nesterov2003introductory,
  title     = {Introductory lectures on convex optimization: A basic course},
  author    = {Nesterov, Yurii},
  year      = {2003},
  publisher = {Springer Science \& Business Media}
}

@inproceedings{ronneberger2015unet,
  title     = {U-Net: Convolutional Networks for Biomedical Image Segmentation},
  author    = {Ronneberger, Olaf and Fischer, Philipp and Brox, Thomas},
  booktitle = {Medical Image Computing and Computer-Assisted Intervention (MICCAI)},
  year      = {2015},
  publisher = {Springer}
}

@article{falk2019unet,
  title   = {U-Net: deep learning for cell counting, detection, and morphometry},
  author  = {Falk, Thorsten and Mai, Dominic and Bensch, Robert and Çiçek, Özgün and Abdulkadir, Ahmed and Marrakchi, Yassine and Böhm, Anton and Deubner, Jan and Jäckel, Zoe and Seiwald, Katharina and others},
  journal = {Nature Methods},
  year    = {2019}
}

@article{silver2016alphago,
  title   = {Mastering the game of Go with deep neural networks and tree search},
  author  = {Silver, David and Huang, Aja and Maddison, Chris J and Guez, Arthur and Sifre, Laurent and van den Driessche, George and Schrittwieser, Julian and Antonoglou, Ioannis and Panneershelvam, Veda and Lanctot, Marc and others},
  journal = {Nature},
  year    = {2016}
}

@article{valiant1984pac,
  title   = {A theory of the learnable},
  author  = {Valiant, Leslie G},
  journal = {Communications of the ACM},
  year    = {1984}
}

@inproceedings{mcallester1999pac,
  title     = {Some PAC-Bayesian theorems},
  author    = {McAllester, David A},
  booktitle = {Proceedings of the Conference on Computational Learning Theory (COLT)},
  year      = {1999}
}

@article{jumper2021alphafold,
  title   = {Highly accurate protein structure prediction with AlphaFold},
  author  = {Jumper, John and Evans, Richard and Pritzel, Alexander and Green, Tim and Figurnov, Michael and Ronneberger, Olaf and Tunyasuvunakool, Kathryn and Bates, Russ and {\v{Z}}{\'\i}dek, Augustin and Potapenko, Anna and others},
  journal = {Nature},
  year    = {2021}
}

@article{sokolic2017robust,
  title   = {{Robust Large Margin Deep Neural Networks}},
  author  = {Sokoli{\'c}, Jure and Giryes, Raja and Sapiro, Guillermo and Rodrigues, Miguel R. D.},
  journal = {IEEE Transactions on Signal Processing},
  year    = {2017}
}

@inproceedings{croce2020reliable,
  title     = {Reliable evaluation of adversarial robustness with an ensemble of diverse parameter-free attacks},
  author    = {Croce, Francesco and Hein, Matthias},
  booktitle = {International Conference on Machine Learning (ICML)},
  year      = {2020}
}

@inproceedings{dwork2006calibrating,
  title     = {Calibrating noise to sensitivity in private data analysis},
  author    = {Dwork, Cynthia and McSherry, Frank and Nissim, Kobbi and Smith, Adam},
  booktitle = {Theory of Cryptography Conference},
  year      = {2006}
}

@article{devroye1979performance_bounds,
  title   = {Distribution-free performance bounds for potential function rules},
  author  = {Devroye, Luc and Wagner, T. J.},
  journal = {IEEE Transactions on Information Theory},
  year    = {1979}
}

@book{bonnans1996perturbation,
  title     = {Perturbation Analysis of Optimization Problems},
  author    = {Bonnans, J. Fr{\'e}d{\'e}ric and Shapiro, Alexander},
  year      = {1996},
  publisher = {Springer-Verlag},
  address   = {New York}
}

@article{vapnik1971uniform,
  author    = {Vapnik, Vladimir N. and Chervonenkis, A. Ya.},
  title     = {On the Uniform Convergence of Relative Frequencies of Events to Their Probabilities},
  journal   = {Theory of Probability and Its Applications},
  year      = {1971},
  publisher = {Society for Industrial and Applied Mathematics}
}

@article{kearns1999stability,
  title   = {Algorithmic stability and sanity-check bounds for leave-one-out cross-validation},
  author  = {Kearns, Michael and Ron, Dana},
  journal = {Neural Computation},
  year    = {1999}
}

@article{tikhonov1943regularization,
  author  = {Tikhonov, Andrey N.},
  title   = {On the Stability of Inverse Problems},
  journal = {Doklady Akademii Nauk SSSR},
  year    = {1943},
  note    = {In Russian}
}

@article{vonneumann1947numerical,
  author  = {Von Neumann, John and Goldstine, Herman H.},
  title   = {Numerical Inverting of Matrices of High Order},
  journal = {Bulletin of the American Mathematical Society},
  year    = {1947}
}

@book{saltelli2000sensitivity,
  author    = {Saltelli, Andrea and Chan, Karen and Scott, E. Marian},
  title     = {Sensitivity Analysis},
  publisher = {John Wiley \& Sons},
  year      = {2000}
}

@book{cramer1946mathematical,
  author    = {Cramér, Harald},
  title     = {Mathematical Methods of Statistics},
  publisher = {Princeton University Press},
  year      = {1946}
}

@book{lyapunov1892stability,
  author    = {Lyapunov, Aleksandr M.},
  title     = {The General Problem of the Stability of Motion},
  year      = {1892},
  publisher = {Taylor \& Francis (Translated in 1992)},
  note      = {Originally published in Russian, Kharkov, 1892}
}

@article{haber2018stable,
  author  = {Haber, Eldad and Ruthotto, Lars},
  title   = {Stable architectures for deep neural networks},
  journal = {Inverse Problems},
  year    = {2018}
}

@article{polyak1964some,
  title   = {Some methods of speeding up the convergence of iteration methods},
  author  = {Polyak, Boris T},
  journal = {USSR Computational Mathematics and Mathematical Physics},
  year    = {1964}
}

@misc{tieleman2012lecture,
  title  = {Lecture 6.5---RMSProp: Divide the gradient by a running average of its recent magnitude},
  author = {Tieleman, Tijmen and Hinton, Geoffrey},
  year   = {2012},
  note   = {Coursera: Neural Networks for Machine Learning, \url{https://www.cs.toronto.edu/~tijmen/csc321/slides/lecture_slides_lec6.pdf}}
}

@inproceedings{duchi2011adaptive,
  title     = {Adaptive subgradient methods for online learning and stochastic optimization},
  author    = {Duchi, John and Hazan, Elad and Singer, Yoram},
  booktitle = {Journal of Machine Learning Research (JMLR)},
  year      = {2011}
}

@misc{yolov5,
  title  = {YOLOv5},
  author = {Glenn Jocher et al.},
  year   = {2020}
}

@inproceedings{weber2023rab,
  author    = {Maurice Weber and Xiaojun Xu and Bojan Karla{\v{s}} and Ce Zhang and Bo Li},
  title     = {RAB: Provable Robustness Against Backdoor Attacks},
  year      = {2023},
  publisher = {IEEE}
}

@misc{haldar2020adversarial,
  author       = {Siddhant Haldar},
  title        = {Gradient-based Adversarial Attacks: An Introduction},
  howpublished = {\url{https://medium.com/swlh/gradient-based-adversarial-attacks-an-introduction-123abc456def}},
  year         = {2020}
}

@inproceedings{yang2020randomized,
  title     = {Randomized Smoothing of All Shapes and Sizes},
  author    = {Yang, Greg and Duan, Tony and Hu, J. Edward and Salman, Hadi and Razenshteyn, Ilya and Li, Jerry},
  booktitle = {Proceedings of the 37th International Conference on Machine Learning},
  year      = {2020}
}

@phdthesis{bethune2024lipschitz,
  author = {Louis B{\'e}thune},
  title  = {Deep learning with Lipschitz constraints},
  year   = {2024},
  school = {Université de Toulouse}
}

@book{bishop2006pattern,
  title     = {Pattern Recognition and Machine Learning},
  author    = {Bishop, Christopher M.},
  year      = {2006},
  publisher = {Springer}
}

@inproceedings{shi2018neural,
  title     = {Neural Lander: Stable Drone Landing Control Using Learned Dynamics},
  author    = {Shi, Guanya and Shi, Xiyang and O’Connell, Michael and Yu, Rose and Azizzadenesheli, Kamyar and Anandkumar, Anima and Yue, Yisong and Chung, Soon-Jo},
  booktitle = {2019 International Conference on Robotics and Automation (ICRA)},
  year      = {2018}
}

@article{brunke2022safe,
  title   = {Safe Learning in Robotics: From Learning-Based Control to Safe Reinforcement Learning},
  author  = {Brunke, Lukas and Greeff, Michael and Hall, Andrew W and Yuan, Zhaocong and Zhou, Siwei and Panerati, Jacopo and Schoellig, Angela P},
  journal = {Annual Review of Control, Robotics, and Autonomous Systems},
  year    = {2022}
}

@book{nocedal2006numerical,
  title     = {Numerical Optimization},
  author    = {Nocedal, Jorge and Wright, Stephen J.},
  year      = {2006},
  publisher = {Springer}
}

@book{dunn1961multiple,
  title     = {Multiple comparisons among means},
  author    = {Dunn, Olive Jean},
  year      = {1961},
  publisher = {Stanford University Press}
}

@book{hochberg1987multiple,
  title     = {Multiple comparison procedures},
  author    = {Hochberg, Yosef and Tamhane, Ajit C},
  year      = {1987},
  publisher = {Wiley}
}

@article{benjamini1995controlling,
  title   = {Controlling the false discovery rate: a practical and powerful approach to multiple testing},
  author  = {Benjamini, Yoav and Hochberg, Yosef},
  journal = {Journal of the Royal Statistical Society: Series B (Methodological)},
  year    = {1995}
}

@book{pontryagin1962mathematical,
  title     = {The Mathematical Theory of Optimal Processes},
  author    = {Pontryagin, Lev Semyonovich and Boltyanskii, Vladimir Grigorievich and Gamkrelidze, Revaz V. and Mishchenko, Evgenii Fedorovich},
  year      = {1962},
  publisher = {John Wiley \& Sons},
  note      = {Translated from the Russian by K. N. Trirogoff and edited by L. W. Neustadt}
}

@inproceedings{lecuyer2019certified,
  title     = {Certified robustness to adversarial examples with differential privacy},
  author    = {Lecuyer, Mathias and Atlidakis, Vaggelis and Geambasu, Roxana and Hsu, Daniel and Jana, Suman},
  booktitle = {IEEE symposium on security and privacy (SP)},
  year      = {2019}
}

@book{stein1970singular,
  title     = {Singular Integrals and Differentiability Properties of Functions},
  author    = {Stein, Elias M.},
  year      = {1970},
  publisher = {Princeton University Press},
  address   = {Princeton, NJ}
}

@book{zayed1996handbook,
  title     = {Handbook of Function and Generalized Function Transformations},
  author    = {Zayed, Ahmed I.},
  year      = {1996},
  publisher = {CRC Press}
}

@article{bousquet2002stability,
  title   = {Stability and generalization},
  author  = {Bousquet, Olivier and Elisseeff, Andr{\'e}},
  journal = {Journal of machine learning research},
  year    = {2002}
}

@article{wang2022robust,
  author  = {Wang, Wei and Dang, Zheng and Hu, Yinlin and Fua, Pascal and Salzmann, Mathieu},
  journal = {IEEE Transactions on Pattern Analysis and Machine Intelligence},
  title   = {Robust Differentiable SVD},
  year    = {2022}
}

@article{kublanovskaya1962algorithms,
  title   = {On some algorithms for the solution of the complete eigenvalue problem},
  journal = {USSR Computational Mathematics and Mathematical Physics},
  year    = {1962},
  author  = {V.N. Kublanovskaya}
}

@article{lanczos1950iteration,
  title   = {An iteration method for the solution of the eigenvalue problem of linear differential and integral operators},
  author  = {Lanczos, Cornelius},
  journal = {Journal of Research of the National Bureau of Standards},
  year    = {1950}
}

@article{lehoucq1996deflation,
  title   = {Deflation techniques for an implicitly restarted {Arnoldi} iteration},
  author  = {Lehoucq, Richard B and others},
  journal = {SIAM Journal on Matrix Analysis and Applications},
  year    = {1996}
}

@article{arnoldi1951principle,
  title   = {The principle of minimized iterations in the solution of the matrix eigenvalue problem},
  author  = {Arnoldi, Walter Edwin},
  journal = {Quarterly of applied mathematics},
  year    = {1951}
}

@misc{openai2023simpleevals,
  author       = {OpenAI},
  title        = {Simple Evals: OpenAI Evaluation Framework},
  howpublished = {\url{https://github.com/openai/simple-evals}},
  year         = {2023}
}

@inproceedings{abadi2016tensorflow,
  title  = {TensorFlow: A system for large-scale machine learning},
  author = {Abadi, Mart{\'\i}n and Barham, Paul and Chen, Jianmin and Chen, Zhifeng and Davis, Andy and Dean, Jeffrey and Devin, Matthieu and Ghemawat, Sanjay and Irving, Geoffrey and Isard, Michael and Kudlur, Manjunath and Levenberg, Josh and Monga, Rajat and Moore, Sherry and Murray, Derek G. and Steiner, Benoit and Tucker, Paul and Vasudevan, Vijay and Warden, Pete and Wicke, Martin and Yu, Yuan and Zheng, Xiaoqiang},
  year   = {2016}
}

@article{nesterov2017random,
  title   = {Random Gradient-Free Minimization of Convex Functions},
  author  = {Yurii Nesterov and Vladimir Spokoiny},
  journal = {Foundations of Computational Mathematics},
  year    = {2017}
}

@inproceedings{yao2020pyhessian,
  title     = {{PyHessian}: Neural Networks through the Lens of the Hessian},
  author    = {Yao, Zhewei and Gholami, Amir and Shen, Sheng and Keutzer, Kurt and Mahoney, Michael W.},
  booktitle = {IEEE International Conference on Big Data},
  year      = {2020}
}

@article{newman2021train,
  author  = {Newman, Elizabeth and Ruthotto, Lars and Hart, Joseph and van Bloemen Waanders, Bart},
  title   = {Train Like a (Var)Pro: Efficient Training of Neural Networks with Variable Projection},
  journal = {SIAM Journal on Mathematics of Data Science},
  year    = {2021}
}

@inproceedings{choromanska2015loss,
  title     = {The loss surfaces of multilayer networks},
  author    = {Choromanska, A. and Henaff, M. and Mathieu, M. and Arous, G. Ben and LeCun, Y.},
  booktitle = {Artificial intelligence and statistics},
  year      = {2015}
}

@article{chaudhari2019entropy,
  title   = {Entropy-sgd: Biasing gradient descent into wide valleys},
  author  = {Chaudhari, P. and Choromanska, A. and Soatto, S. and LeCun, Y. and Baldassi, C. and Borgs, C.and Chayes, J. and Sagun, L. and Zecchina, R.},
  journal = {Journal of Statistical Mechanics: Theory and Experiment},
  year    = {2019}
}

@article{hochreiter1997flat,
  title   = {Flat minima},
  author  = {Hochreiter, Sepp and Schmidhuber, J{\"u}rgen},
  journal = {Neural computation},
  year    = {1997}
}

@article{srivastava2014dropout,
  title   = {Dropout: A simple way to prevent neural networks from overfitting},
  author  = {Srivastava, Nitish and Hinton, Geoffrey and Krizhevsky, Alex and Sutskever, Ilya and Salakhutdinov, Ruslan},
  journal = {Journal of Machine Learning Research},
  year    = {2014}
}

@article{bishop1995training,
  author  = {Bishop, Chris M.},
  journal = {Neural Computation},
  title   = {Training with Noise is Equivalent to Tikhonov Regularization},
  year    = {1995}
}

@article{stein1981annals,
  author = {Charles M. Stein},
  title  = {Estimation of the Mean of a Multivariate Normal Distribution},
  year   = {1981}
}

@article{henderson1981TheVM,
  title   = {The Vec-Permutation Matrix, the Vec Operator and Kronecker Products: A Review},
  author  = {Harold V. Henderson and Shayle R. Searle},
  journal = {Linear \& Multilinear Algebra},
  year    = {1981}
}

@inproceedings{massart2007concentration,
  title     = {Concentration inequalities and model selection},
  author    = {Pascal Massart},
  year      = {2007},
  booktitle = {{\'E}cole d'{\'e}t{\'e} de probabilit{\'e}s de Saint-Flour}
}

@article{maurer2009empirical,
  title   = {Empirical bernstein bounds and sample variance penalization},
  author  = {Maurer, Andreas and Pontil, Massimiliano},
  journal = {Conference on Learning Theory},
  year    = {2009}
}

@article{pal2023understanding,
  title   = {Understanding {Noise}-{Augmented} {Training} for {Randomized} {Smoothing}},
  journal = {Transactions on Machine Learning Research},
  author  = {Pal, Ambar and Sulam, Jeremias},
  year    = {2023}
}

@article{gouk2021regularisation,
  title   = {Regularisation of neural networks by enforcing lipschitz continuity},
  author  = {Gouk, Henry and Frank, Eibe and Pfahringer, Bernhard and Cree, Michael J},
  journal = {Machine Learning},
  year    = {2021}
}

@inproceedings{conneau2017very,
  address   = {Valencia, Spain},
  title     = {Very {Deep} {Convolutional} {Networks} for {Text} {Classification}},
  booktitle = {Proceedings of the 15th {Conference} of the {European} {Chapter} of the {Association} for {Computational} {Linguistics}: {Volume} 1, {Long} {Papers}},
  author    = {Conneau, Alexis and Schwenk, Holger and Barrault, Loïc and Lecun, Yann},
  year      = {2017}
}

@book{boucheron2013concentration,
  author    = {Boucheron, Stéphane and Lugosi, Gábor and Massart, Pascal},
  title     = {{Concentration Inequalities: A Nonasymptotic Theory of Independence}},
  publisher = {Oxford University Press},
  year      = {2013}
}

@article{levine2019certifiably,
  author = {Alexander Levine and
            Sahil Singla and
            Soheil Feizi},
  year   = {2019},
  title  = {Certifiably Robust Interpretation in Deep Learning},
  url    = {http://arxiv.org/abs/1905.12105}
}

@article{pfister2019bounding,
  title   = {Bounding multivariate trigonometric polynomials},
  author  = {Luke Pfister and Yoram Bresler},
  year    = {2019},
  journal = {IEEE Transactions on Signal Processing}
}

@article{el_mehdi2022existence,
  author  = {El Mehdi Achour and François Malgouyres and Franck Mamalet},
  title   = {Existence, Stability and Scalability of Orthogonal Convolutional Neural Networks},
  journal = {Journal of Machine Learning Research},
  year    = {2022}
}

@article{li2022survey,
  author  = {Li, Zewen and Liu, Fan and Yang, Wenjie and Peng, Shouheng and Zhou, Jun},
  journal = {IEEE Trans Neural Netw Learn Syst},
  title   = {A Survey of Convolutional Neural Networks: Analysis, Applications, and Prospects},
  year    = {2022}
}

@article{zhang2019graph,
  author  = {Zhang, Si and Tong, Hanghang and Xu, Jiejun and Maciejewski, Ross},
  year    = {2019},
  title   = {Graph convolutional networks: a comprehensive review},
  journal = {Computational Social Networks}
}

@article{mallat2016understanding,
  title   = {Understanding deep convolutional networks},
  journal = {Philosophical Transactions of the Royal Society A: Mathematical, Physical and Engineering Sciences},
  author  = {Mallat, Stéphane},
  year    = {2016}
}

@book{goodfellow2016deep,
  title     = {Deep Learning},
  author    = {Ian Goodfellow and Yoshua Bengio and Aaron Courville},
  publisher = {MIT Press},
  year      = {2016}
}

@book{jain1989fundamentals,
  title     = {Fundamentals of digital image processing},
  author    = {Jain, Anil K},
  publisher = {Englewood Cliffs, NJ: Prentice Hall},
  year      = {1989}
}

@inproceedings{carlini2017towards,
  title     = {Towards evaluating the robustness of neural networks},
  author    = {Carlini, Nicholas and Wagner, David},
  booktitle = {IEEE Symposium on Security and Privacy},
  year      = {2017}
}

@article{lecun1998gradient,
  title   = {Gradient-based learning applied to document recognition},
  author  = {LeCun, Yann and Bottou, L{\'e}on and Bengio, Yoshua and Haffner, Patrick},
  journal = {Proceedings of the IEEE},
  year    = {1998}
}

@inproceedings{lecuyer2018certified,
  author    = {M. Lecuyer and V. Atlidakais and R. Geambasu and D. Hsu and S. Jana},
  booktitle = {2019 IEEE Symposium on Security and Privacy (SP)},
  title     = {Certified Robustness to Adversarial Examples with Differential Privacy},
  year      = {2018}
}

@inproceedings{devlin2019bert,
  title     = {BERT: Pre-training of Deep Bidirectional Transformers for Language Understanding},
  author    = {Devlin, Jacob  and Chang, Ming-Wei  and Lee, Kenton  and Toutanova, Kristina},
  booktitle = {Proceedings of the 2019 Conference of the North American Chapter of the Association for Computational Linguistics: Human Language Technologies \emph{(NAACL)}},
  year      = {2019}
}

@book{golub1996matrix,
  title     = {Matrix Computations},
  author    = {Golub, G.H. and Van Loan, C.F. and Van Loan, C.F. and Van Loan, P.C.F.},
  series    = {Johns Hopkins Studies in the Mathematical Sciences},
  publisher = {Johns Hopkins University Press},
  year      = {1996}
}

@article{ahmad2022developing,
  title   = {Developing future human-centered smart cities: Critical analysis of smart city security, Data management, and Ethical challenges},
  journal = {Computer Science Review},
  year    = {2022},
  author  = {Kashif Ahmad and Majdi Maabreh and Mohamed Ghaly and Khalil Khan and Junaid Qadir and Ala Al-Fuqaha}
}

@article{bjorck1971iterative,
  title     = {An iterative algorithm for computing the best estimate of an orthogonal matrix},
  author    = {Bj{\"o}rck, {\AA}ke and Bowie, Clazett},
  journal   = {SIAM Journal on Numerical Analysis},
  publisher = {SIAM},
  year      = {1971}
}

@article{raffel2020exploring,
  title   = {Exploring the Limits of Transfer Learning with a Unified Text-to-Text Transformer},
  author  = {Colin Raffel and Noam Shazeer and Adam Roberts and Katherine Lee and Sharan Narang and Michael Matena and Yanqi Zhou and Wei Li and Peter J. Liu},
  journal = {Journal of Machine Learning Research},
  year    = {2020}
}

@techreport{page1999pagerank,
  title       = {The PageRank citation ranking: Bringing order to the web.},
  author      = {Page, Lawrence and Brin, Sergey and Motwani, Rajeev and Winograd, Terry},
  institution = {Stanford InfoLab},
  year        = {1999}
}

@inproceedings{croce2021robustbench,
  title     = {RobustBench: a standardized adversarial robustness benchmark},
  author    = {Francesco Croce and Maksym Andriushchenko and Vikash Sehwag and Edoardo Debenedetti and Nicolas Flammarion and Mung Chiang and Prateek Mittal and Matthias Hein},
  booktitle = {Thirty-fifth Conference on Neural Information Processing Systems Datasets and Benchmarks Track (Round 2)},
  year      = {2021},
  url       = {https://openreview.net/forum?id=SSKZPJCt7B}
}

@book{golub2012matrix,
  author    = {Gene H. Golub and Charles F. Van Loan},
  title     = {Matrix Computations},
  publisher = {Johns Hopkins University Press},
  year      = {2012}
}

@inproceedings{verine2023expressivity,
  author    = {A. Verine and B. Negrevergne and Y. Chevaleyre and F. Rossi},
  title     = {On the expressivity of bi-Lipschitz normalizing flows},
  booktitle = {Proceedings of the Asian Conference on Machine Learning (ACML)},
  year      = {2023}
}

@inproceedings{katz2017reluplex,
  author    = {Guy Katz and Clark Barrett and David L. Dill and Kyle Julian and Mykel J. Kochenderfer},
  editor    = {Rupak Majumdar and Viktor Kun{\v{c}}ak},
  title     = {Reluplex: An Efficient SMT Solver for Verifying Deep Neural Networks},
  booktitle = {Computer Aided Verification},
  year      = {2017}
}


%% file: bibliography/springer.bib
@book{bauschke2017convex,
  title     = {Convex Analysis and Monotone Operator Theory in Hilbert Spaces},
  author    = {Bauschke, Heinz H. and Combettes, Patrick L.},
  year      = {2017},
  edition   = {2nd},
  publisher = {Springer}
}

@article{condat2016fast,
  title     = {Fast projection onto the simplex and the $\ell_1$ ball},
  author    = {Condat, Laurent},
  journal   = {Mathematical Programming},
  volume    = {158},
  number    = {1},
  pages     = {575--585},
  year      = {2016},
  publisher = {Springer}
}

@article{held1974validation,
  title     = {Validation of subgradient optimization},
  author    = {Held, Michael and Wolfe, Philip and Crowder, Henry P},
  journal   = {Mathematical Programming},
  volume    = {6},
  number    = {1},
  pages     = {62--88},
  year      = {1974},
  publisher = {Springer}
}

@book{zhang2011matrix,
  title     = {Matrix theory: basic results and techniques},
  author    = {Zhang, Fuzhen},
  year      = {2011},
  publisher = {Springer Science \& Business Media}
}


%% file: references.bib
@inproceedings{araujo_lipschitz_2020,
  title     = {On {Lipschitz} {Regularization} of {Convolutional} {Layers} using {Toeplitz} {Matrix} {Theory}},
  url       = {http://arxiv.org/abs/2006.08391},
  urldate   = {2021-10-15},
  booktitle = {Proceedings of the 35th {AAAI} {Conference} on {Artificial} {Intelligence}},
  author    = {Araujo, Alexandre and Negrevergne, Benjamin and Chevaleyre, Yann and Atif, Jamal},
  month     = nov,
  year      = {2020},
  note      = {arXiv: 2006.08391}
}
